\documentclass[preprint,12pt,nopreprintline]{elsarticle}

\usepackage[T1]{fontenc}
\usepackage[utf8]{inputenc}
\usepackage{textcomp}
\usepackage{amsmath,amssymb}
\usepackage{graphicx}
\usepackage{adjustbox}
\usepackage{algorithm}
\usepackage{algpseudocode}
\usepackage{multirow}
\usepackage{subcaption}
\usepackage{booktabs}
\usepackage{makecell}
\usepackage{bm}
\usepackage[version=4]{mhchem}
\usepackage{siunitx}
\usepackage{longtable,tabularx}
\usepackage{xcolor}
\usepackage[hyphens]{url}
\usepackage[hidelinks,unicode]{hyperref}

\biboptions{sort&compress}

\begin{document}
\begin{frontmatter}
\title{Condition-Enhanced Generative Adversarial Network for Aerodynamic Configuration Design of Commercial Aircraft}

\author[aff1,aff2,aff3]{Youzheng Yu}
\author[aff1,aff2,aff3]{Jiaqing Kou\corref{cor1}}
\ead{jqkou@nwpu.edu.cn}
\author[aff1,aff2,aff3]{Weiwei Zhang}

\affiliation[aff1]{organization={School of Aeronautics, Northwestern Polytechnical University},
                  city={Xi'an},
                  postcode={710072},
                  country={People's Republic of China}}
\affiliation[aff2]{organization={International Joint Institute of Artificial Intelligence on Fluid Mechanics, Northwestern Polytechnical University},
                  city={Xi'an},
                  postcode={710072},
                  country={People's Republic of China}}
\affiliation[aff3]{organization={National Key Laboratory of Aircraft Configuration Design},
                  city={Xi'an},
                  postcode={710072},
                  country={People's Republic of China}}
\cortext[cor1]{Corresponding author.}

\begin{abstract}
Generative machine learning provides an end-to-end framework for aerodynamic design optimization by generating diverse aerodynamic configurations conditioned on target performance. However, the generated designs may violate geometric constraints or deviate from the prescribed aerodynamic performance. Therefore, we propose a Condition-Enhanced Generative Adversarial Network (CEGAN) that embeds geometric and aerodynamic information into the generative process. A condition-informed loss function is developed by combining a geometry-guided feasibility penalty with an aerodynamic performance penalty evaluated by surrogate models trained on the same dataset. Additionally, an output filtering mechanism discards infeasible designs using these surrogate models, ensuring compliance with prescribed conditions. The method is demonstrated on aircraft configurations similar to the Cessna Citation X. Each configuration is represented by 28 shape parameters, whose aerodynamic performance is evaluated using OpenVSP. Interpolation and extrapolation tests are conducted for different network architectures and conditioning strategies. Compared with the baseline model, CEGAN reduces the average generation error by at least 36\% for interpolated conditions and by at least 46\% under extreme extrapolation conditions, while successfully producing diverse and physically feasible designs that extend beyond the training distribution. Therefore, the proposed CEGAN framework offers a robust and accurate tool for rapid aerodynamic configuration exploration during the conceptual design phase. 

\end{abstract}


\end{frontmatter}

\section*{Nomenclature}

{\renewcommand\arraystretch{1.0}
\noindent\begin{longtable}{@{}l @{\quad=\quad} p{0.65\linewidth}@{}}
$\alpha$ & angle of attack \\
$a$ & tolerance threshold for geometric penalty \\
$AR$ & aspect ratio \\
$\bm{c}$ & condition vector \\
$C_L$& lift coefficient \\
$C_D$ & drag coefficient \\
$C_M$ & pitching moment coefficient \\
$g_{k,\mathrm{cond}}$, $C_{i,\mathrm{cond}}$ & specified geometric and aerodynamic condition values \\
$C_i^{\mathrm{VSP}}(\bm{x})$ & VSPAERO result for the same configuration \\
$\epsilon$, $\epsilon_{\mathrm{MRE}}$ & numerical stability constants \\
$\zeta$ & random interpolation coefficient in the gradient penalty \\\
$D$ & discriminator \\
$G$ & generator \\
$L$ & main wing span \\
$L/D$ & lift-to-drag ratio \\
$Ma$  & Mach number \\
$Re$ & Reynolds number \\
$S$ & main wing area\\
$\bm{x}$ & aerodynamic configuration parameters \\
$\bm{\theta}_G$ ,$\bm{\theta}_D$& trainable parameters of the generator and discriminator \\
$\bm{z}$ & random noise\\
$\Lambda$ & mean sweep angle of main wing\\
$\lambda_{\mathrm{GP}}$ & gradient penalty coefficient \\
$\lambda_k$, $\lambda_{C_i}$ & loss weighting coefficients \\
$\mathcal{L}_{\text{geo}} $, $\mathcal{L}_{\text{aero}} $  & geometry-guided and aerodynamic performance-guided loss terms \\
\end{longtable}}

\section{Introduction}
The conventional aircraft design process typically encompasses three phases: conceptual design, preliminary design, and detailed design~\cite{Nicolai2010, raymer2012aircraft}. Among these phases, conceptual design focuses primarily on determining the aerodynamic configuration based on the prescribed mission requirements. The goal of this phase is to rapidly generate and screen optimal configurations to establish the overall baseline for subsequent design stages. Although this phase accounts for only 20\%--30\% of the total design cycle, it mainly determines the aircraft's fundamental performance characteristics and life-cycle cost. Consequently, it represents the most strategically critical phase throughout the entire development process. However, existing conceptual design methodologies rely heavily on expert knowledge for the selection of initial configurations~\cite{Price2011}. Constrained by the subjective preferences of designers, the initial configuration may deviate from the optimal design space dictated by mission requirements, leading to prolonged design iteration cycles and increased development cost~\cite{SUn2014}. Therefore, developing intelligent conceptual design methodologies is essential to overcome these limitations and accelerate the aircraft development process.

Current methodologies for aerodynamic optimization design are primarily categorized into direct design and inverse design approaches. Direct design focuses on minimizing or maximizing aerodynamic objectives subject to constraints, such as airfoil design aimed at maximizing the lift-to-drag ratio~\cite{hicks1974,Jameson1995,Lyu2014,Lyu2015,zhangyufei2021,Zingg2026joa}. In contrast, inverse design aims to directly generate optimal aerodynamic shapes that achieve target performance characteristics, such as geometric design given a specified airfoil pressure coefficient distribution~\cite{lighthill1945new,taka1985,DU2019371}. In principle, existing inverse design methodologies still focus on solving an optimization problem for matching performance requirements. Subsequently, algorithms employed to solve these optimization problems can be broadly classified into gradient-based and gradient-free methods~\cite{multi2013}. Common gradient-based methods include the adjoint method~\cite{PETER2010373} and finite difference method~\cite{Luo2014}. These methods rely on gradient information of the objective function to guide the iterative optimization, offering advantages such as rapid convergence rates and high computational efficiency. However, they suffer from sensitivity to initial guesses and tend to converge to local optima. Furthermore, gradient-based methods require the objective function to be continuously differentiable, making them challenging to apply to complex nonlinear problems. Gradient-free methods mainly include so-called global optimization methods, which can be achieved through heuristic algorithms, such as Genetic Algorithms (GA)~\cite{holland1992adaptation} and Particle Swarm Optimization (PSO)~\cite{pso2012}, and Bayesian optimization~\cite{bayes2025,Leifur2025opti,Leifur2025bayes,Zingg2026bayes}. While capable of exploring global optima and suitable for multi-objective problems, they incur higher computational cost and exhibit slower convergence rates, particularly suffering from the curse of dimensionality in high-dimensional design spaces. However, both aerodynamic optimization methods are iterative processes that converge to a single optimal solution. Consequently, they are well-suited for the detailed design phase where configuration refinement is paramount, but impractical for the extensive design space exploration required during the conceptual design phase.

In recent years, artificial intelligence, machine learning and data-driven methods have been widely adopted in aerodynamics~\cite{koupas,lipas,vepas,nature2023,tang2026,shan1aiaa,shan2aiaa,wangxuaiaa}. Traditional aerodynamic shape optimization benefited from machine learning in many aspects, including lower computational cost, fewer design parameters, etc., while maintaining reasonable accuracy. To achieve efficient design space exploration required during the conceptual design phase, significant attentions has been drawn towards data-driven generative design paradigms~\cite{Zhaohuan2026}. Generative models have been widely applied to scientific computing ~\cite{computer2022}, including materials synthesis~\cite{materials2023}, meteorological forecasting~\cite{nowcast2025}, and mechanical design~\cite{regenwetter2022}. Moreover, their inherent end-to-end generation capability offers an efficient new pathway for the conceptual design phase of aircraft. Commonly employed generative models include Generative Adversarial Networks (GANs)~\cite{goodfellow2014generative}, Variational Auto-Encoders (VAEs)~\cite{Kingma2014}, and Diffusion Models~\cite{SohlDickstein2015}. As discussed in~\cite{models2025}, GANs generate high-quality and realistic samples through adversarial training between a generator and a discriminator; however, their training process can be unstable. VAEs utilize probabilistic modeling of the latent space, offering favorable continuity and interpretability, though they typically exhibit inferior performance in sample detail compared to GANs. Diffusion models are renowned for generating high-quality samples, particularly suited for complex data distributions, yet they incur higher training and inference cost.

Recently, Generative models have been extensively applied to aerodynamic optimization design~\cite{Chen2019,Yilmaz2020,Chen2021,alonso2023,alonso2026}. In airfoil design, Li et al.~\cite{li2020efficient} proposed a parameterization method based on a Deep Convolutional Generative Adversarial Network (DCGAN) to enhance optimization efficiency. Wang et al.~\cite{Wangxu2025} utilized an improved conditional VAE to achieve diverse airfoil generation given a particular lift coefficient. Anand et al.~\cite{Anand2026} proposed a tandem neural network architecture, providing a valuable reference for constraint embedding in data-driven inverse design. For three-dimensional configuration representation, Chen et al.~\cite{Chenyan2026} achieved dimensionality reduction of geometric parameters for supersonic aircraft based on GAN. Wei et al.~\cite{Wei2026} proposed a parameterization scheme termed Deep Geometric Mapping (DeepGeo), which effectively mitigates the curse of dimensionality commonly encountered in aerodynamic shape optimization. Bamford et al.~\cite{Bamford2026} developed a generative Adversarial AutoEncoder (AAE) for direct generation of three-dimensional smooth aerodynamic configurations. To further satisfy engineering application requirements, Lou et al.~\cite{Lou2026} developed a conditional design method for three-dimensional axisymmetric flight vehicle configurations based on diffusion models. Yu et al.~\cite{Yu2026Aerodynamic} implemented conditional generative design for conventional configuration of aircraft based on Conditional Wasserstein Generative Adversarial Network with Gradient Penalty (CWGAN-GP). However, although CWGAN-GP effectively mitigates training instability and mode collapse, its generation accuracy under multiple coupled design conditions remains a subject for further improvement. Previous studies have advanced generative inverse design through performance guidance and conditional generation. Chen and Ahmed~\cite{chenwei2021} incorporated performance evaluation and a diversity-promoting term into GAN training to improve the performance and diversity of generated design datasets. Wang et al.~\cite{Wang2022NacelleGAN} combined an improved GAN with neural networks to establish a pressure-distribution-based framework for nacelle inverse design. Lin et al.~\cite{Lin2026} proposed a velocity-prediction lightweight diffusion model for three-dimensional aircraft inverse design. Building on these advances, further investigation is needed to improve generation accuracy under prescribed conditions, particularly when geometric and aerodynamic conditions must be satisfied simultaneously.

To further enhance the performance of generative models while simultaneously exploring diverse configurations during the conceptual design phase, this work introduces a Condition-Enhanced Generative Adversarial Network (CEGAN) framework. The proposed model is built upon a robust CWGAN-GP structure, which leverages the Wasserstein distance with gradient penalty to guarantee effective and stable training. To ensure reasonable accuracy under multiple design conditions, we argument the loss function with two strategies: a geometry-guided enhancement and an aerodynamic performance-guided enhancement that incorporates a surrogate model. During inference, the same surrogate model is employed as an output filter to discard physically inconsistent designs. Consequently, the proposed CEGAN framework significantly enhances generation accuracy under multiple conditions, preserves configuration diversity, and overcomes the degradation in generation quality under extrapolated conditions.

The rest of this paper is organized as follows. Section~\ref{sec:method} presents the proposed methodology, encompassing the various enhancement strategies of the CEGAN framework and the model evaluation metrics. Section~\ref{sec:3} introduces the test case, details the specific model configurations, and presents the generation results. Finally, conclusions are summarized in Section~\ref{sec:conclusions}.

\section{Methodology}
\label{sec:method}

This section introduces the overall framework of the proposed approach. Sec.~\ref{sec:CWGAN-GP} introduces the standard CWGAN-GP architecture and its loss functions to achieve stable model training. Building upon this baseline, Sec.~\ref{sec:CEGAN-enhanced} elaborates on the condition enhancement strategies achieved through the loss function. Sec.~\ref{sec:filter} presents the output filtering method employed during the generation phase. The overall CEGAN framework is then summarized in Sec.~\ref{sec:overall}, followed by a description of the model evaluation metrics in Sec.~\ref{sec:Model Evaluation}. As a preliminary step, we first define the input and output of the generative model. The objective of generative design is to learn a generative mapping function $G$, as defined in Eq.~\ref{eq:gen}. With fixed learned parameters $\bm{\theta}_G$, this function takes a latent variable $\bm{z}$ and a condition vector $\bm{c}$ as inputs to generate diverse design parameters $\bm{x}$. Here, $\bm{c}$ specifies the geometric and aerodynamic conditions that the generated designs are expected to satisfy.

\begin{equation}
\bm{x} = G(\bm{z}, \bm{c}; \bm{\theta}_G)
\label{eq:gen}
\end{equation}

\subsection{CWGAN-GP}
\label{sec:CWGAN-GP}

GANs are a class of deep learning models rooted in game theory, first introduced by Goodfellow et al. \cite{goodfellow2014generative} in 2014. The core principle of GANs involves adversarial training between two neural networks to achieve data generation: the Generator ($G$) aims to produce realistic data samples from random noise inputs, while the Discriminator ($D$) aims to distinguish between generated and real samples. During training, these two networks compete against each other, ultimately enabling the generator to produce samples that align closely with the real data distribution. The loss function is formulated as follows.

\begin{equation}
\min _{G} \max _{D} V(D, G)=
\underbrace{\mathbb{E}_{\boldsymbol{x} \sim P_{\text {data }}}[\log D(\boldsymbol{x})]}_{\text{Real}} \\
+\underbrace{\mathbb{E}_{\bm{z} \sim P_{\bm{z}}}[\log (1-D(G(\bm{z})))]}_{\text{Fake}}
\label{eq:gan}
\end{equation}
where $\boldsymbol{x}$ and $\bm{z}$ denote real training samples and random noise inputs, respectively. $G$ and $D$ represent the generator and discriminator networks. As the theory and applications of GANs have continuously expanded, leading to the development of various improved models designed to address issues inherent in the original GAN, such as mode collapse and training instability~\cite{arxiv2017}. Mode collapse refers to the phenomenon where the generator produces a limited variety of samples that fail to cover the full diversity of the training data distribution.

\subsubsection{CGAN}

\begin{figure}[hbt!]
\centering
\includegraphics[width=.75\textwidth]{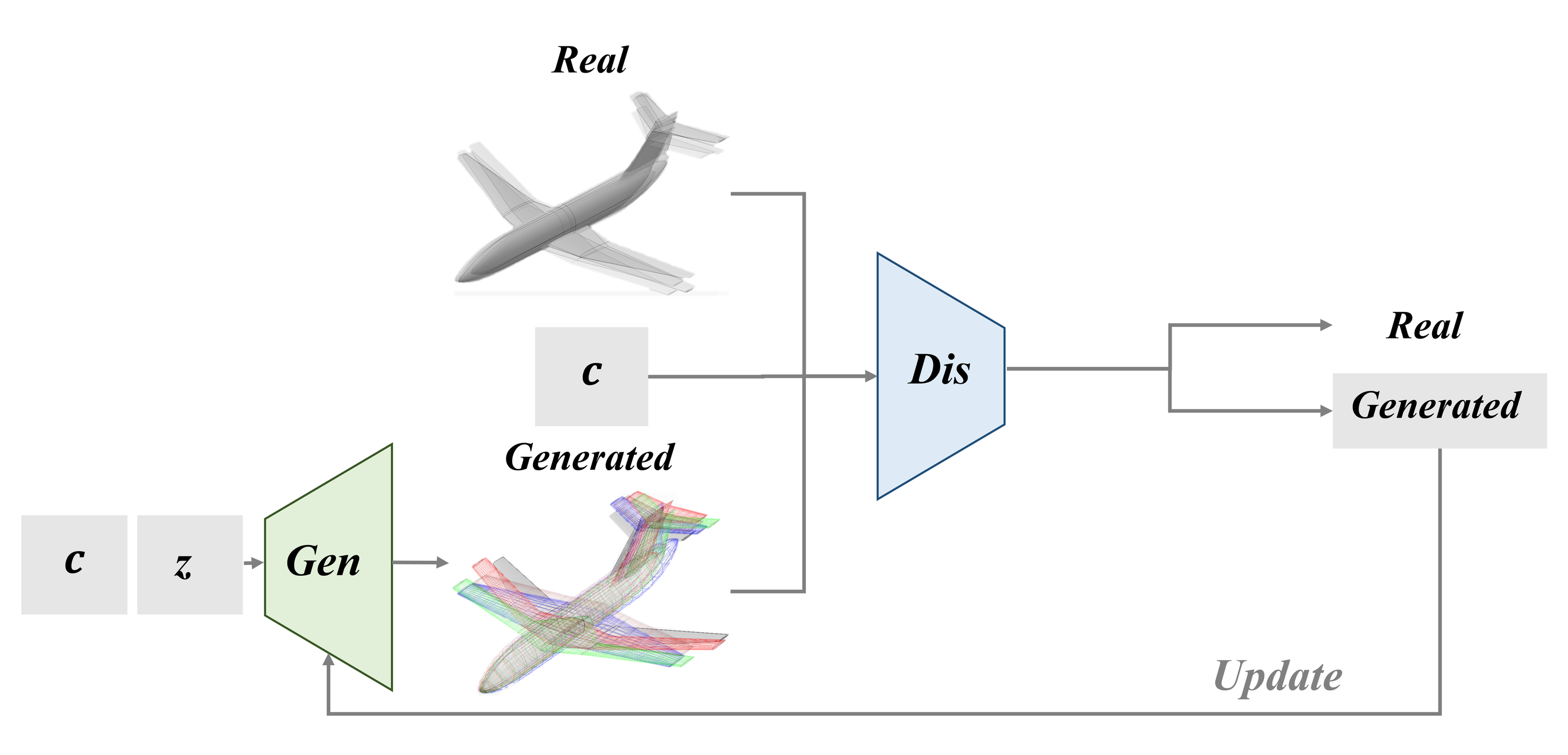}
\caption{Model structure of CGAN.}
\label{fig:CGAN}
\end{figure}

The Conditional GAN (CGAN) was proposed by Mirza et al.~\cite{mirza2014conditional} in 2014. As illustrated in Figure~\ref{fig:CGAN}, the inputs to both the generator and the discriminator in a CGAN include not only random noise $\bm{z}$ but also a condition vector $\bm{c}$. This condition vector specifies the design requirements. Providing these design conditions guides the generator to produce designs that approximately satisfy them, depending on the accuracy of the model. The loss function of the CGAN is expressed as follows.

\begin{equation}
\min _{G} \max _{D} V(D, G)=
\underbrace{\mathbb{E}_{\boldsymbol{x} \sim P_{\text {data }}}[\log D(\boldsymbol{x} \mid \boldsymbol{c})]}_{\text{Real}} \\
+\underbrace{\mathbb{E}_{\boldsymbol{z} \sim P_{\boldsymbol{z}}}[\log (1-D(G(\boldsymbol{z} \mid \boldsymbol{c})))]}_{\text{Fake}}
\label{eq:cgan}
\end{equation}
where $\boldsymbol{c}$ is the condition vector (e.g., aerodynamic coefficient values $[C_{L,\mathrm{cond}},C_{D,\mathrm{cond}},C_{M,\mathrm{cond}}]^\top$), and other symbols remain consistent with Eq.~\ref{eq:gan}. CGANs enable the generation of valid designs under conditions, which is important in practical applications. However, they essentially retain the loss function structure of the original GAN, failing to fundamentally mitigate the inherent issues of training instability and mode collapse. In practice, particularly when the condition space is complex or the data distribution is sparse, the generator still exhibits insufficient diversity among the generated samples.

\subsubsection{CWGAN-GP}

\begin{figure}[hbt!]
\centering
\includegraphics[width=.75\textwidth]{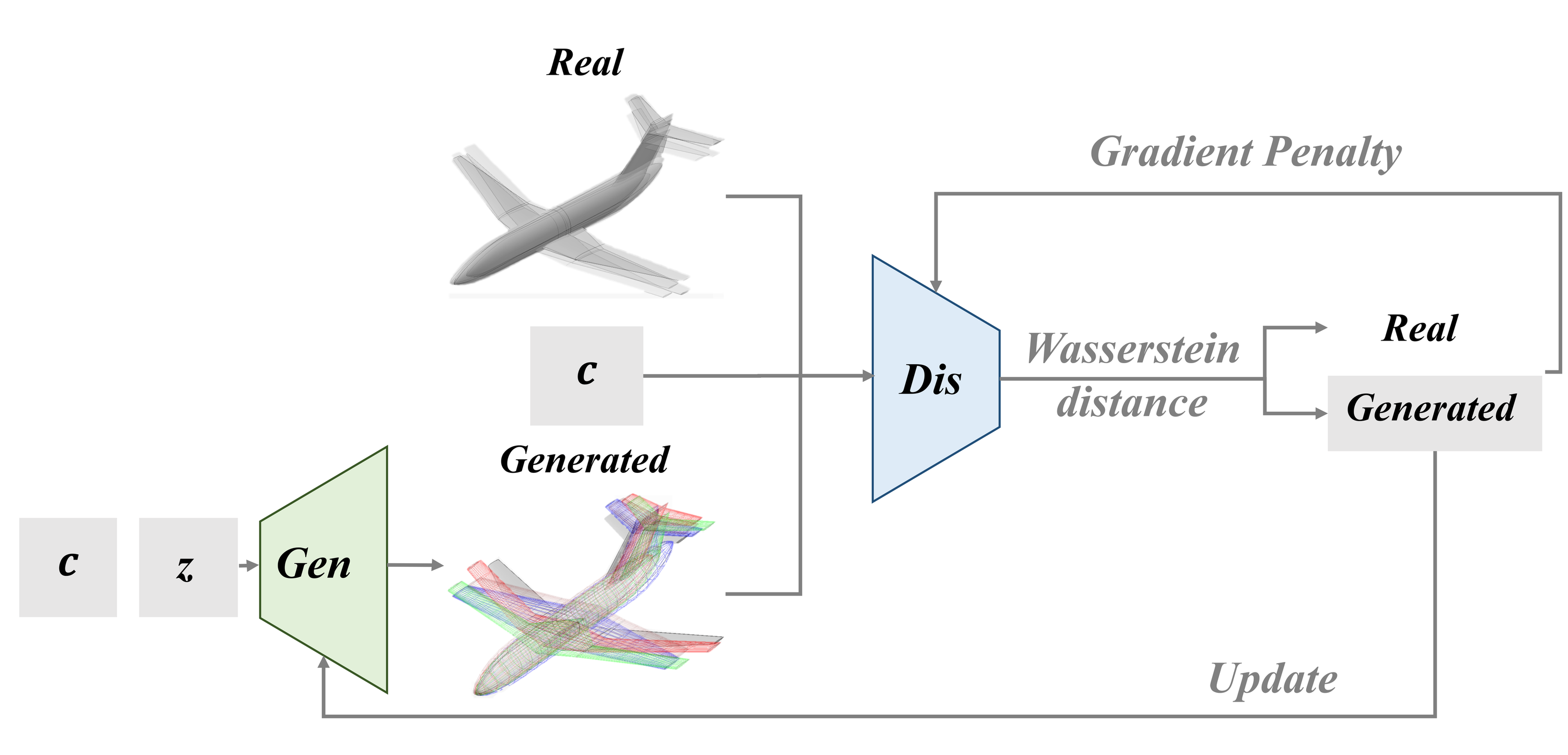}
\caption{Model structure of CWGAN-GP.}
\label{fig:CWGAN-GP}
\end{figure}

CWGAN-GP is a generative model that combines CGAN with Wasserstein GAN (WGAN) and Gradient Penalty (WGAN-GP)~\cite{arxiv2017}. In contrast to CGAN, CWGAN-GP employs the Wasserstein distance and a gradient penalty term to measure the discrepancy between generated and real samples, thereby mitigating training instability and improving model performance. 
In the original WGAN, network weights are explicitly confined to the interval $[-\xi, \xi]$ to enforce the Lipschitz condition. The gradient penalty term defined in Eq.~\ref{eq:gradient_penalty} effectively alleviates the adverse effects of this weight clipping strategy.

\begin{equation}
\mathcal{L}_{\text{GP}} = \mathbb{E}_{\hat{\boldsymbol{x}} \sim P_{\hat{\boldsymbol{x}}}} \left[ \left( \|\nabla_{\hat{\boldsymbol{x}}} D(\hat{\boldsymbol{x}} \mid \boldsymbol{c}) \|_2 - 1 \right)^2 \right]
\label{eq:gradient_penalty}
\end{equation}
where $\hat{\boldsymbol{x}} = \zeta \boldsymbol{x} + (1-\zeta) G(\boldsymbol{z} \mid \boldsymbol{c})$ denotes interpolated samples weighted linearly between real data $\boldsymbol{x}$ and generated samples $G(\boldsymbol{z} \mid \boldsymbol{c})$, with $\zeta \sim \mathcal{U}(0,1)$; $\nabla_{\hat{\boldsymbol{x}}} D(\cdot)$ represents the gradient of the output from the Critic with respect to $\hat{\boldsymbol{x}}$ (in the WGAN framework, the discriminator is referred to as the Critic because it outputs a scalar score rather than a probability). $\|\cdot\|_2$ denotes the $L_2$ norm. The penalty enforces the 1-Lipschitz constraint by encouraging the gradient norm to remain close to 1. The final loss function of the CWGAN-GP model is formulated as Eq.~\ref{eq:cwgan_gp}.

\begin{equation}
\begin{gathered}
\min _{G} \max _{D} V(D, G)= \\
\underbrace{\mathbb{E}_{\boldsymbol{x} \sim P_{\text {data }}}[D(\boldsymbol{x} \mid \boldsymbol{c})]
-\mathbb{E}_{z \sim P_{\boldsymbol{z}}}[D(G(\boldsymbol{z} \mid \boldsymbol{c}) \mid \boldsymbol{c})]}_{\text{Wasserstein Distance}} \\
-\underbrace{\lambda_{\mathrm{GP}} \mathbb{E}_{\hat{\boldsymbol{x}} \sim P_{\hat{\boldsymbol{x}}}} \left[ \left( \|\nabla_{\hat{\boldsymbol{x}}} D(\hat{\boldsymbol{x}} \mid \boldsymbol{c}) \|_2 - 1 \right)^2 \right]}_{\text{Gradient Penalty}}
\end{gathered}
\label{eq:cwgan_gp}
\end{equation}
where $\boldsymbol{x}$, $\bm{z}$, $\boldsymbol{c}$, $G$, and $D$ are consistent with Eq.~\ref{eq:cgan}, and $\hat{\boldsymbol{x}}$ is defined in Eq.~\ref{eq:gradient_penalty}. $\lambda_{\mathrm{GP}}$ denotes the gradient penalty coefficient (set to 10 in this study).

\subsection{Condition-Informed Loss Function}
\label{sec:CEGAN-enhanced}
\subsubsection{Geometry-Guided Enhancement}
In aircraft conceptual design, geometric parameters influence not only aerodynamic performance but also manufacturing feasibility and operational constraints. For instance, wing span is often constrained by airport runway widths or parking space. Wing area affects wing loading, structural weight, and manufacturing cost. Mean sweep angle is closely linked to transonic performance and wing stiffness design. To enhance the engineering applicability of generated samples and incorporate an interpretable model improvement, this study explicitly introduces a geometric penalty mechanism in the loss function of the generator, aiming to improve generation accuracy under geometric conditions. During model training, the penalty is imposed when the relative error of generated geometric parameters (e.g., wing area, span, sweep angle) with respect to their geometric condition values exceeds a predefined tolerance $a$; otherwise, no penalty is imposed.

Let $\bm g(\bm x)=[g_1(\bm x),\ldots,g_K(\bm x)]^\top$ contain the $K$ geometric quantities evaluated on a generated configuration. Their specified values form $\bm g_{\mathrm{cond}}=[g_{1,\mathrm{cond}},\ldots,g_{K,\mathrm{cond}}]^\top$. The subscript $\mathrm{cond}$ denotes a specified condition value.

The geometry-guided loss is
\begin{equation}
\mathcal L_{\text{geo}} = \sum_{k=1}^{K}\lambda_k
\left[\max\left(
\frac{|g_k(\bm x)-g_{k,\mathrm{cond}}|}{g_{k,\mathrm{cond}}+\epsilon}-a,\,0
\right)\right]
\label{eq:geo}
\end{equation}
Here, $\epsilon=10^{-8}$ prevents division by zero, and $a$ is the tolerance in the geometry-guided loss.

\subsubsection{Aerodynamic Performance-Guided Enhancement}

In aircraft conceptual design, the engineering applicability of generated samples depends not only on geometric feasibility but also critically on how accurate the aerodynamic performance is satisfied, such as the lift coefficient $C_L$ and drag coefficient $C_D$. Although CWGAN-GP models can learn conditional distributions in a statistical sense, the generator lacks proper mechanism to ensure the generated data satisfy the imposed condition. This absence of supervision on errors relative to specific aerodynamic conditions limits the generation accuracy, highlighting the need for further improvement.

\begin{figure}[hbt!]
\centering
\includegraphics[width=.75\textwidth]{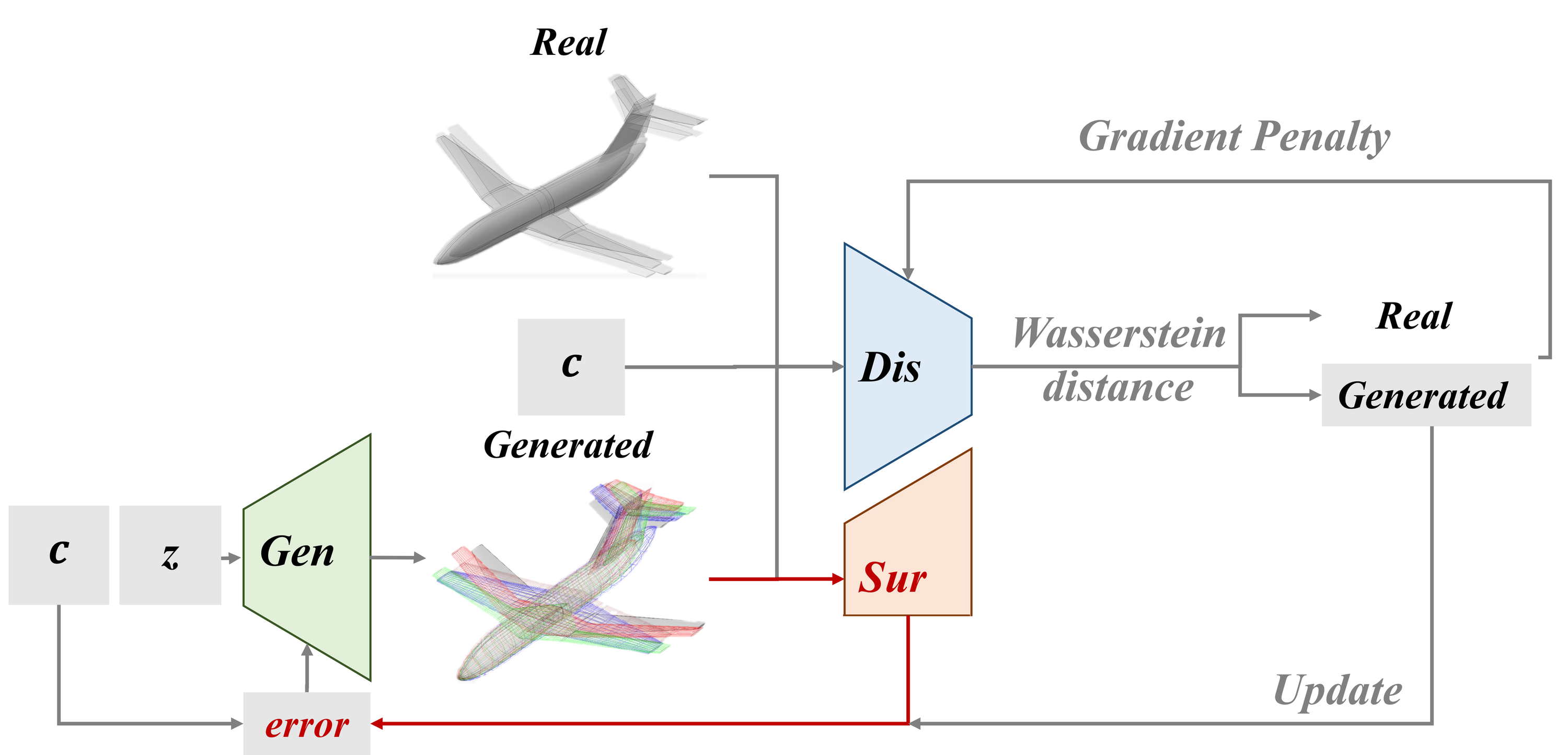}
\caption{Aerodynamic performance-guided enhancement applied to CWGAN-GP.}
\label{fig:regression-enhanced}
\end{figure}

As illustrated in Figure~\ref{fig:regression-enhanced}, to enhance the model accuracy in satisfying conditions related to aerodynamic performance, an aerodynamic performance-guided enhancement strategy is introduced in this work. Specifically, a surrogate model is developed to establish the nonlinear mapping between geometric parameters and aerodynamic performance. Subsequently, the discrepancy between the model predictions and the aerodynamic conditions is embedded into the loss function of the generator during model training, thus improving the accuracy of data generation. The surrogate model is trained on the same dataset as the generative model, making sufficient use of the training data. As outlined in Algorithm~\ref{alg:mlp_surrogate}, we employ multiple independent single-output feedforward neural networks to construct surrogate models for each aerodynamic coefficient, including the lift coefficient $C_L$, drag coefficient $C_D$, and moment coefficient $C_M$. 
Each surrogate takes the geometric parameter vector $\bm x\in\mathbb R^n$ as input and predicts one aerodynamic coefficient. An independent MLP is trained for each coefficient as an initial proof of concept.

\begin{algorithm}[htbp]
\renewcommand{\algorithmicrequire}{\textbf{Input:}}
\renewcommand{\algorithmicensure}{\textbf{Output:}}
\caption{Training of a Single-Output Surrogate Model based on MLP}
\label{alg:mlp_surrogate}
\begin{algorithmic}[1]
\Require
    One aerodynamic coefficient $i\in\{L,D,M\}$; \\
    Training dataset $\mathcal D_i=\{(\bm x^{(j)},C_i^{\mathrm{VSP}}(\bm x^{(j)}))\}_{j=1}^{N}$, where: \\
    \quad $\bm x^{(j)}\in\mathbb R^n$ is the geometry of sample $j$; \\
    \quad $C_i^{\mathrm{VSP}}(\bm x^{(j)})\in\mathbb R$ is its VSPAERO reference value; \\
    \quad $N$ is the number of training samples and $n$ is the input dimension.
\Ensure
    Trained surrogate $f_i:\mathbb R^n\to\mathbb R$, with prediction $\hat C_i(\bm x)=f_i(\bm x)$.
\State Initialize an MLP $f_i$ with $n$ inputs and one scalar output.
\State Define the mean squared error loss:
    \[
    \mathcal L_{\mathrm{MSE},i}=\frac{1}{N}\sum_{j=1}^{N}
    \left[f_i(\bm x^{(j)})-C_i^{\mathrm{VSP}}(\bm x^{(j)})\right]^2.
    \]
\State Train $f_i$ by minimizing $\mathcal L_{\mathrm{MSE},i}$ using Adam until convergence.
\State \Return $f_i$
\end{algorithmic}
\end{algorithm}

An aerodynamic loss guides generated samples toward the specified aerodynamic conditions. For $i\in\{L,D,M\}$, $\hat C_i(\bm x)=f_i(\bm x)$ is the surrogate prediction and $C_{i,\mathrm{cond}}$ is the specified condition value. The VSPAERO result for the same geometry is denoted by $C_i^{\mathrm{VSP}}(\bm x)$. The loss penalizes differences between the predicted coefficients and their specified condition values:

\begin{equation}
\mathcal L_{\text{aero}} = \sum_{i\in\{L,D,M\}}\lambda_{C_i}
\sqrt{\left[\left(\hat C_i(\bm x)-C_{i,\mathrm{cond}}\right)^2\right]}
\label{eq:aero}
\end{equation}

\subsection{Output Filtering}
\label{sec:filter}

\begin{figure}[hbt!]
\centering
\includegraphics[width=.75\textwidth]{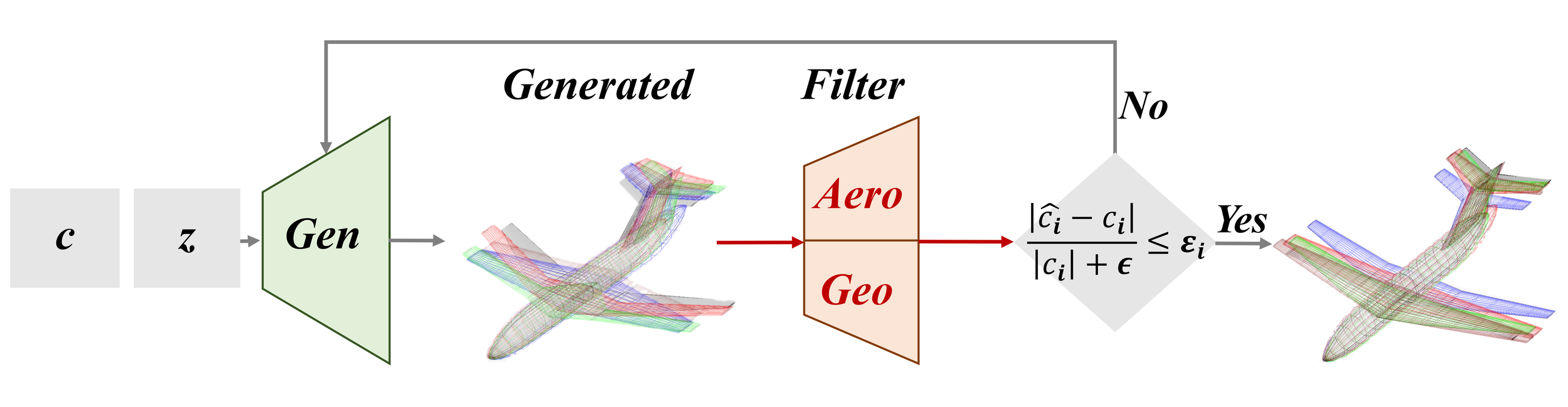}
\caption{Output filtering applied to generator.}
\label{fig:Output Filtering}
\end{figure}

To ensure the quality of generated results, we further introduce an output filtering procedure after the trained generator to ensure the accuracy and quality of the generated data. As shown in Figure~\ref{fig:Output Filtering}, this procedure employs the trained surrogate models as a filter to discard infeasible samples from the generator that do not properly satisfy the imposed aerodynamic conditions. Furthermore, the key geometric constraints (e.g., wing area $S$, span $L$, and sweep angle $\Lambda$) are also evaluated to check whether the imposed geometric conditions are satisfied. Specifically, a generated sample is considered valid only if the following criterion is met for all design conditions:
\begin{equation}
\frac{|\hat c_i-c_i|}{|c_i|+\epsilon}\leq\varepsilon_i
\label{eq:filter}
\end{equation}
In Eq.~\ref{eq:filter}, the index $i$ runs over all design quantities. The value $c_i$ is the specified condition, and $\hat c_i$ is its estimated value for a generated configuration. Geometric quantities are evaluated directly, whereas aerodynamic quantities are predicted by the surrogate models. The tolerance is $\varepsilon_i$, and $\epsilon=10^{-8}$ prevents division by zero. Filtering continues until enough valid samples have been obtained. It uses the already trained surrogates.

In Appendix A, the proposed output filtering approach has been successfully validated on a conditional generative model with two aerodynamic conditions. When combined with an aerodynamic performance-guided enhancement strategy, it further enhances the accuracy of the generated designs. 

\subsection{Overall Model Structure}
\label{sec:overall}
\begin{figure}[hbt!]
\centering
\includegraphics[width=\linewidth]{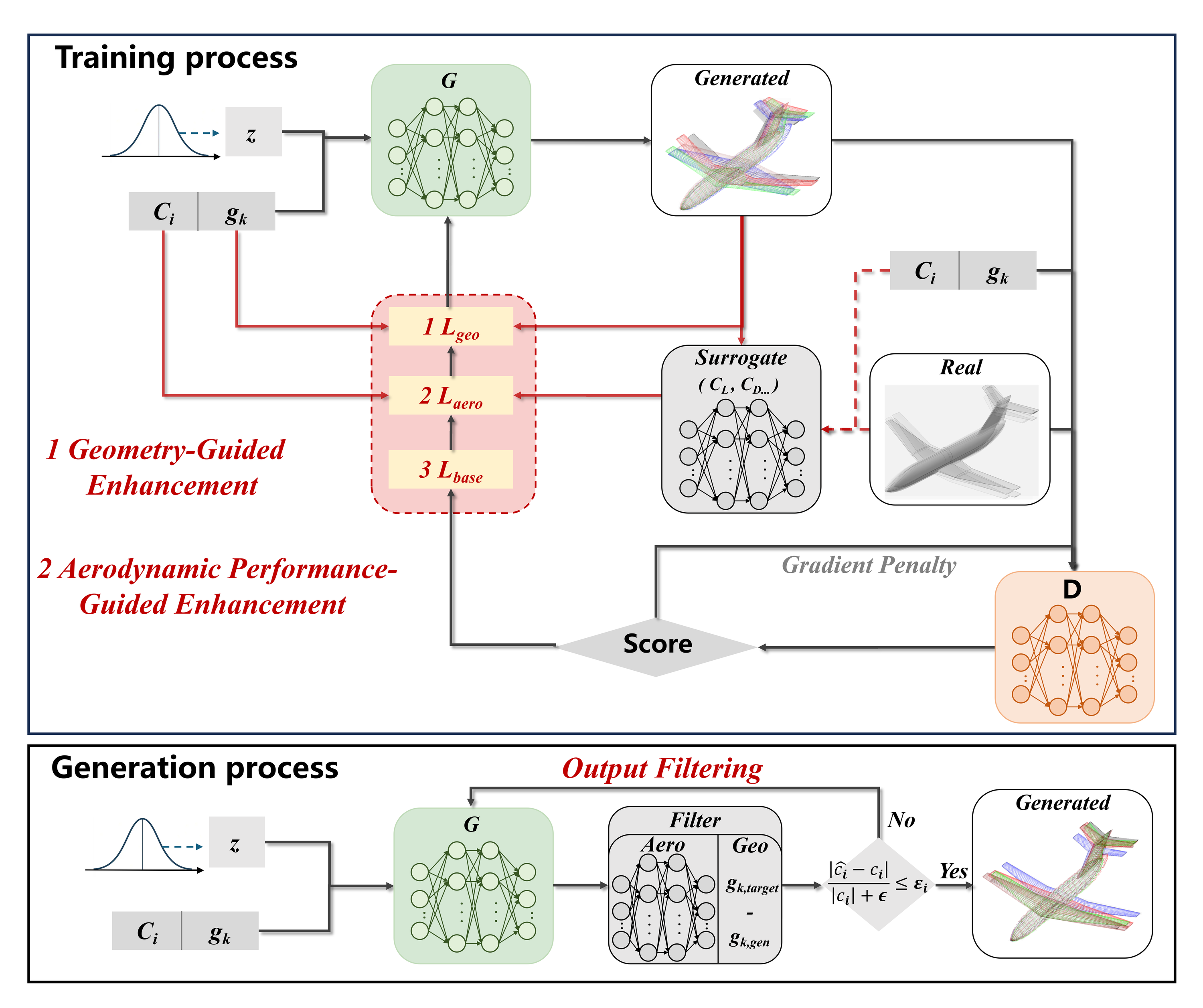}
\caption{Overall structure of CEGAN.}
\label{fig:CEGAN}
\end{figure}

The proposed Condition-Enhanced Generative Adversarial Network (CEGAN) arguments the loss functions of the CWGAN-GP by incorporating two key components: a geometry-guided enhancement to enforce geometric constraints during training, and an aerodynamic performance-guided enhancement to align generated samples with aerodynamic conditions. Furthermore, an output filtering approach is applied to the generation phase to further improve the quality and accuracy of the generator. Together, these methods form a complete condition-enhanced generative framework in this study. The overall structure is illustrated in Figure~\ref{fig:CEGAN}.

During training, the generator takes a Gaussian noise vector $\bm z$ and a condition vector $\bm c=[g_{1,\mathrm{cond}},\ldots,g_{K,\mathrm{cond}},C_{L,\mathrm{cond}},C_{D,\mathrm{cond}},C_{M,\mathrm{cond}}]^\top$. It generates geometric parameters $\bm x_g=G(\bm z,\bm c)$. The number and types of conditions depend on the design problem. Its loss is
\begin{equation}
\begin{gathered}
\mathcal L_G=
\underbrace{-\mathbb E_{\bm z\sim P_{\bm z}}[D(G(\bm z\mid\bm c)\mid\bm c)]}_{\mathcal L_{\text{base}}}\\
+\underbrace{\sum_{k=1}^{K}\lambda_k
\left[\max\left(
\frac{|g_k(\bm x_g)-g_{k,\mathrm{cond}}|}{g_{k,\mathrm{cond}}+\epsilon}-a,\,0
\right)\right]}_{\mathcal L_{\text{geo}}}\\
+\underbrace{\sum_{i\in\{L,D,M\}}\lambda_{C_i}
\sqrt{\left[\left(\hat C_i(\bm x_g)-C_{i,\mathrm{cond}}\right)^2\right]}}_{\mathcal L_{\text{aero}}}.
\end{gathered}
\label{eq:all-g}
\end{equation}
The first term is the generator's adversarial loss. The second penalizes geometric deviations beyond $a$, and the third reduces the differences between surrogate predictions and the specified aerodynamic conditions. Here, $\hat C_i(\bm x_g)=f_i(\bm x_g)$, and all condition values are taken from $\bm c$. The generator update minimizes Eq.~\ref{eq:all-g}. The gradient penalty is used only when updating the discriminator. The complete objective is given in Eq.~\ref{eq:all}. Algorithms~\ref{alg:cegan_training} and~\ref{alg:cegan_filtering} summarize training and generation, respectively.

\begin{equation}
\begin{gathered}
\min_G\max_D V(D,G)=\\
\underbrace{\mathbb E_{\bm x\sim P_{\text{data}}}[D(\bm x\mid\bm c)]
-\mathbb E_{\bm z\sim P_{\bm z}}[D(G(\bm z\mid\bm c)\mid\bm c)]}_{\text{Wasserstein Distance}}\\
-\underbrace{\lambda_{\mathrm{GP}}\mathbb E_{\hat{\bm x}\sim P_{\hat{\bm x}}}
\left[(\|\nabla_{\hat{\bm x}}D(\hat{\bm x}\mid\bm c)\|_2-1)^2\right]}_{\text{Gradient Penalty}}\\
+\underbrace{\sum_{k=1}^{K}\lambda_k
\left[\max\left(
\frac{|g_k(\bm x_g)-g_{k,\mathrm{cond}}|}{g_{k,\mathrm{cond}}+\epsilon}-a,\,0
\right)\right]}_{\mathcal L_{\text{geo}}}\\
+\underbrace{\sum_{i\in\{L,D,M\}}\lambda_{C_i}
\sqrt{\left[\left(\hat C_i(\bm x_g)-C_{i,\mathrm{cond}}\right)^2\right]}}_{\mathcal L_{\text{aero}}}.
\end{gathered}
\label{eq:all}
\end{equation}

\begin{algorithm}[htbp]
\renewcommand{\algorithmicrequire}{\textbf{Input:}}
\renewcommand{\algorithmicensure}{\textbf{Output:}}
\caption{Training Procedure of CEGAN}
\label{alg:cegan_training}
\linespread{.9}\selectfont
\begin{algorithmic}[1]
\Require
    Training dataset $\mathcal D=\{(\bm x^{(j)},\bm c^{(j)})\}_{j=1}^{N}$; \\
    \quad $\bm x^{(j)}\in\mathbb R^n$ is the geometry of sample $j$; \\
    \quad $\bm c^{(j)}$ contains $g_{k,\mathrm{cond}}^{(j)}$ and $C_{i,\mathrm{cond}}^{(j)}$; \\
    \quad for training samples, $g_{k,\mathrm{cond}}^{(j)}=g_k(\bm x^{(j)})$ and $C_{i,\mathrm{cond}}^{(j)}=C_i^{\mathrm{VSP}}(\bm x^{(j)})$; \\
    Pre-trained surrogates $f_L,f_D,f_M$; trainable parameters $\bm\theta_G,\bm\theta_D$; \\
    Loss weights $\lambda_{\mathrm{GP}},\{\lambda_k\}_{k=1}^{K},\{\lambda_{C_i}\}_{i\in\{L,D,M\}}$; geometric tolerance $a$; \\
    Learning rates $\eta_G,\eta_D$; discriminator updates $n_{\mathrm{dis}}$; iterations $N_{\mathrm{iter}}$.
\Ensure Trained generator $G$ and discriminator $D$.
\State Keep surrogate weights fixed and retain gradients with respect to their inputs.
\For{$s=1$ \textbf{to} $N_{\mathrm{iter}}$}
    \For{$t=1$ \textbf{to} $n_{\mathrm{dis}}$}
        \State Sample real sample--condition pairs $(\bm x_r,\bm c_r)$ from $\mathcal D$.
        \State Sample $\bm z\sim\mathcal N(0,I)$ and generate $\bm x_f=G(\bm z,\bm c_r)$.
        \State $\mathcal L_D=\mathbb E[D(\bm x_f,\bm c_r)]-\mathbb E[D(\bm x_r,\bm c_r)]$
        \State Sample $\zeta\sim\mathcal U(0,1)$ and form $\hat{\bm x}=\zeta\bm x_r+(1-\zeta)\bm x_f$.
        \State $\mathcal L_{\mathrm{GP}}=\mathbb E[(\|\nabla_{\hat{\bm x}}D(\hat{\bm x},\bm c_r)\|_2-1)^2]$
        \State $\bm\theta_D\gets\bm\theta_D-\eta_D\nabla_{\bm\theta_D}(\mathcal L_D+\lambda_{\mathrm{GP}}\mathcal L_{\mathrm{GP}})$
    \EndFor
    \State Sample condition vectors from $\{\bm c^{(j)}\}_{j=1}^{N}$ and noise $\bm z\sim\mathcal N(0,I)$.
    \State Generate $\bm x_g=G(\bm z,\bm c)$.
    \State $\mathcal L_{\mathrm{base}}=-\mathbb E[D(\bm x_g,\bm c)]$
    \State $\mathcal L_{\mathrm{geo}}=\sum_{k=1}^{K}\lambda_k\left[\max\left(\frac{|g_k(\bm x_g)-g_{k,\mathrm{cond}}|}{g_{k,\mathrm{cond}}+\epsilon}-a,0\right)\right]$
    \State Predict $\hat C_i(\bm x_g)=f_i(\bm x_g)$ for $i\in\{L,D,M\}$.
    \State $\mathcal L_{\mathrm{aero}}=\sum_{i\in\{L,D,M\}}\lambda_{C_i}\sqrt{\left[(\hat C_i(\bm x_g)-C_{i,\mathrm{cond}})^2\right]}$
    \State $\mathcal L_G=\mathcal L_{\mathrm{base}}+\mathcal L_{\mathrm{geo}}+\mathcal L_{\mathrm{aero}}$
    \State $\bm\theta_G\gets\bm\theta_G-\eta_G\nabla_{\bm\theta_G}\mathcal L_G$
\EndFor
\State \Return $G,D$
\end{algorithmic}
\end{algorithm}

\begin{algorithm}[htbp]
\renewcommand{\algorithmicrequire}{\textbf{Input:}}
\renewcommand{\algorithmicensure}{\textbf{Output:}}
\caption{Generation Procedure with Output Filtering}
\label{alg:cegan_filtering}
\begin{algorithmic}[1]
\Require
    Trained generator $G$ and surrogates $f_L,f_D,f_M$; \\
    Condition vector $\bm c=[g_{1,\mathrm{cond}},\ldots,g_{K,\mathrm{cond}},C_{L,\mathrm{cond}},C_{D,\mathrm{cond}},C_{M,\mathrm{cond}}]^\top$; \\
    Tolerances $\varepsilon_{g_k}$ ($k=1,\ldots,K$) and $\varepsilon_{C_i}$ ($i\in\{L,D,M\}$); \\
    Desired number of valid samples $N_{\mathrm{valid}}$.
\Ensure A set $\mathcal S$ of generated geometries satisfying the specified conditions within their tolerances.
\State Initialize $\mathcal S=\emptyset$.
\While{$|\mathcal S|<N_{\mathrm{valid}}$}
    \State Sample $\bm z\sim\mathcal N(0,I)$ and generate $\bm x=G(\bm z,\bm c)$.
    \State Predict $\hat C_i(\bm x)=f_i(\bm x)$ for $i\in\{L,D,M\}$.
    \State Compute $\delta_{C_i}=\frac{|\hat C_i(\bm x)-C_{i,\mathrm{cond}}|}{|C_{i,\mathrm{cond}}|+\epsilon}$ for $i\in\{L,D,M\}$.
    \State Evaluate geometric quantities $g_k(\bm x)$ for $k=1,\ldots,K$.
    \State Compute $\delta_{g_k}=\frac{|g_k(\bm x)-g_{k,\mathrm{cond}}|}{|g_{k,\mathrm{cond}}|+\epsilon}$ for $k=1,\ldots,K$.
    \If{$\delta_{C_i}\leq\varepsilon_{C_i}$ for all $i\in\{L,D,M\}$ \textbf{and} $\delta_{g_k}\leq\varepsilon_{g_k}$ for all $k=1,\ldots,K$}
        \State $\mathcal S\gets\mathcal S\cup\{\bm x\}$.
    \EndIf
\EndWhile
\State \Return $\mathcal S$
\end{algorithmic}
\end{algorithm}

With the proposed enhancement strategies, the accuracy of the conditional generation can be flexibly tuned by adjusting the hyperparameters $\lambda_k$ and $\lambda_{C_i}$, depending on specific engineering requirements. When the task emphasizes geometric feasibility (e.g., compatibility with airport runway constraints), $\lambda_k$ can be increased; when aerodynamic precision is prioritized (e.g., quantitative improvement of the lift coefficient), $\lambda_{C_i}$ can be increased accordingly. To further improve the overall fidelity of the generated designs with respect to the conditions, the relative weighting of the two hyperparameters can also be jointly tuned. Note that the output filtering procedure is applied only to the generation phase to perform a secondary validation of the results, as detailed in Algorithm~\ref{alg:cegan_filtering}. It should be noted that the training dataset is sufficiently used in this work. It is used for the training of both the generative model and the surrogate model, while the surrogate model is also used during the output filtering process. This conditional generation framework is general, and can be extended to other design conditions in future work, such as payload or range requirements. Representative generation results are presented in Section~\ref{sec:3}.

\subsection{Model Evaluation}
\label{sec:Model Evaluation}
To comprehensively evaluate the performance of the generative model, quantitative analysis needs to be conducted from two aspects: accuracy and diversity. In terms of accuracy, the Mean Relative Error (MRE) between the generated geometric or aerodynamic values and their prescribed values is calculated to assess the accuracy in satisfying the design conditions. The calculation formula is given as follows

\begingroup
\begin{align}
\mathrm{MRE}_i &=\frac{1}{N}\sum_{j=1}^{N}
\left|\frac{y_{i,\mathrm{gen}}^{(j)}-c_i^{(j)}}{c_i^{(j)}+\epsilon_{\mathrm{MRE}}}\right|.
\end{align}
\endgroup
For a given design quantity $i$, $j$ indexes the $N$ generated samples. The value $c_i^{(j)}$ is the specified condition, and $y_{i,\mathrm{gen}}^{(j)}$ is the value evaluated on the generated geometry. Geometric quantities are computed directly; aerodynamic coefficients are evaluated using VSPAERO. Here, $\epsilon_{\mathrm{MRE}}=10^{-6}$. The resulting fraction is multiplied by 100 when reported as a percentage.

Regarding diversity, the Euclidean distance between each pair of generated samples is computed based on their geometric parameters $\bm{x}\in\mathbb{R}^n$. For each sample, the distance to its nearest neighbor is identified, and the Mean Nearest Neighbor Distance (MNND) is then computed as the average of these minimum distances. The calculation formula is given as follows

\begin{equation}
\mathrm{MNND} = \frac{1}{N} \sum_{j=1}^{N} \min_{\substack{p=1 \\ p \neq j}}^{N} \left\| \bm{x}^{(j)} - \bm{x}^{(p)} \right\|_2,
\end{equation}
Here, $\bm{x}^{(j)}\in\mathbb R^n$ is the normalized geometric parameter vector of generated sample $j$. A larger MNND indicates that the generated results are widely distributed within the design space, corresponding to enhanced sample diversity. This metric is sufficiently sensitive to the mode collapse phenomenon frequently encountered in GANs; specifically, the MNND decreases significantly when multiple samples cluster together. This metric is highly sensitive to mode collapse---a common challenge in GAN training---as MNND decreases substantially when samples concentrate in localized clusters rather than spreading uniformly.

\begin{figure}[htbp]
\centering
\begin{subfigure}[b]{0.48\textwidth}
    \centering
    \includegraphics[width=\textwidth]{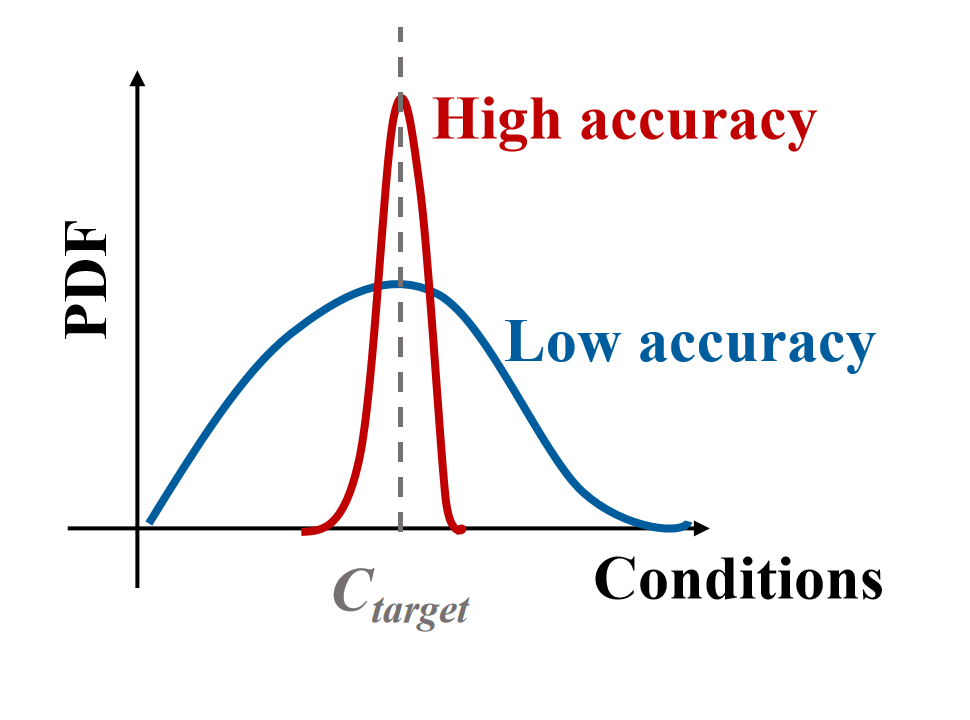}
    \caption{Accuracy}
    \label{fig:准确}
\end{subfigure}
\hfill 
\begin{subfigure}[b]{0.48\textwidth}
    \centering
    \includegraphics[width=\textwidth]{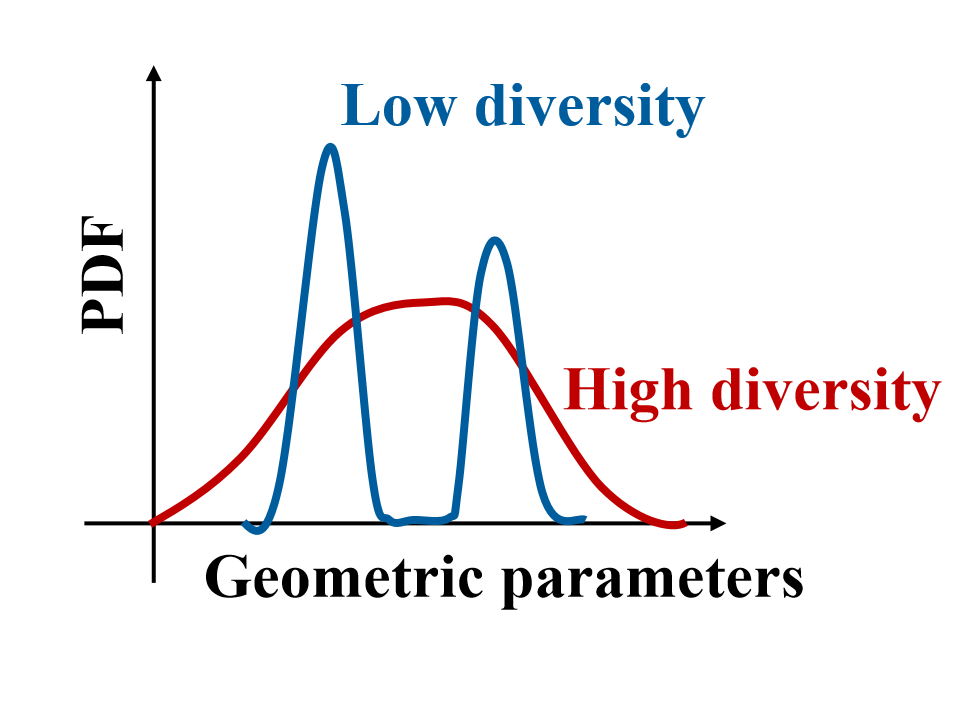}
    \caption{Diversity}
    \label{fig:多样}
\end{subfigure}
\caption{Accuracy and diversity evaluation of generative models via probability density distributions.}
\label{fig:多样与准确}
\end{figure}

To better understand the evaluation process, Figure~\ref{fig:多样与准确} visualizes the Probability Distribution Functions (PDFs) of the generated results in terms of accuracy and diversity. In terms of accuracy (Figure~\ref{fig:准确}), the horizontal axis represents aerodynamic parameters (e.g., $C_L$), and the vertical axis denotes probability density; a sharper and denser distribution indicates greater accuracy in satisfying the aerodynamic conditions. In terms of diversity (Figure~\ref{fig:多样}), the horizontal axis represents the key geometric parameter, and the vertical axis denotes probability density; a broader and flatter distribution indicates that, for the same aerodynamic conditions, the model can generate more diverse feasible design solutions. Together, these two figures jointly serve to assess the relative merits of different models.

\section{Results and Analysis}
\label{sec:3}
To systematically evaluate the effectiveness of the proposed CEGAN framework, this section tests the proposed model to achieve the aerodynamic configuration design of a three-dimensional aircraft. Specifically, Sec~\ref{sec:Dataset Construction} introduces the dataset information, including the parametric representation of the reference aircraft and the validation of the aerodynamic simulation. Sec~\ref{sec:Optimal Network Architecture Selection} describes the selection of the optimal CWGAN-GP architecture as the baseline model. Sec~\ref{sec:Test configurations} defines the model inputs, outputs, and the sets of design conditions under investigation, while Sec~\ref{sec:results} presents the generation results obtained with different enhancement strategies. Collectively, these results highlight the improved performance of the CEGAN framework in terms of diversity, accuracy, and engineering applicability.

\subsection{Dataset Construction}
\label{sec:Dataset Construction}
To test the proposed method on realistic three-dimensional aircraft, this study employs the small general aviation aircraft Cessna Citation X as a representative case \cite{segui2018design}. Previous work by Segui et al.\cite{segui2018design} demonstrated that aerodynamic simulations based on simplified aircraft models constructed using OpenVSP can produce results consistent with experimental case. OpenVSP is a parametric aircraft geometry tool. It allows the user to create a 3D model of an aircraft defined by common engineering parameters~\cite{vsp2022}. This model can be processed into formats suitable for engineering analysis. Consequently, OpenVSP is used in this study to represent the typical aerodynamic evaluation tool in the conceptual design phase. We further parameterize the aerodynamic configuration of the aircraft model established by Segui et al.\cite{segui2018design} and validate the proposed method.

\begin{figure}[htbp]
\centering
\begin{subfigure}[b]{0.48\textwidth}
    \centering
    \includegraphics[width=\textwidth]{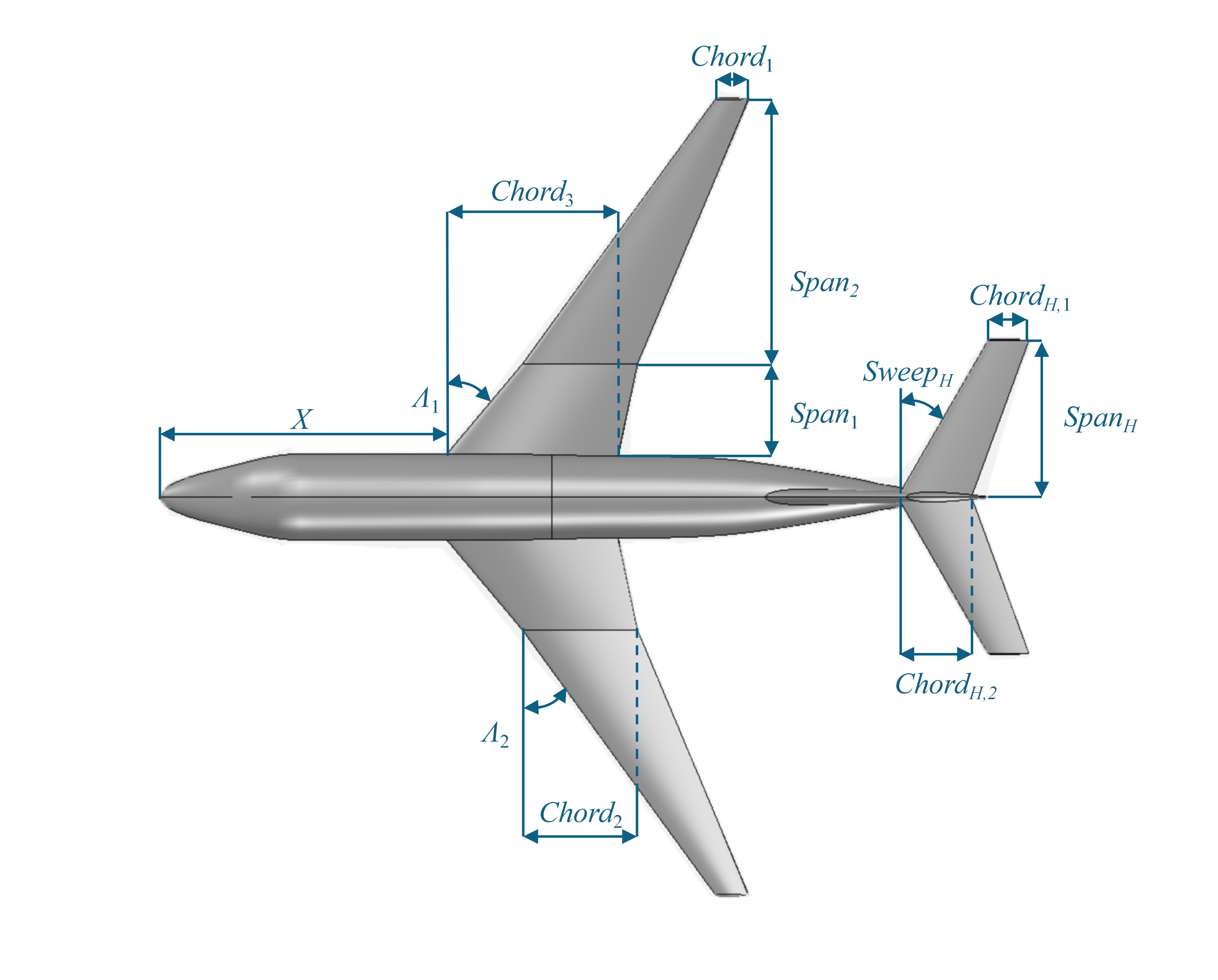}
    \caption{Top view}
    \label{fig:primary view}
\end{subfigure}
\hfill 
\begin{subfigure}[b]{0.48\textwidth}
    \centering
    \includegraphics[width=\textwidth]{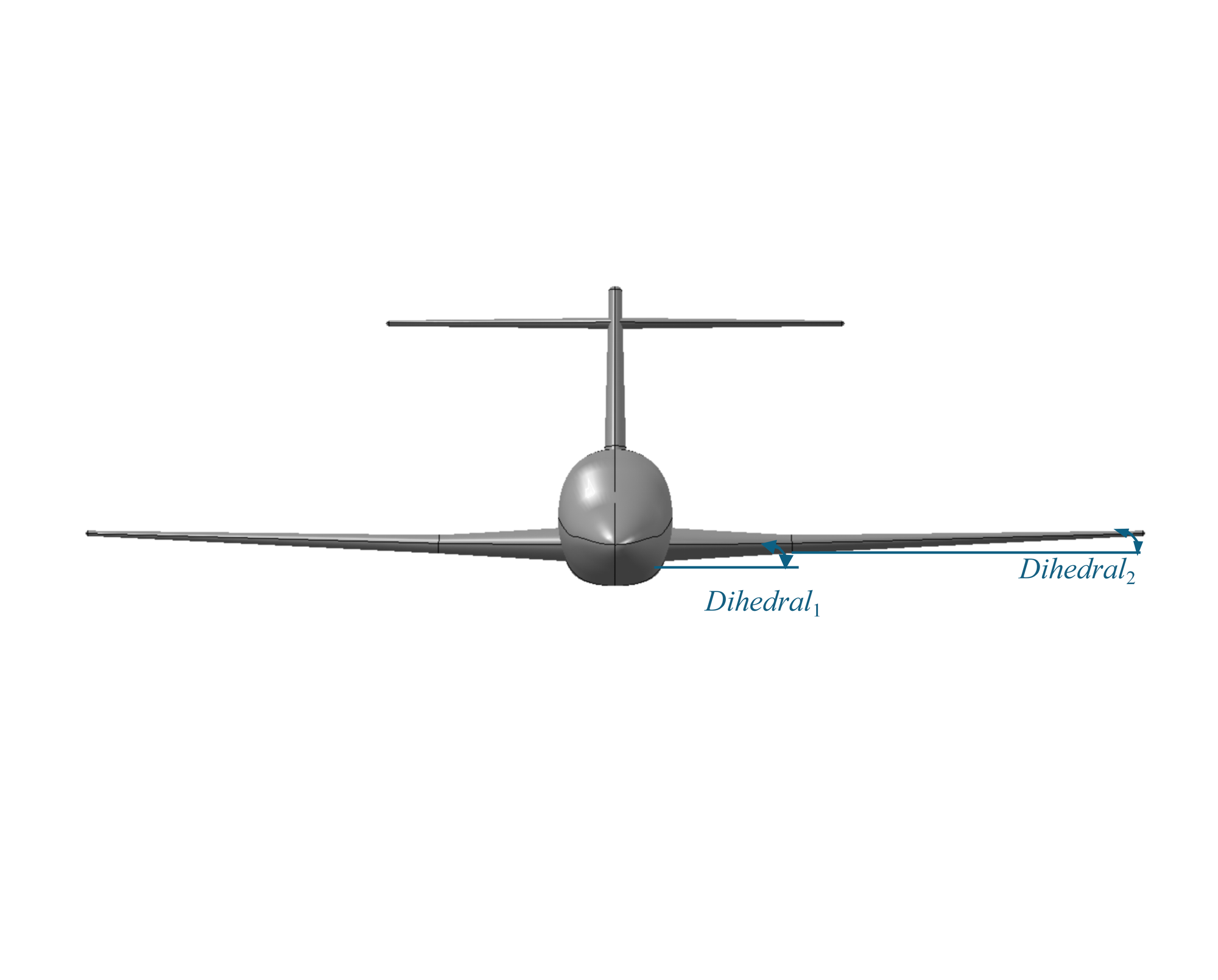}
    \caption{Front view}
    \label{fig:Front view}
\end{subfigure}

\vspace{0.5em} 

\begin{subfigure}[b]{0.48\textwidth}
    \centering
    \includegraphics[width=\textwidth]{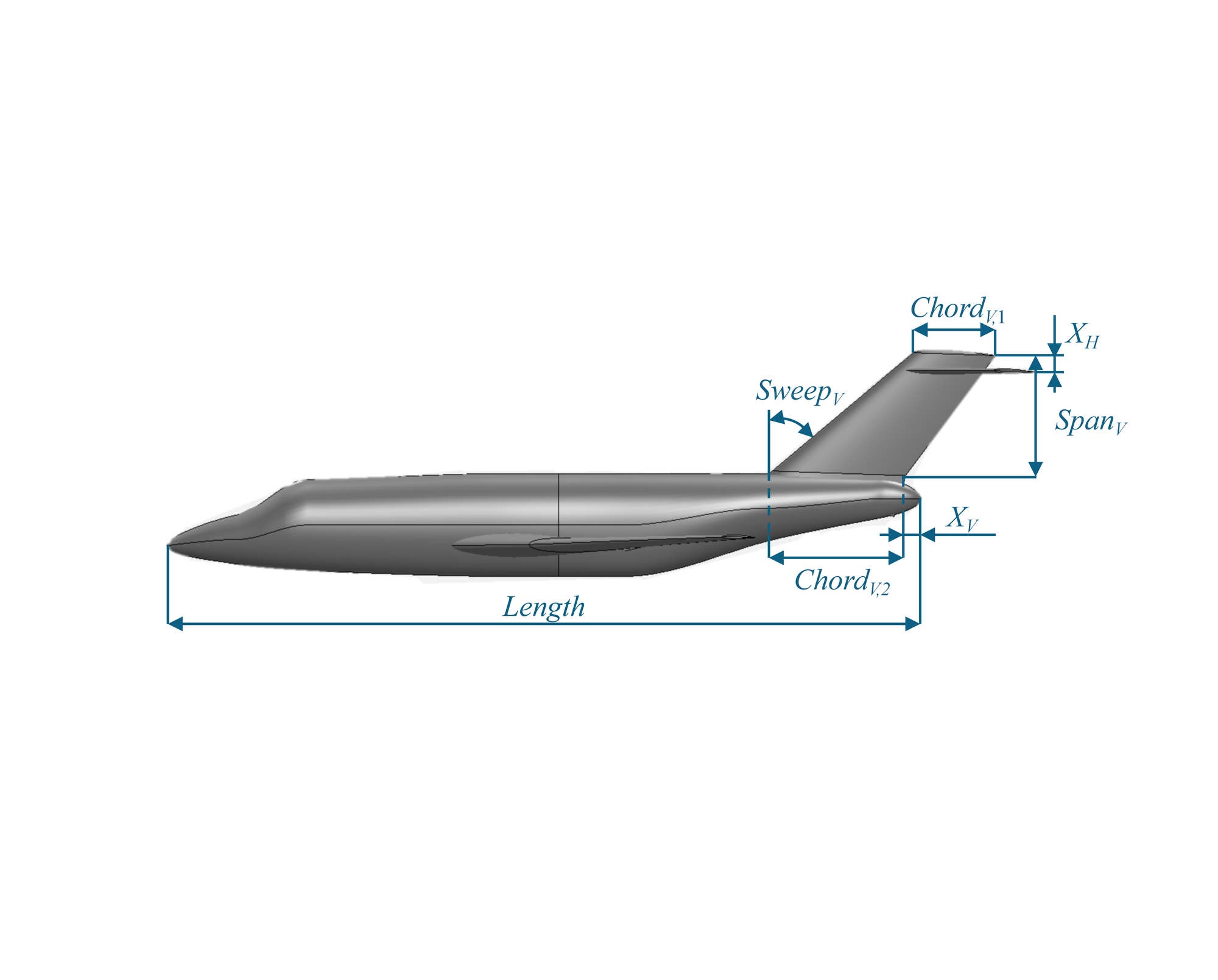}
    \caption{Side view}
    \label{fig:Left view}  
\end{subfigure}
\hfill 
\begin{subfigure}[b]{0.48\textwidth}
    \centering
\end{subfigure}

\caption{Three-view drawing and geometric parameters of the small general aviation aircraft, derived from the Cessna Citation X model in \cite{segui2018design}.}
\label{fig:combined}
\end{figure}

To obtain aerodynamic coefficients, the area and the mean aerodynamic chord of the main wing are adopted as the reference area and reference length, respectively. To mimic the cruise state under typical high-subsonic conditions, the Prandtl--Glauert compressibility correction is employed to enhance accuracy. A refined discretization strategy is applied to each lifting surface to balance computational efficiency and accuracy. Following the work of Rosas-Cordova et al.\cite{TAUFIK2025649}, the main wing is discretized into 48 panels along the span and 32 panels along the chord; the horizontal tail is discretized into 24 panels along the span and 16 panels along the chord; and the vertical tail is discretized into 24 panels along the span and 16 panels along the chord. Mesh refinement is applied to the leading and trailing edges of all lifting surfaces. Specifically, the \texttt{LECluster} and \texttt{TECluster} parameters in OpenVSP are set to 0.25 to capture the aerodynamic load distribution more accurately. Through convergence studies, the number of iterations is set to 40. Figure~\ref{fig:combined} illustrates the three-view drawing and the geometric parameters constructed using the physics-based parameterization method employed in this study. Note that the airfoil parameters at each wing tip are not depicted in the figure.

\begin{figure}[hbt!]
\centering
\includegraphics[width=0.75\textwidth]{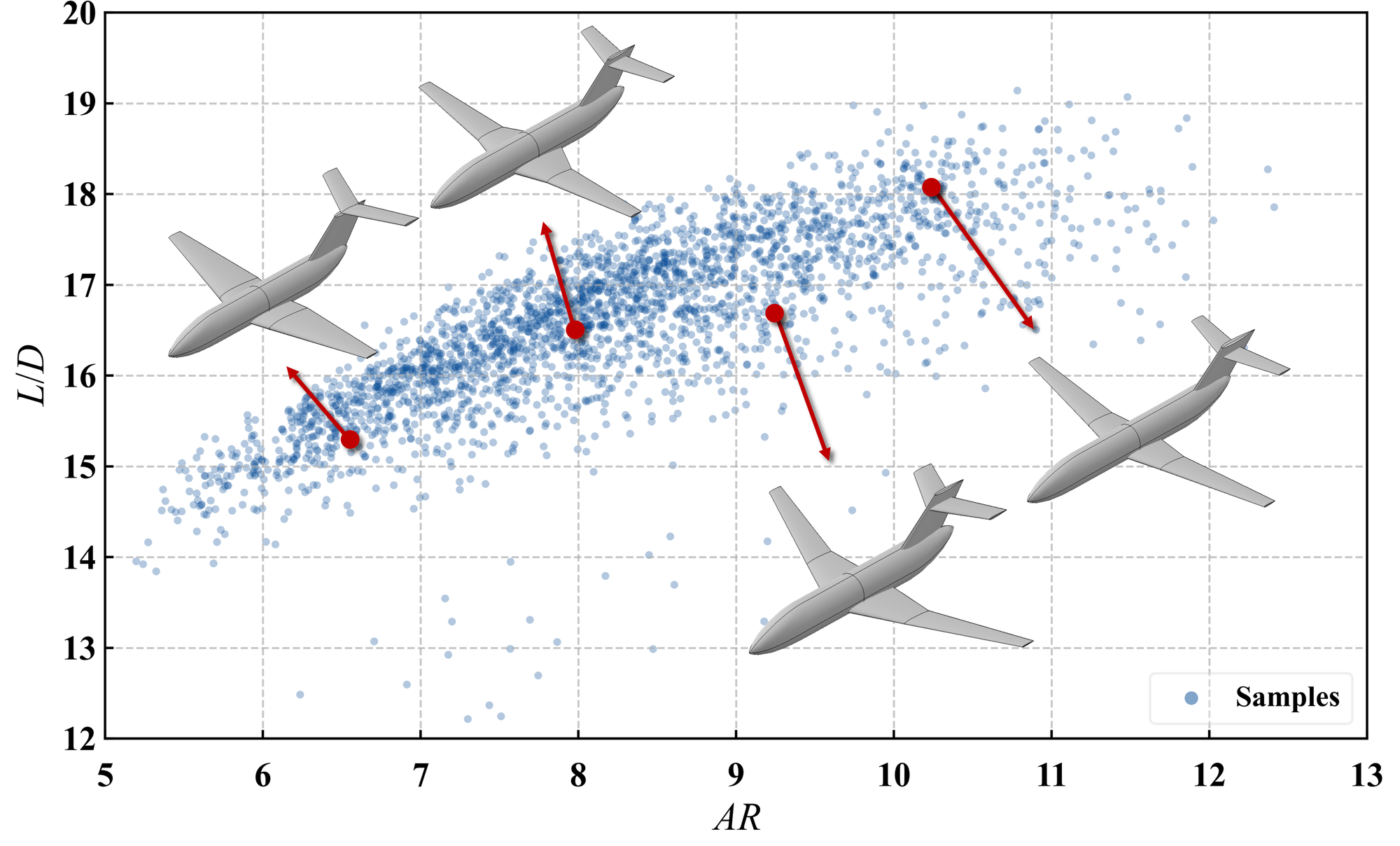}
\caption{Distribution of training samples in the geometric-aerodynamic condition space, composed of the aspect ratio and the lift-to-drag ratio.} 
\label{fig:train}
\end{figure}

Table~\ref{tab:design_variables} summarizes the selected geometric parameters and their ranges for the general aviation aircraft. The aircraft geometry is defined by 28 design variables. Following Segui et al. \cite{segui2018design}, constraints on the upper and lower bounds of wing area, horizontal tail area, and vertical tail area are imposed during the training sample generation. For model training, Latin Hypercube Sampling (LHS) is employed to obtain 2,500 distinct sets of geometric parameters. For each parameter set, the aerodynamic coefficients ($C_L$, $C_D$, and $C_M$) are computed under the flow conditions of Mach number $Ma=0.6$, Reynolds number $Re=8 \times 10^{6}$, and angle of attack $\alpha=2^\circ$. Figure~\ref{fig:train} illustrates the sample distribution and variations among different samples within the aerodynamic design space $AR$-$L/D$. The resulting geometric dataset is subsequently used to train the various generative models.

\begin{table}[htbp]
\centering
\caption{Design variables, bounds, and reference values.}
\label{tab:design_variables}
\small
\setlength{\tabcolsep}{4pt}
\begin{adjustbox}{max width=\linewidth}
\begin{tabular}{l p{0.42\linewidth} ccc}
\hline\hline
Variable & Description & \makecell{Lower \\ Bound} & \makecell{Upper \\ Bound} & \makecell{Reference\\Value\cite{segui2018design}} \\
\midrule
$Span_1$   & Wing segment 1 span (half) / m       & 2   & 4   &  \multirow{2}{*}{19.38} \\
$Span_2$   & Wing segment 2 span (half) / m       & 6   & 8   &  \\
$Chord_3$  & Wing root chord / m                  & 3.5 & 5.5 & \multirow{3}{*}{\makecell{Aspect ratio \\ = 7.8}} \\
$Chord_1$  & Wing tip chord / m                   & 0.8 & 1.5 &  \\
$Chord_2$  & Chord at wing kink / m               & ---   & ---   &  \\
Area      & Wing area / m$^2$                    & 47  & 50  & 48.96 \\
$\Lambda_1$  & Sweep angle of wing segment 1        & $0^\circ$ & $40^\circ$ & --- \\
$\Lambda_2$  & Sweep angle of wing segment 2        & $0^\circ$ & $40^\circ$ & --- \\
$Dihedral_1$ & Dihedral angle of wing segment 1   & $0^\circ$ & $2^\circ$  & \multirow{2}{*}{$2^\circ$} \\
$Dihedral_2$ & Dihedral angle of wing segment 2   & $0^\circ$ & $3^\circ$  &  \\
$t/c_1$    & Max thickness-to-chord ratio at tip  & 0.11 & 0.24 & --- \\
$t/c_2$    & Max thickness-to-chord ratio at kink & 0.11 & 0.24 & --- \\
$t/c_3$    & Max thickness-to-chord ratio at root & 0.11 & 0.24 & --- \\
$X$        & Fore edge location relative to fuselage & 25\% & 45\% & --- \\
\midrule
$Span_V$   & Vertical tail span / m               & 2.5   & 4 & --- \\
$Chord_{V,2}$ & Vertical tail root chord / m       & 3.5   & 4.5 & --- \\
$Chord_{V,1}$ & Vertical tail tip chord / m        & 2 & 4.5  & --- \\
$Area_V$   & Vertical tail area / m$^2$           & 9   & 11  & 10.31 \\
$t/c_{V,2}$ & Max t/c at vertical tail root        & 0.11 & 0.24 & --- \\
$t/c_{V,1}$ & Max t/c at vertical tail tip         & 0.11 & 0.24 & --- \\
$\Lambda_V$  & Vertical tail sweep angle            & $40^\circ$ & $55^\circ$ & --- \\
$X_V$      & Distance from vertical tail TE to fuselage aft end / m & 0.1 & 1.5 & --- \\
\midrule
$Span_H$  & Horizontal tail span (half) / m      & 3 & 4.5 & --- \\
$Chord_{H,2}$ & Horizontal tail root chord / m     & 1.5   & 3   & --- \\
$Chord_{H,1}$ & Horizontal tail tip chord / m      & 1   & 2 & --- \\
$Area_H$   & Horizontal tail area / m$^2$         & 9   & 12  & 11.15 \\
$t/c_{H,2}$ & Max t/c at horizontal tail root      & 0.11 & 0.24 & --- \\
$t/c_{H,1}$ & Max t/c at horizontal tail tip       & 0.11 & 0.24 & --- \\
$\Lambda_H$  & Horizontal tail sweep angle          & $20^\circ$ & $35^\circ$ & --- \\
$X_{H}$      & Horizontal tail position relative to vertical tail tip & 10\% & 20\% & --- \\
$Length$    & Fuselage length / m                  & 17  & 20  & --- \\
\hline\hline
\end{tabular}
\end{adjustbox}
\end{table}

\begin{figure}[htbp]
\centering
\begin{subfigure}[b]{0.48\textwidth}
    \centering
    \includegraphics[width=\textwidth]{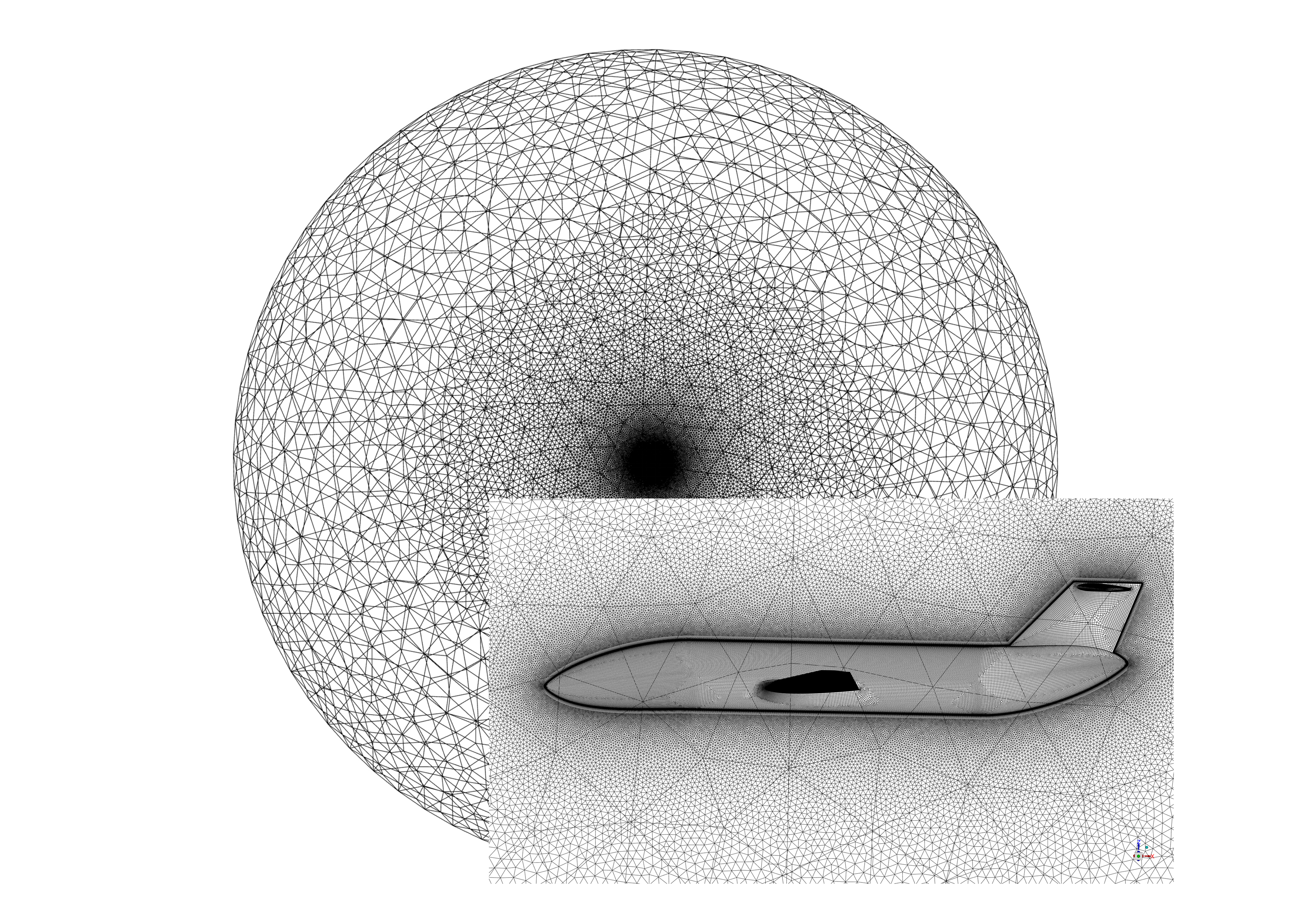}
    \caption{Volume mesh}
    \label{fig:volume mesh}
\end{subfigure}
\hfill 
\begin{subfigure}[b]{0.48\textwidth}
    \centering
    \includegraphics[width=\textwidth]{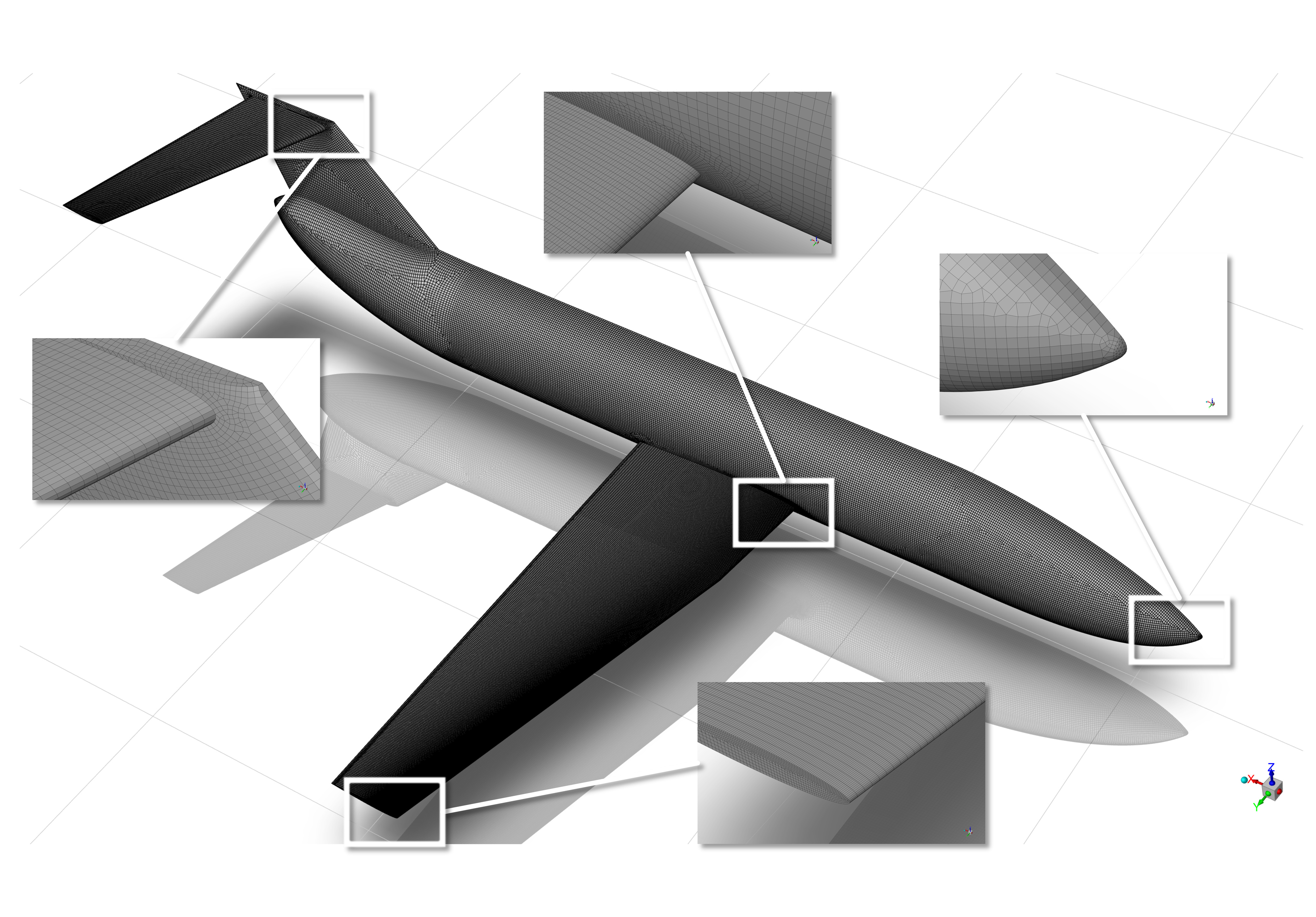}
    \caption{Surface mesh}
    \label{fig:surface mesh}
\end{subfigure}
\caption{Computational grid of the aircraft model for validation.}
\label{fig:mesh}
\end{figure}

\begin{figure}[htbp]
\centering
\begin{subfigure}[b]{0.48\textwidth}
    \centering
    \includegraphics[width=\textwidth]{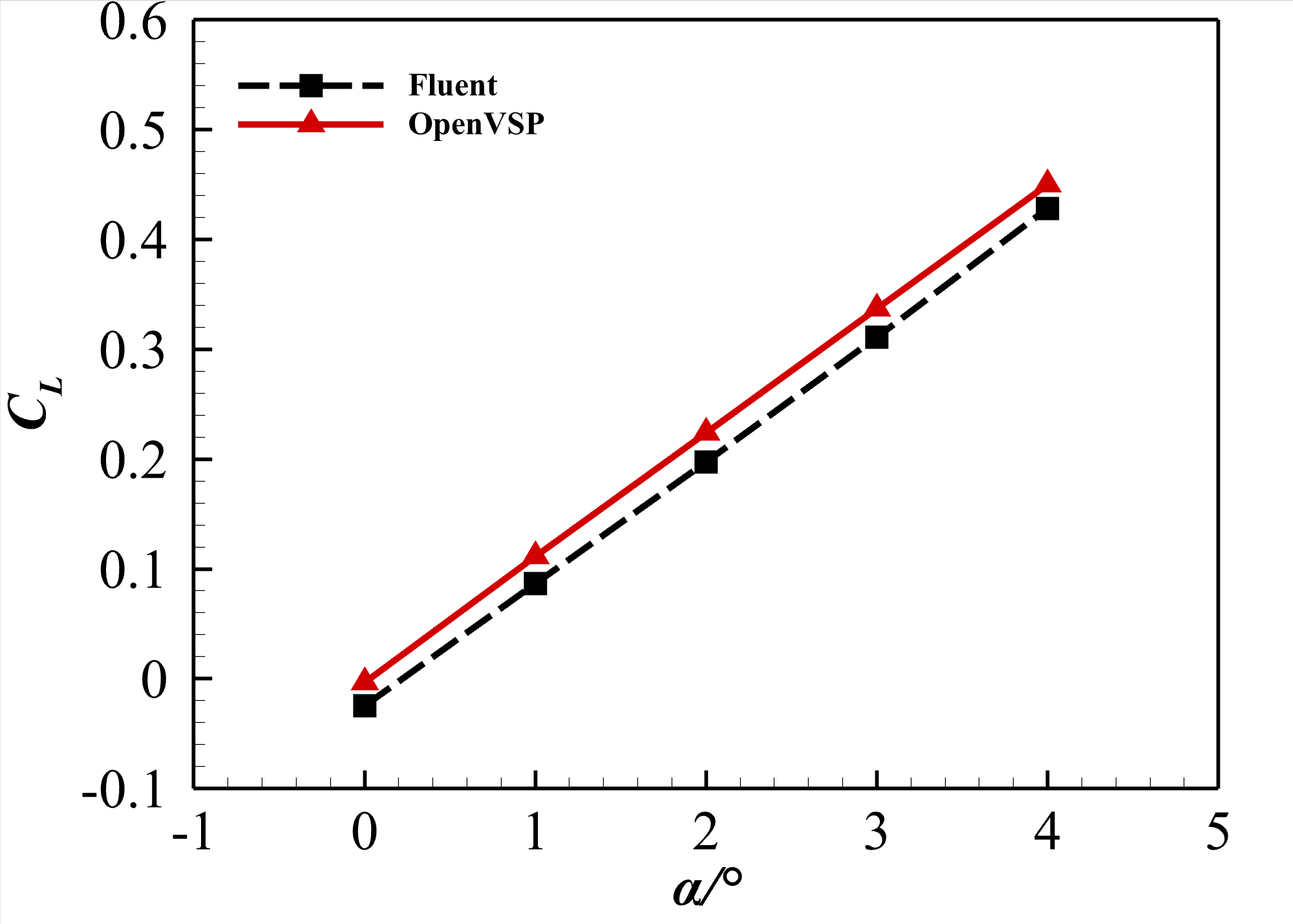}
    \caption{$C_L$}
    \label{fig:CLjingdu}
\end{subfigure}
\hfill 
\begin{subfigure}[b]{0.48\textwidth}
    \centering
    \includegraphics[width=\textwidth]{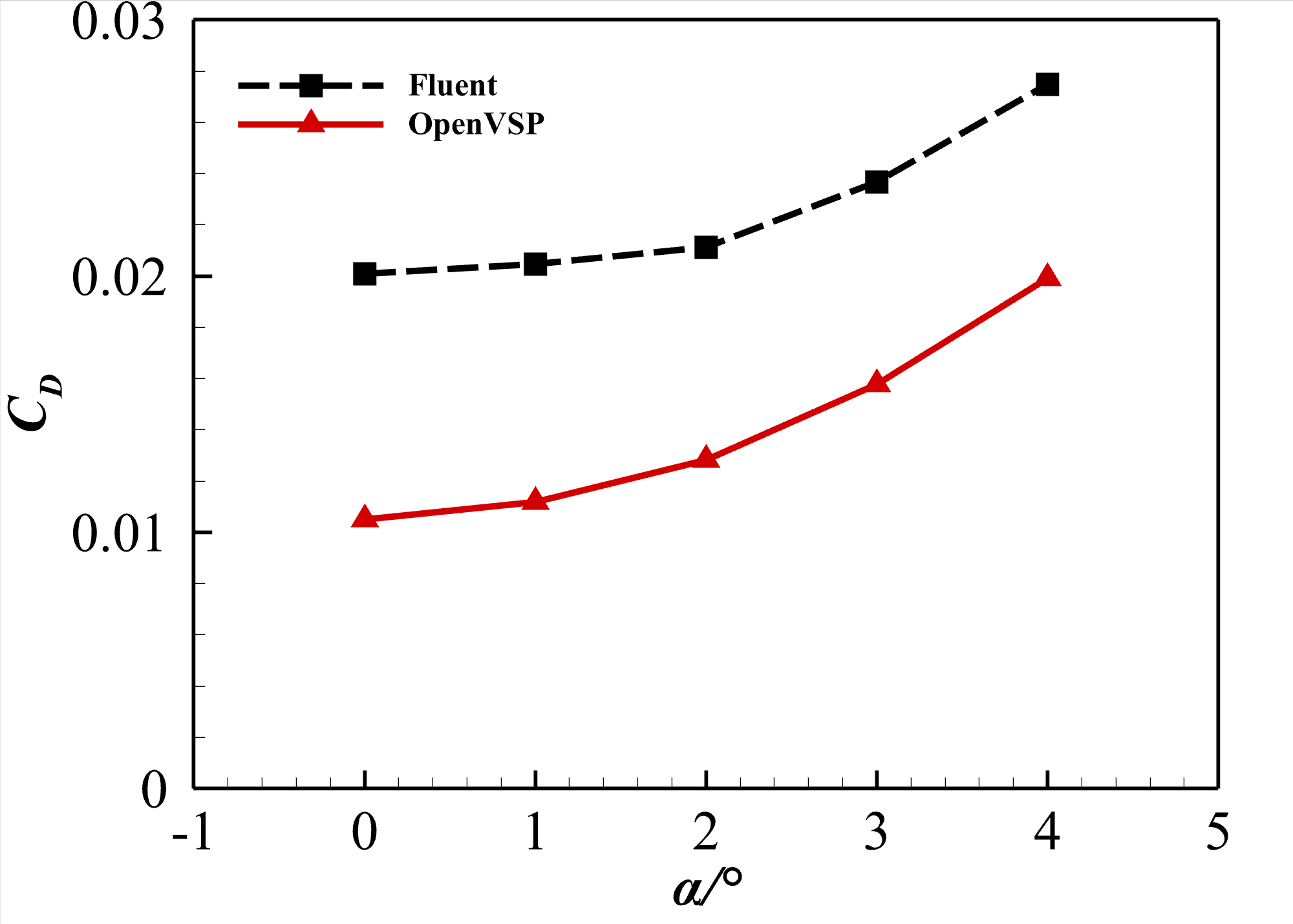}
    \caption{$C_D$}
    \label{fig:CDjingdu}
\end{subfigure}
\caption{Comparison of numerical simulation results between CFD and OpenVSP.}
\label{fig:Fluent and VSP}
\end{figure}

To validate the accuracy of aerodynamic analysis, lift and drag coefficients for a specific configuration selected from the dataset were computed using both OpenVSP and Ansys Fluent at the angles of attack ranging from 0\textdegree{} to 4\textdegree{}. For the Ansys Fluent simulations, the computational mesh is composed of approximately 287,500 surface elements and 11.57 million volume cells as illustrated in Figure~\ref{fig:mesh}. The simulation is based on the Reynolds Average Navier-Stokes (RANS) equations, where the compressible Spalart-Allmaras (SA) turbulence model is employed. The results are compared in Figure~\ref{fig:Fluent and VSP}, indicating that OpenVSP predictions for the lift coefficient agree well with CFD results, while maintaining a mean relative error within 10\%. Although the drag coefficient exhibits larger discrepancies, both methods capture consistent variation trends. The deviation primarily stems from the inherent accuracy limitations of the Vortex Lattice Method (VLM) employed by OpenVSP. Nevertheless, for the purpose of conceptual design, this low-fidelity computational approach remains capable of analyzing aerodynamic characteristics across different configurations.

\subsection{Optimal Network Architecture Selection}
\label{sec:Optimal Network Architecture Selection}
To establish a baseline for evaluating the proposed CEGAN model, optimal network structure should be determined before further investigation. Note that the baseline CWGAN-GP, different variations of this model, and the CEGAN model adopt the same model structure for a fair comparison. For the generative model, the noise dimension of $\bm{z}$ is fixed at 100, guided by the parameters in Li et al. \cite{li2020efficient}. Subsequently, drawing upon related studies \cite{Yu2026Aerodynamic,WANG202262, Chenyan2026, Lin2026}, the discriminator architecture is fixed with hidden layers comprising 128, 64, and 32 neurons. Based on this configuration, a systematic evaluation was conducted on training datasets including 2,500 and 20,000 samples to assess the impact of network depth, width, and activation functions (for both generator and discriminator) on the generation accuracy, thereby determining the optimal network architecture. The generative model under consideration is conditioned on two aerodynamic coefficients ($C_L$, $C_D$). Model performance is primarily evaluated using the MRE between the generated and prescribed values of these two coefficients. Detailed comparison of model performance is discussed in Appendix B.

\begin{figure}[hbt!]
\centering
\includegraphics[width=1\textwidth]{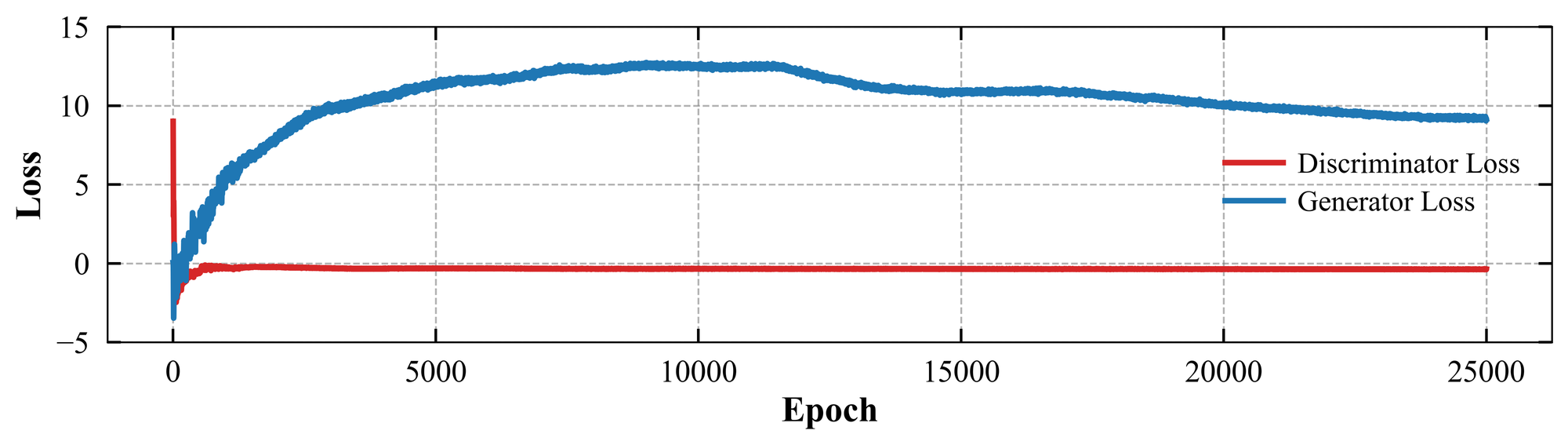}
\caption{An example of convergence process of the loss function (CWGAN-GP).}
\label{fig:loss}
\end{figure}

\begin{figure}[htbp]
\centering
\begin{subfigure}{0.48\textwidth}
    \centering
    \includegraphics[width=\linewidth]{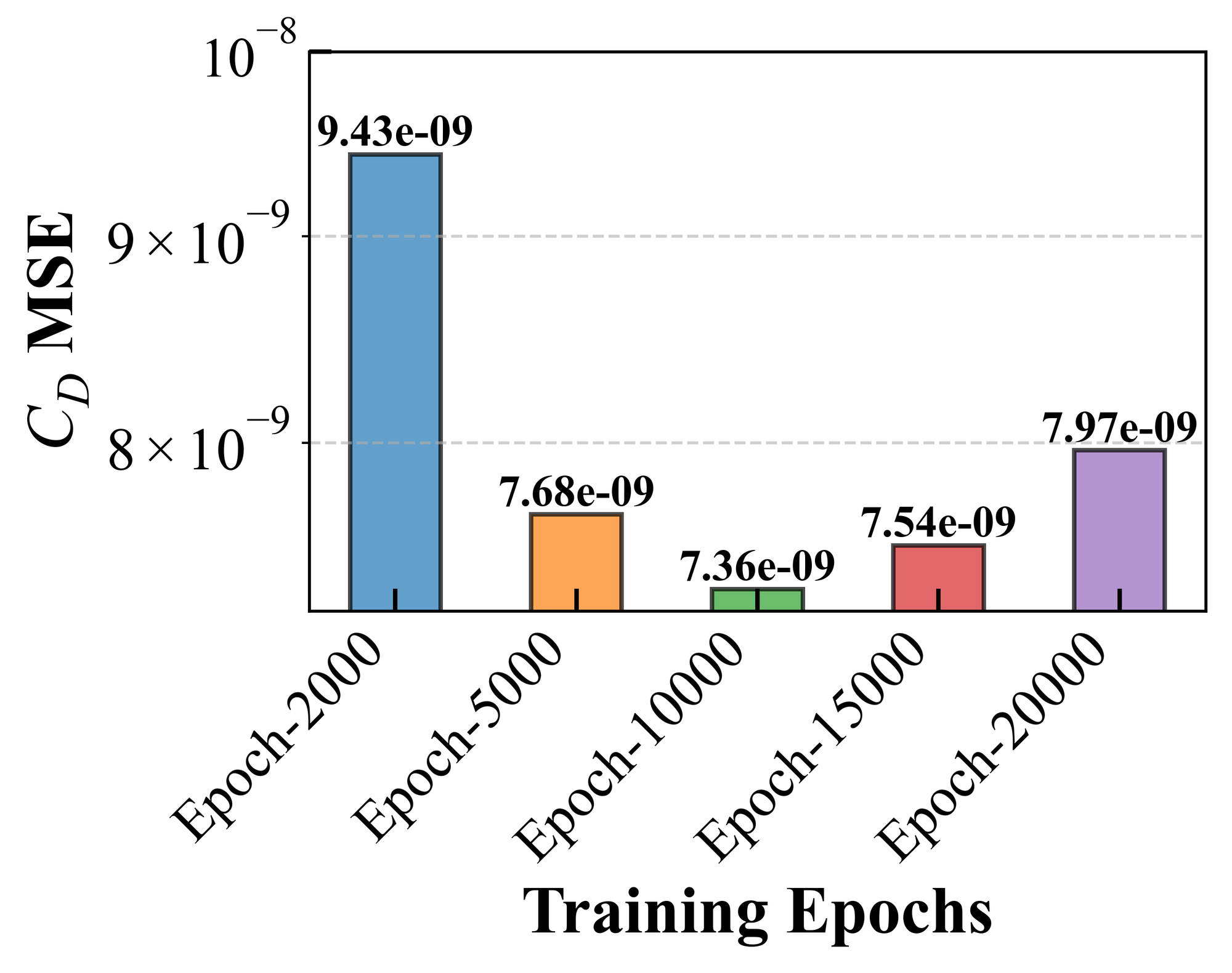}
    \caption{$C_D$}
    \label{fig:loss-cd}
\end{subfigure}
\hfill
\begin{subfigure}{0.48\textwidth}
    \centering
    \includegraphics[width=\linewidth]{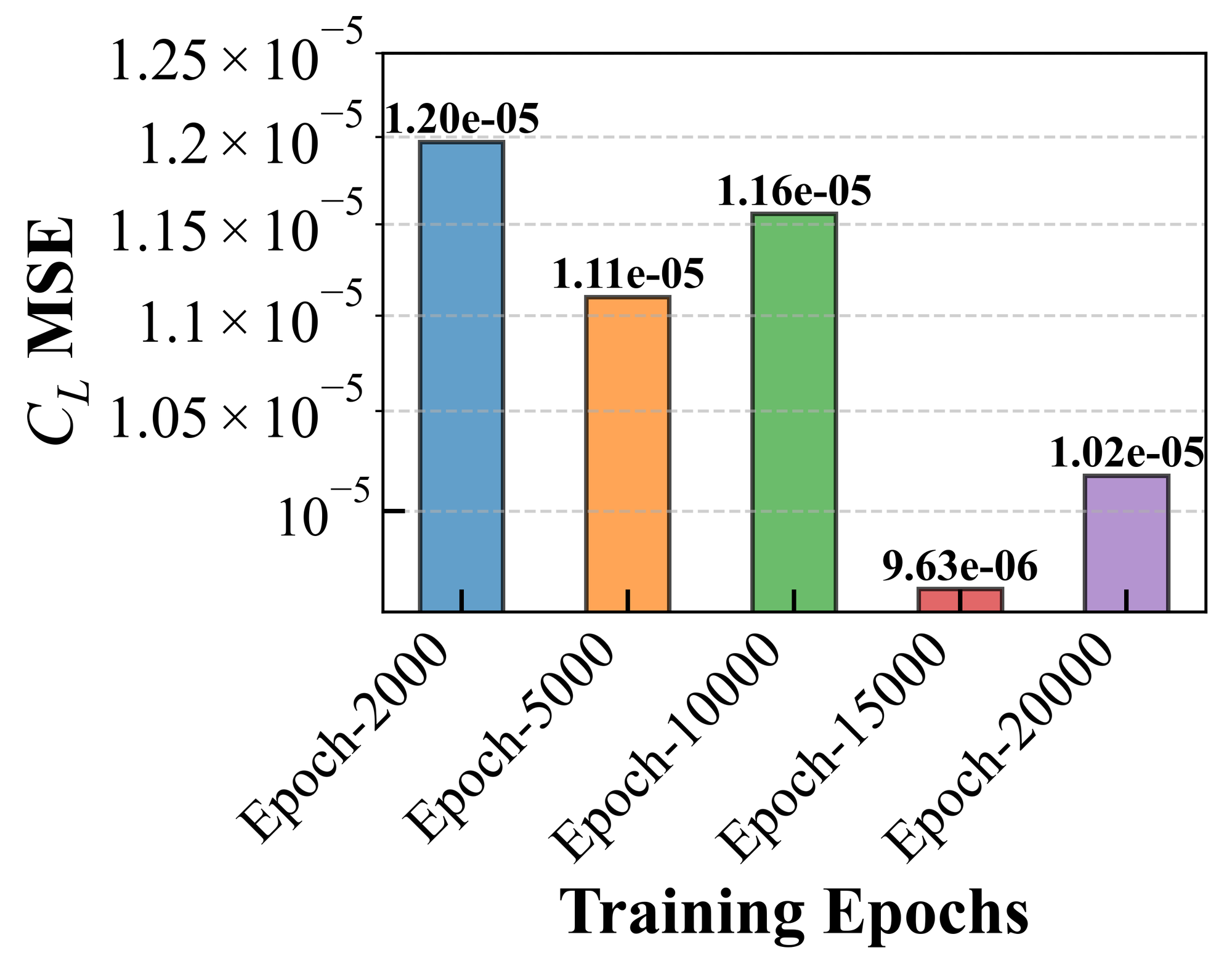}
    \caption{$C_L$}
    \label{fig:loss-cl}
\end{subfigure}
\hfill

\vspace{1em} 

\begin{subfigure}{0.48\textwidth}
    \centering
    \includegraphics[width=\linewidth]{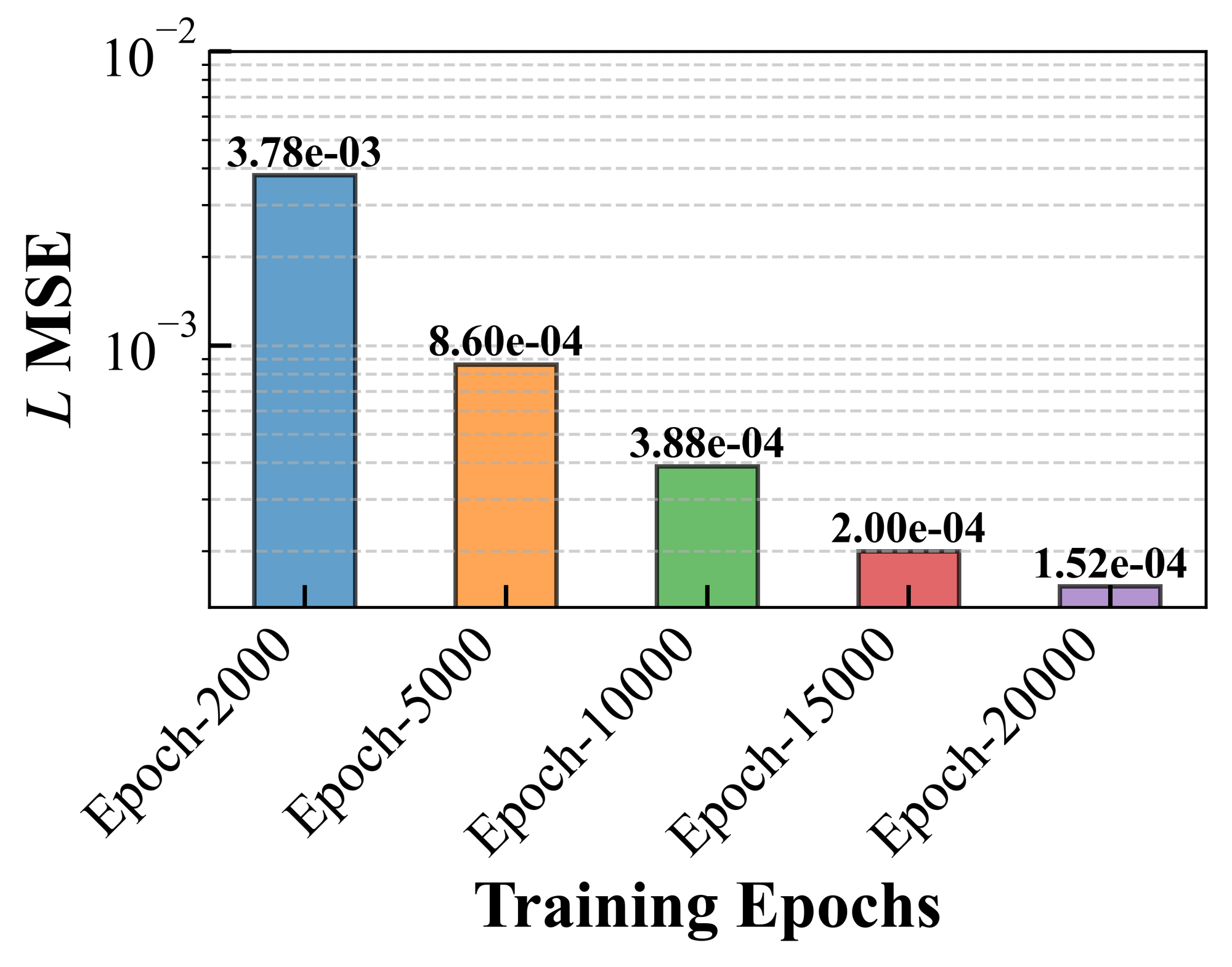}
    \caption{$L$}
    \label{fig:loss-l}
\end{subfigure}
\begin{subfigure}{0.48\textwidth}
    \centering
    \includegraphics[width=\linewidth]{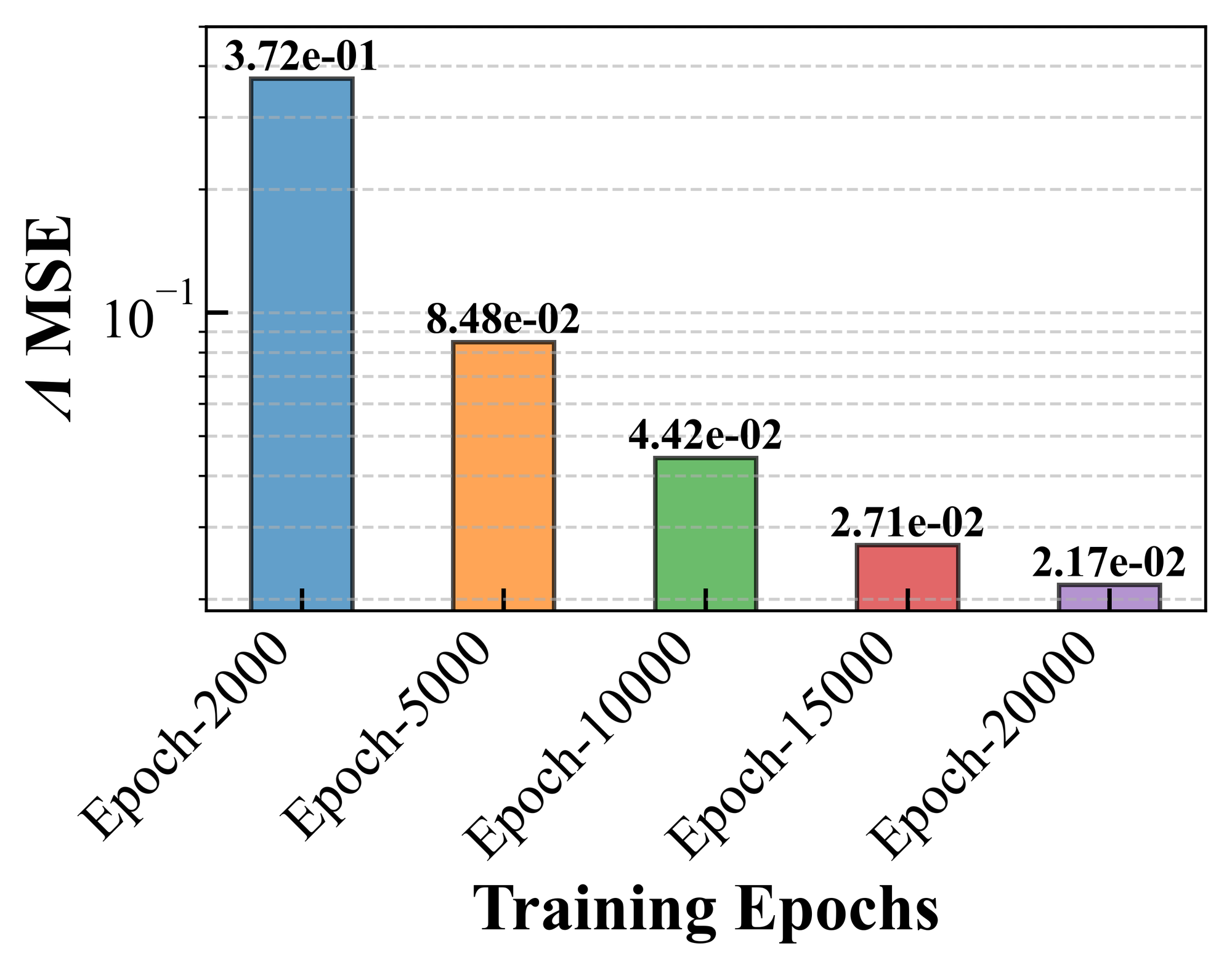}
    \caption{$\Lambda$}
    \label{fig:loss-sweep}
\end{subfigure}
\caption{
    Error of generator across different iteration steps.
}
\label{fig:iteration_accuracy}
\end{figure}

Experimental results indicate that the selection of activation functions and the generator network architecture significantly influence generation accuracy. Across different sample sizes, the combination of Tanh function for the Generator and LeakyReLU function for the Discriminator consistently demonstrated optimal performance. Combined with specific network width and depth configurations (Generator: 256-128-64, Discriminator: 128-64-32; see Table~\ref{tab:Network_design}), this architecture achieved lower errors in satisfying the aerodynamic conditions compared to other architectures when trained on 2,500 samples. Furthermore, Figure~\ref{fig:loss} shows a typical convergence plot of loss functions in the CWGAN-GP model. Different from traditional neural networks, the loss function of GANs include both the discriminator loss and the generator loss. In particular, the discriminator loss usually converges fast while the generator loss keeps changing as the iteration continues. Figure~\ref{fig:iteration_accuracy} further shows the error of the generator across different iteration steps. It indicates that increasing the number of iteration steps significantly enhances the accuracy of conditional generation, while it may also reduce the accuracy for some conditions. Therefore, to determine the optimal baseline model, this study evaluated the accuracy of the generator across different iteration steps and ultimately selected 20,000 steps as the optimal number of training iteration. Finally, all subsequent models utilize the optimized model structure and training approach detailed in this section.

\subsection{Test configurations}
\label{sec:Test configurations}
\begin{figure}[hbt!]
\centering
\includegraphics[width=0.5\textwidth]{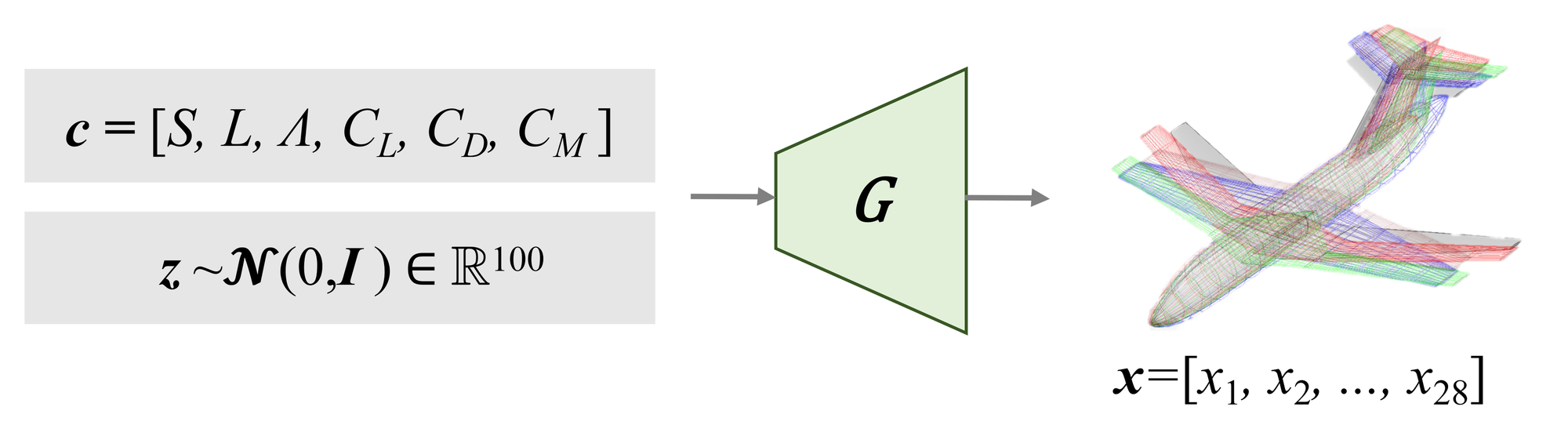}
\caption{Input and Output of the Generator.} 
\label{fig:model detail}
\end{figure}

To generate configurations satisfying the aerodynamic conditions, the lift coefficient $C_L$, drag coefficient $C_D$, and pitching moment coefficient $C_M$ (with the nose as the moment reference point) are selected as the aerodynamic condition variables. These parameters determine cruise efficiency, takeoff and landing performance, and longitudinal static stability, respectively. Meanwhile, the main wing area $S$, span $L$, and mean sweep angle $\Lambda$ (${\Lambda}$ = ($\Lambda_{1}$ + $\Lambda_{2}$) / 2) are selected as additional geometric condition variables, which are associated with structural weight, airport compatibility, and transonic aerodynamic characteristics, respectively. Consequently, the condition vector and generated geometric parameter vector in Eq.~\ref{eq:gen} are defined as $\bm{c}=[S, L, \Lambda, C_L, C_D, C_M]$ and $\bm{x}=[x_1, \dots, x_{28}]$, as illustrated in Figure~\ref{fig:model detail}.

\begin{figure}[htbp]
\centering
\begin{subfigure}{0.32\textwidth}
    \centering
    \includegraphics[width=\linewidth, keepaspectratio]{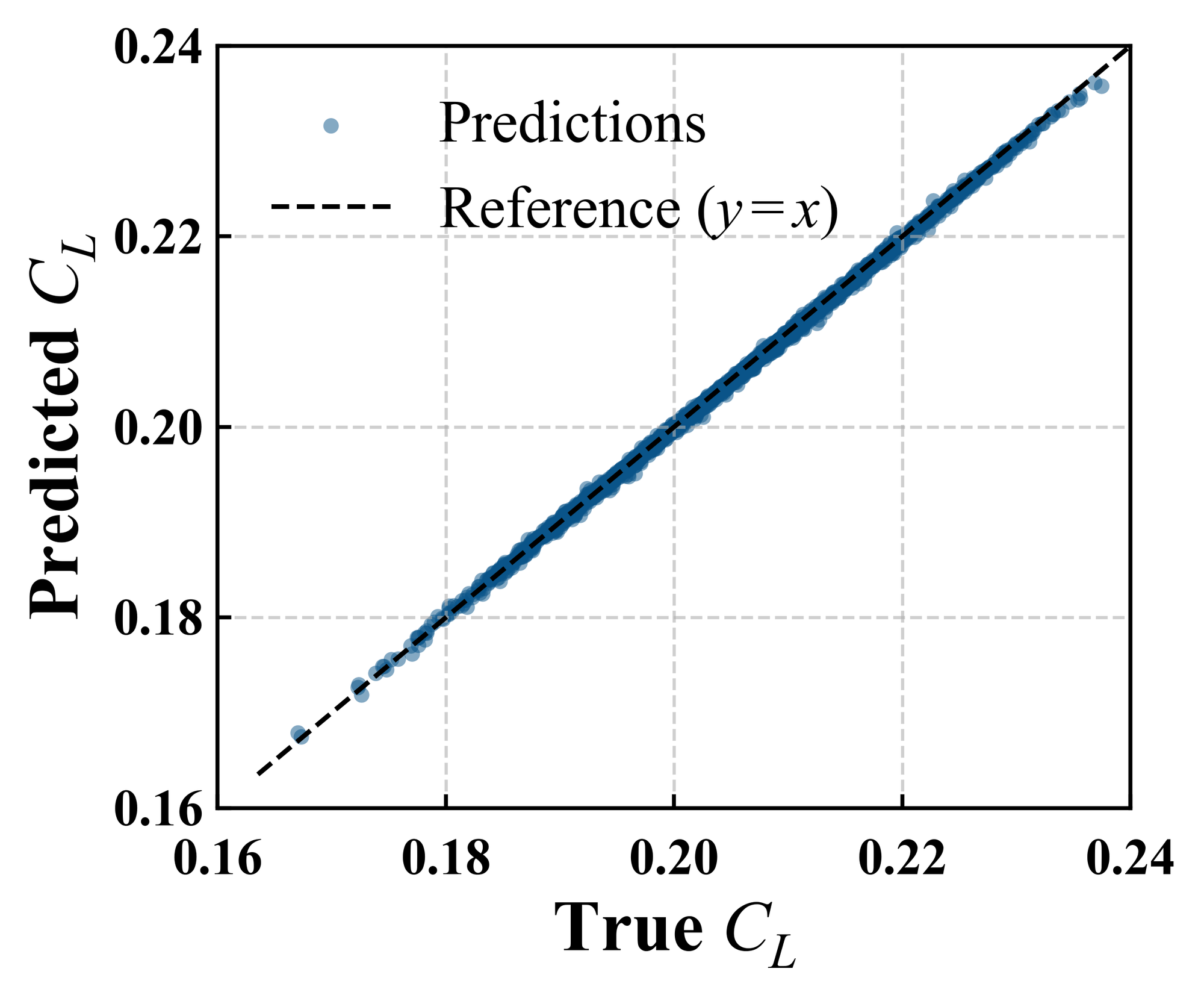}
    \caption{$C_L$}
    \label{fig:truecl}
\end{subfigure}
\hfill
\begin{subfigure}{0.32\textwidth}
    \centering
    \includegraphics[width=\linewidth, keepaspectratio]{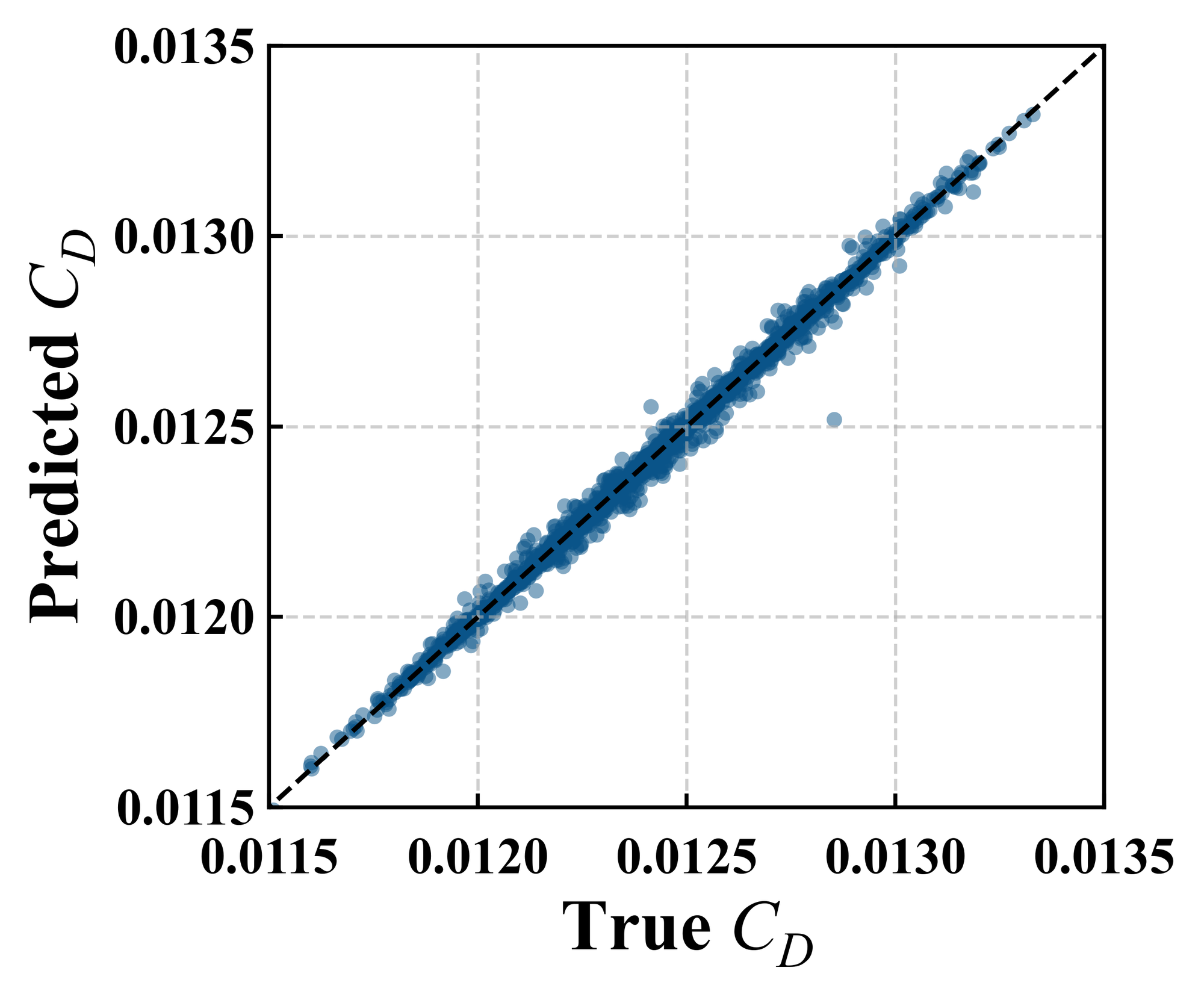}
    \caption{$C_D$}
    \label{fig:truecd}
\end{subfigure}
\hfill
\begin{subfigure}{0.32\textwidth}
    \centering
    \includegraphics[width=\linewidth, keepaspectratio]{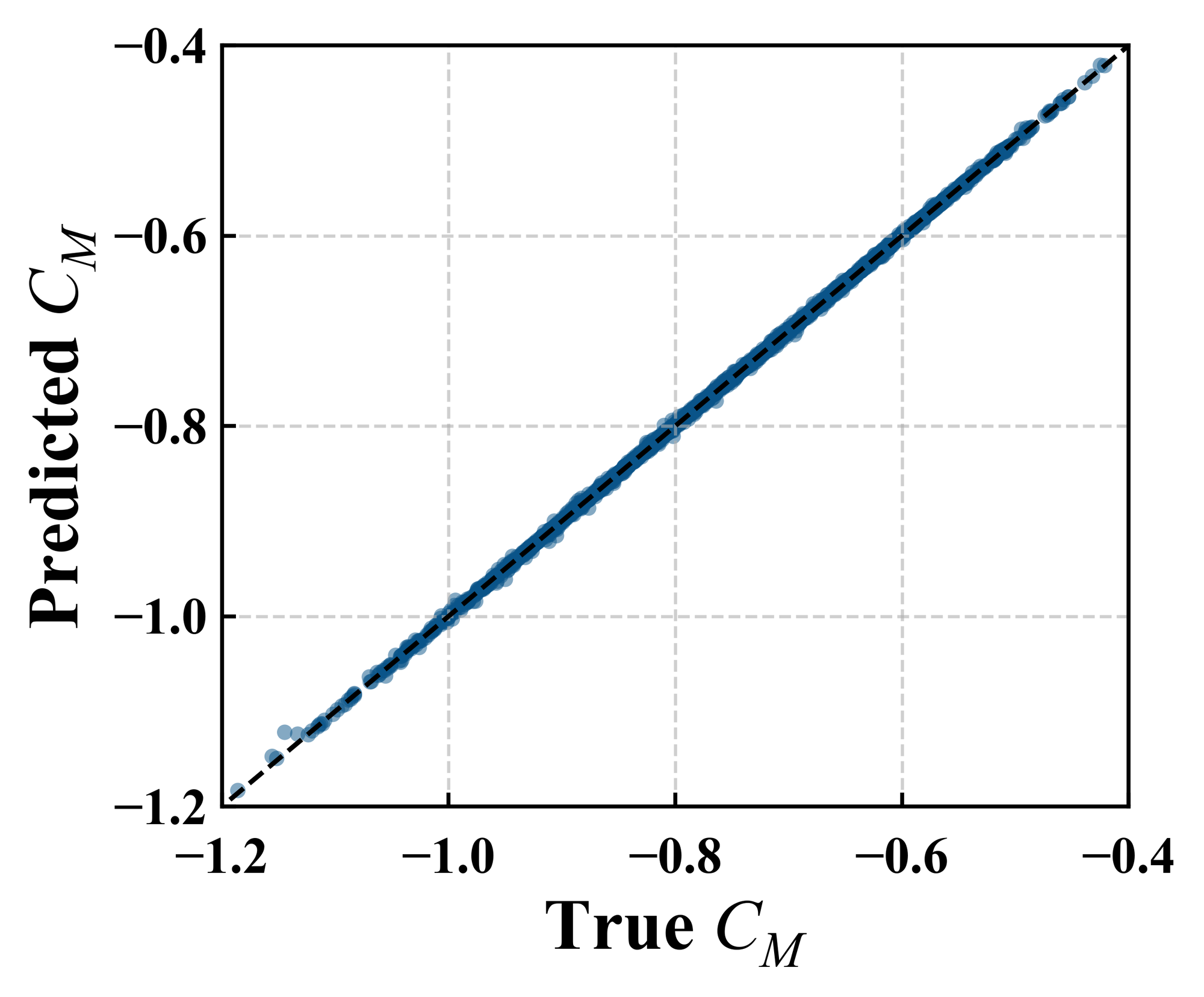}
    \caption{$C_M$}
    \label{fig:truecm}
\end{subfigure}
\caption{
    Scatter plots of predicted versus true aerodynamic coefficients. The dashed line indicates perfect prediction ($y = x$). Points closer to the line denote higher accuracy of the surrogate models.
}
\label{fig:surrogate_accuracy}
\end{figure}

The surrogate models are trained using the same dataset used for the generative models, with all architectures constructed based on MLP. The final network structures are determined following systematic hyperparameter tuning, where the parametric studies are omitted here for brevity. Table~\ref{tab:Surrogate_model} summarizes the final architectures of the surrogate models, and Figure~\ref{fig:surrogate_accuracy} presents the prediction accuracy for various aerodynamic parameters, which confirms that the surrogate models achieve acceptable accuracy. The main comparisons use the same set of surrogate models for all condition-enhanced generators. The separate sensitivity study in \ref{app:surrogate_validation} deliberately varies surrogate accuracy.

\begin{table}[htbp]
\centering
\caption{Final architecture and hyperparameters of the aerodynamic surrogate models.}
\label{tab:Surrogate_model}
\small
\setlength{\tabcolsep}{4pt}
\begin{adjustbox}{max width=\linewidth}
\begin{tabular}{cccccc}
\toprule
Aerodynamic & Hidden Layers & Activation & Learning & Batch & Test \\
Coefficient & (neurons)     & Function   & Rate     & Size  & MSE \\
\midrule
$C_L$ & 512-128 & \multirow{3}{*}{Swish} & $5 \times 10^{-4}$ & 16 & $8.29 \times 10^{-7}$ \\
$C_D$ & 512-128 &   & $1 \times 10^{-3}$ & 64 & $3.97 \times 10^{-10}$ \\
$C_M$ & 256-16  &   & $5 \times 10^{-4}$ & 32 & $8.46 \times 10^{-6}$ \\
\bottomrule
\end{tabular}
\end{adjustbox}
\end{table}

\begin{figure}[htbp]
\centering
\begin{subfigure}[b]{0.48\textwidth}
    \centering
    \includegraphics[width=\textwidth]{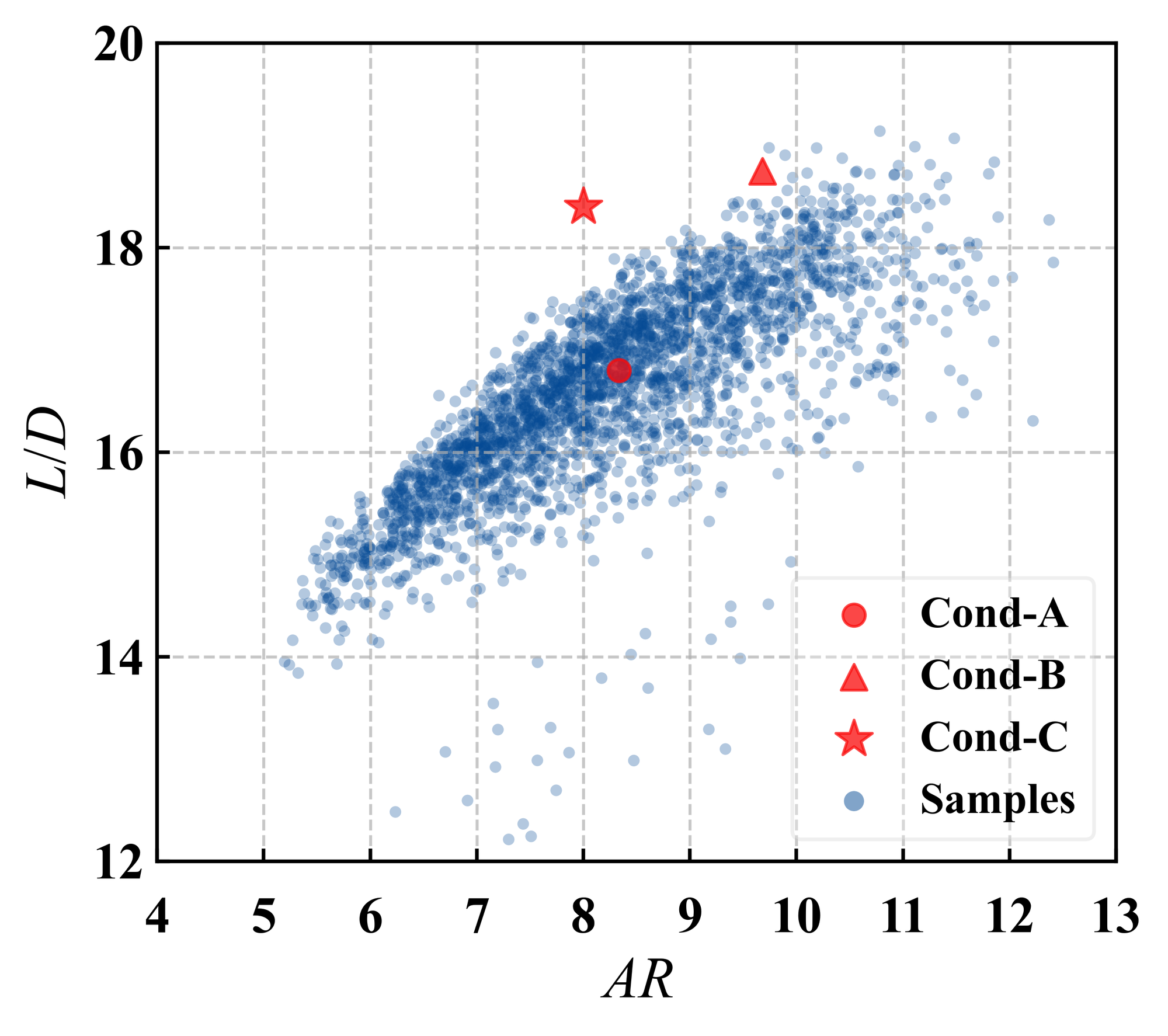}
    \caption{Cond-A, Cond-B and Cond-C}
    \label{fig:目标与样本散点}
\end{subfigure}
\hfill
\begin{subfigure}[b]{0.48\textwidth}
    \centering
    \includegraphics[width=\textwidth]{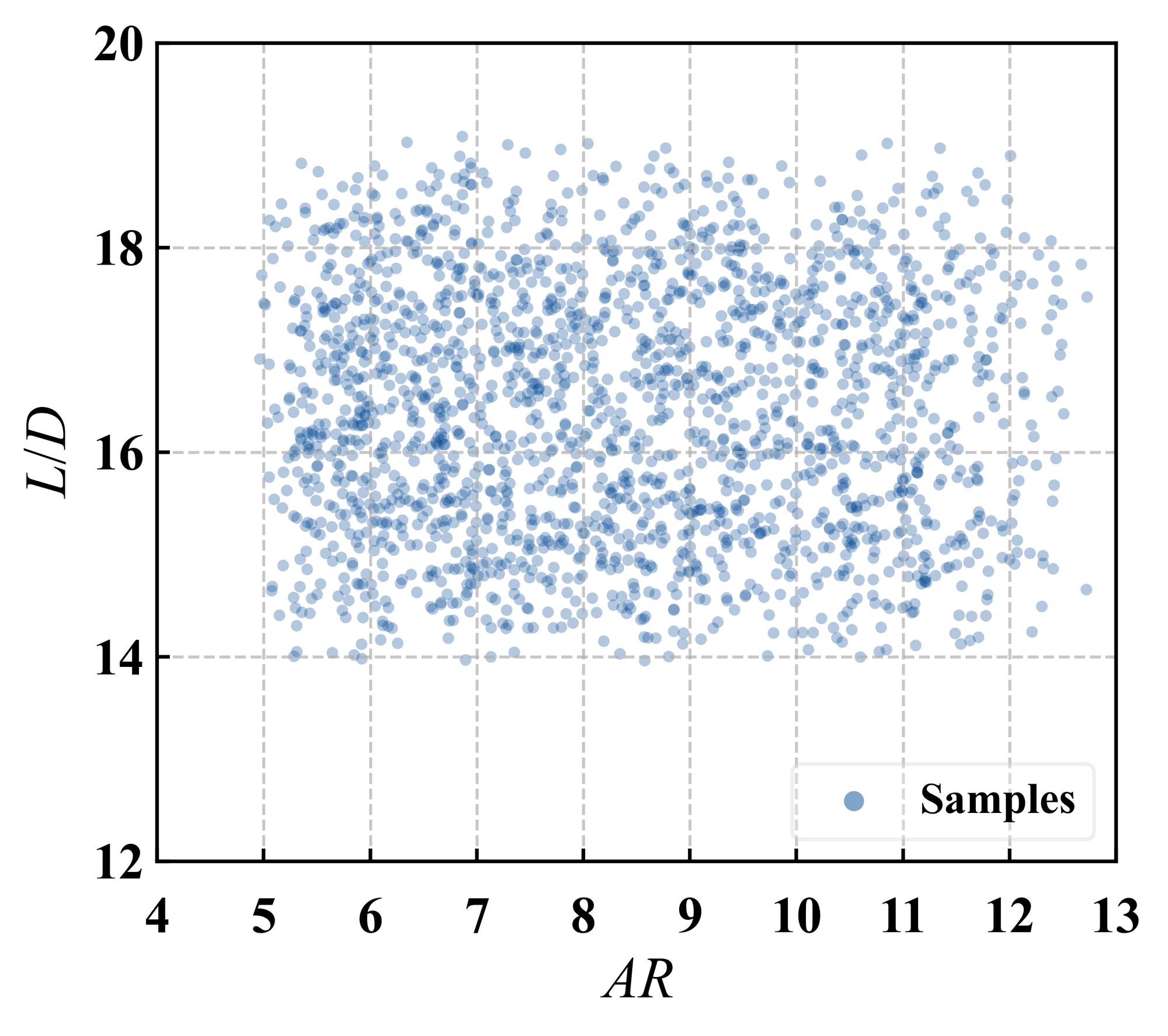}
    \caption{Cond-D}
    \label{fig:LHS目标散点位置}
\end{subfigure}

\caption{Distribution of test conditions in the geometric-aerodynamic condition space.}
\label{fig:全局条件对比}
\end{figure}

To comprehensively evaluate the generation capability of the model, four distinct configurations of conditions are introduced to the generator, as summarized in Table~\ref{tab:target}. It should be noted that although the prescribed condition values ($C_L$, $C_D$, area of the main wing $S$, and span of the main wing $L$) all remain within the variable ranges of the training dataset, their joint configurations may still constitute extrapolation scenarios. For example, the lift-to-drag ratio ($L/D = C_L/C_D$) and the aspect ratio ($AR = L^2/S$) typically exhibit a positive correlation in aerodynamic design. If we simultaneously specify a high lift-to-drag ratio and a low aspect ratio, such samples may never appear in the training data, even though each condition value falls within the range of the training dataset. To this end, distributions of the training data and the test conditions are shown in the $AR-L/D$ space illustrated in Figure~\ref{fig:目标与样本散点} and Figure~\ref{fig:LHS目标散点位置}. Cond-A is set at the center of the sample distribution, representing a typical interpolated case. Cond-B is located at the boundary of the sample space. To challenge the model's capability under extreme conditions, Cond-C further reduces the span based on Cond-B to mimic the stringent demands of engineering design, which can be regarded as an extrapolated case. Cond-D employs Latin Hypercube Sampling (LHS) over predefined ranges of the six design conditions, thereby encompassing both interpolation and extrapolation cases across the global design space.

\begin{table}[hbt!]
\caption{\label{tab:target} Different configurations of the conditions for model validation.}
\centering
\begin{tabular}{ccccccc}
\hline\hline
 Condition & S/$m^2$ & $L/m$ & $\Lambda/^\circ$ & $C_M$ & $C_D$ & $C_L$ \\
\hline
Cond-A & 48 & 10 & 20 & -0.75 & 0.0125 & 0.21 \\
Cond-B & 50 & 11 & 10 & -0.8  & 0.012  & 0.225 \\
Cond-C & 50 & 10 & 10 & -0.8  & 0.0125 & 0.23 \\
Cond-D & \multicolumn{6}{c}{LHS} \\
\hline\hline
\end{tabular}
\end{table}

\ref{app:generated_surrogate_accuracy} compares surrogate predictions with VSPAERO results for the same generated geometries under Cond-A--D. For Cond-A--C, the errors of the surrogate models are below 5\% for all samples and all three aerodynamic coefficients. The drag surrogate is less accurate under Cond-D, possibly due to more demanding combinations of conditions sampled by LHS. Nevertheless, its errors remain below 10\% for 92.0\% of the generated samples.

\subsection{Results}
\label{sec:results}

This section evaluates the generation performance of different models under various design conditions. Section~\ref{sec:preliminary} presents a preliminary demonstration of the performance improvements achieved by applying different condition enhancement strategies, while Section~\ref{sec:full_evaluation} provides a comprehensive assessment of the accuracy and diversity of CEGAN, its intermediate variants, and the baseline CWGAN-GP model. The five models involved in this section and their corresponding method settings are summarized in Table~\ref{tab:model_summary}.

\begin{table}[htbp]
\centering
\caption{Specific configurations of different models.}
\label{tab:model_summary}
\begin{tabular}{lcccc}
\hline\hline
\multirow{2}{*}{Model} & \multicolumn{3}{c}{Loss} & \multirow{2}{*}{Output Filtering} \\
\cmidrule(lr){2-4}
& $\mathcal{L}_{\text{base}}$ & $\mathcal{L}_{\text{geo}}$ & $\mathcal{L}_{\text{aero}}$ &\\
\midrule
CWGAN-GP (Baseline) & $\checkmark$ & & & \\
CE-geo & $\checkmark$ & $\checkmark$ & & \\
CE-aero & $\checkmark$ &  & $\checkmark$ &  \\
CEGAN-Unfiltered & $\checkmark$ & $\checkmark$ & $\checkmark$ &  \\
CEGAN & $\checkmark$ & $\checkmark$ & $\checkmark$ & $\checkmark$ \\
\hline\hline
\end{tabular}
\end{table}

\subsubsection{Condition-Informed Loss Function}
\label{sec:preliminary}

Initially, we implemented the condition-informed loss strategy introduced in Sec.~\ref{sec:CEGAN-enhanced} to construct two condition-enhanced generative models: one for the geometry-guided enhancement and the other for the aerodynamic performance-guided enhancement. These results do not yet incorporate the output filtering mechanism, which only serves to validate the two enhancement strategies used in the training phase. Since geometric parameters possess distinct dimensionalities, value ranges, and sensitivities, employing identical $\lambda_k$ values in Eq.~\ref{eq:geo} would bias the model towards conditions with larger loss magnitudes. This leads to an imbalance in model enhancement across multiple conditions. Furthermore, generative design problems are inherently non-unique. Furthermore, without introducing $a$ in $\mathcal{L}_{\text{geo}}$ will force the geometric shapes, thus limiting the diversity of generated results. Therefore, when imposing the geometry-guided loss function $\mathcal{L}_{\text{geo}}$, we introduced four hyperparameters for experimentation: three weight coefficients $\lambda_L$, $\lambda_S$, and $\lambda_{\Lambda}$ assigned to the losses of different geometric conditions, and a tolerance threshold coefficient $a$ to penalize the errors of conditions. The hyperparameter configurations for the different models are summarized in Table~\ref{tab:Geo-hyperparam}.

\begin{table}[hbt!]
\caption{\label{tab:Geo-hyperparam} Hyperparameter settings for different models.}
\centering
\begin{tabular}{ccccc}
\hline\hline
Configuration & $a$& $\lambda_S$& $\lambda_L$& $\lambda_{\Lambda}$\\\hline
Geo-1& 0& 10& 10& 10\\
Geo-2& 0& 30& 20& 10\\
Geo-3& 0.02& 10& 10& 10\\
Geo-4& 0.02& 10& 20& 30\\
\hline\hline
\end{tabular}
\end{table}

\begin{figure}[hbt!]
\centering
\includegraphics[width=\textwidth]{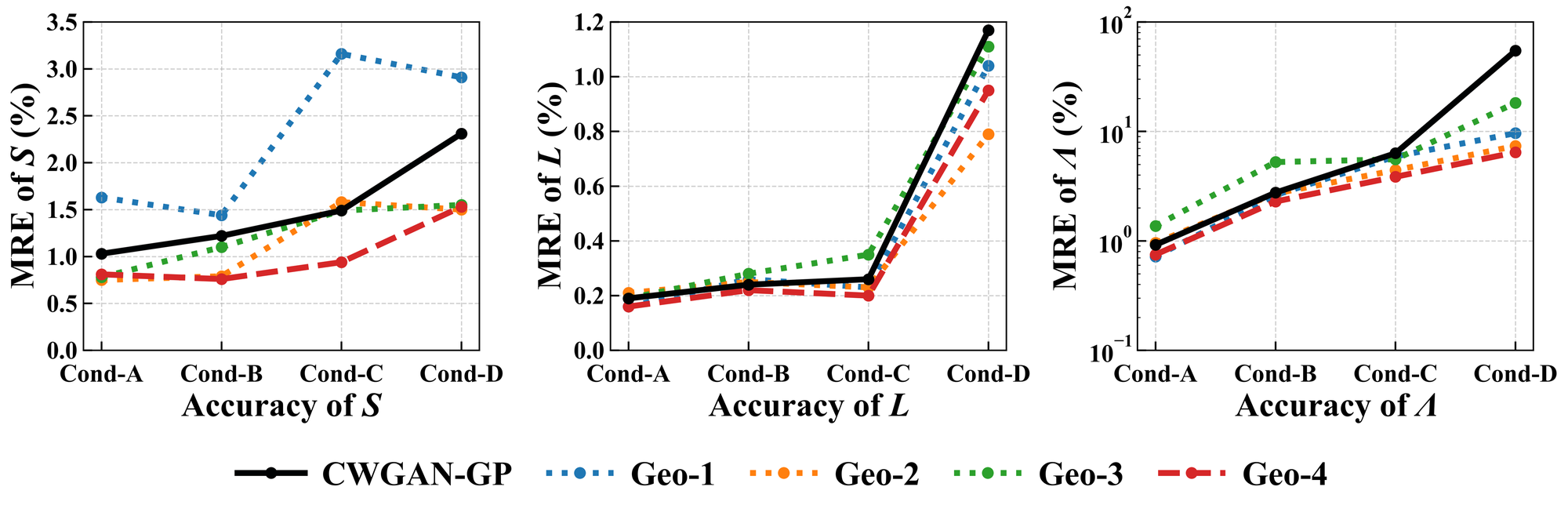}
\caption{MRE comparison of geometric conditions under different hyperparameter configurations, for the CE-geo model framework. Configurations Geo-1 to Geo-4 correspond to Table~\ref{tab:Geo-hyperparam}.}
\label{fig:GeoLoss}
\end{figure}

Figure~\ref{fig:GeoLoss} compares the generation performance of the CE-geo model with different hyperparameter configurations. The Geo-4 configuration model exhibits better generation accuracy compared to the baseline CWGAN-GP model across all tested cases. Therefore, the hyperparameter combination corresponding to Geo-4 is selected as the optimal configuration for the geometry-guided loss function. The weighting coefficients $\lambda_{C_i}$ in the aerodynamic performance-guided enhancement loss function (Eq.~\ref{eq:aero}) were determined using the same hyperparameter optimization procedure as that employed for the geometric conditions, while the detailed tuning process is omitted for brevity.

\begin{figure}[htbp]
\centering

\begin{subfigure}[b]{0.3\textwidth}
    \centering
    \includegraphics[width=\textwidth]{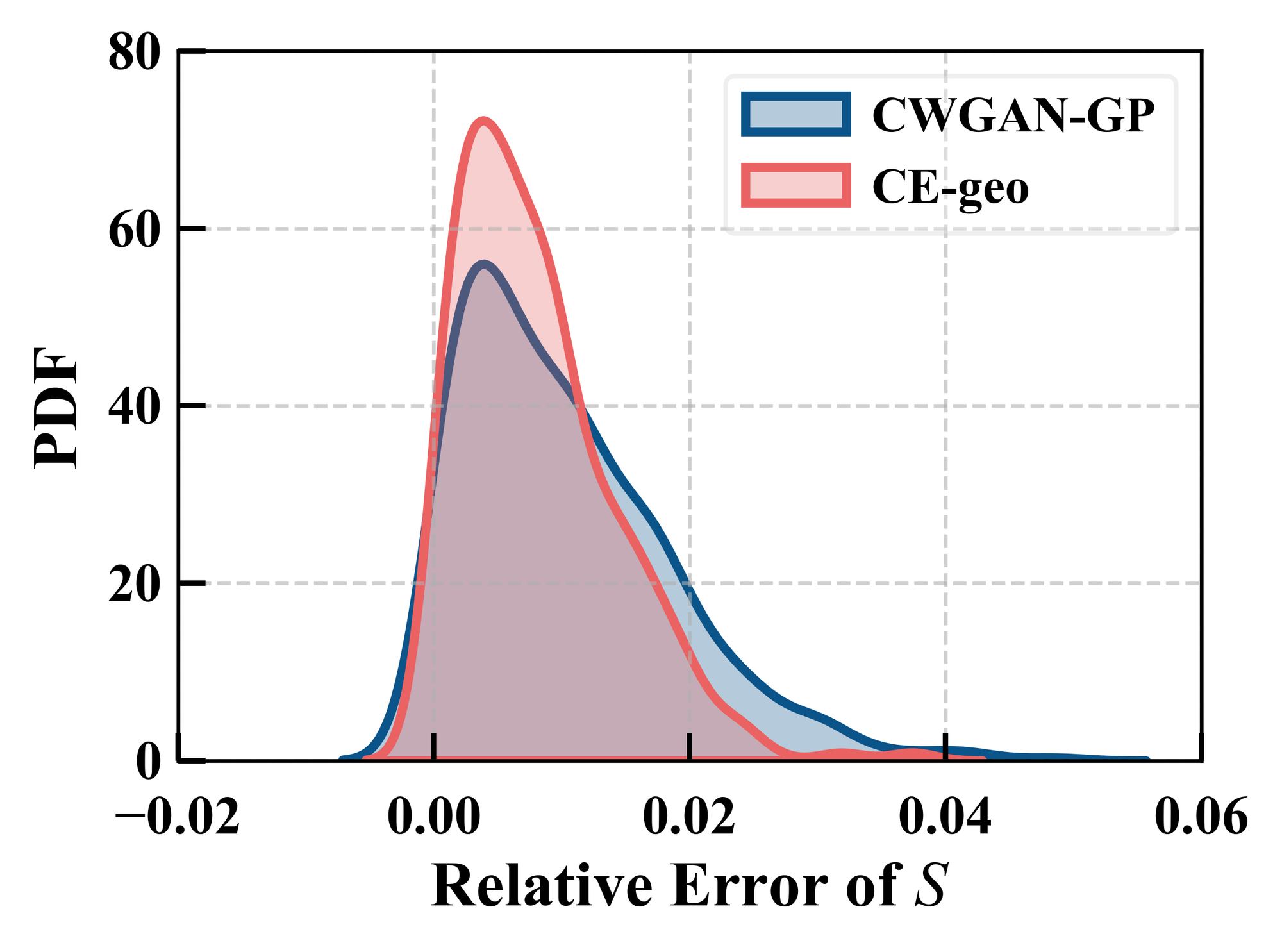}
    \caption{$S$}
    \label{fig:1a}
\end{subfigure}
\hfill
\begin{subfigure}[b]{0.3\textwidth}
    \centering
    \includegraphics[width=\textwidth]{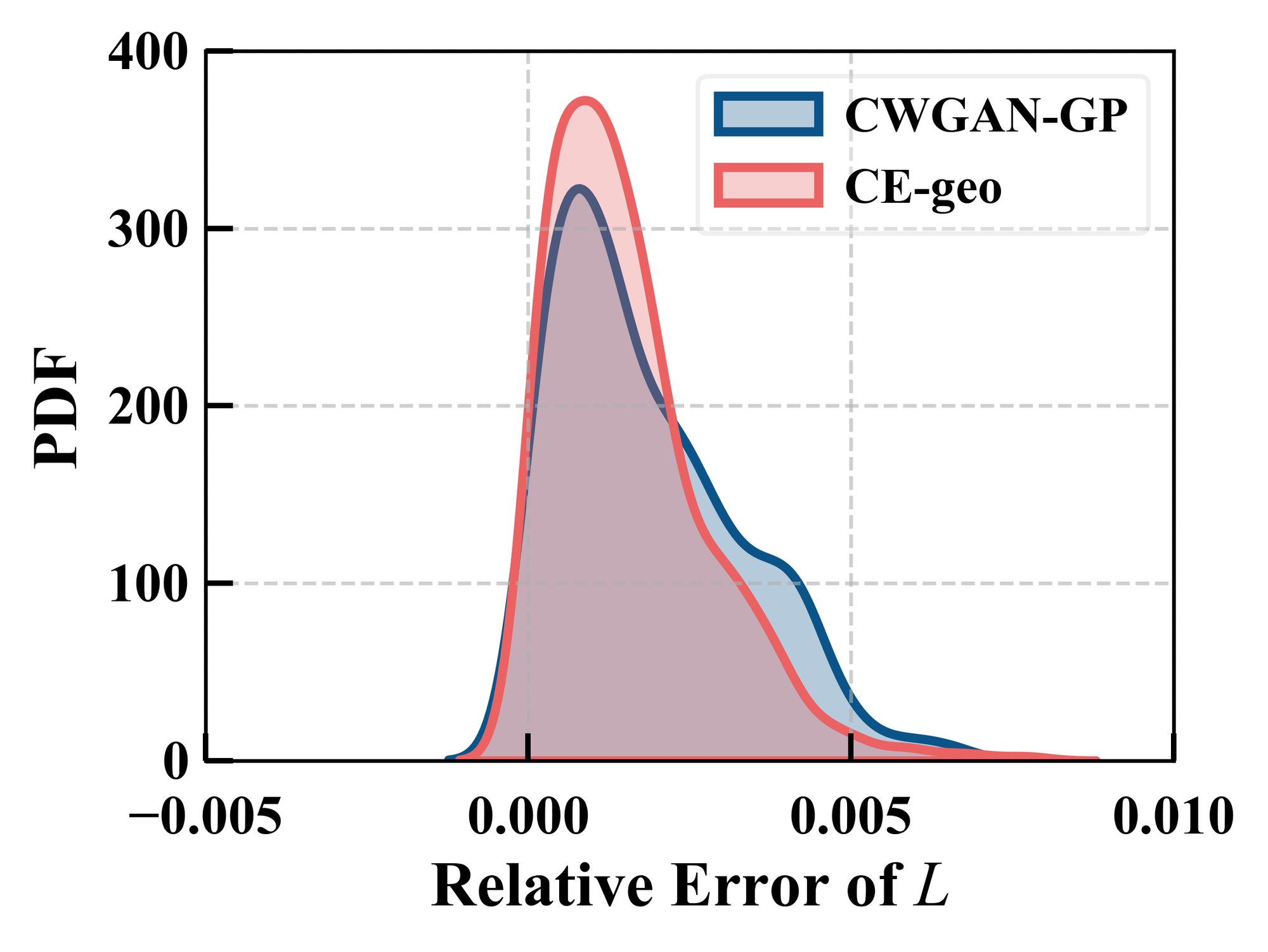}
    \caption{$L$}
    \label{fig:1b}
\end{subfigure}
\hfill
\begin{subfigure}[b]{0.3\textwidth}
    \centering
    \includegraphics[width=\textwidth]{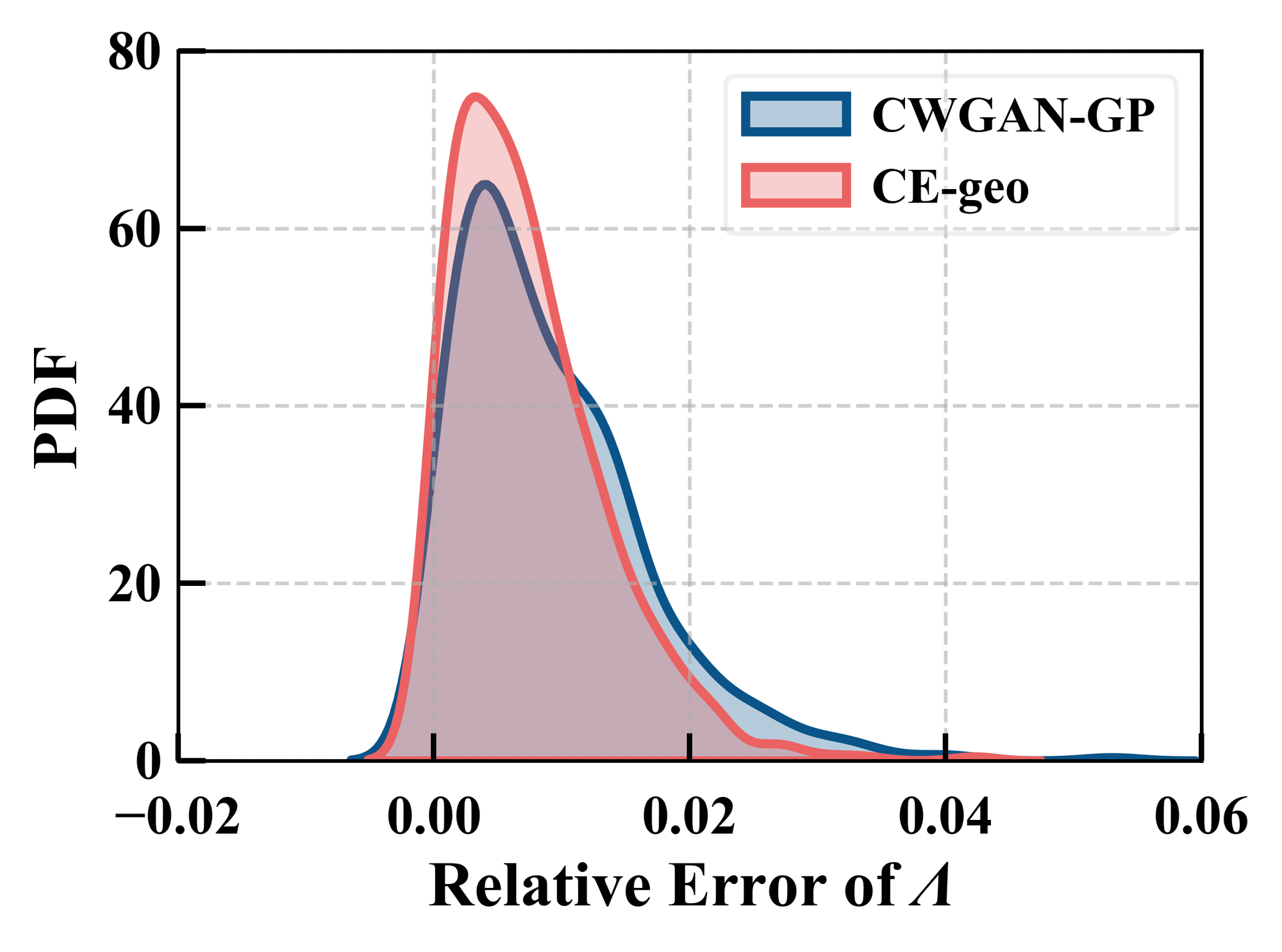}
    \caption{$\Lambda$}
    \label{fig:1c}
\end{subfigure}

\vspace{1em} 

\begin{subfigure}[b]{0.3\textwidth}
    \centering
    \includegraphics[width=\textwidth]{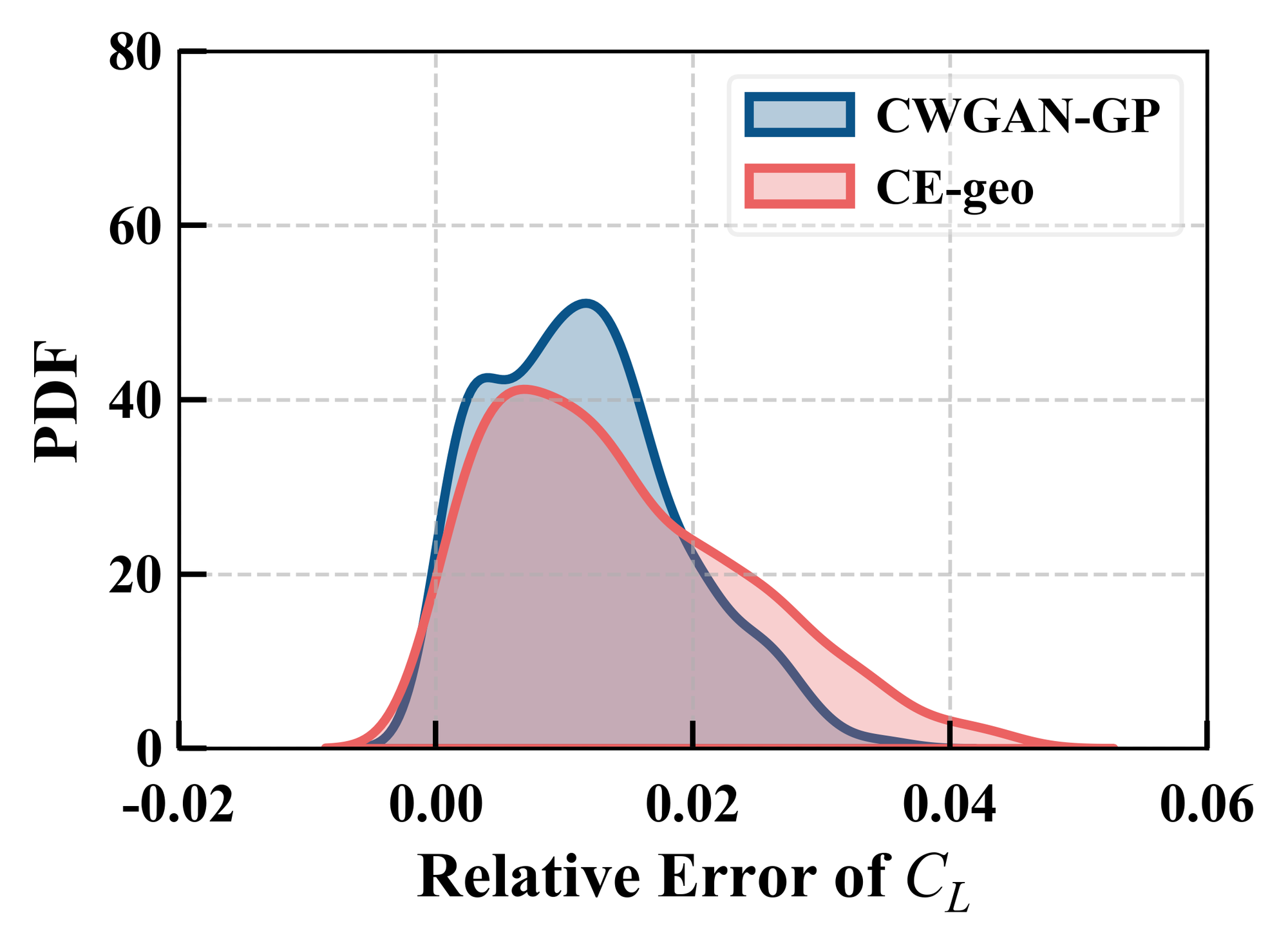}
    \caption{$C_L$}
    \label{fig:1d}
\end{subfigure}
\hfill
\begin{subfigure}[b]{0.3\textwidth}
    \centering
    \includegraphics[width=\textwidth]{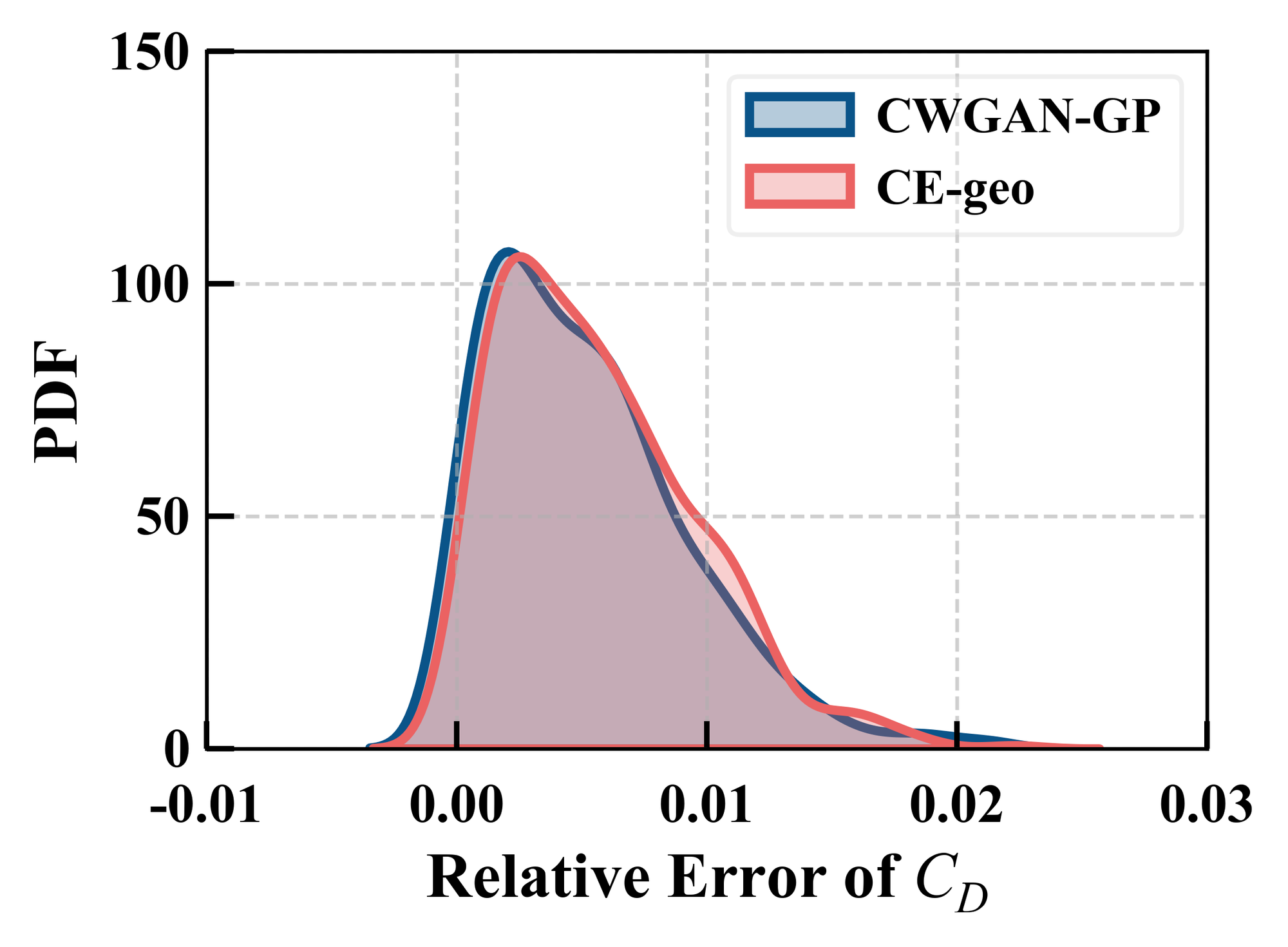}
    \caption{$C_D$}
    \label{fig:1e}
\end{subfigure}
\hfill
\begin{subfigure}[b]{0.3\textwidth}
    \centering
    \includegraphics[width=\textwidth]{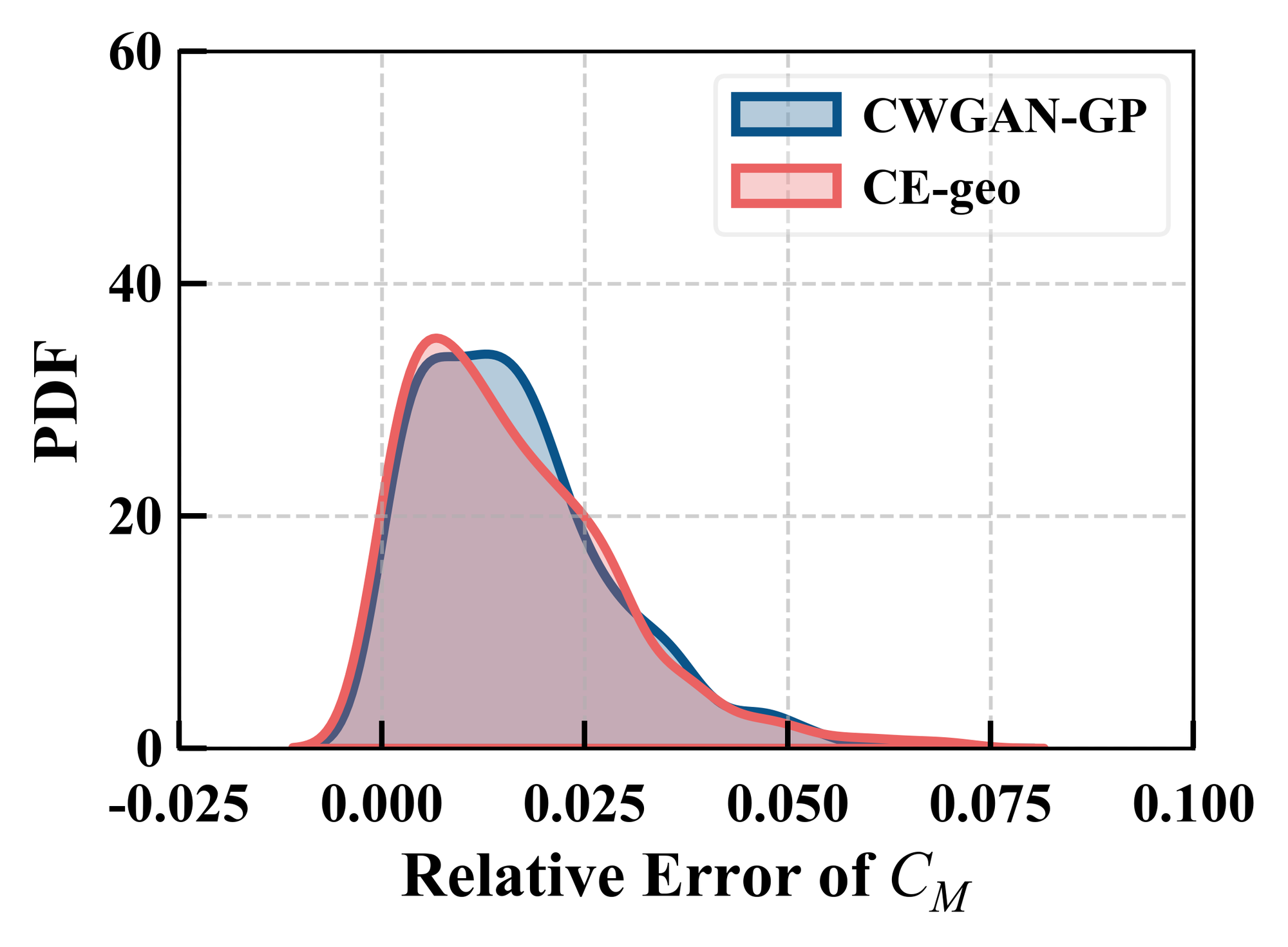}
    \caption{$C_M$}
    \label{fig:1f}
\end{subfigure}

\begin{subfigure}[b]{0.3\textwidth}
    \centering
    \includegraphics[width=\textwidth]{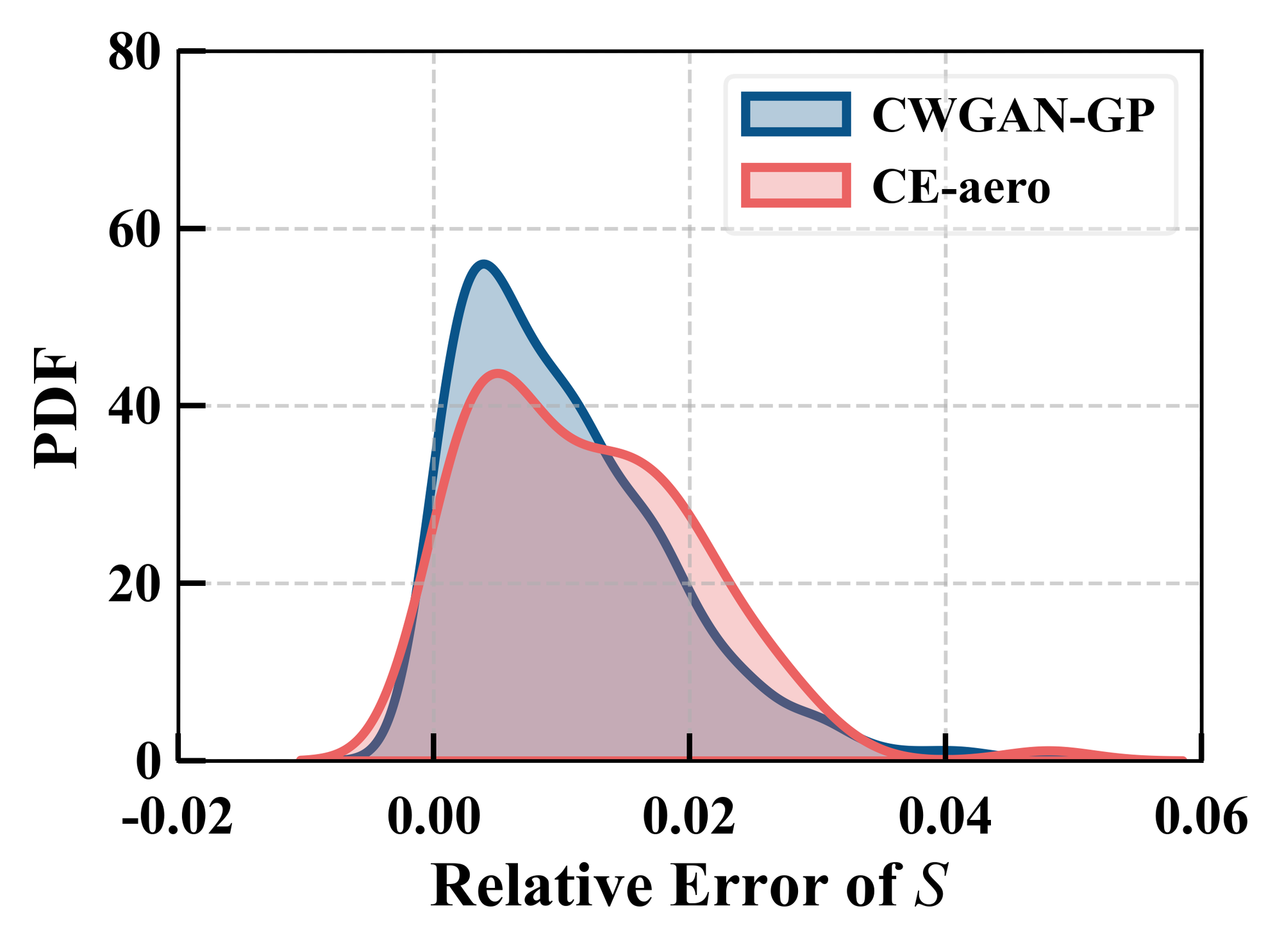}
    \caption{$S$}
    \label{fig:1h}
\end{subfigure}
\hfill
\begin{subfigure}[b]{0.3\textwidth}
    \centering
    \includegraphics[width=\textwidth]{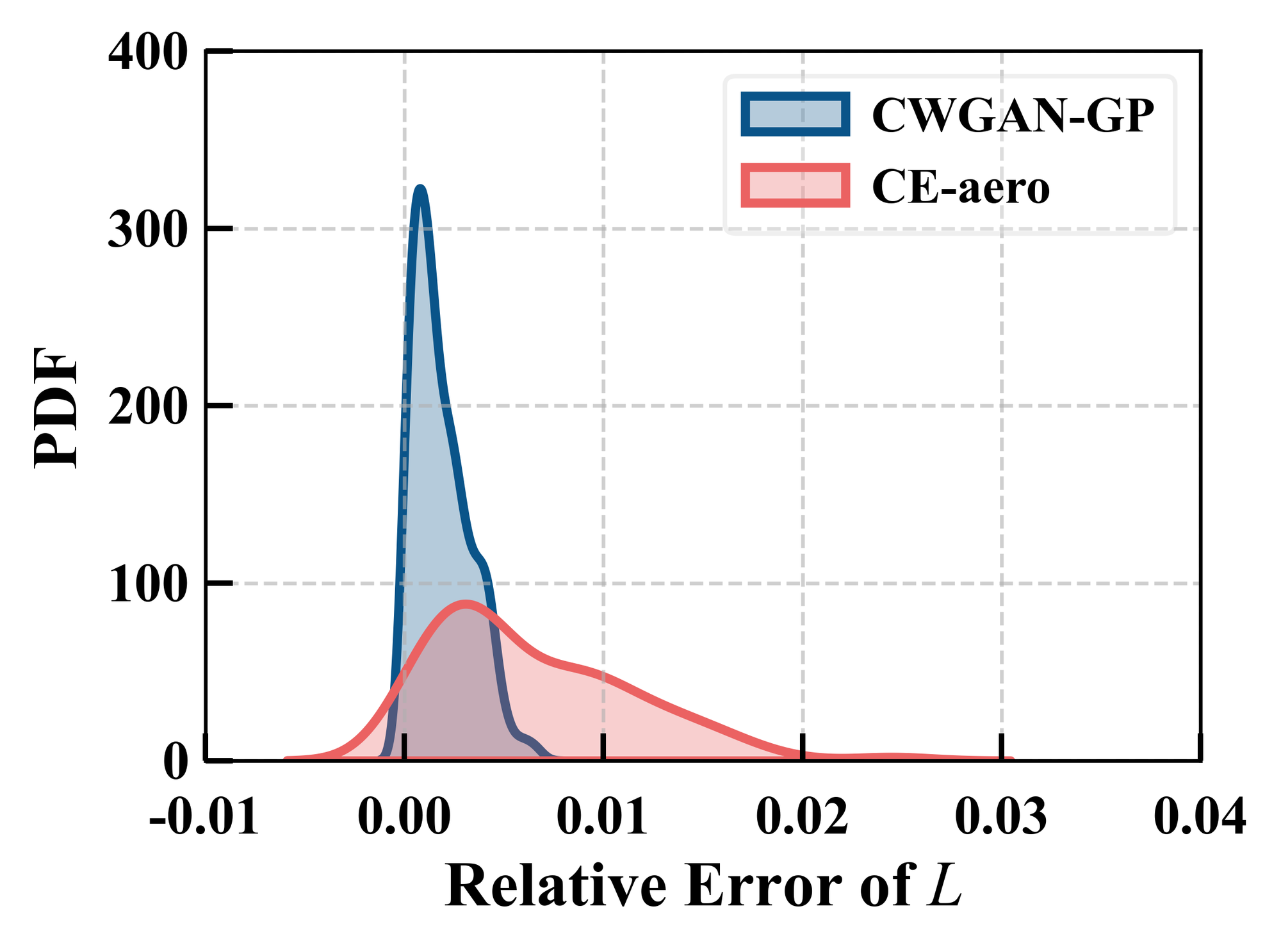}
    \caption{$L$}
    \label{fig:1i}
\end{subfigure}
\hfill
\begin{subfigure}[b]{0.3\textwidth}
    \centering
    \includegraphics[width=\textwidth]{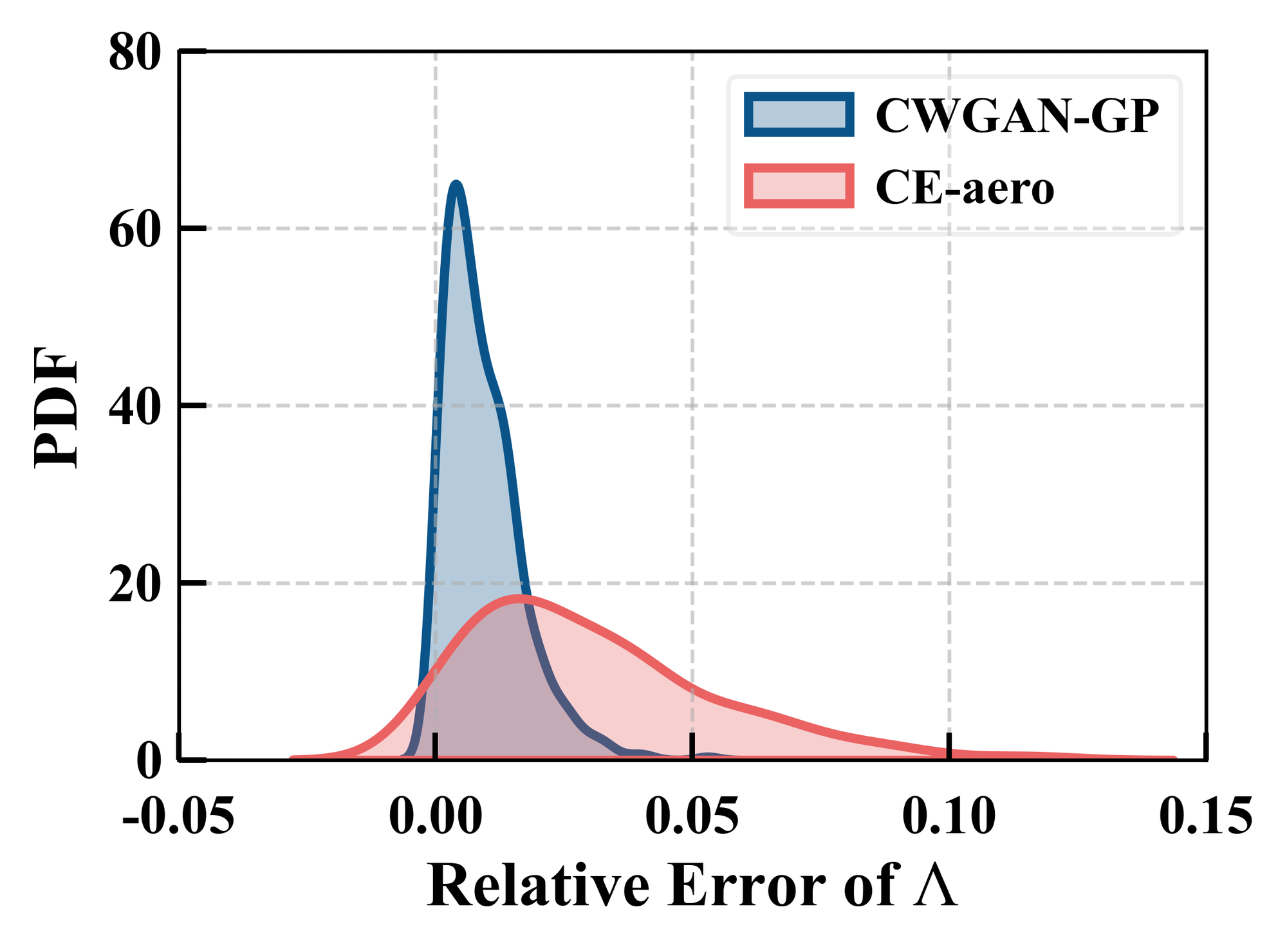}
    \caption{$\Lambda$}
    \label{fig:1j}
\end{subfigure}

\vspace{1em} 

\begin{subfigure}[b]{0.3\textwidth}
    \centering
    \includegraphics[width=\textwidth]{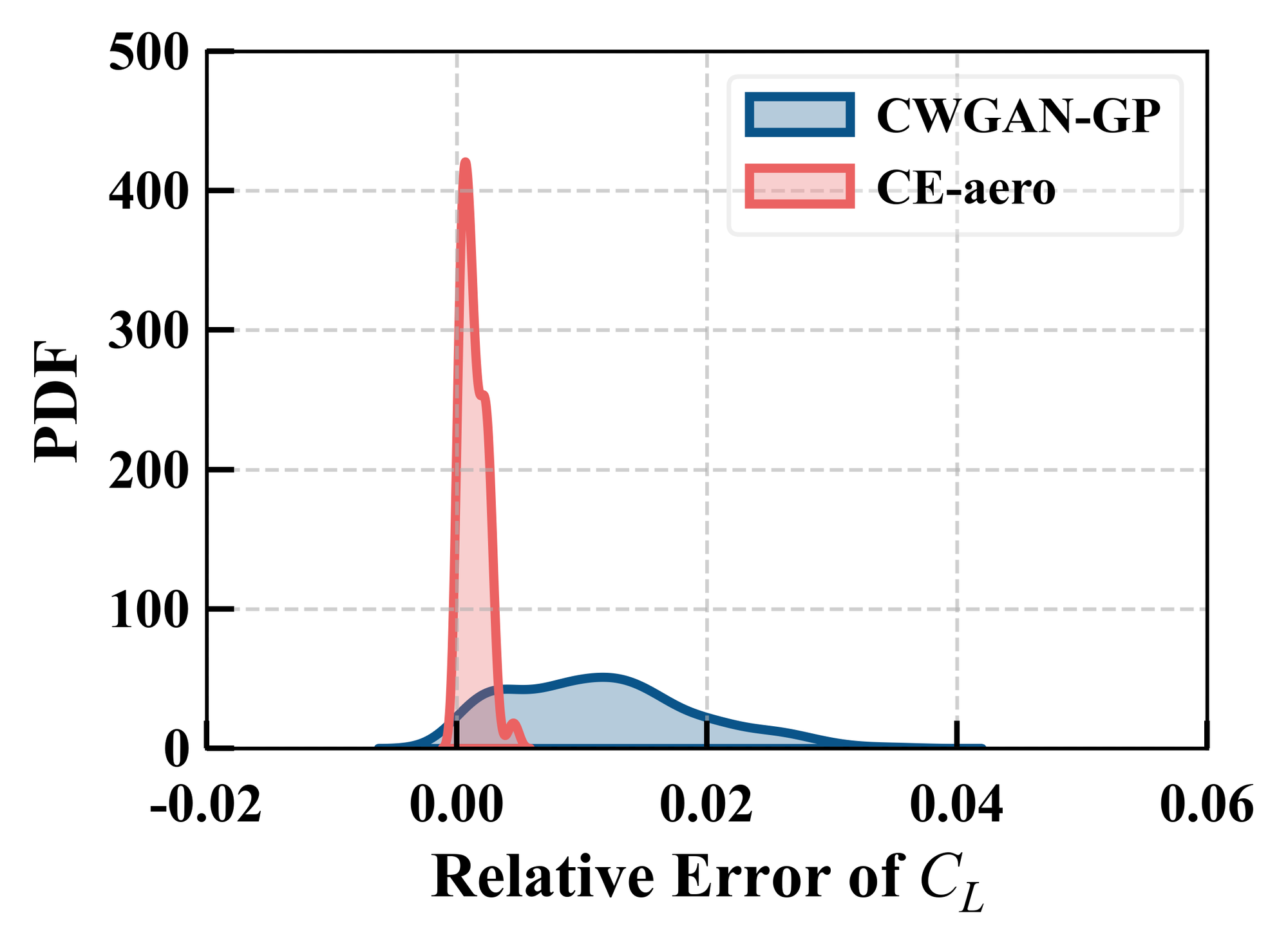}
    \caption{$C_L$}
    \label{fig:1k}
\end{subfigure}
\hfill
\begin{subfigure}[b]{0.3\textwidth}
    \centering
    \includegraphics[width=\textwidth]{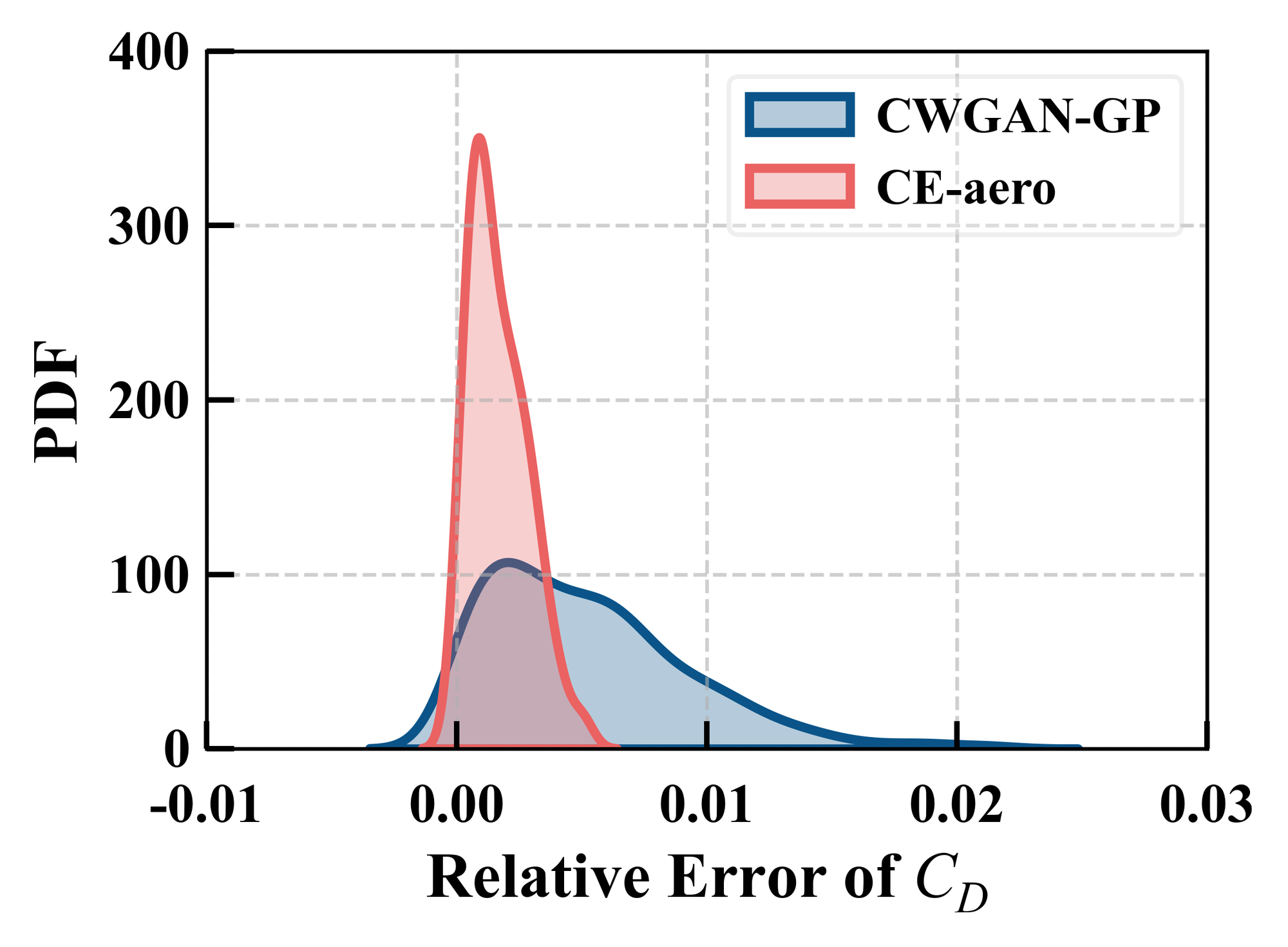}
    \caption{$C_D$}
    \label{fig:1l}
\end{subfigure}
\hfill
\begin{subfigure}[b]{0.3\textwidth}
    \centering
    \includegraphics[width=\textwidth]{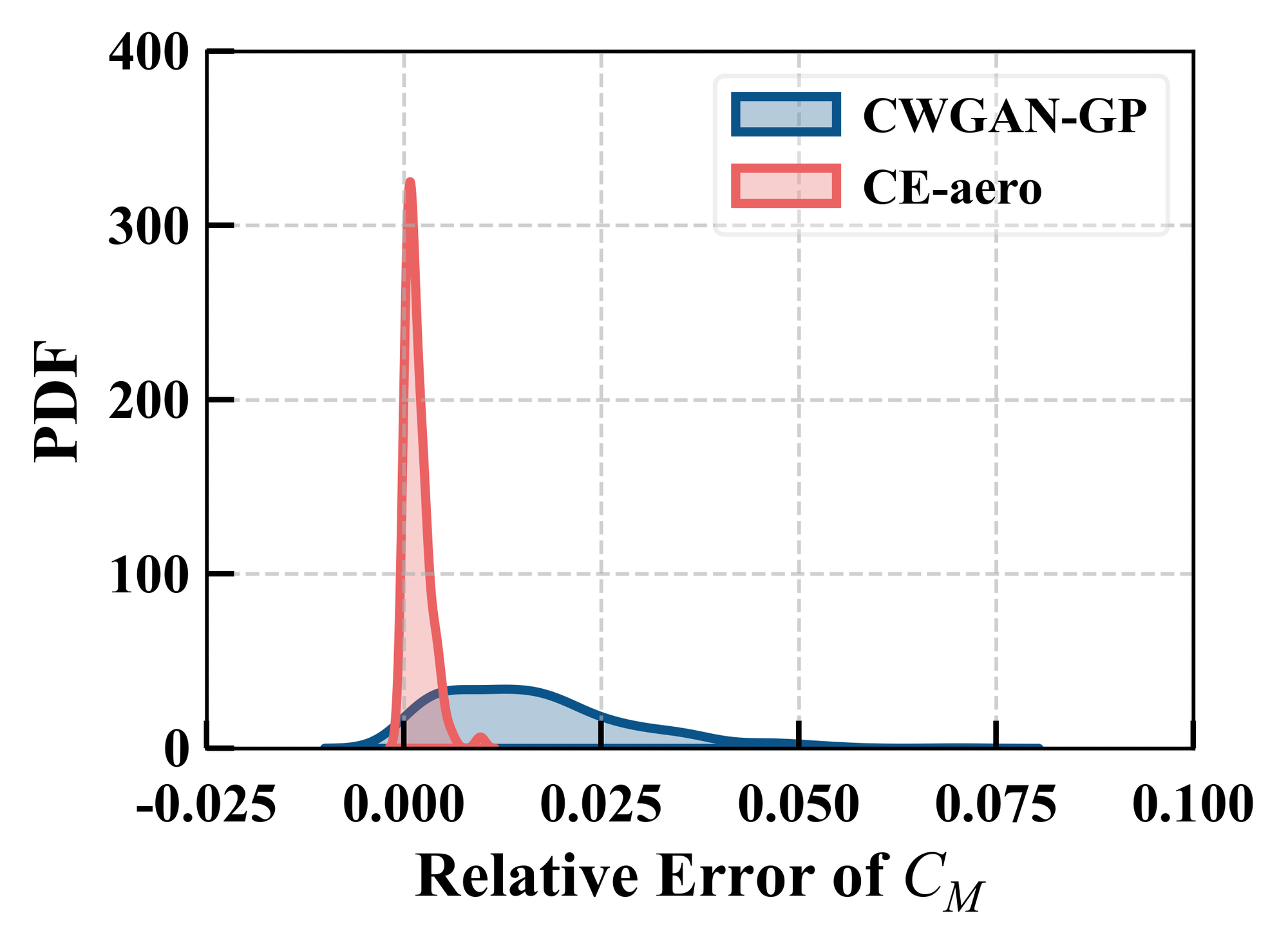}
    \caption{$C_M$}
    \label{fig:1m}
\end{subfigure}

\caption{PDFs of the absolute relative errors for the six design conditions under Cond-A. The first two rows compare CE-geo with CWGAN-GP: (a--c) geometric conditions ($S$, $L$, $\Lambda$) and (d--f) aerodynamic conditions ($C_L$, $C_D$, $C_M$). The last two rows compare CE-aero with CWGAN-GP: (g--i) geometric conditions and (j--l) aerodynamic conditions.}

\label{fig:pdf_comparison}
\end{figure}

Figure~\ref{fig:pdf_comparison} compares the PDFs of the absolute relative errors for different models across six design conditions under Cond-A. The first two rows present the results of CE-geo, whereas the last two rows present those of CE-aero, with CWGAN-GP serving as the baseline in each comparison. For Cond-A, CE-geo reduces the errors in the geometric conditions, but does not provide consistent improvements in the aerodynamic conditions. Conversely, CE-aero substantially reduces the errors in all three aerodynamic coefficients, but increases the errors in the geometric conditions. Each enhancement term improves the matching accuracy of its associated conditions. However, it may also increase errors in the other conditions. These results support combining the geometry-guided and aerodynamic performance-guided losses in CEGAN. The combined loss aims to improve accuracy for both geometric and aerodynamic conditions.

A further study using CE-aero evaluates how surrogate accuracy affects aerodynamic enhancement (\ref{app:surrogate_sensitivity}). In this aircraft case, surrogate models with a mean validation $R^2$ of about 0.95 or higher provide clearer improvements across all three aerodynamic coefficients. Generation accuracy also depends on the feasibility of the specified conditions. Some samples may therefore have accurate surrogate predictions but poor agreement with these conditions. Overall, the results indicate that improving surrogate accuracy can provide more effective guidance for the generator.

\begin{figure}[htbp]
\centering
\begin{subfigure}[b]{0.32\textwidth}
    \centering
    \includegraphics[width=\textwidth]{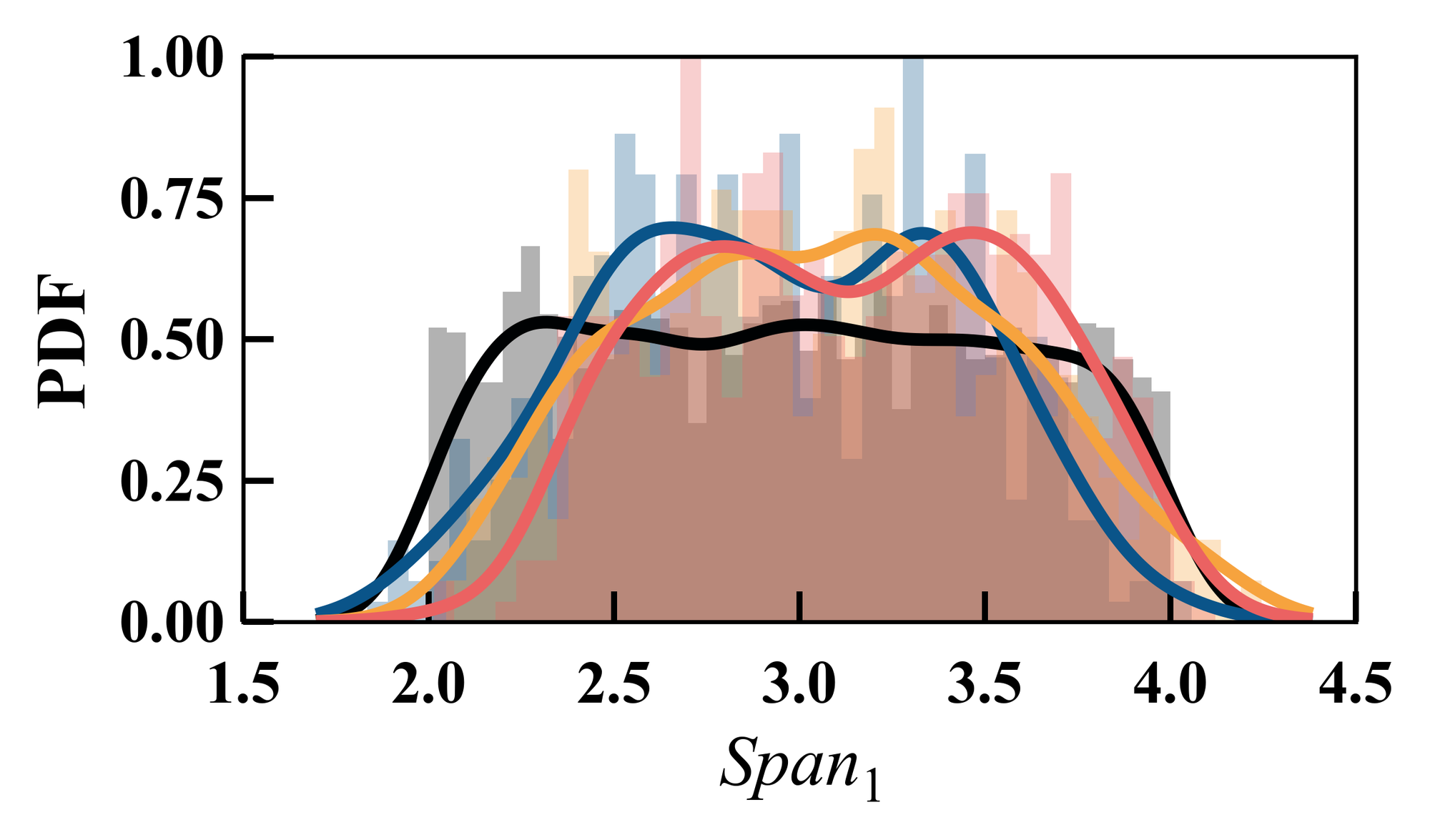}
\end{subfigure}
\hfill
\begin{subfigure}[b]{0.32\textwidth}
    \centering
    \includegraphics[width=\textwidth]{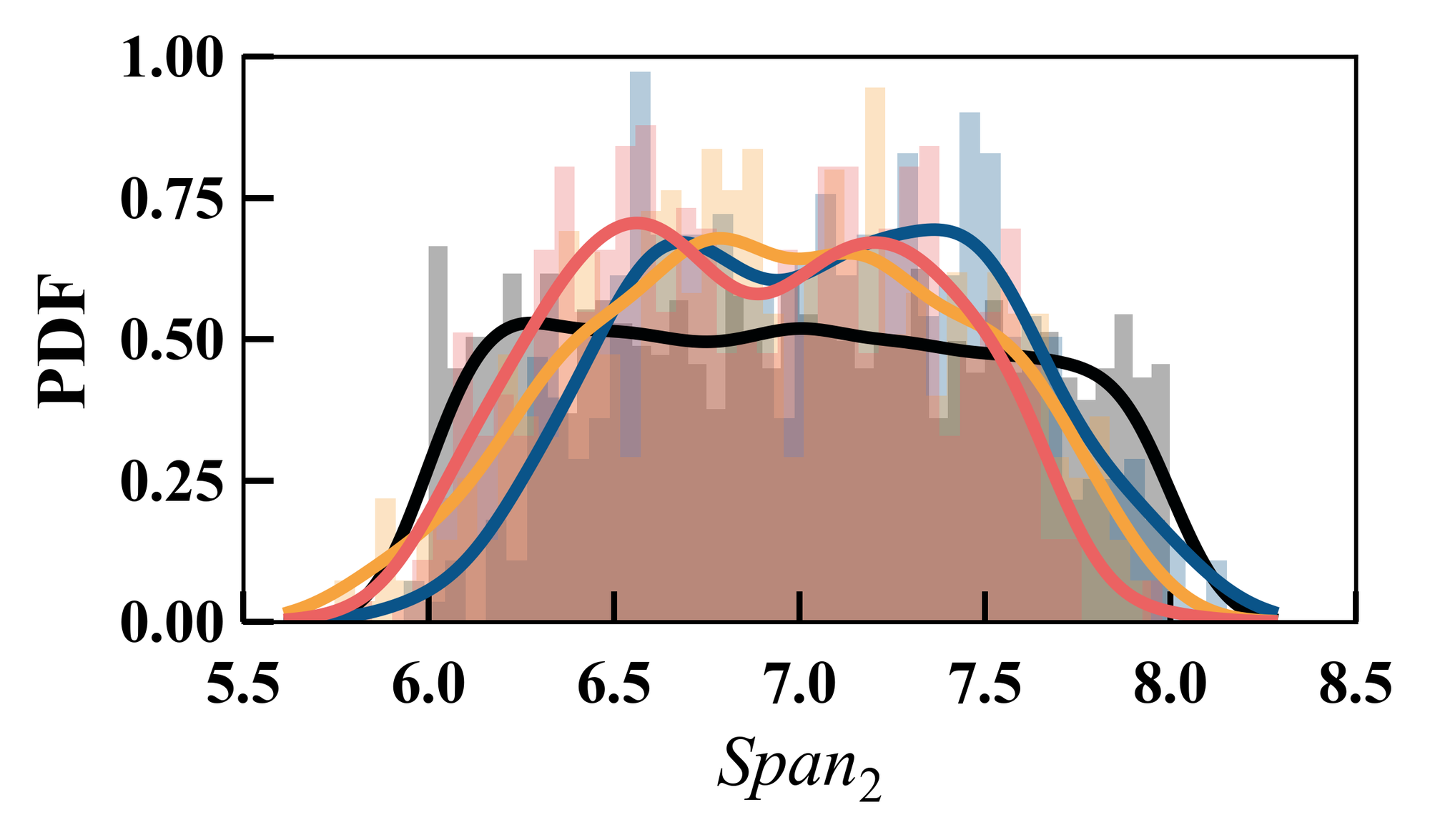}
\end{subfigure}
\hfill
\begin{subfigure}[b]{0.32\textwidth}
    \centering
    \includegraphics[width=\textwidth]{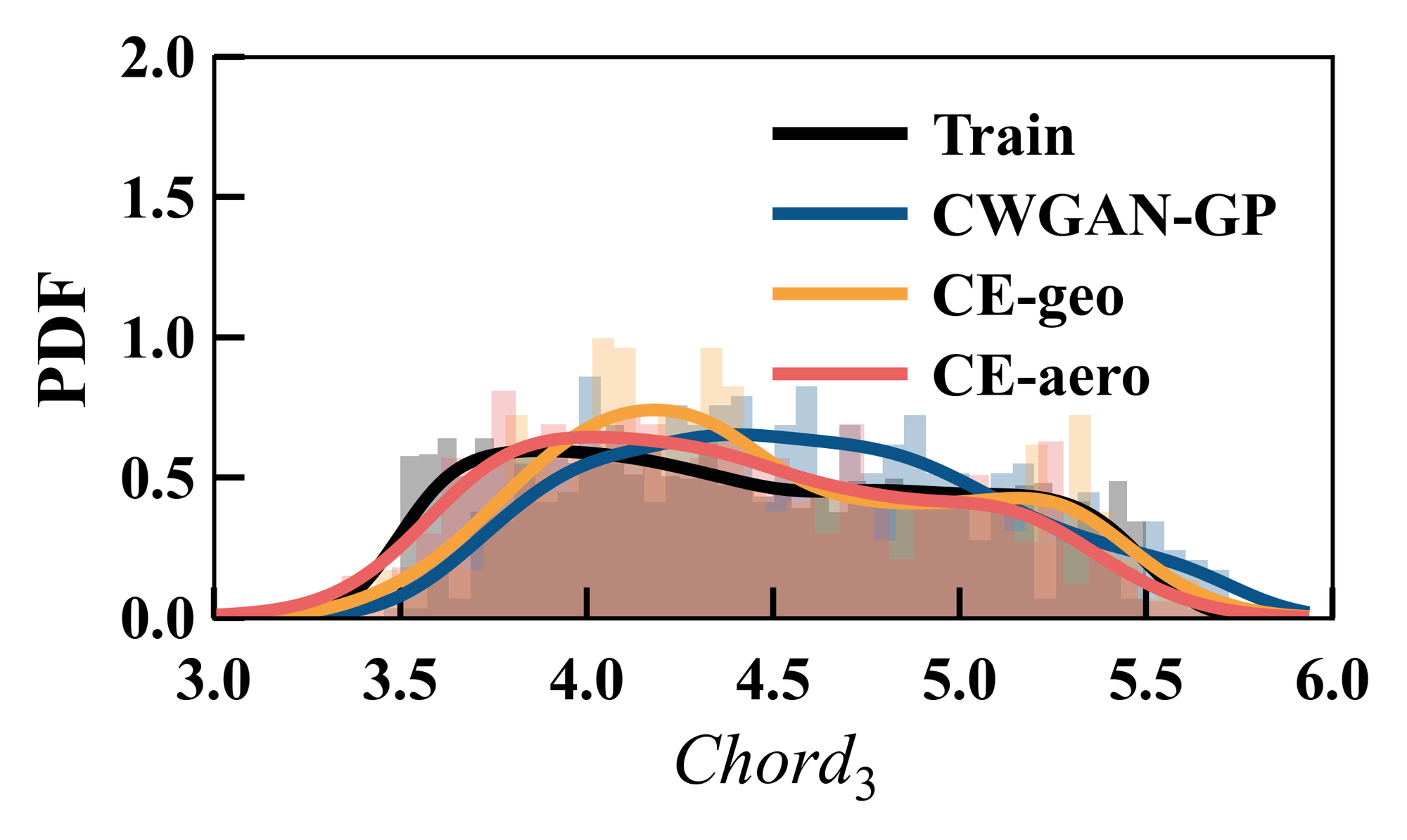}
\end{subfigure}

\vspace{0.5em} 

\begin{subfigure}[b]{0.32\textwidth}
    \centering
    \includegraphics[width=\textwidth]{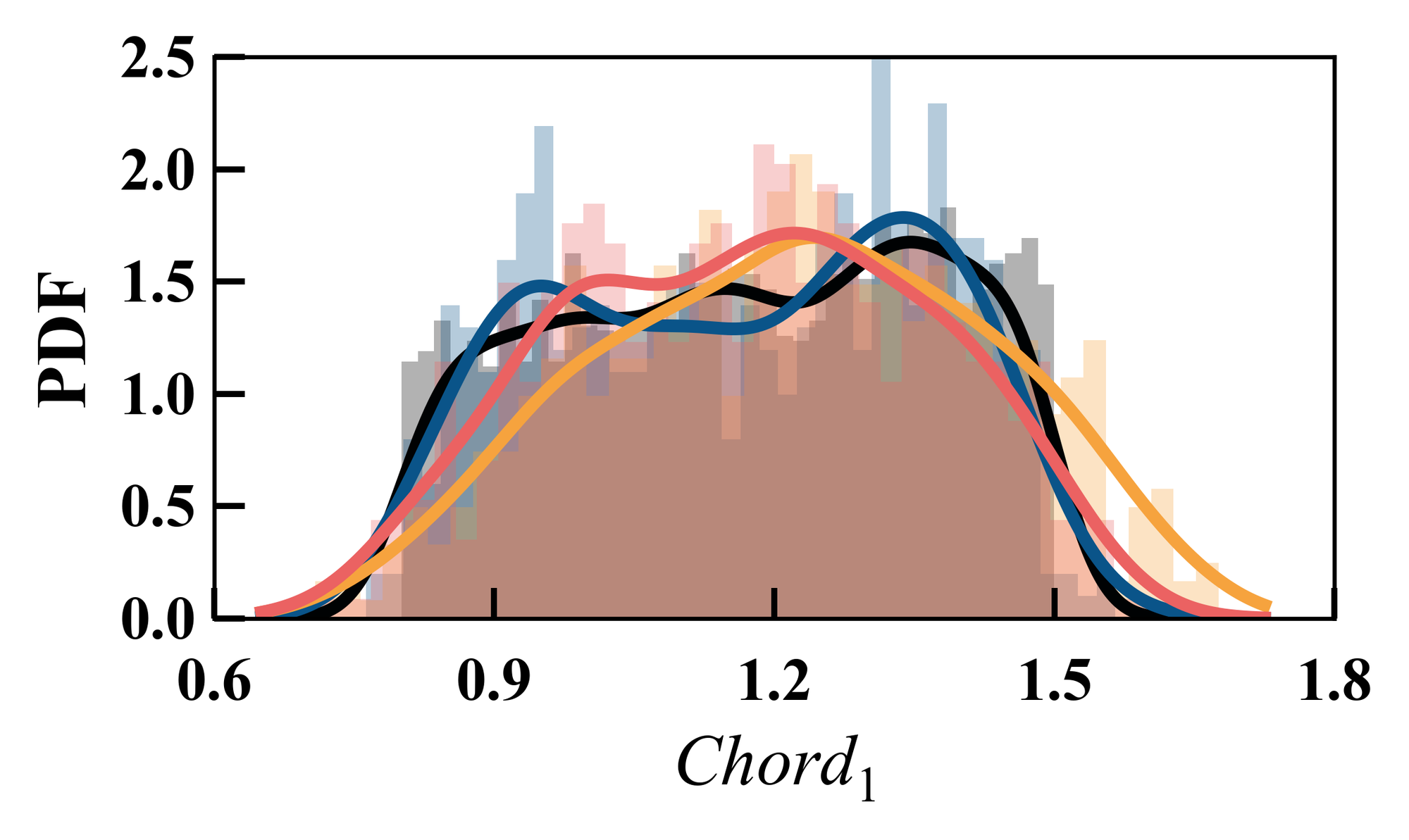}
\end{subfigure}
\hfill
\begin{subfigure}[b]{0.32\textwidth}
    \centering
    \includegraphics[width=\textwidth]{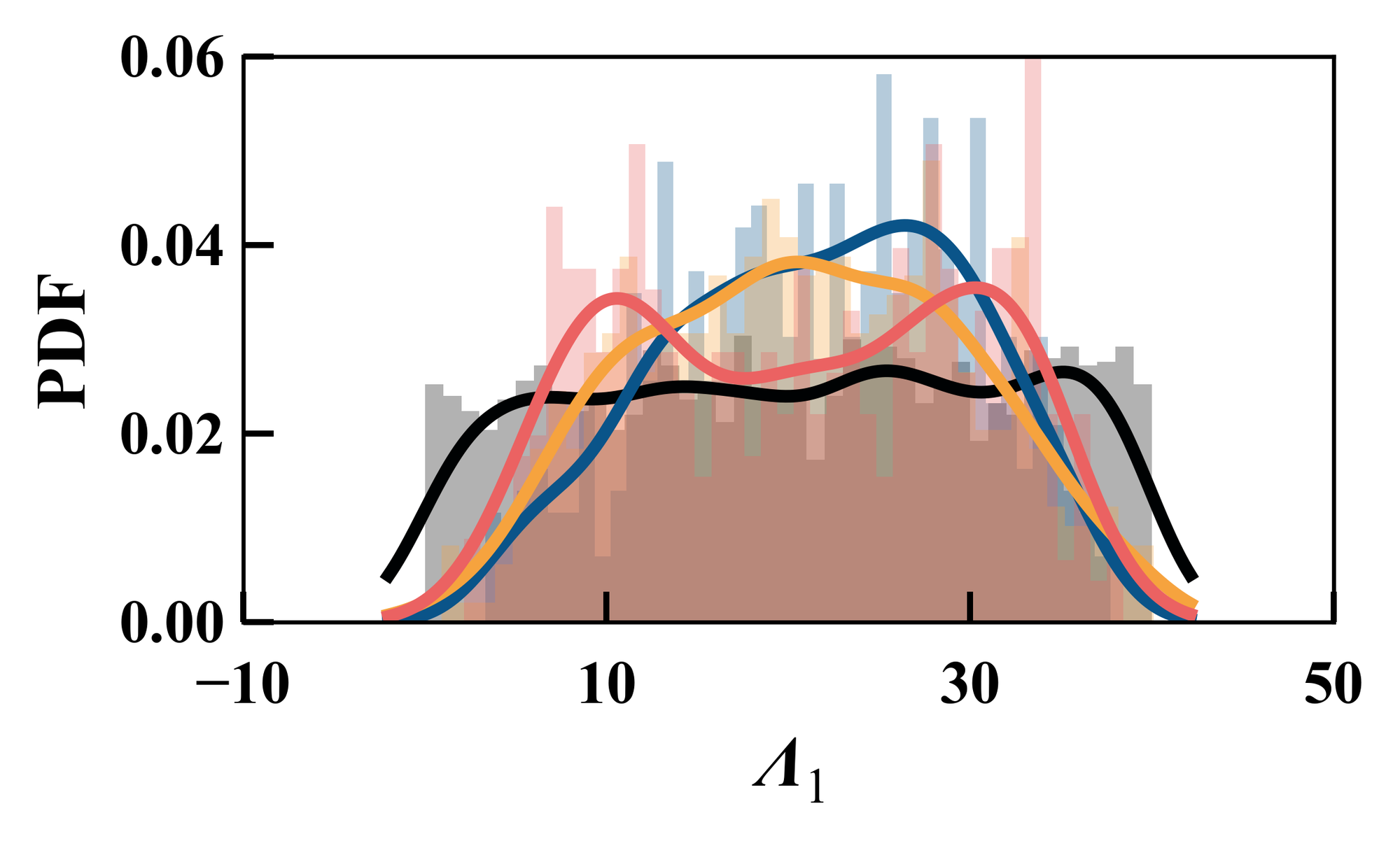}
\end{subfigure}
\hfill
\begin{subfigure}[b]{0.32\textwidth}
    \centering
    \includegraphics[width=\textwidth]{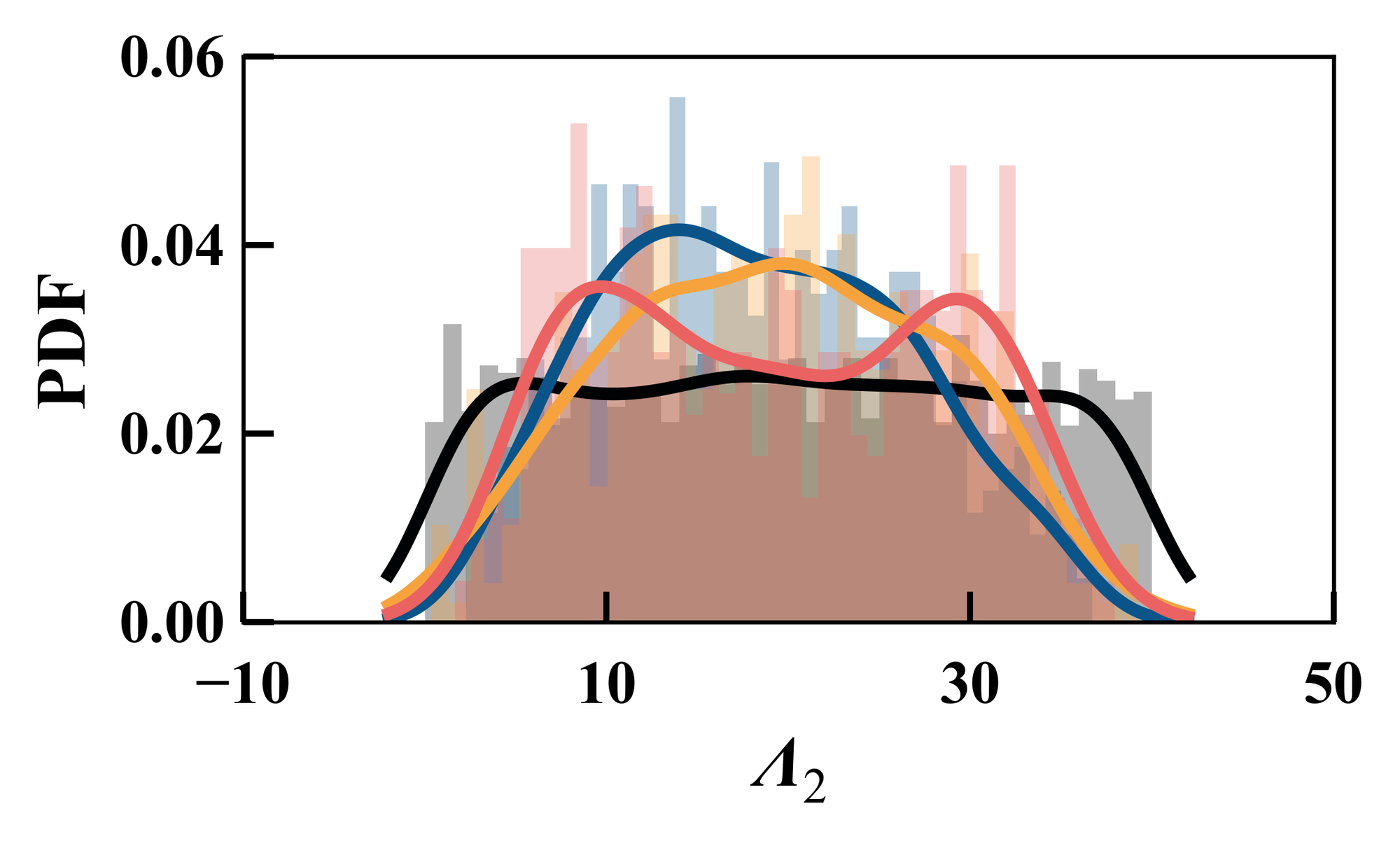}
\end{subfigure}

\vspace{0.5em} 

\begin{subfigure}[b]{0.32\textwidth}
    \centering
    \includegraphics[width=\textwidth]{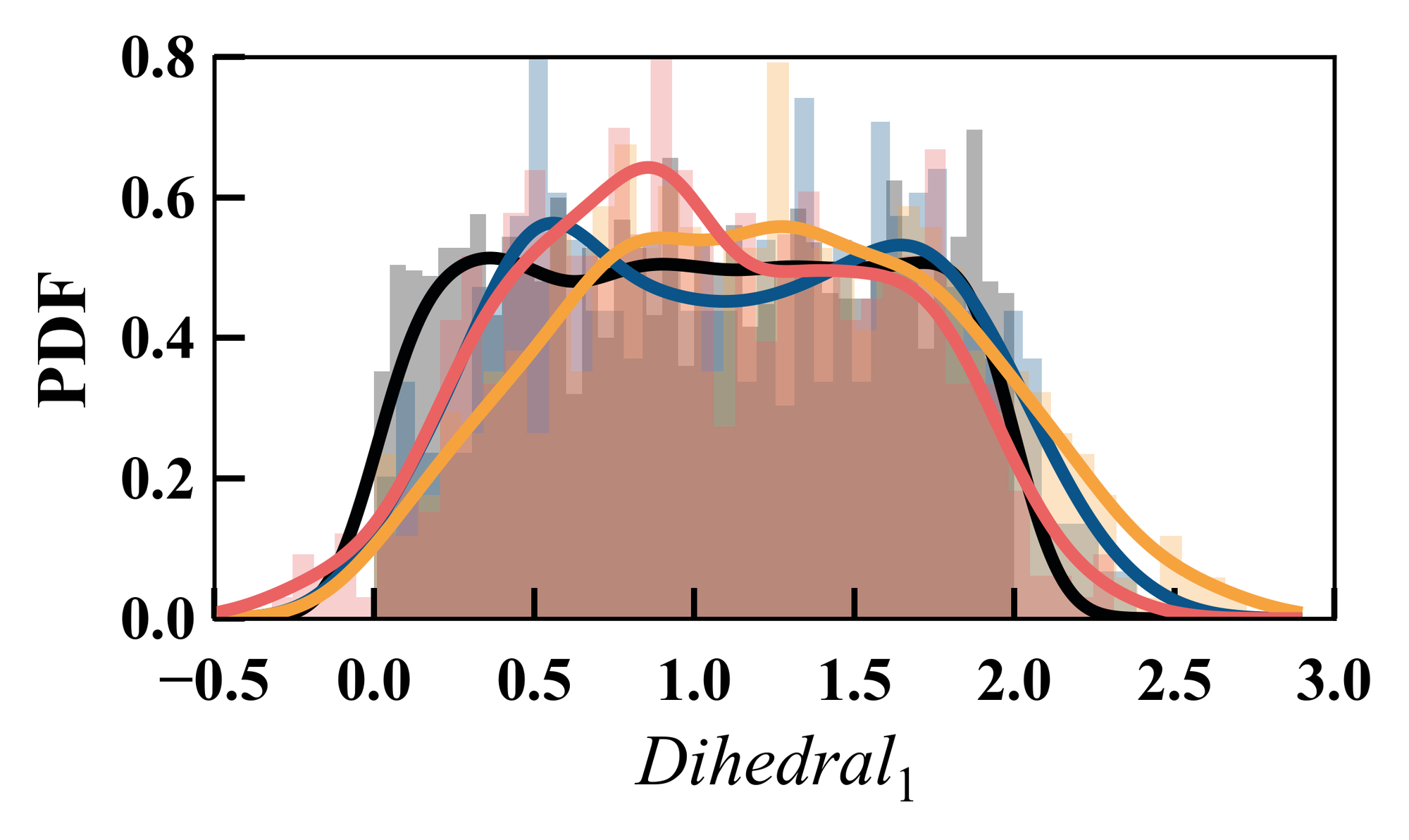}
\end{subfigure}
\hfill
\begin{subfigure}[b]{0.32\textwidth}
    \centering
    \includegraphics[width=\textwidth]{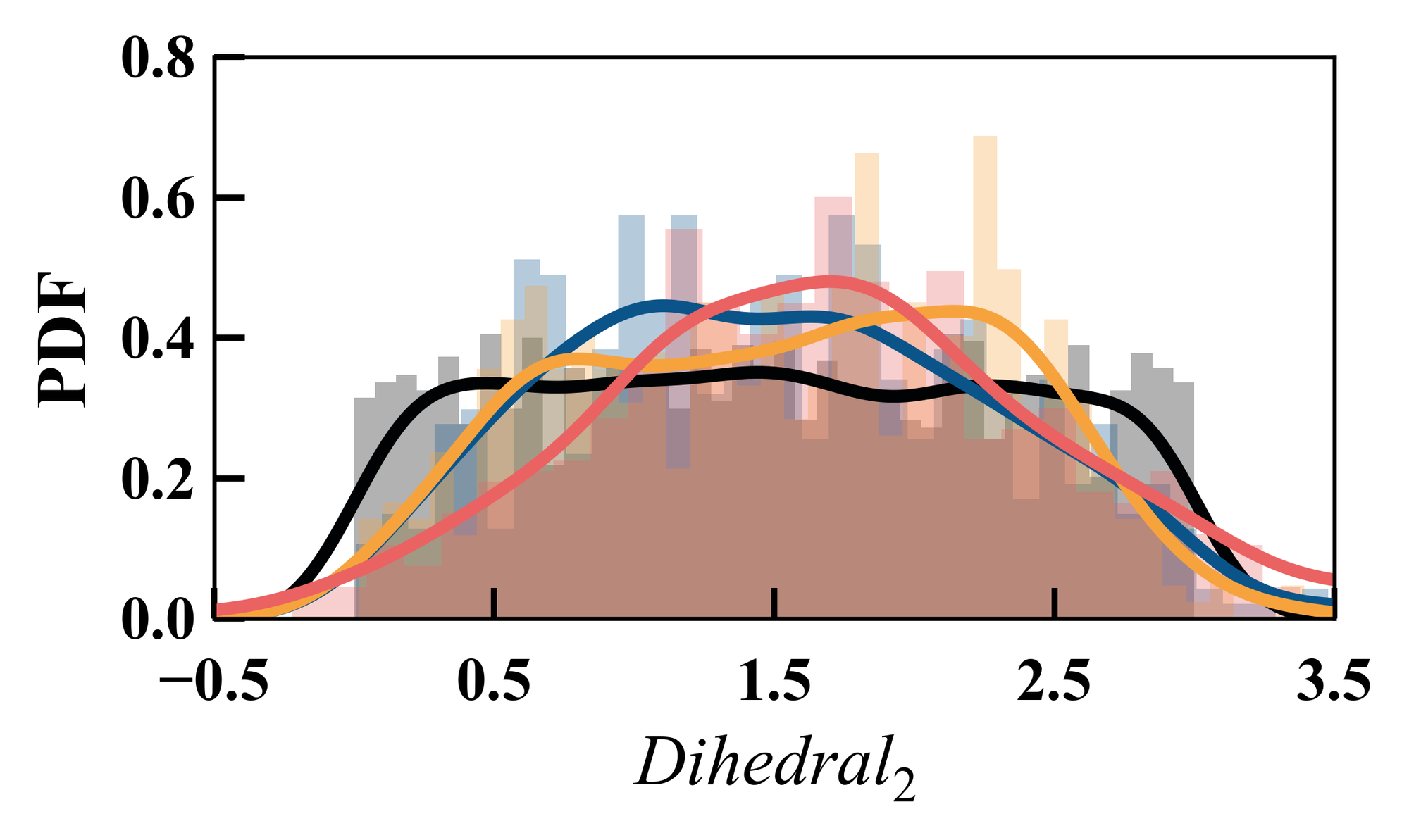}
\end{subfigure}
\hfill
\begin{subfigure}[b]{0.32\textwidth}
    \centering
    \includegraphics[width=\textwidth]{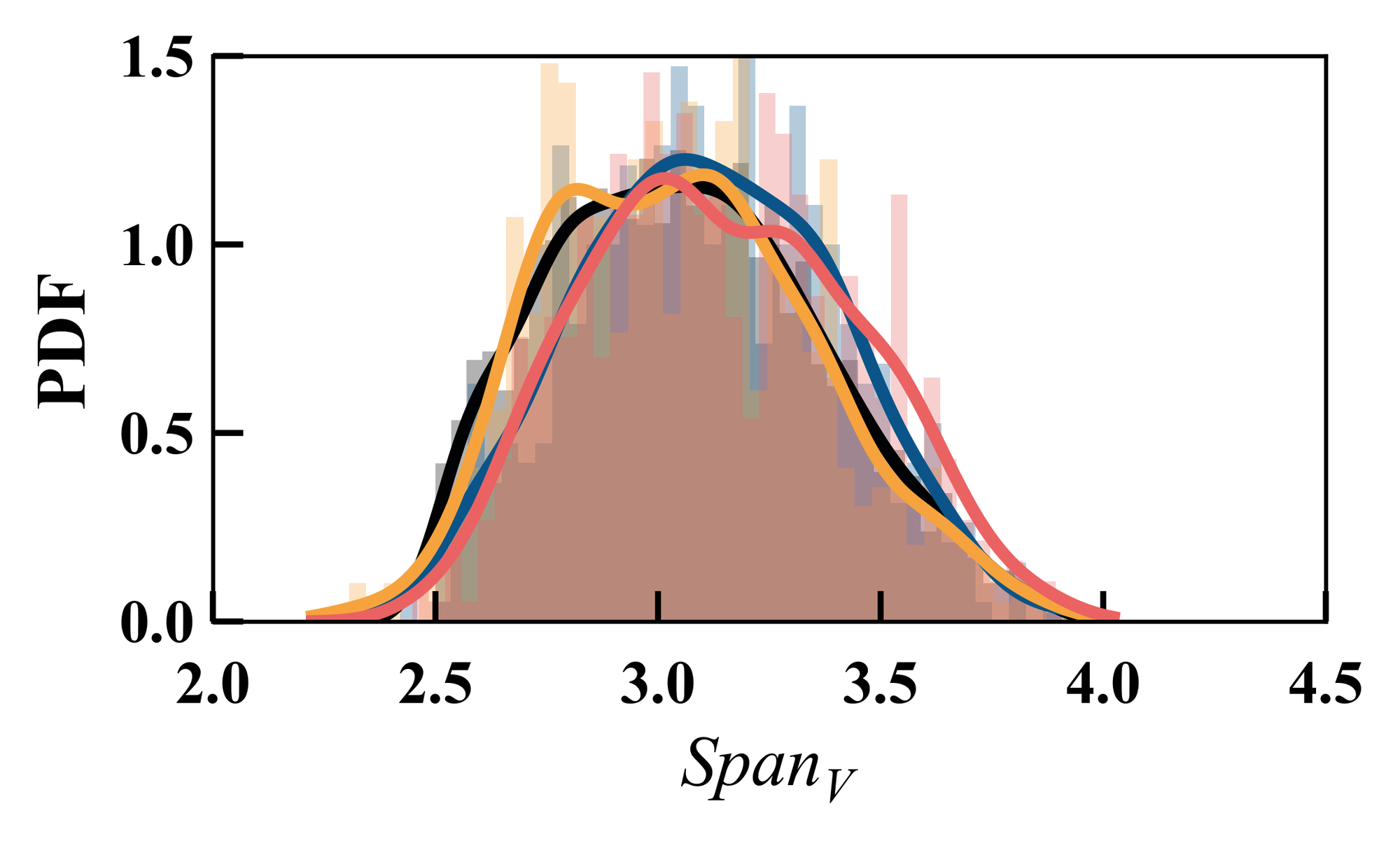}
\end{subfigure}

\vspace{0.5em} 

\begin{subfigure}[b]{0.32\textwidth}
    \centering
    \includegraphics[width=\textwidth]{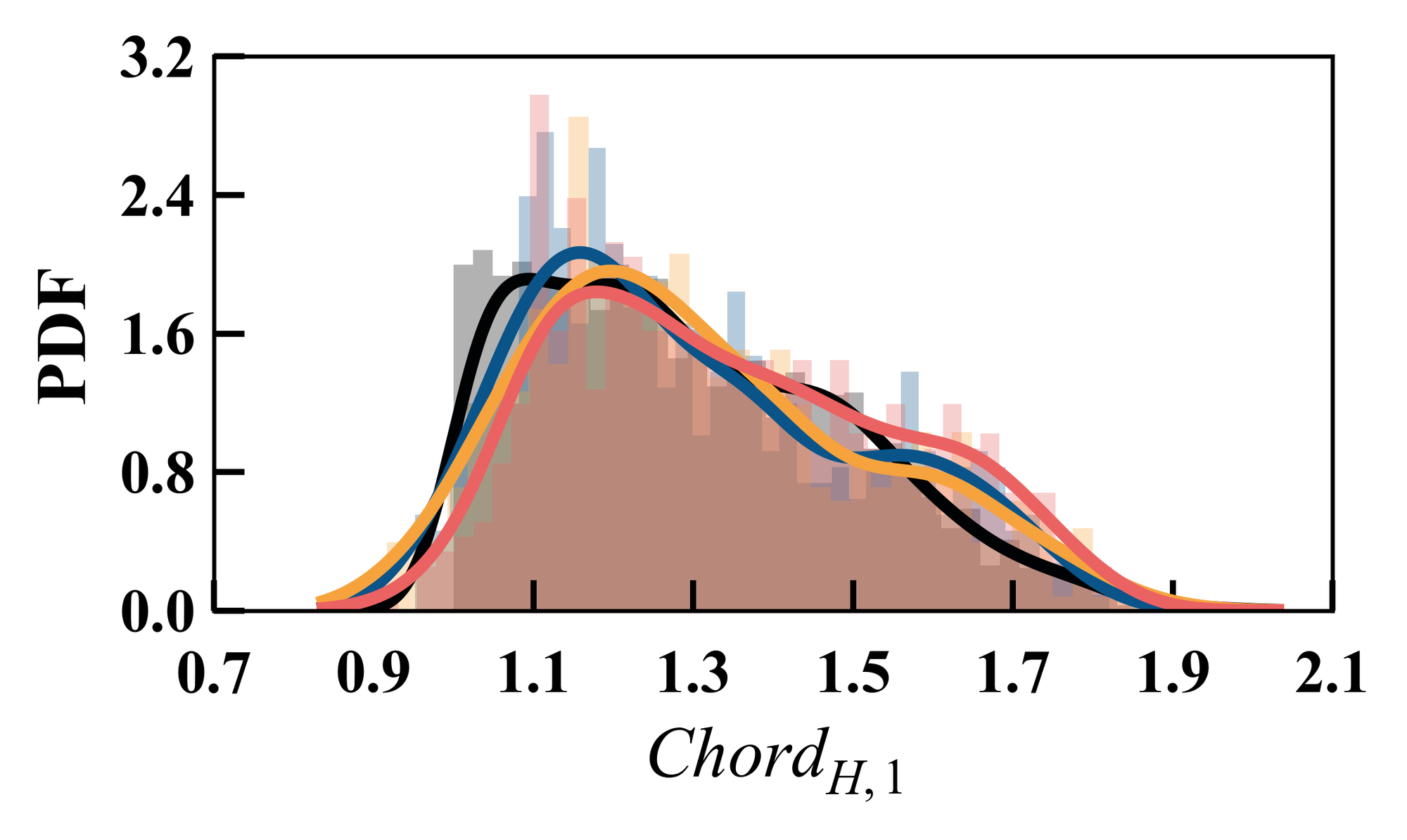}
\end{subfigure}
\hfill
\begin{subfigure}[b]{0.32\textwidth}
    \centering
    \includegraphics[width=\textwidth]{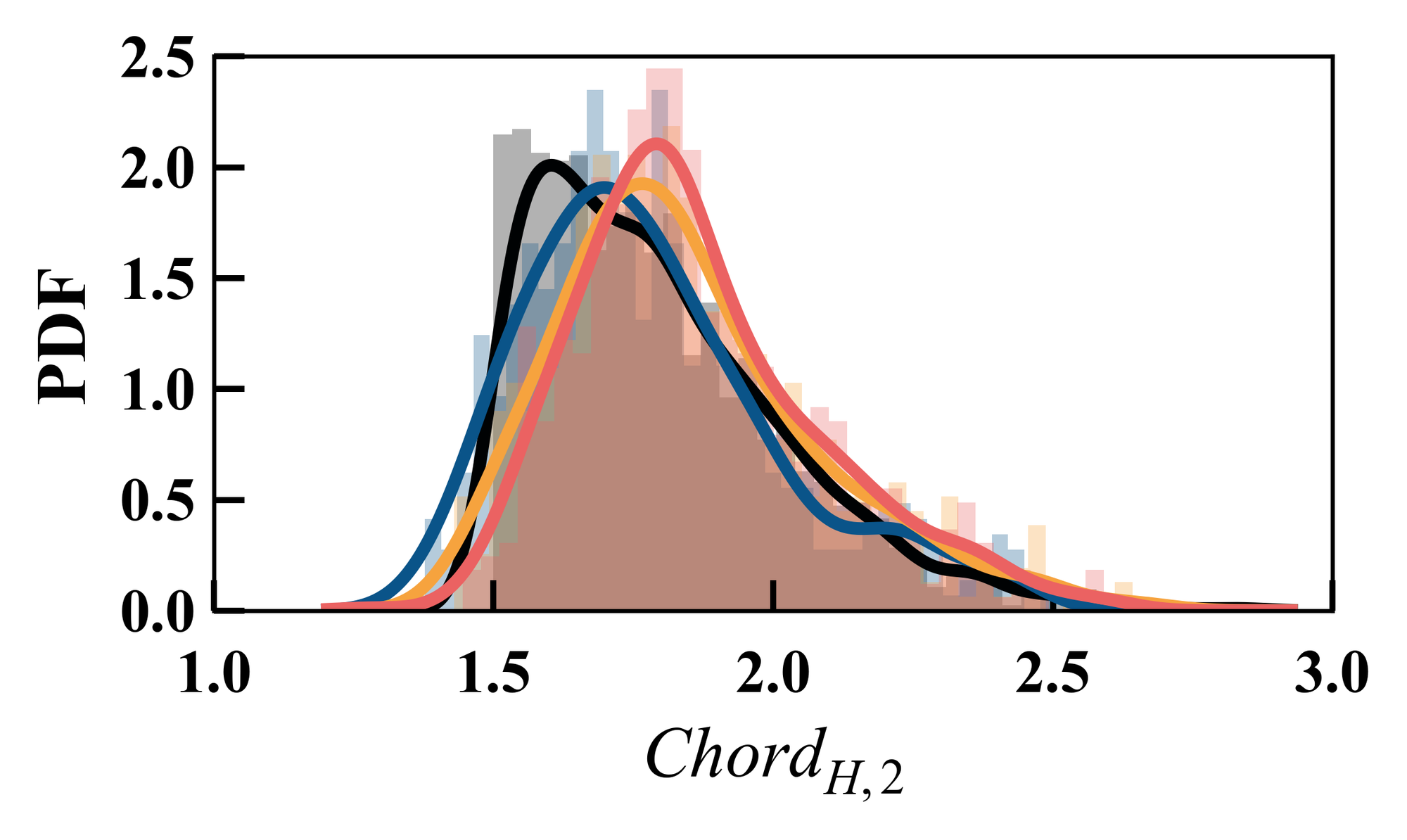}
\end{subfigure}
\hfill
\begin{subfigure}[b]{0.32\textwidth}
    \centering
    \includegraphics[width=\textwidth]{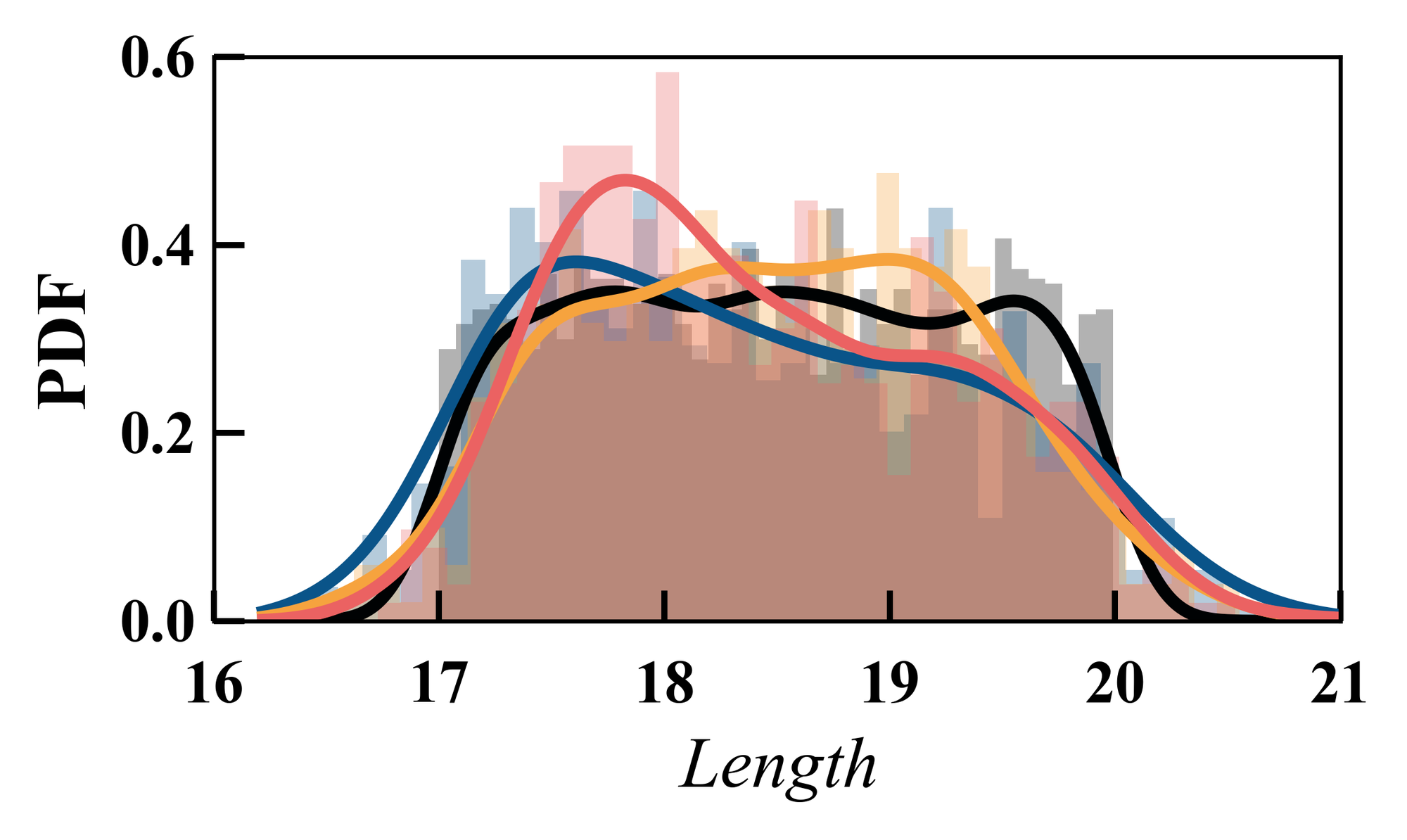}
\end{subfigure}

\caption{Distribution of generated geometric parameters for different models under Cond-A.}
\label{fig:feature-mean}
\end{figure}

Figure~\ref{fig:feature-mean} presents the distribution of selected geometric parameters for the generated samples under Cond-A. The results indicate that geometric diversity is effectively maintained even with the introduction of geometry-guided enhancement or aerodynamic performance-guided enhancement strategies. This confirms that the proposed condition-enhanced training mechanism does not induce the mode collapse issue commonly encountered in GANs, retaining the exploration capability of the design space while improving generation accuracy. Note that the training data shows the best diversity since it includes samples under multiple conditions, rather than a specific condition vector (i.e., Cond-A here).

\subsubsection{Output Filtering}
\label{sec:full_evaluation}

\begin{figure}[htbp]
\centering

\begin{subfigure}[b]{0.3\textwidth}
    \centering
    \includegraphics[width=\textwidth]{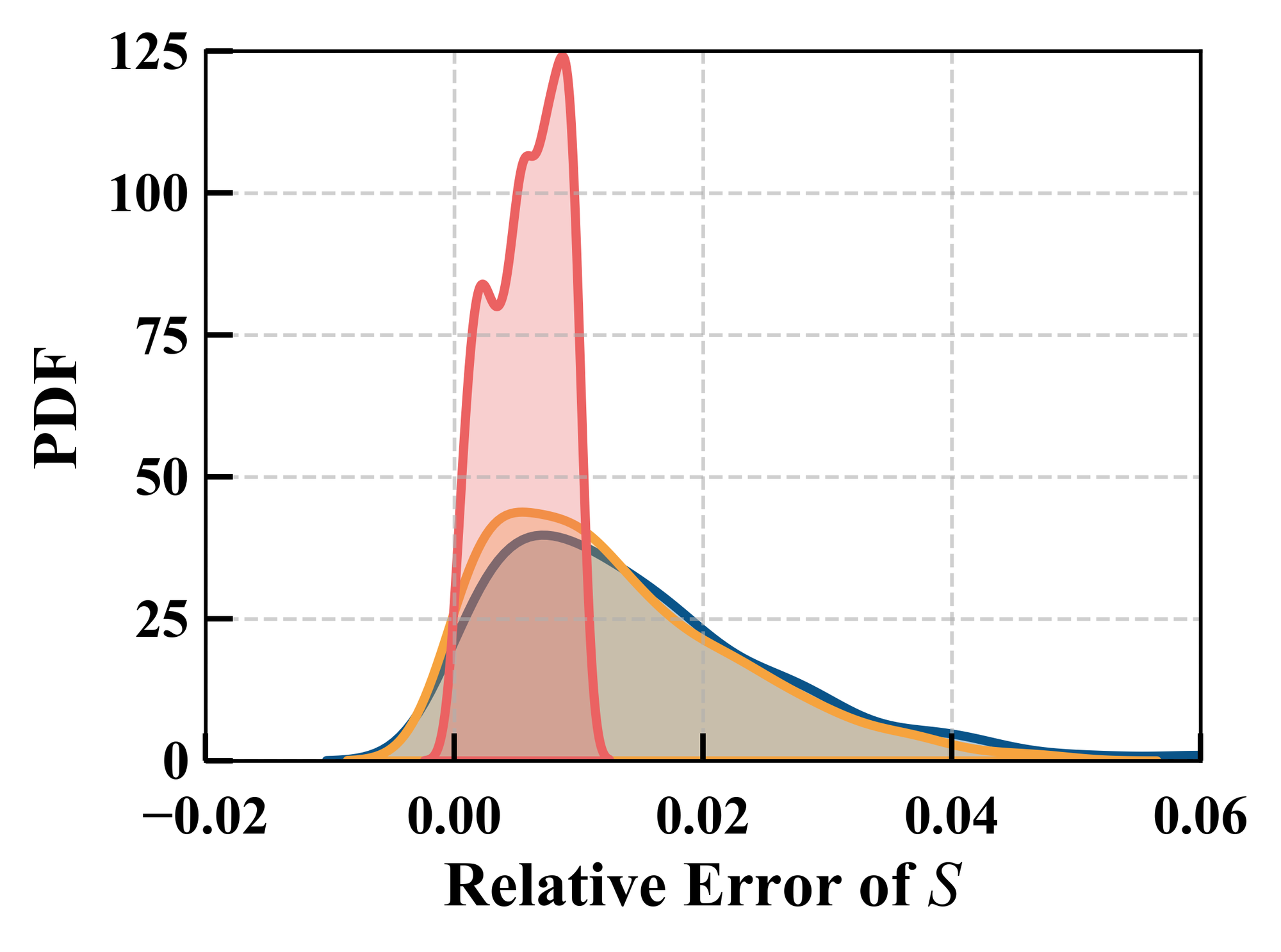}
    \caption{$S$}
    \label{fig:sura}
\end{subfigure}
\hfill
\begin{subfigure}[b]{0.3\textwidth}
    \centering
    \includegraphics[width=\textwidth]{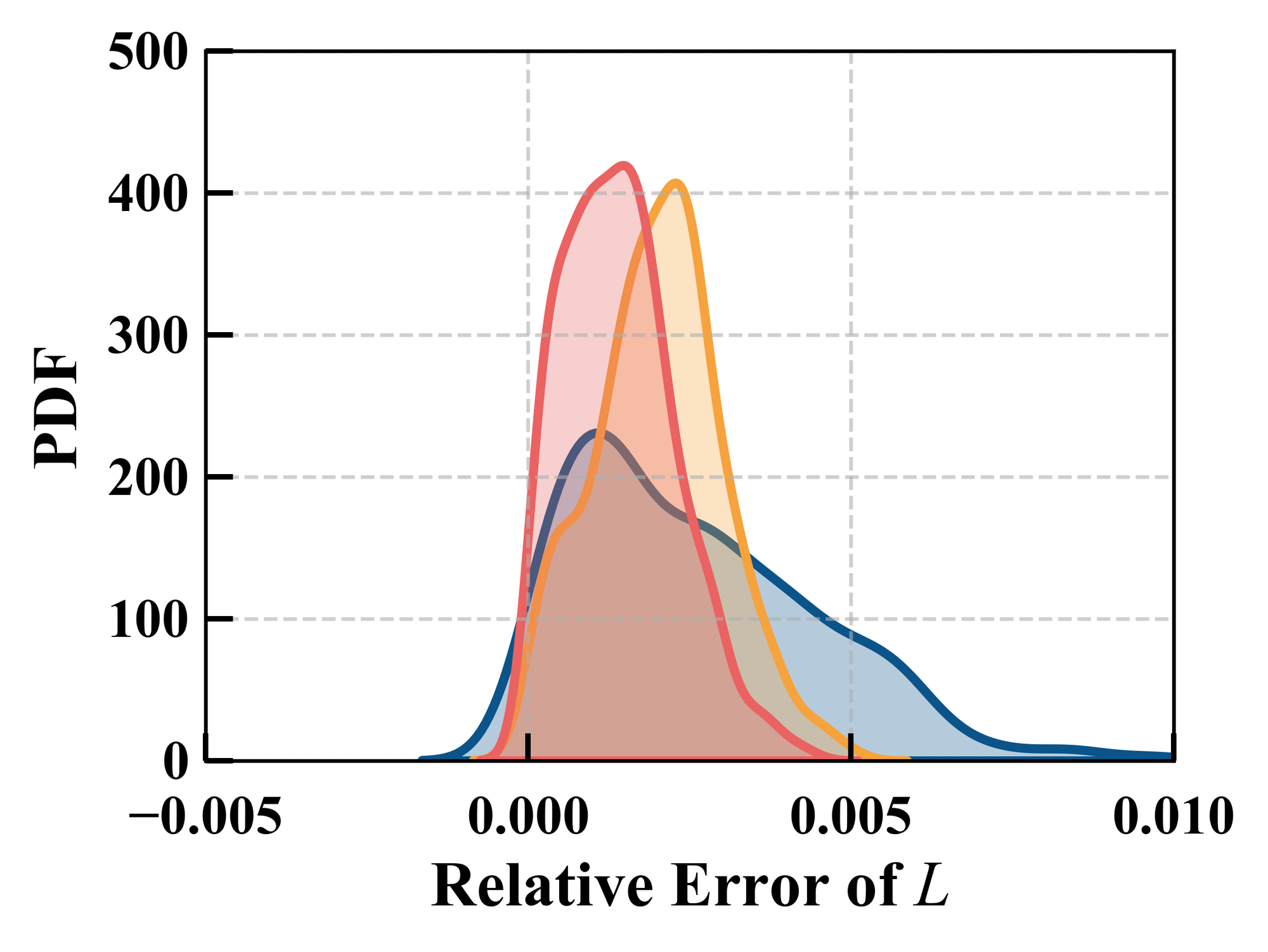}
    \caption{$L$}
    \label{fig:surb}
\end{subfigure}
\hfill
\begin{subfigure}[b]{0.3\textwidth}
    \centering
    \includegraphics[width=\textwidth]{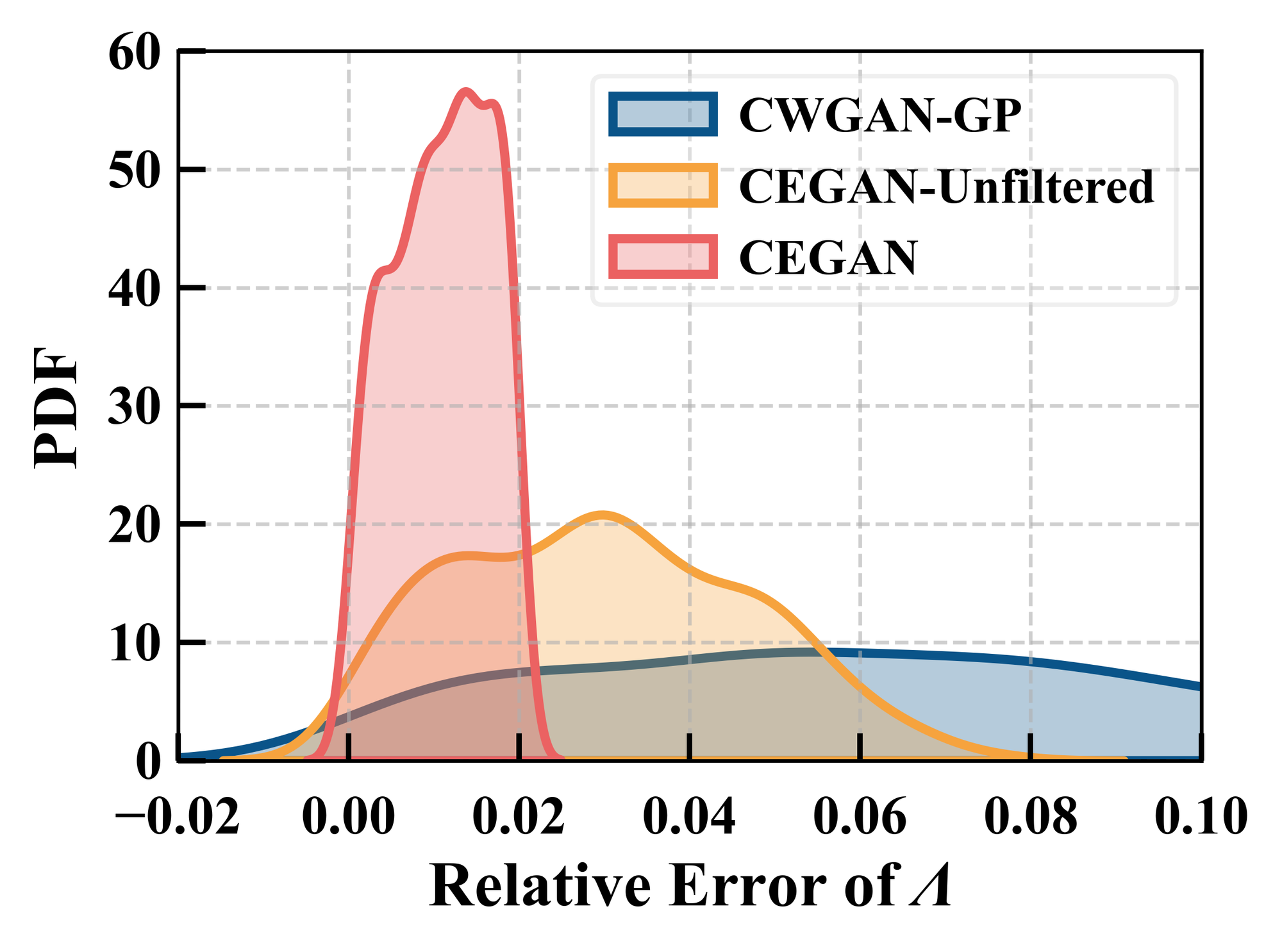}
    \caption{$\Lambda$}
    \label{fig:surc}
\end{subfigure}

\vspace{1em} 

\begin{subfigure}[b]{0.3\textwidth}
    \centering
    \includegraphics[width=\textwidth]{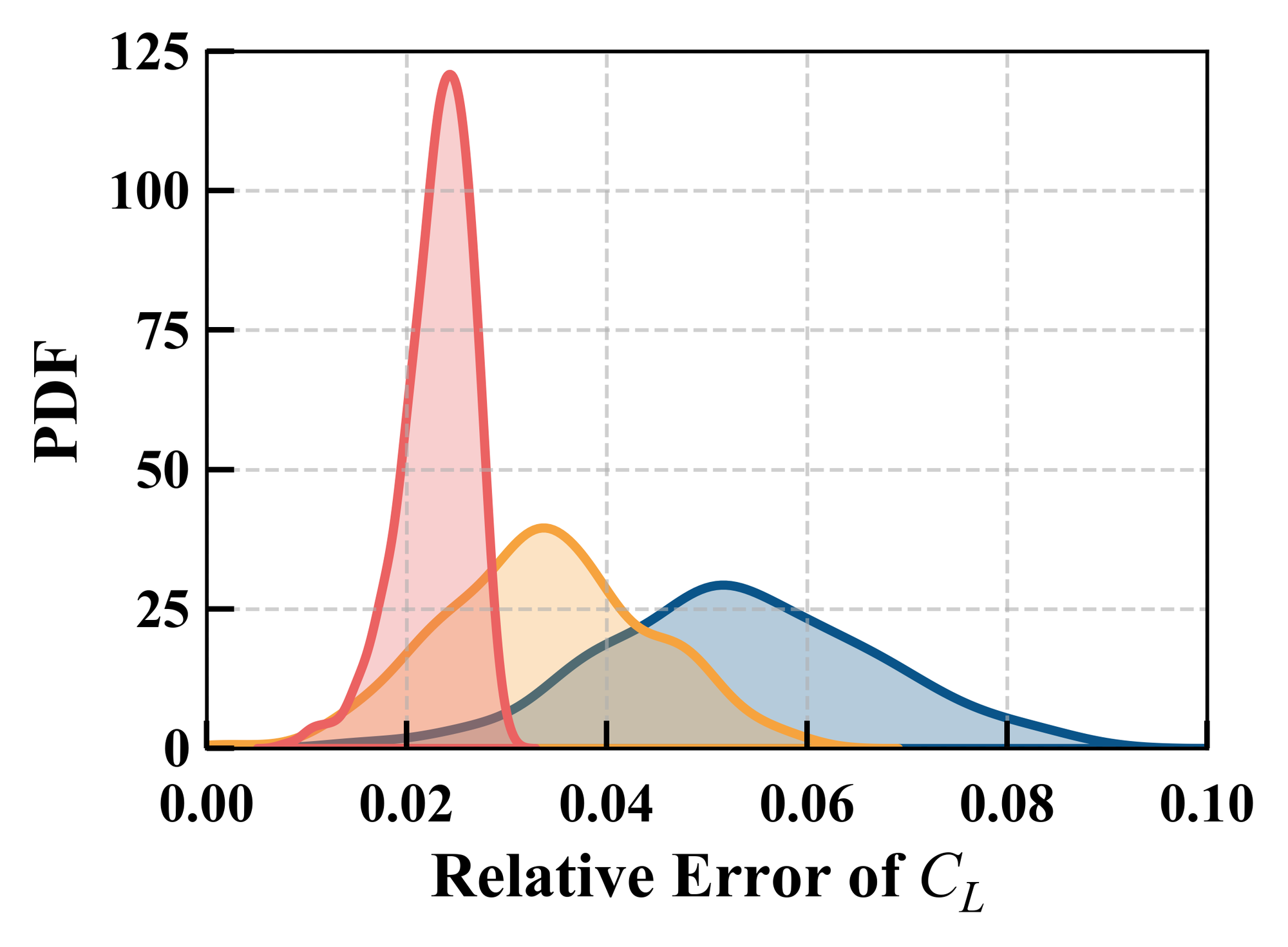}
    \caption{$C_L$}
    \label{fig:surd}
\end{subfigure}
\hfill
\begin{subfigure}[b]{0.3\textwidth}
    \centering
    \includegraphics[width=\textwidth]{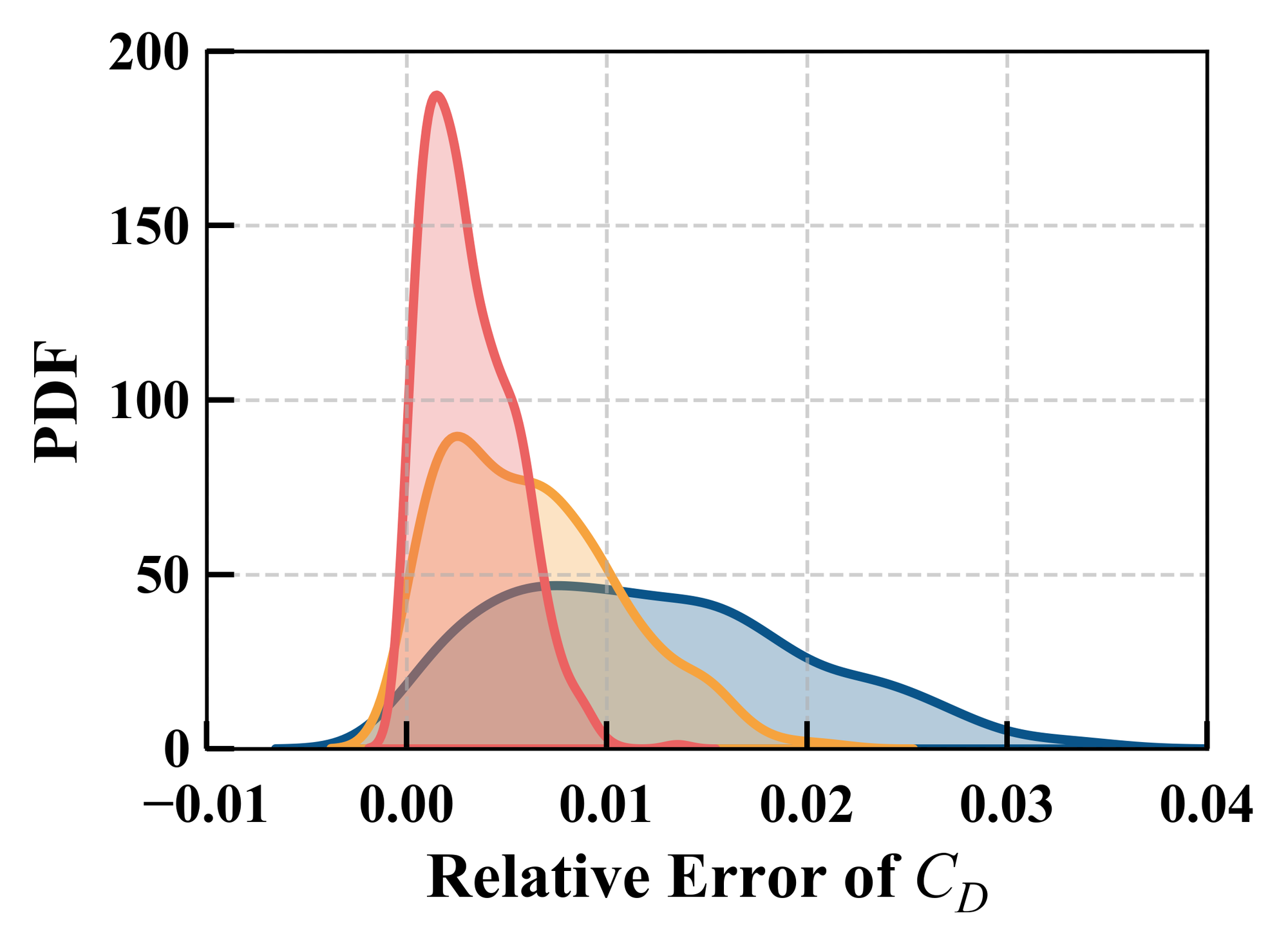}
    \caption{$C_D$}
    \label{fig:sure}
\end{subfigure}
\hfill
\begin{subfigure}[b]{0.3\textwidth}
    \centering
    \includegraphics[width=\textwidth]{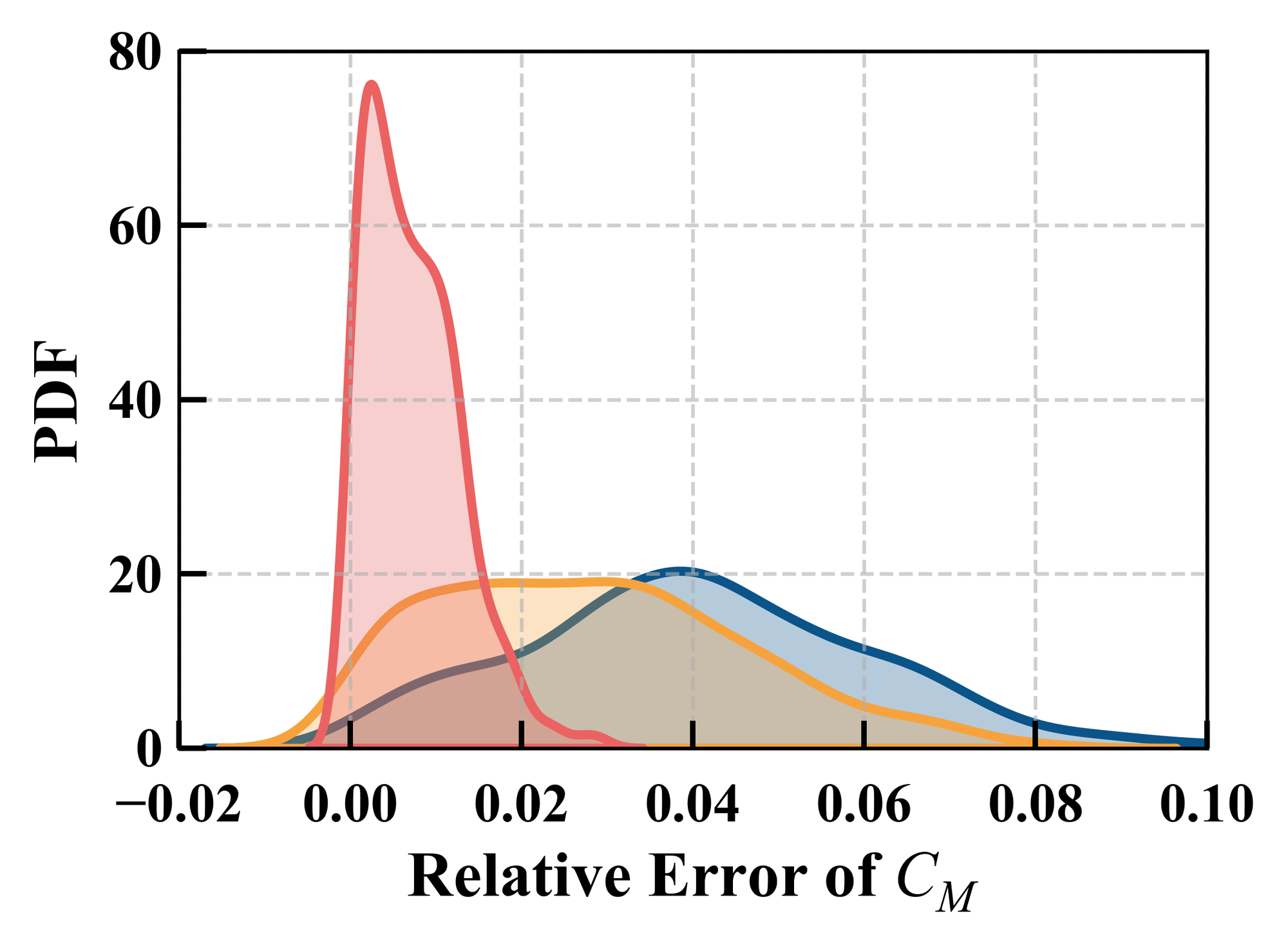}
    \caption{$C_M$}
    \label{fig:surf}
\end{subfigure}

\caption{Generation accuracy under extreme design conditions (Cond-C).}
\label{fig:CEGAN框架完整效果}
\end{figure}

In this section, both enhancement strategies are jointly introduced, and the performance evaluation of the complete CEGAN model is conducted. Additionally, the effect of output filtering is also compared. Validation focuses on the extreme combination of conditions in Cond-C. Figure~\ref{fig:CEGAN框架完整效果} compares the accuracy of model generation. It can be observed that, relative to the CWGAN-GP model, the CEGAN model achieves effective improvements in accuracy across all aspects, even without the output filtering module. Furthermore, the complete CEGAN framework yields generation results with higher accuracy, compared to the one without the output filtering module. Similar improvement is also observed for Cond-B, while the detailed results of which are provided in Appendix~C for brevity of the main text.

\begin{figure}[htbp]
\centering

\begin{subfigure}[b]{0.3\textwidth}
    \centering
    \includegraphics[width=\textwidth]{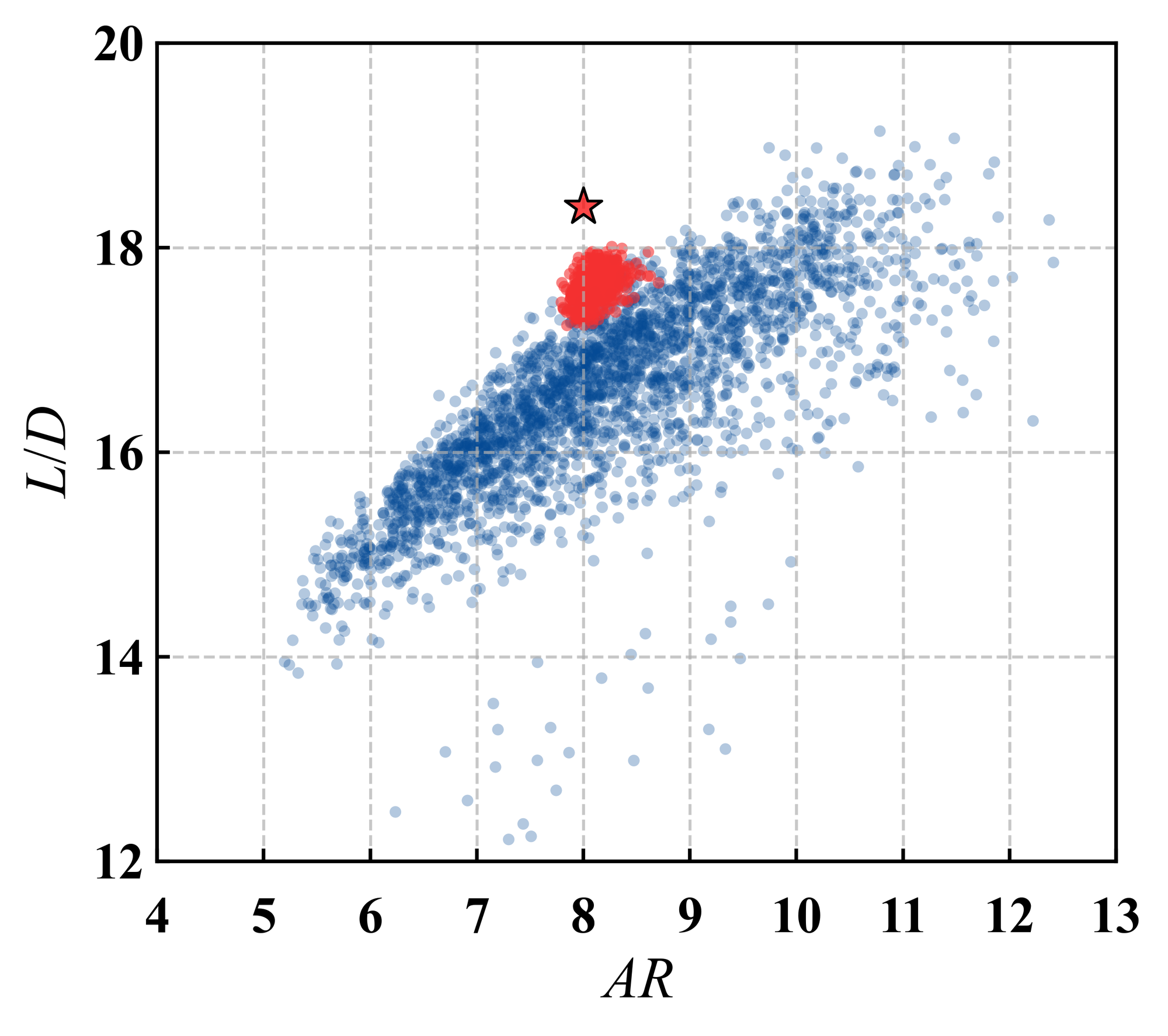}
    \caption{}
    \label{fig:sandiana}
\end{subfigure}
\hfill
\begin{subfigure}[b]{0.3\textwidth}
    \centering
    \includegraphics[width=\textwidth]{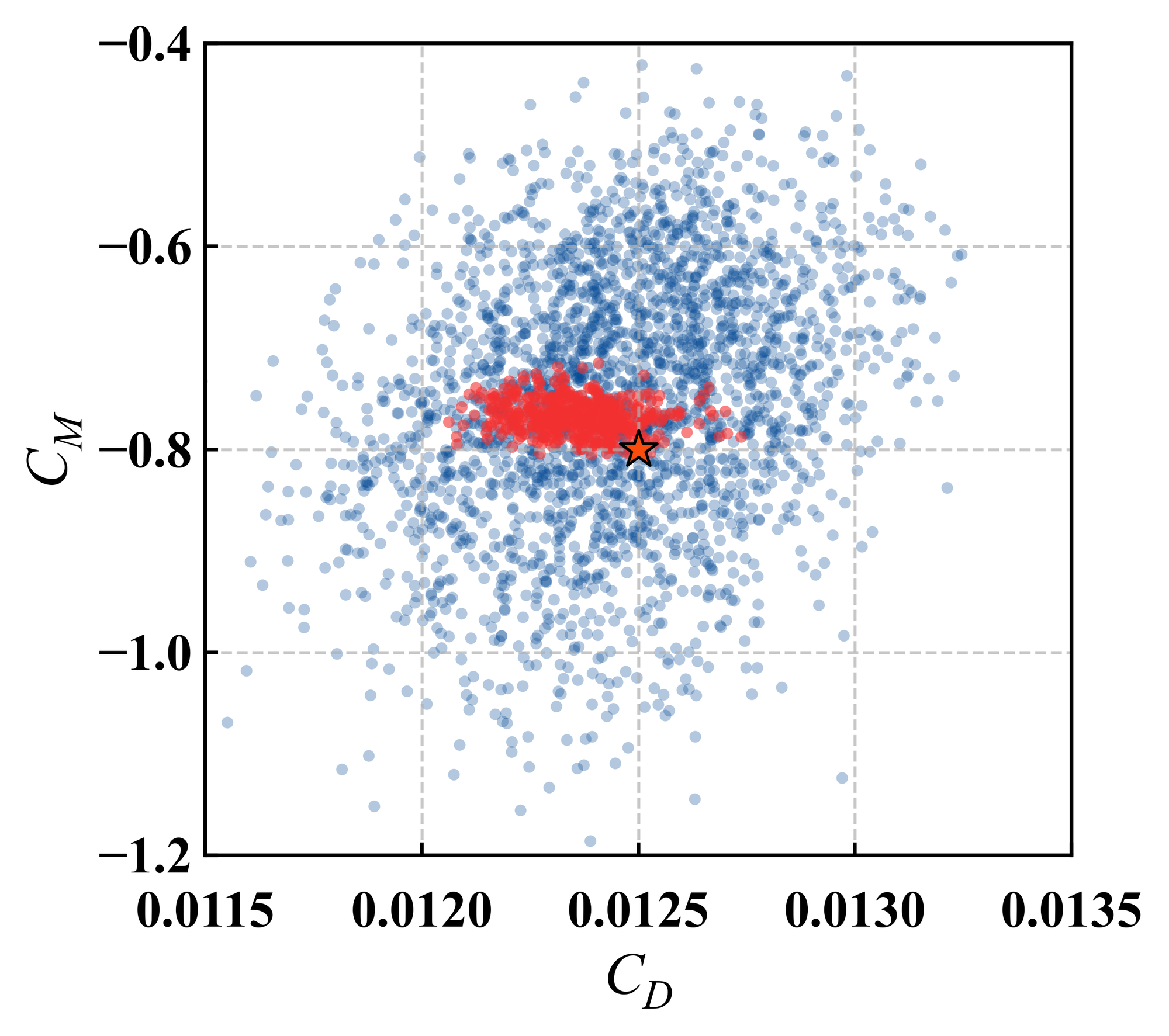}
    \caption{}
    \label{fig:sandianb}
\end{subfigure}
\hfill
\begin{subfigure}[b]{0.3\textwidth}
    \centering
    \includegraphics[width=\textwidth]{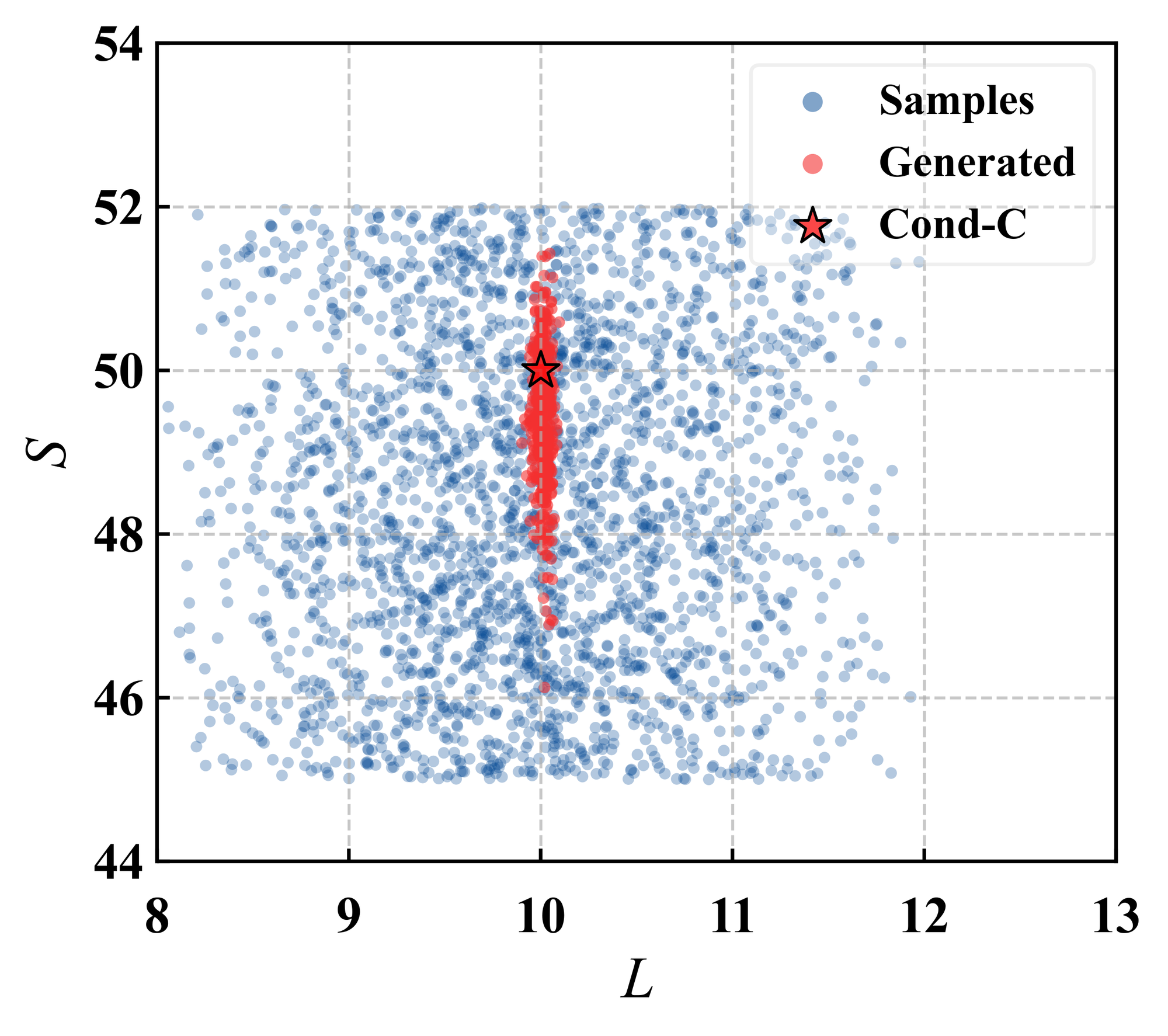}
    \caption{}
    \label{fig:sandianc}
\end{subfigure}

\vspace{1em} 

\begin{subfigure}[b]{0.3\textwidth}
    \centering
    \includegraphics[width=\textwidth]{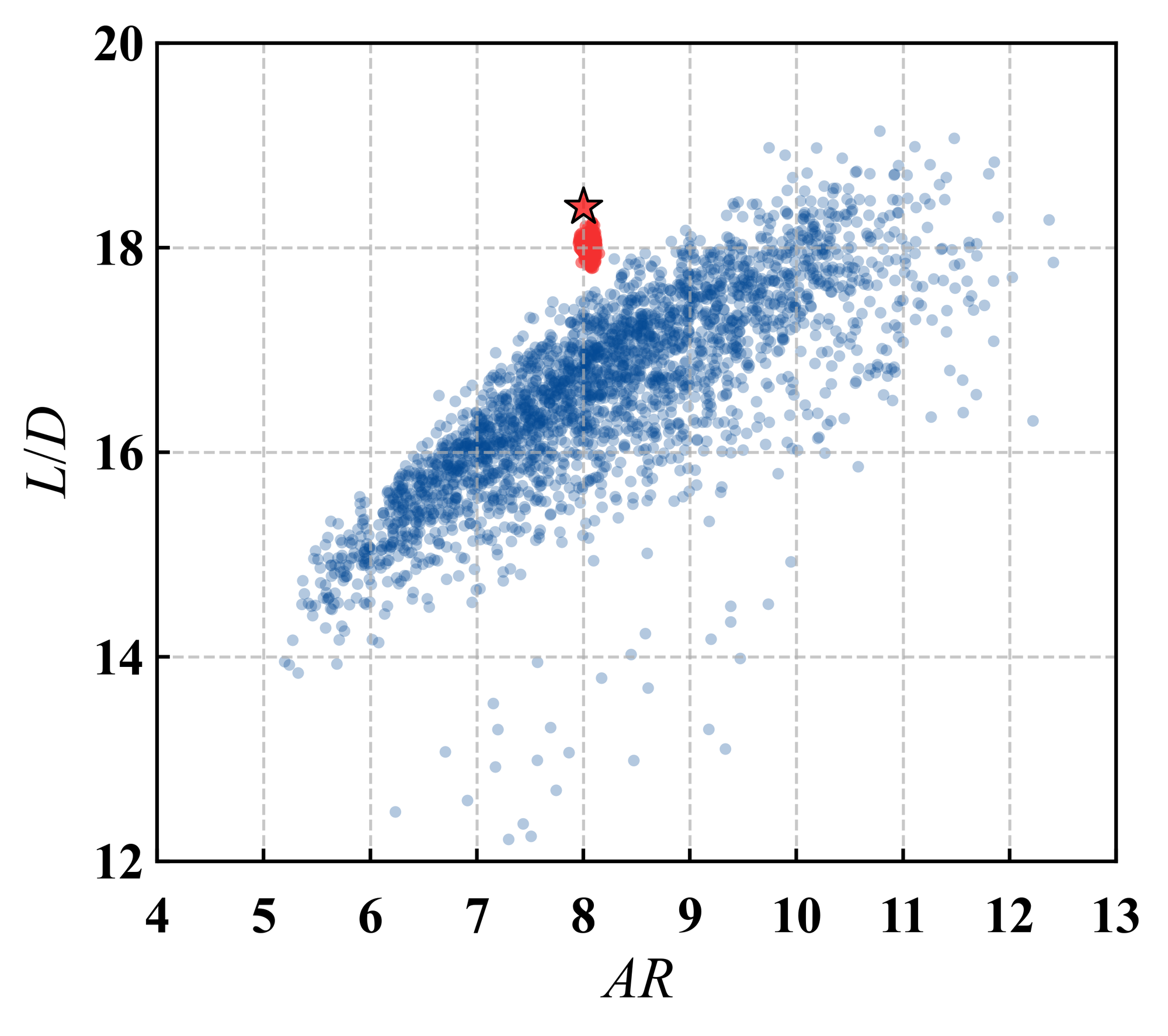}
    \caption{}
    \label{fig:sandiand}
\end{subfigure}
\hfill
\begin{subfigure}[b]{0.3\textwidth}
    \centering
    \includegraphics[width=\textwidth]{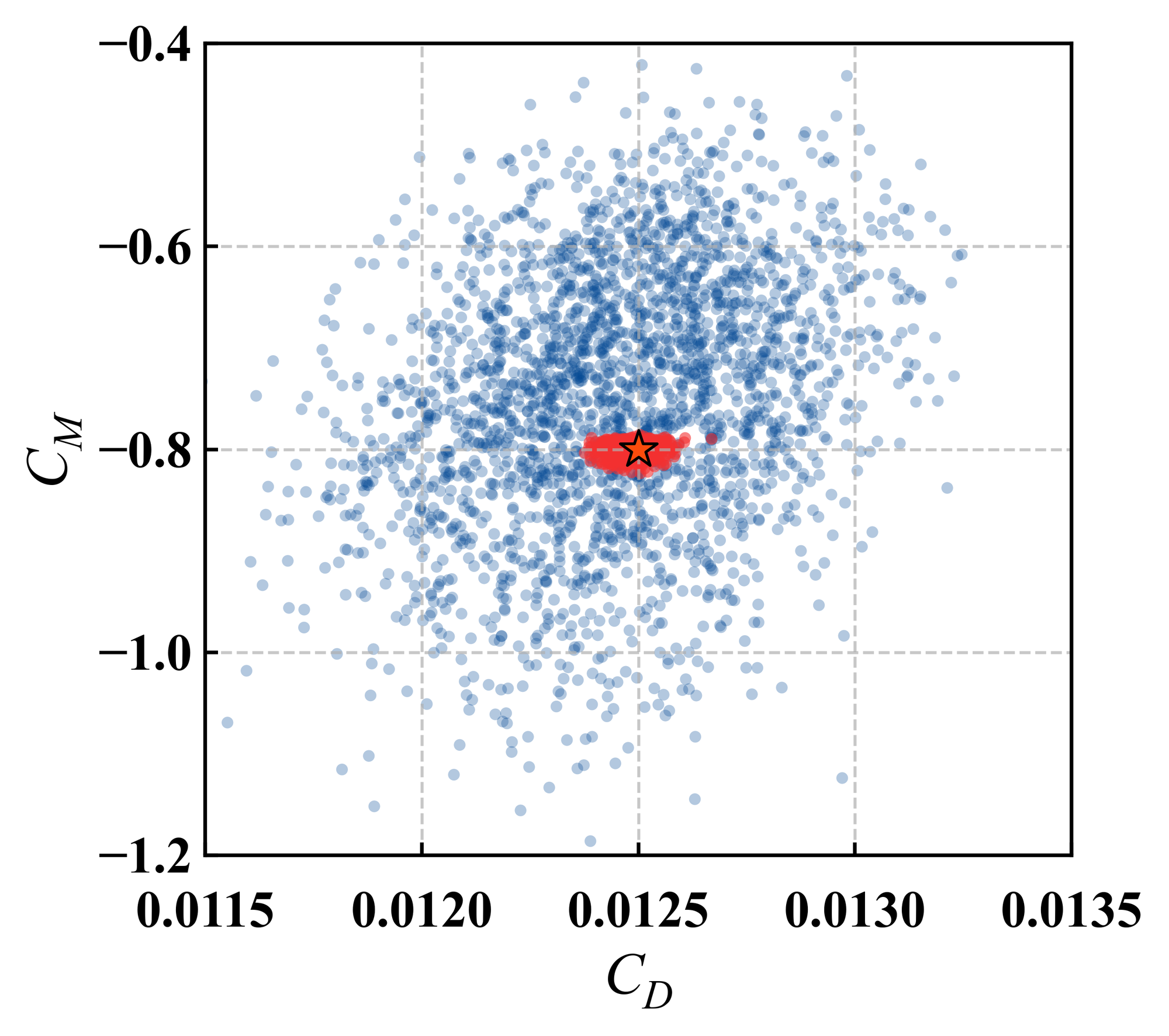}
    \caption{}
    \label{fig:sandiane}
\end{subfigure}
\hfill
\begin{subfigure}[b]{0.3\textwidth}
    \centering
    \includegraphics[width=\textwidth]{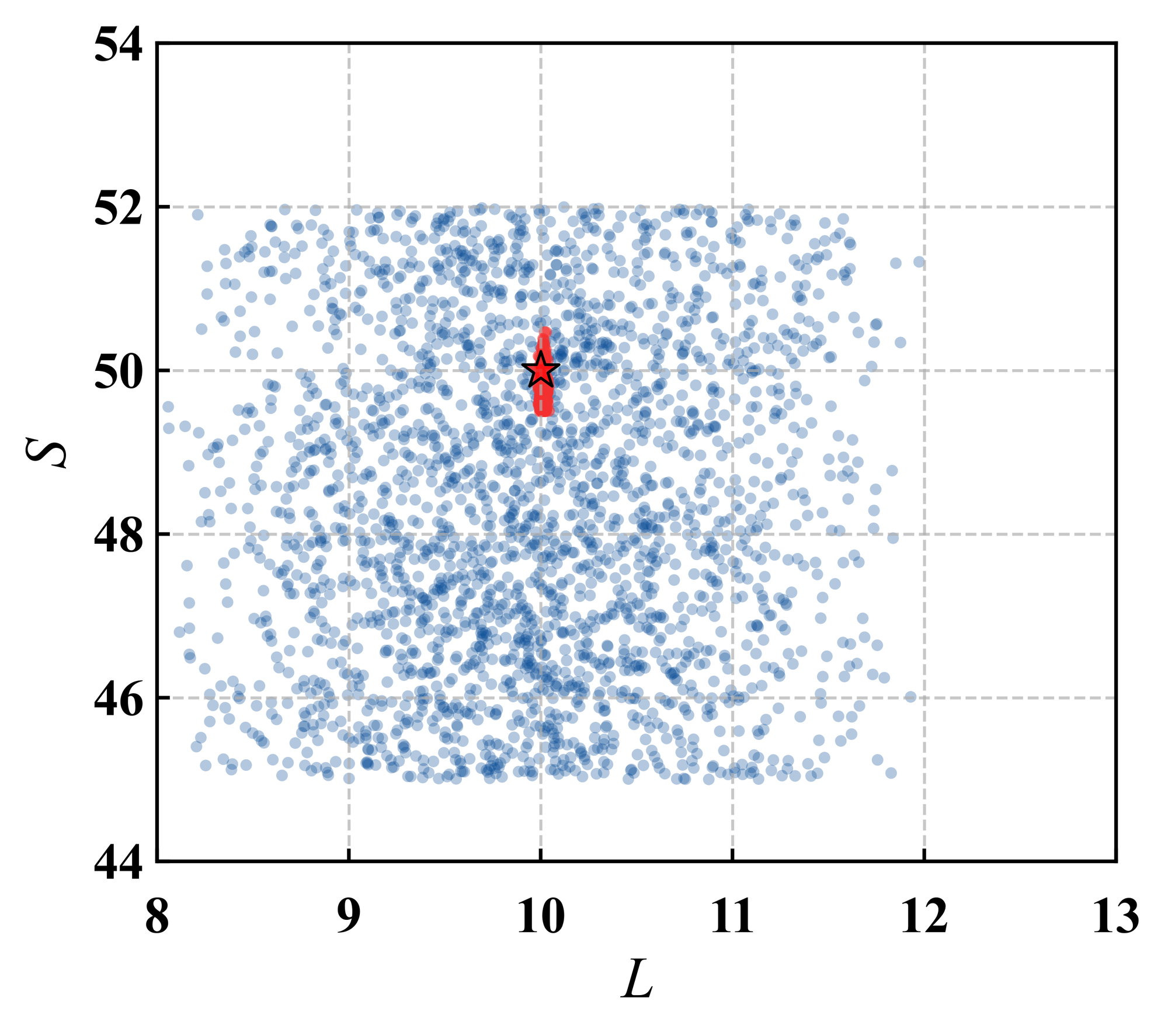}
    \caption{}
    \label{fig:sandianf}
\end{subfigure}

\caption{Generation results of CWGAN-GP vs. CEGAN in different condition spaces under extreme design conditions (Cond-C). (a)--(c) CWGAN-GP; (d)--(f) CEGAN.}
\label{fig:极限散点对比}
\end{figure}

To provide a more intuitive comparison of the conditional generation capabilities of different models, Figure~\ref{fig:极限散点对比} presents the generation results of CEGAN and CWGAN-GP for extreme combinations of conditions across different condition spaces. It can be observed that Cond-C may appear as interpolation points when examined in the individual aerodynamic space or geometric space alone. However, in the actual design process, this combination of conditions represents an extreme case because it lies outside the joint aerodynamic-geometric condition space, as shown in Figure~\ref{fig:极限散点对比}a and Figure~\ref{fig:极限散点对比}d. Results demonstrate that CEGAN exhibits significantly better generation capability compared to CWGAN-GP for these extreme coupled conditions.

\begin{figure}[htbp]
\centering
\begin{subfigure}[b]{0.32\textwidth}
    \centering
    \includegraphics[width=\textwidth]{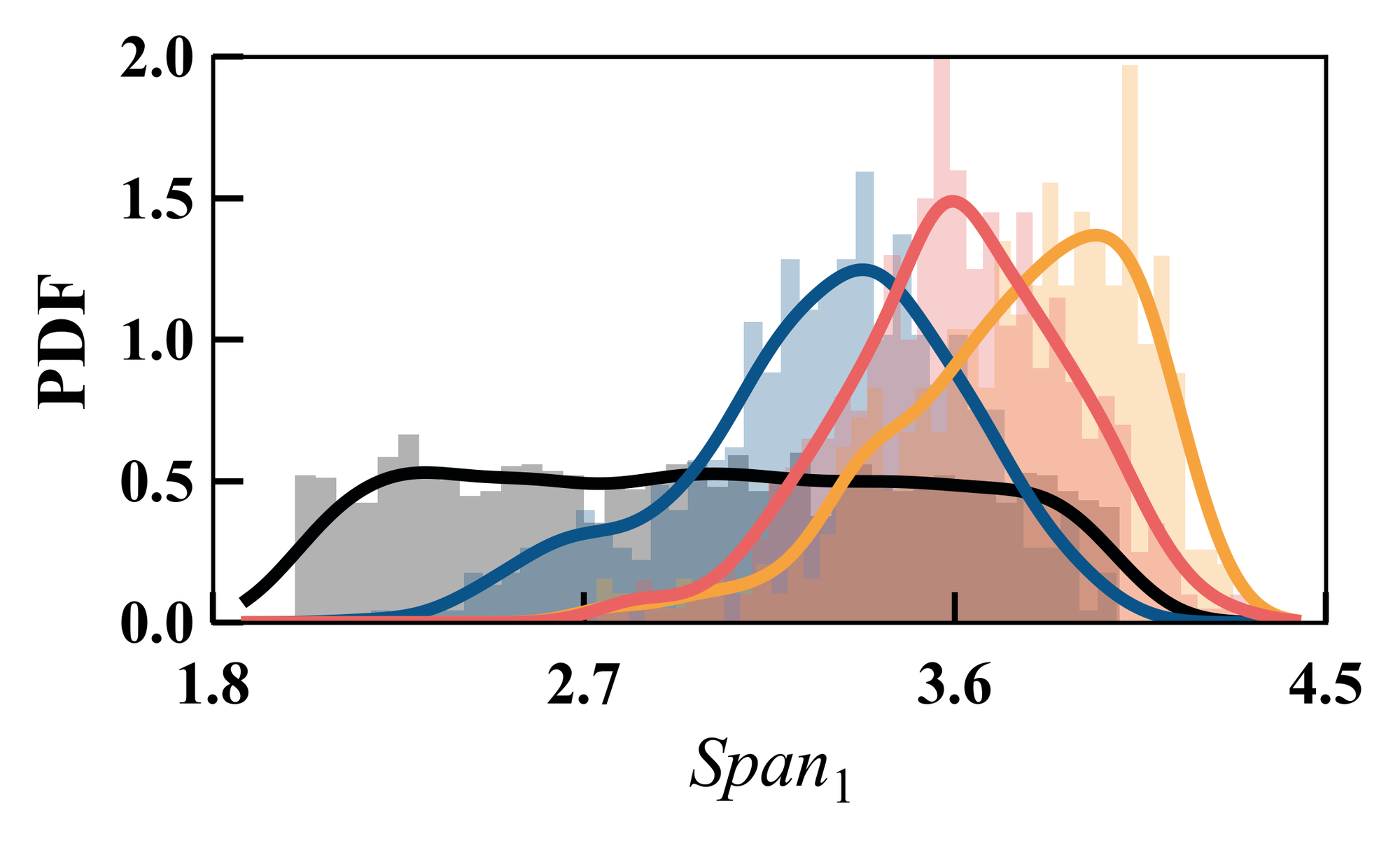}
\end{subfigure}
\hfill
\begin{subfigure}[b]{0.32\textwidth}
    \centering
    \includegraphics[width=\textwidth]{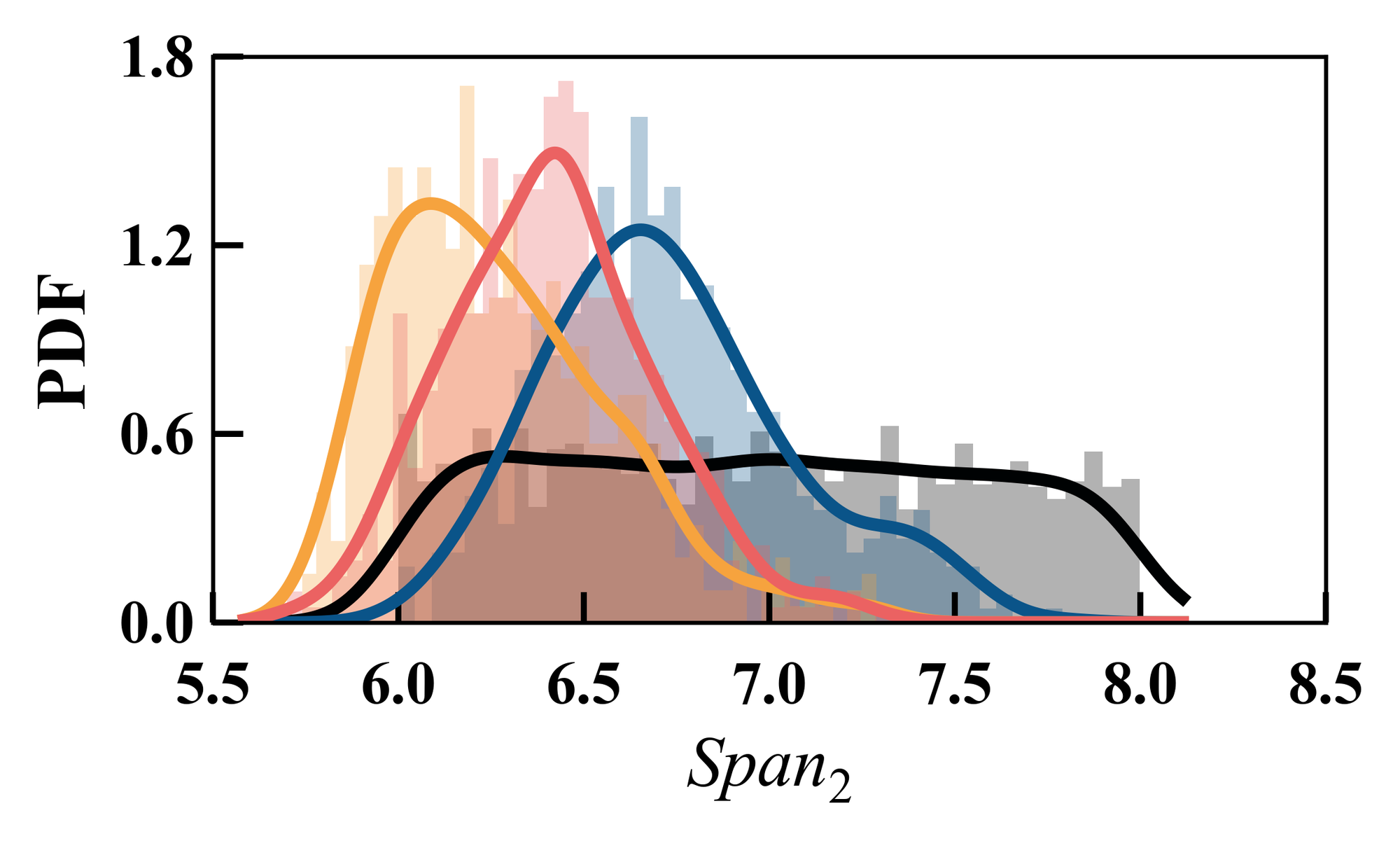}
\end{subfigure}
\hfill
\begin{subfigure}[b]{0.32\textwidth}
    \centering
    \includegraphics[width=\textwidth]{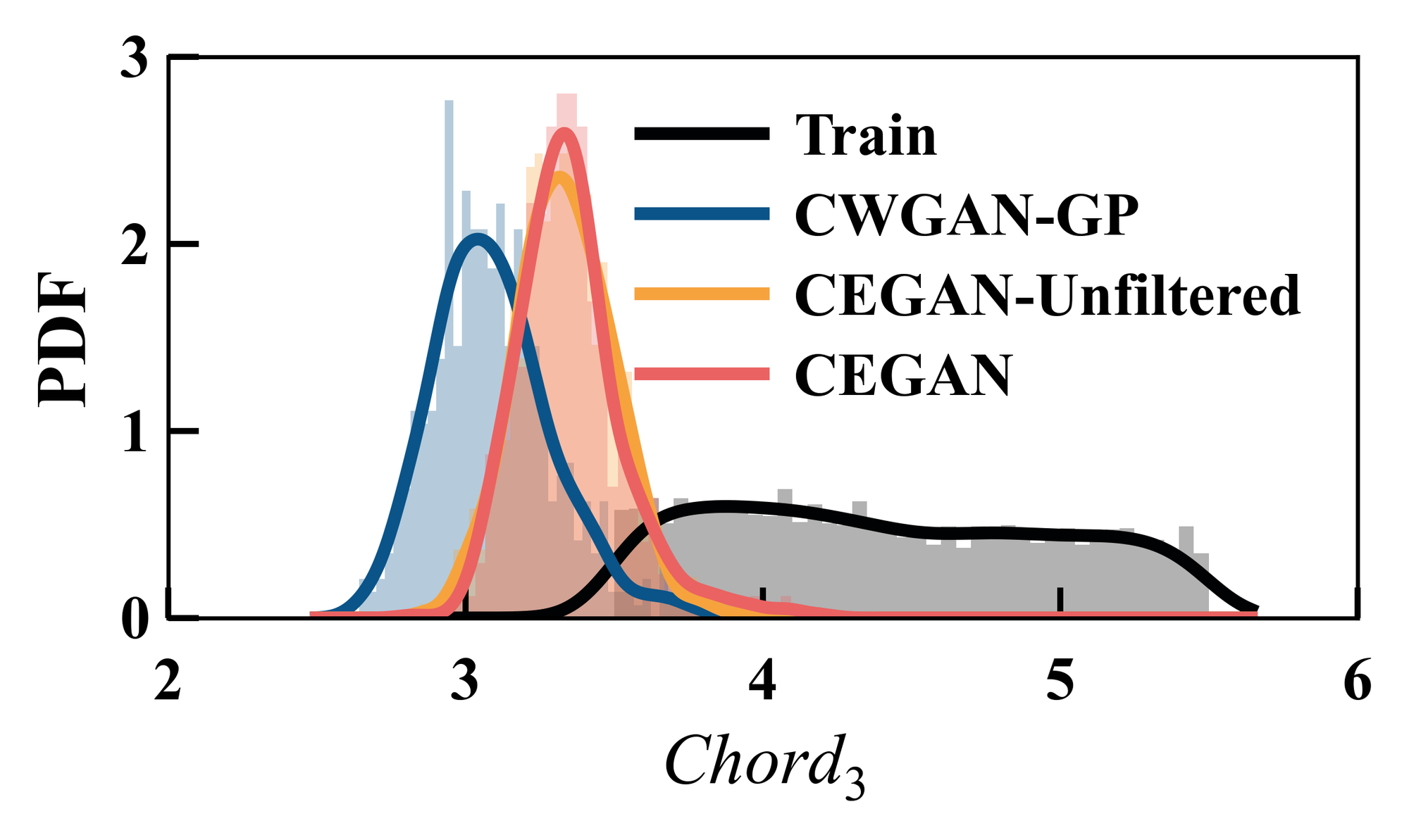}
\end{subfigure}

\vspace{0.5em} 

\begin{subfigure}[b]{0.32\textwidth}
    \centering
    \includegraphics[width=\textwidth]{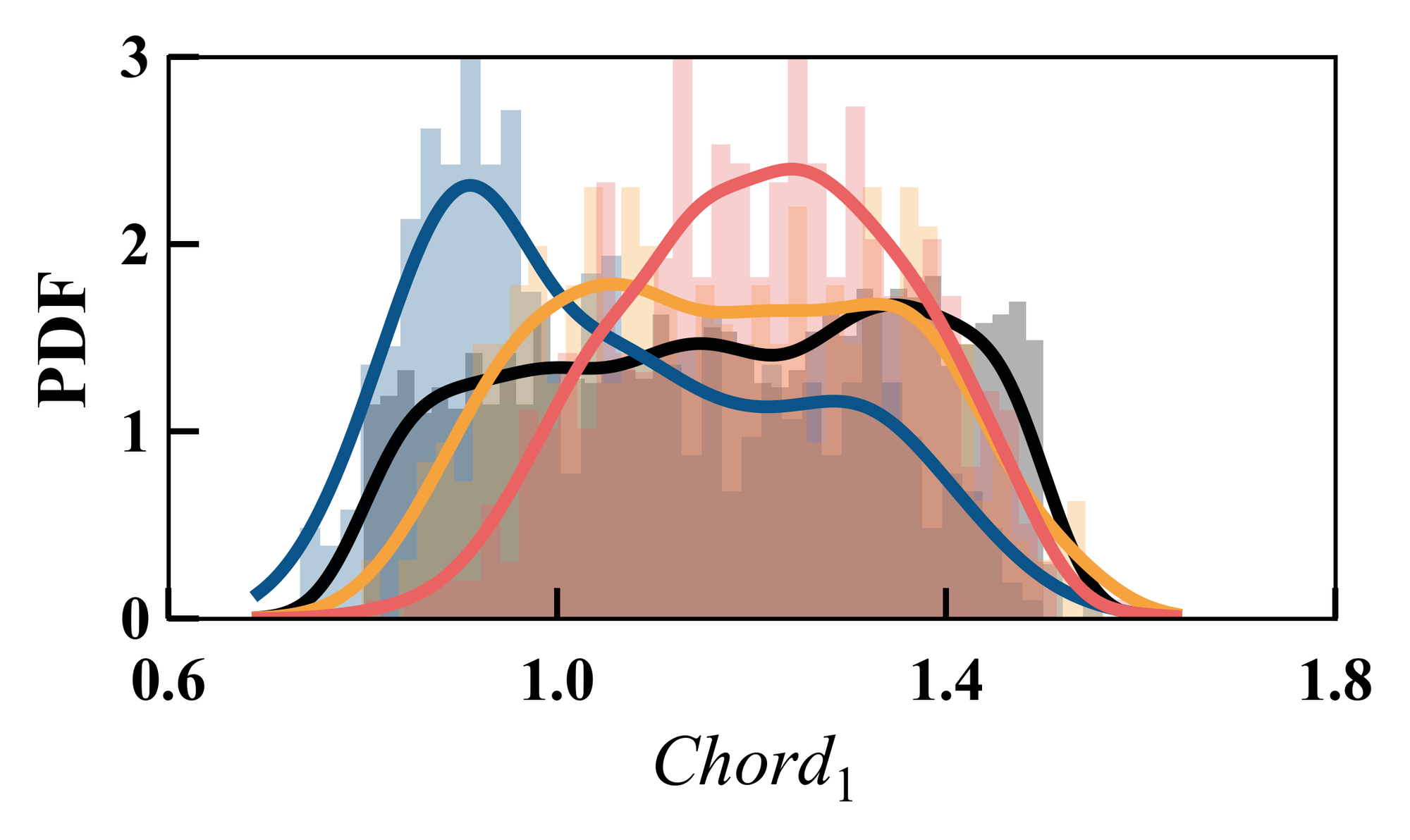}
\end{subfigure}
\hfill
\begin{subfigure}[b]{0.32\textwidth}
    \centering
    \includegraphics[width=\textwidth]{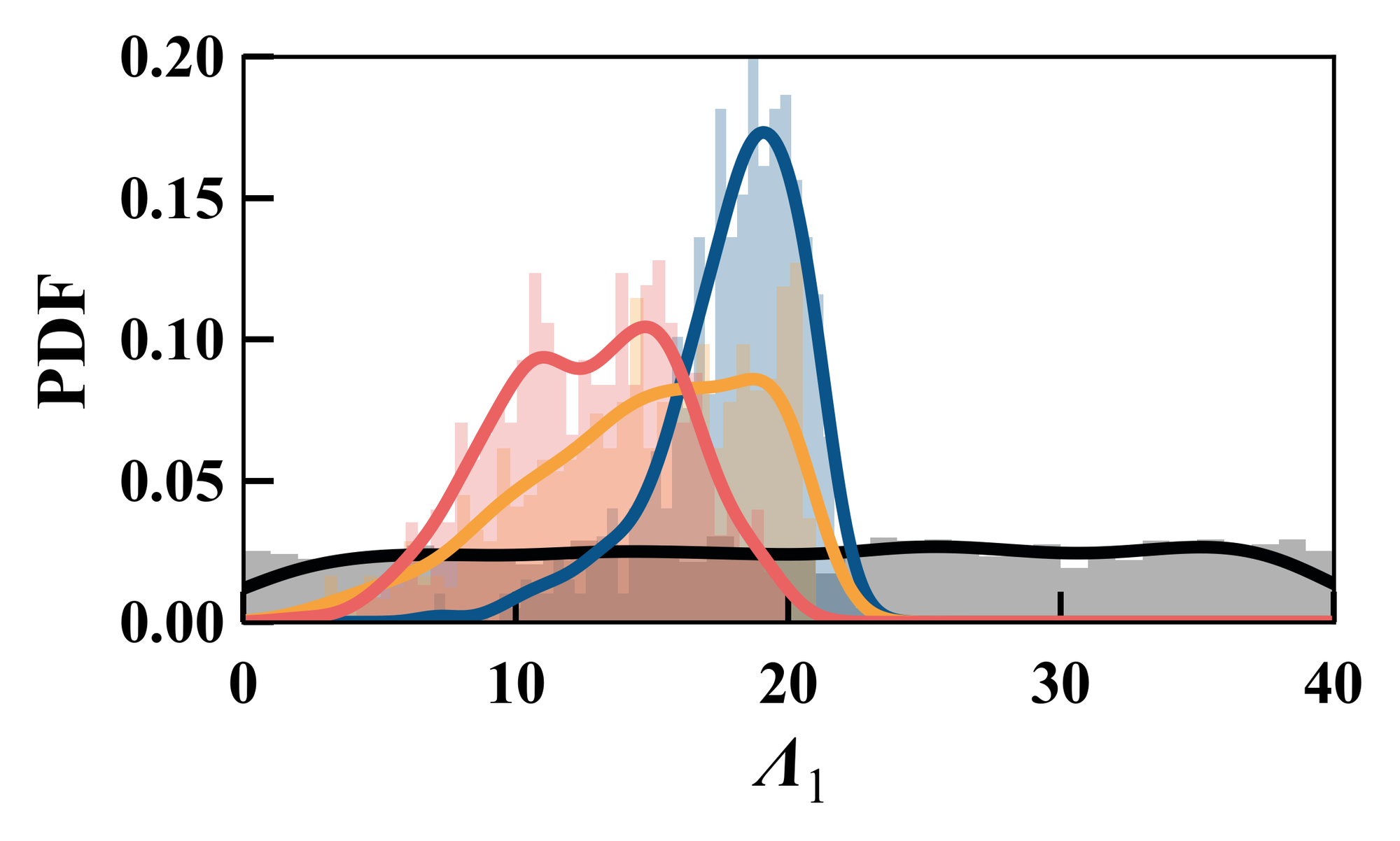}
\end{subfigure}
\hfill
\begin{subfigure}[b]{0.32\textwidth}
    \centering
    \includegraphics[width=\textwidth]{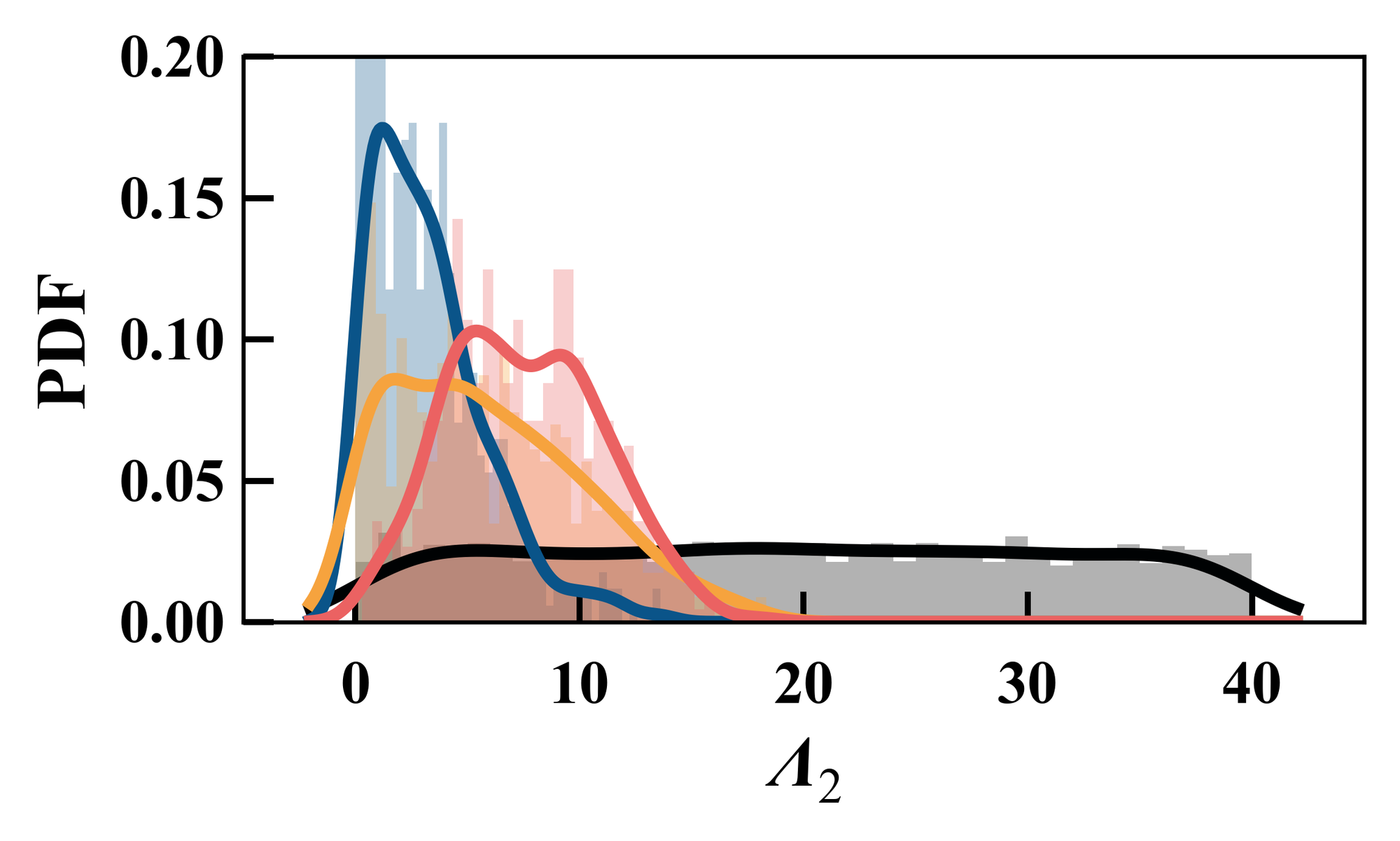}
\end{subfigure}

\vspace{0.5em} 

\begin{subfigure}[b]{0.32\textwidth}
    \centering
    \includegraphics[width=\textwidth]{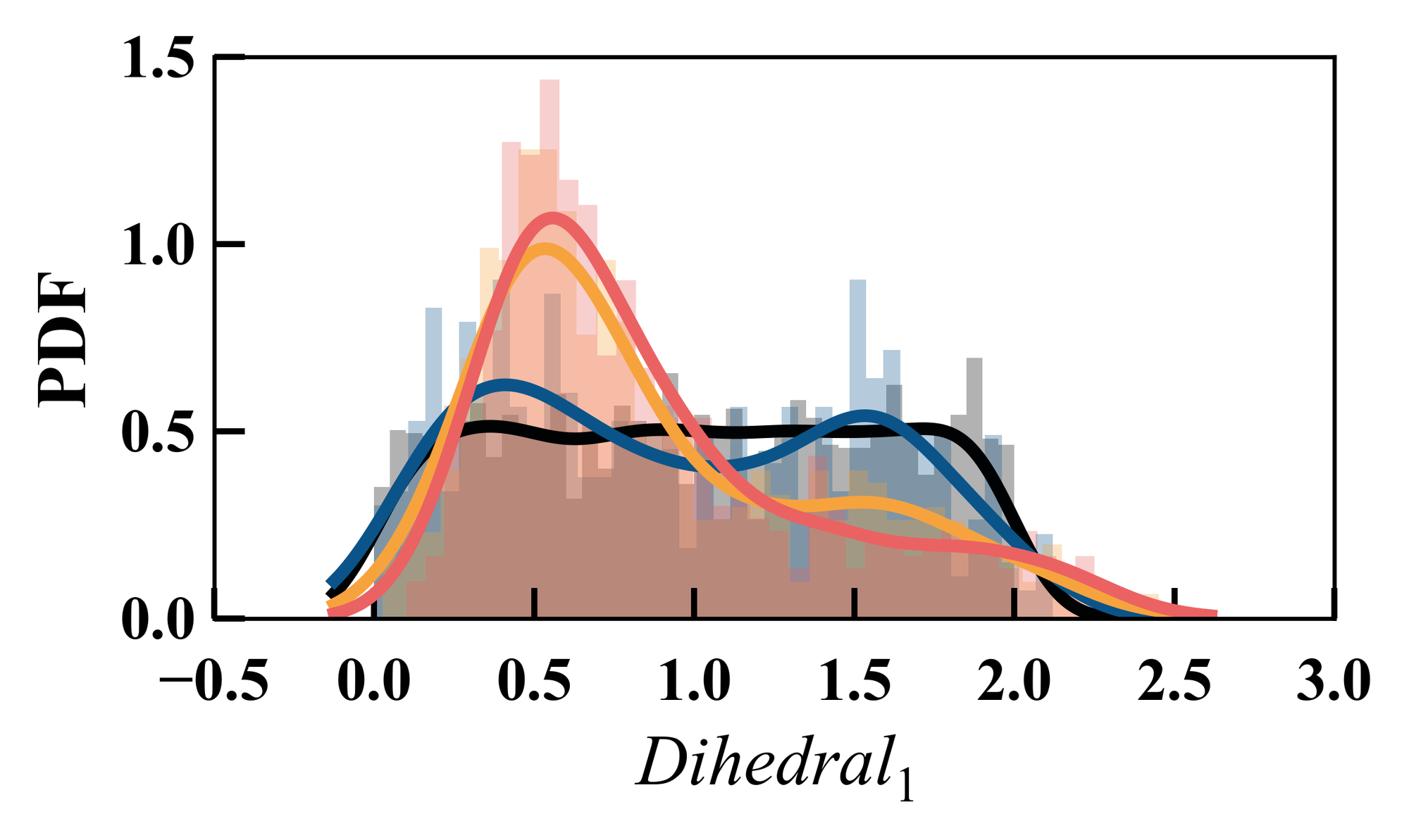}
\end{subfigure}
\hfill
\begin{subfigure}[b]{0.32\textwidth}
    \centering
    \includegraphics[width=\textwidth]{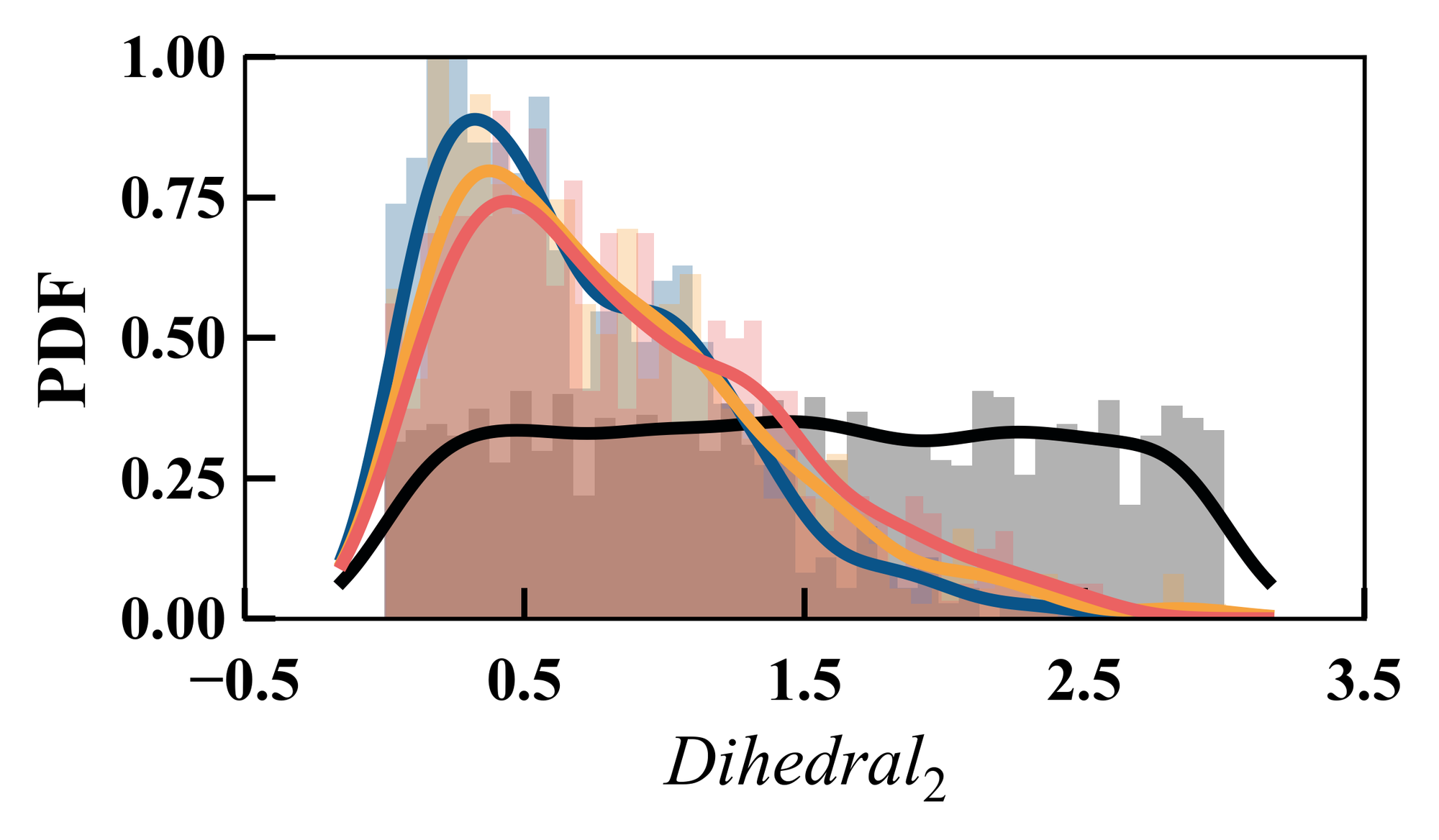}
\end{subfigure}
\hfill
\begin{subfigure}[b]{0.32\textwidth}
    \centering
    \includegraphics[width=\textwidth]{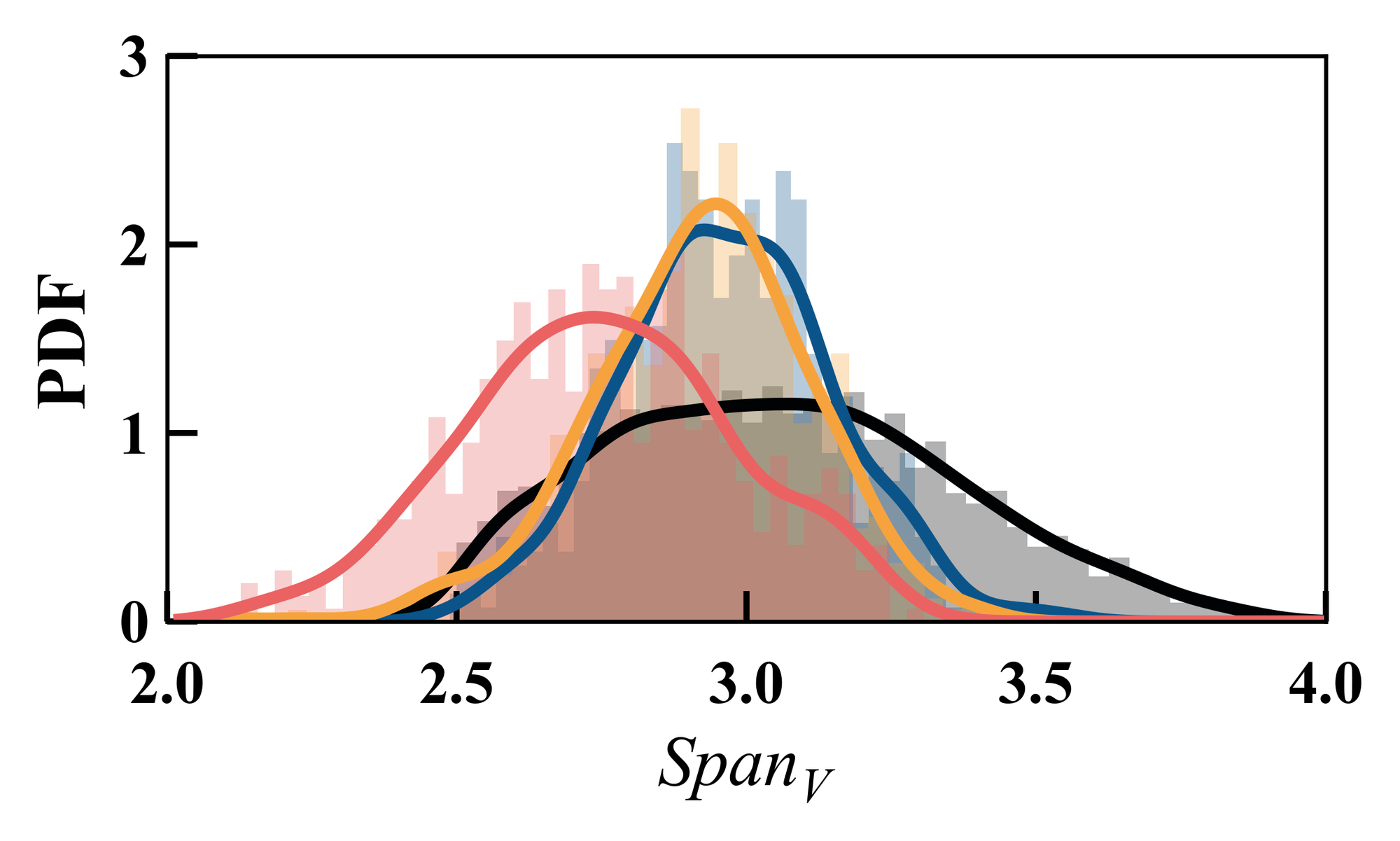}
\end{subfigure}

\vspace{0.5em} 

\begin{subfigure}[b]{0.32\textwidth}
    \centering
    \includegraphics[width=\textwidth]{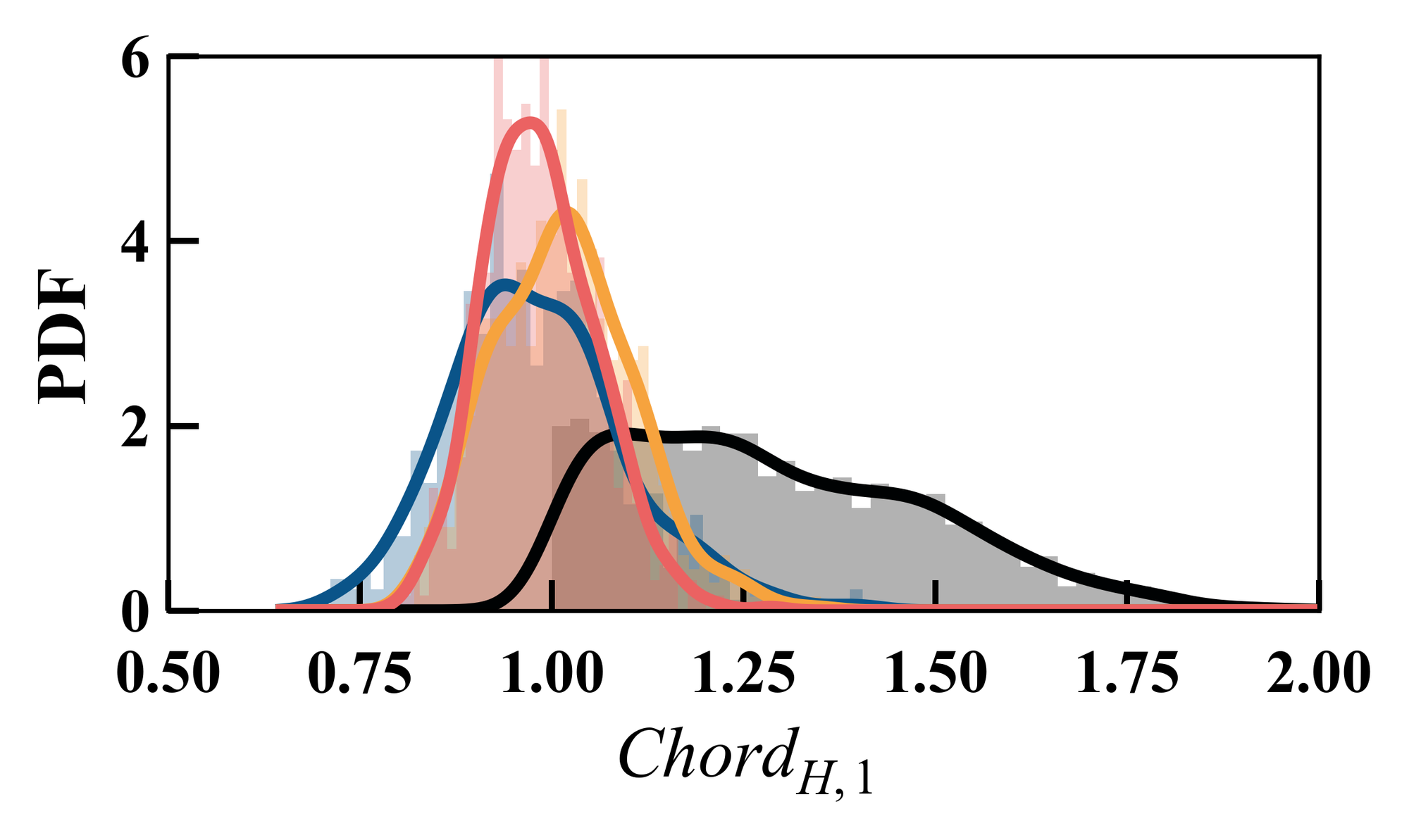}
\end{subfigure}
\hfill
\begin{subfigure}[b]{0.32\textwidth}
    \centering
    \includegraphics[width=\textwidth]{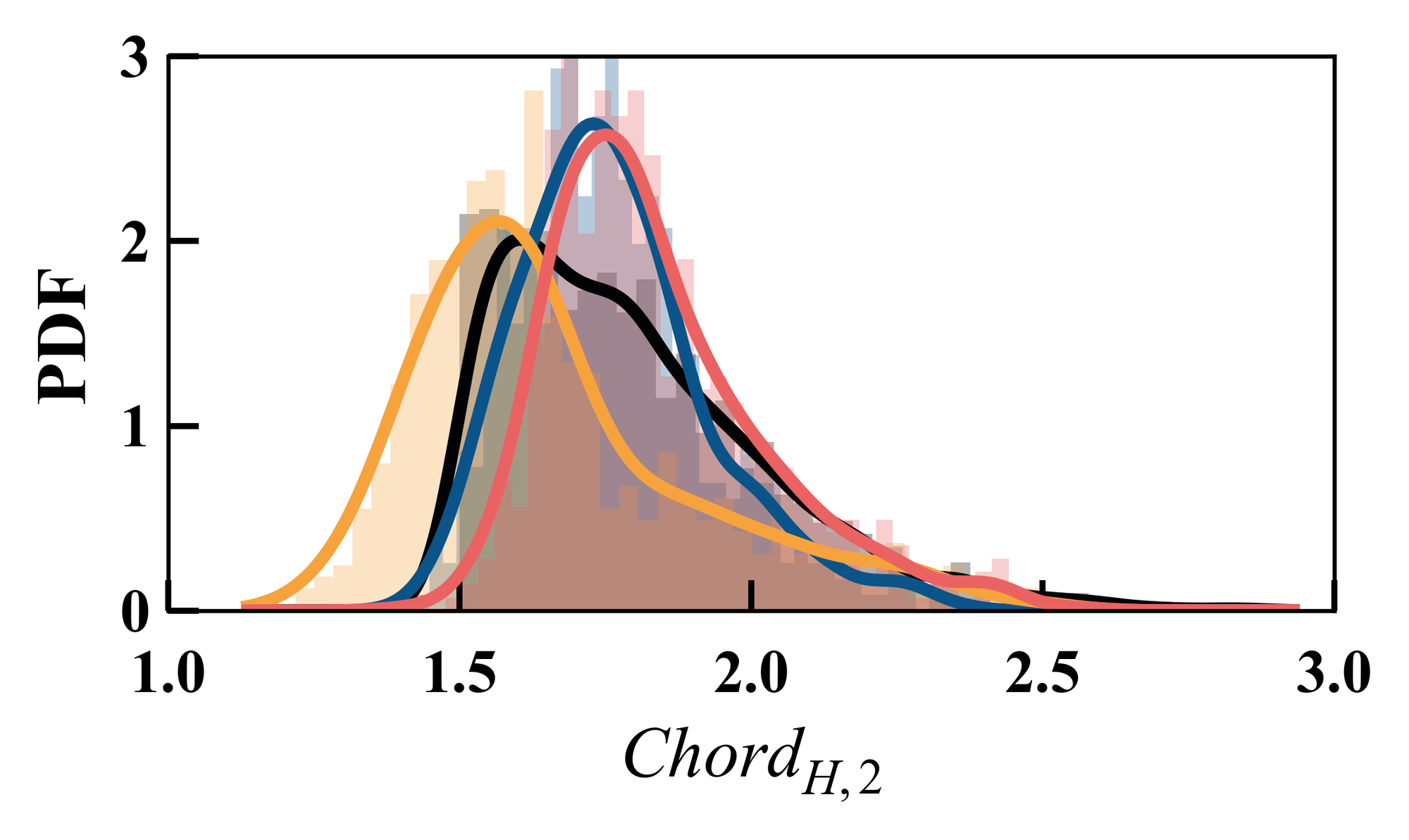}
\end{subfigure}
\hfill
\begin{subfigure}[b]{0.32\textwidth}
    \centering
    \includegraphics[width=\textwidth]{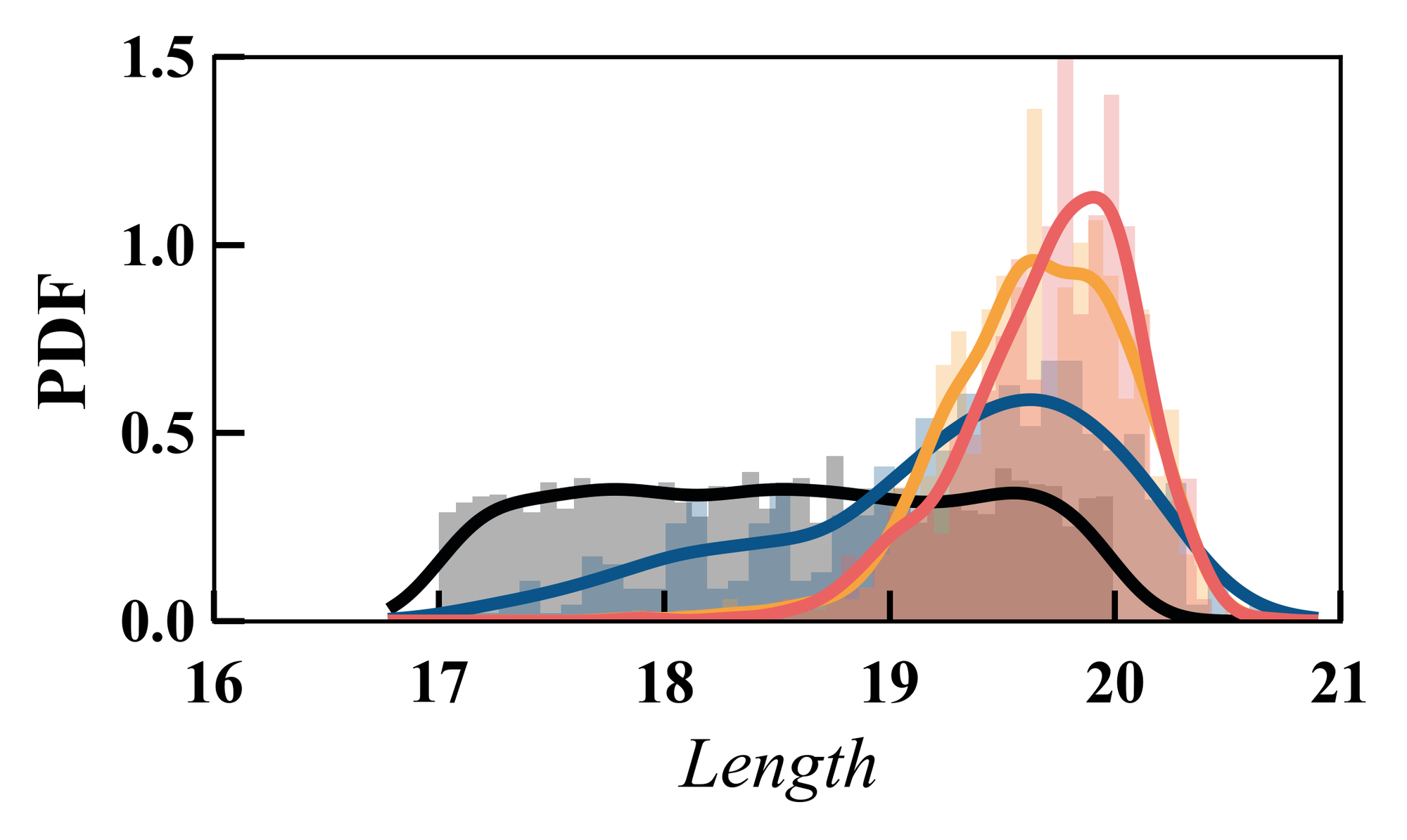}
\end{subfigure}

\caption{Distribution of generated geometric parameters for different models under extreme design conditions (Cond-C).}
\label{fig:feature}
\end{figure}

\begin{figure}[htbp]
\centering
\begin{subfigure}[b]{0.48\textwidth}
    \centering
    \includegraphics[width=\textwidth]{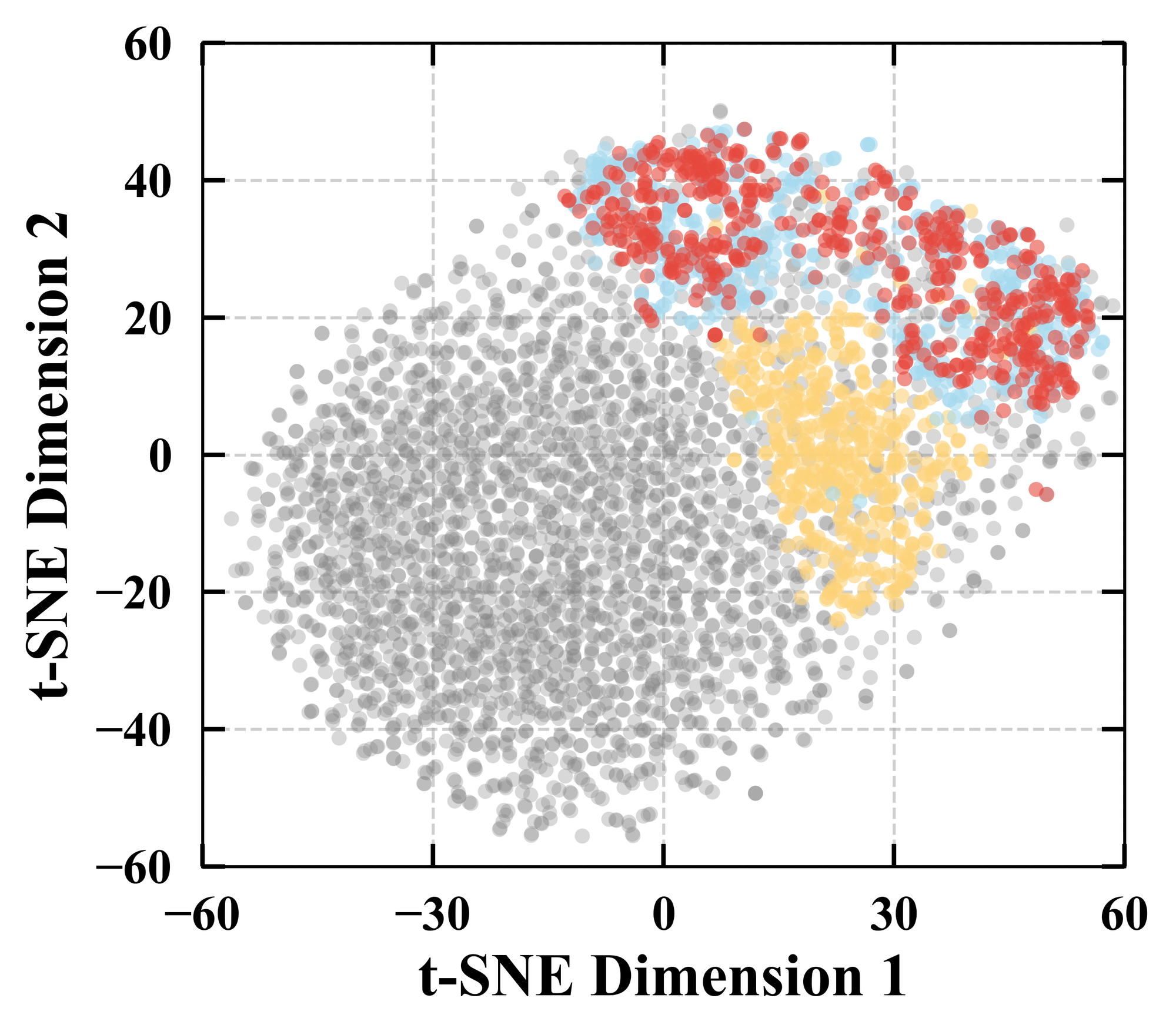}
    \caption{Cond-B}
    \label{fig:边缘tSNE}
\end{subfigure}
\hfill
\begin{subfigure}[b]{0.48\textwidth}
    \centering
    \includegraphics[width=\textwidth]{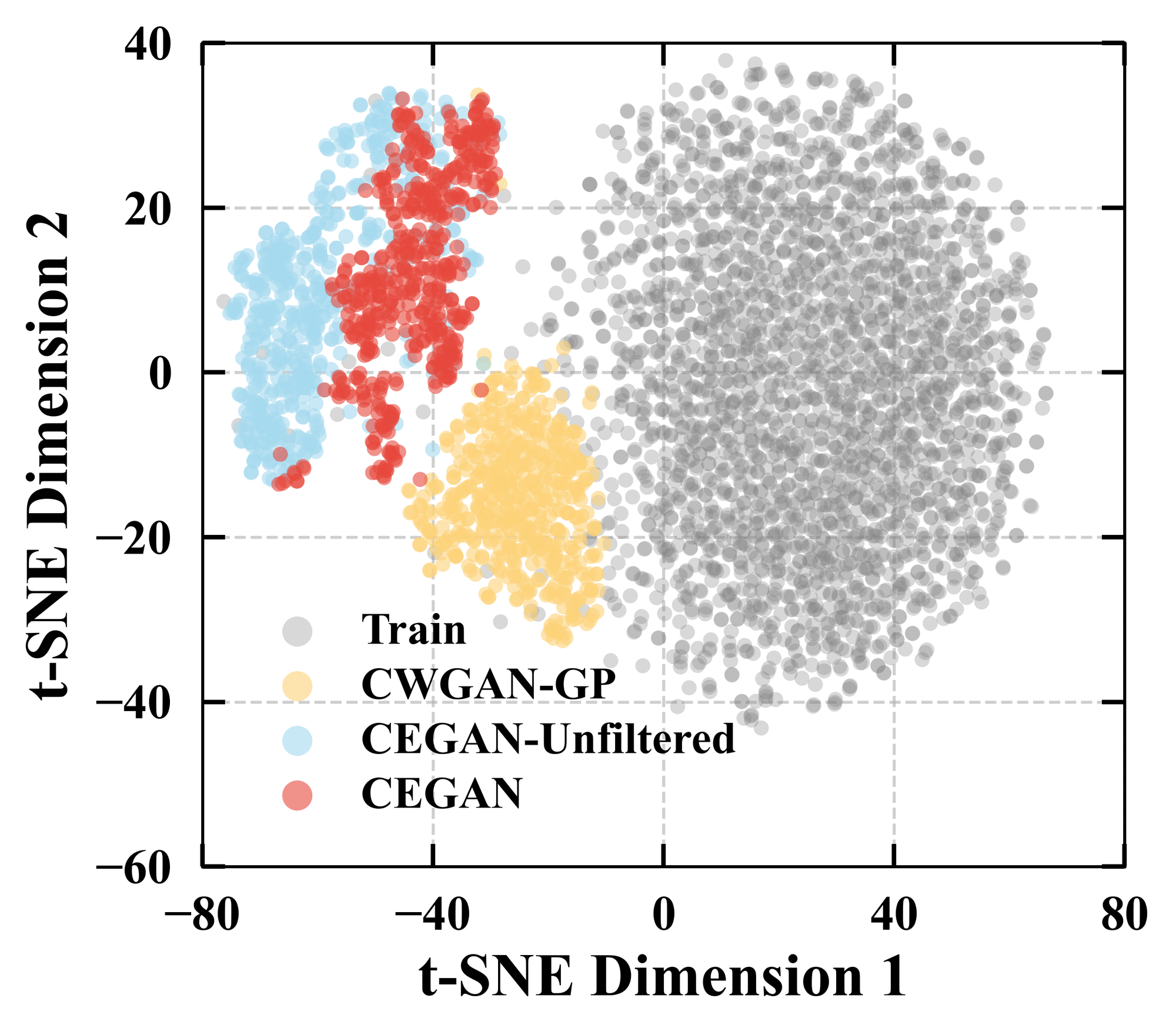}
    \caption{Cond-C}
    \label{fig:极限tSNE}
\end{subfigure}

\caption{Low-dimensional visualization of generated geometries by different models under different sets of design conditions.}
\label{fig:创新性探索}
\end{figure}

\begin{figure}[hbt!]
\centering
\includegraphics[width=1\textwidth]{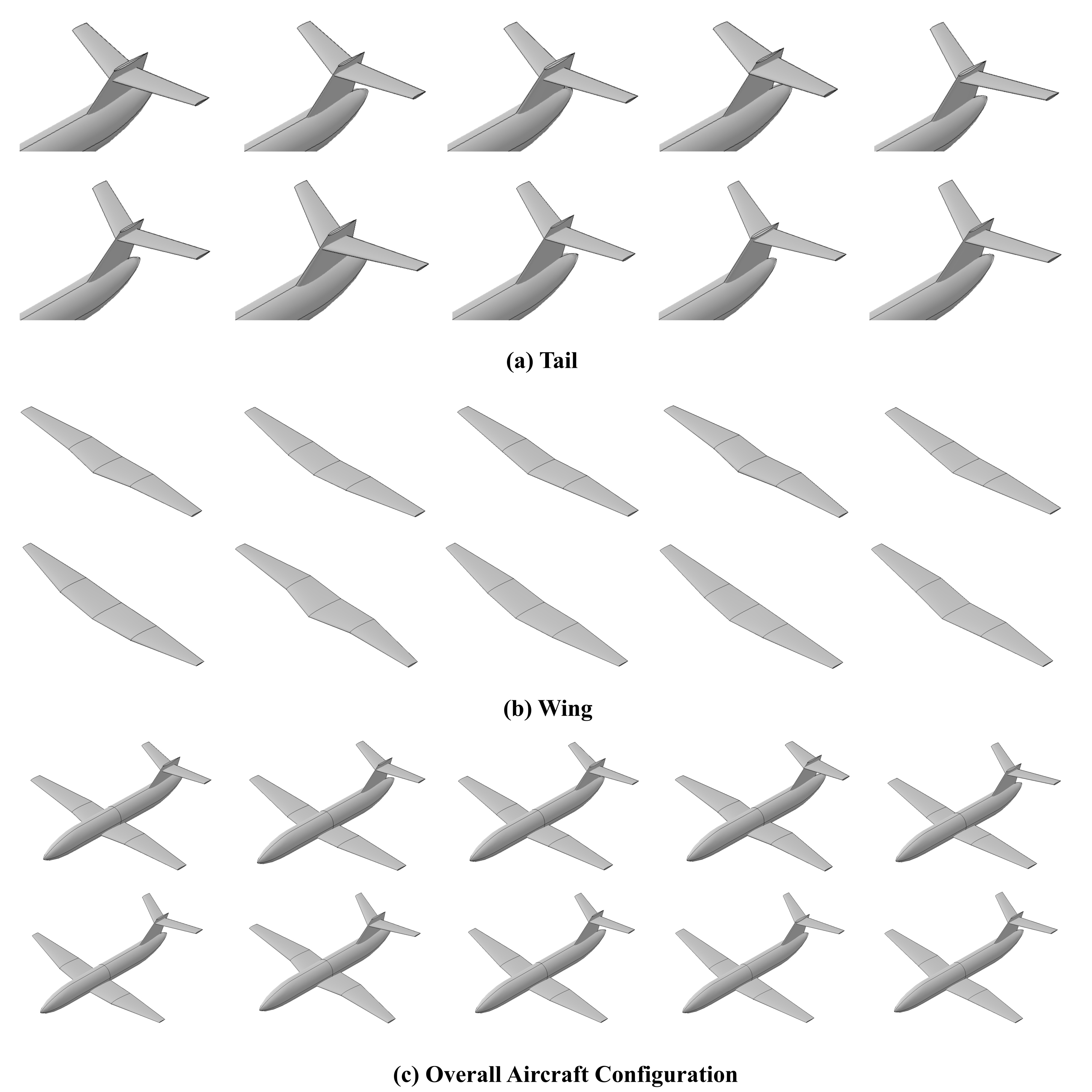}
\caption{Visualization of generated geometries based on CEGAN.}
\label{fig:CEGAN显示}
\end{figure}

Figure~\ref{fig:feature} presents the generation results of selected geometric parameters for both the training samples and CEGAN. It can be observed that CEGAN also preserves the diversity of the generated geometries. Moreover, the generated geometric parameters exhibit a significant tendency in the distribution. Figure~\ref{fig:创新性探索} uses the t-SNE algorithm to visualize, via dimensionality reduction of the generated geometric parameters, the geometrical feature distribution of the original training samples and the samples generated by different models~\cite{t-SNE2008}. The purpose is to evaluate how the generated samples are distributed relative to the original samples under various design conditions. Each point in the figure represents one sample, while its position in the two-dimensional space reflects the similarity of high-dimensional features, as determined by the t-SNE mapping. For Cond-B at the boundary of the distribution of the training conditions, as shown in Figure~\ref{fig:边缘tSNE}, a small number of training samples already satisfy this condition. Consequently, both CWGAN-GP and CEGAN are able to identify viable solutions within the original design space. Nevertheless, CEGAN achieves higher accuracy while maintaining comparable generation diversity. For strongly extrapolated conditions that lie outside the distribution of the training samples, as shown in Figure~\ref{fig:极限tSNE}, the performance of the two models differs markedly. CWGAN-GP is limited by its weak exploration capability beyond the training distribution, thus most of its generated samples remain clustered near the training data. In contrast, no matter whether output filtering is introduced, CEGAN actively explores design spaces outside the training space and discovers geometric configurations that better satisfy the prescribed conditions. Moreover, the output filtering module effectively discards samples that exhibit high exploratory potential but insufficient accuracy. This mechanism offers CEGAN with both extrapolation capability and high generation accuracy.  Figure~\ref{fig:CEGAN显示} directly visualizes representative generated shapes of the full-aircraft configuration, main wing, and tail under extreme conditions using CEGAN, further demonstrating the generation diversity.

\begin{figure}[htbp]
\centering
\begin{subfigure}[b]{0.45\textwidth}
    \centering
    \includegraphics[width=\textwidth]{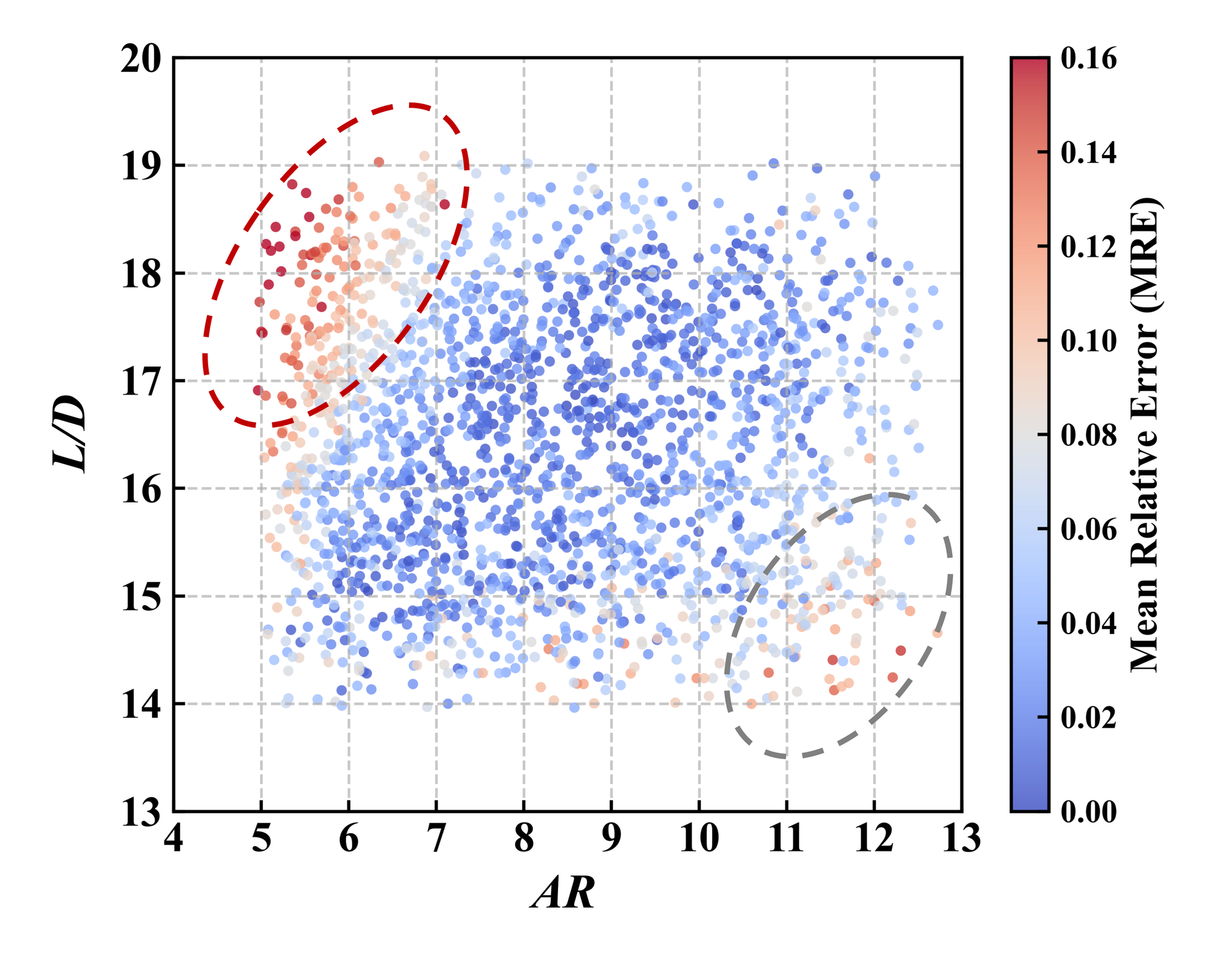}
    \caption{CWGAN-GP}
    \label{fig:目标与样本散点CWGAN-GP}
\end{subfigure}
\hfill
\begin{subfigure}[b]{0.45\textwidth}
    \centering
    \includegraphics[width=\textwidth]{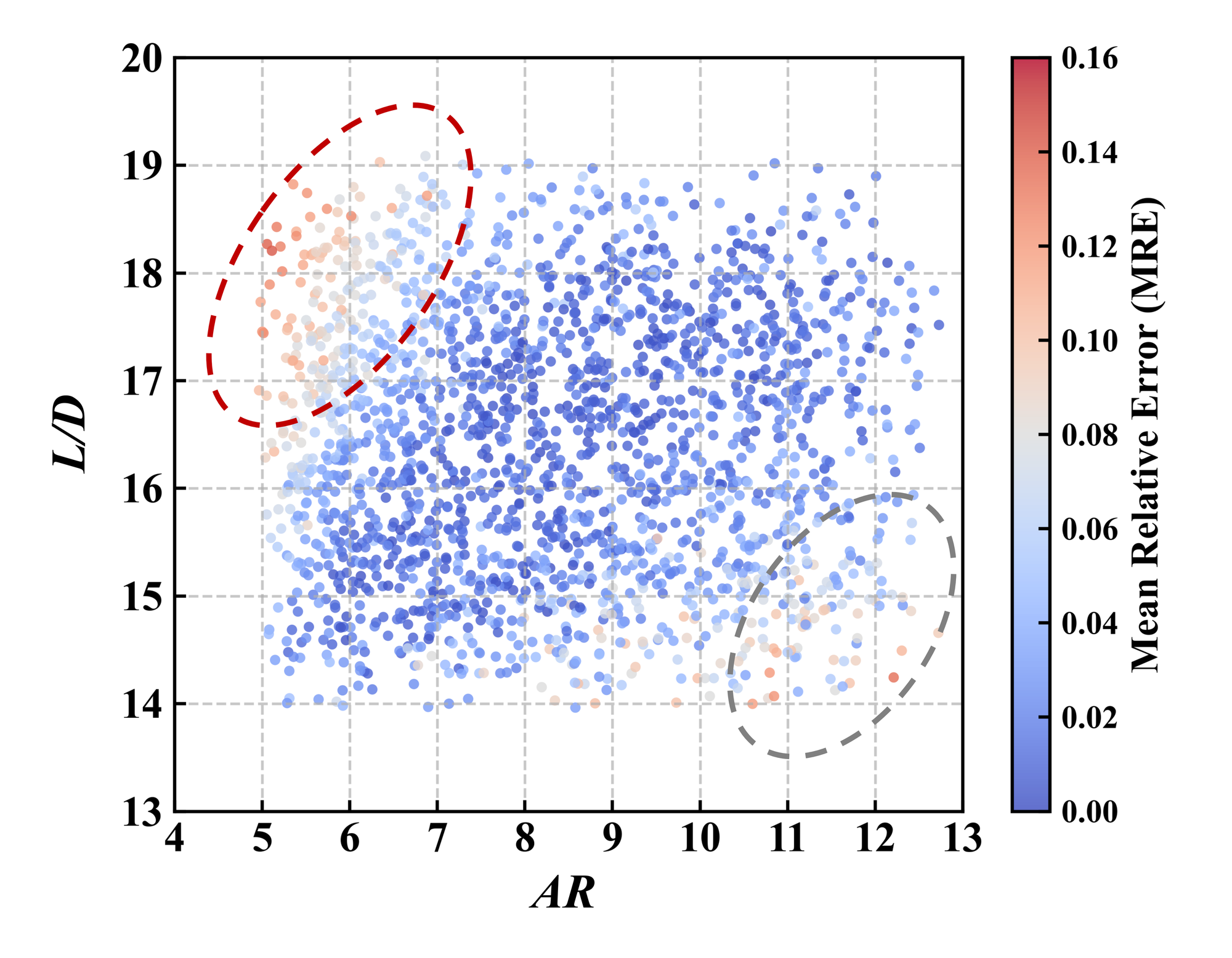}
    \caption{CEGAN}
    \label{fig:目标与样本散点CEGAN}
\end{subfigure}

\caption{Error Distribution of CWGAN-GP and CEGAN (Cond-D).}
\label{fig:全局条件精度对比}
\end{figure}

To further validate the generation quality and generalization capability of the CEGAN model across the global design space, Figure~\ref{fig:全局条件精度对比} presents a visualization of generation errors based on 2,000 LHS points within the aerodynamic-geometric coupled space (Cond-D), characterized by lift-to-drag ratio and aspect ratio. For Cond-D, since each sampled condition vector corresponds to only a single generated sample, the results lack sufficient statistical support and are susceptible to individual outliers. Consequently, the improvement in Mean Relative Error (MRE) for certain parameters (e.g., $\Lambda$, $C_M$, $C_L$) is not statistically representative but are still meaningful.  It can be observed that the CEGAN model significantly improves generation accuracy when facing strongly coupled conditions, demonstrating superior robustness. Notably, for the extrapolation conditions located in the upper-left region, which correspond to configurations with a high lift-to-drag ratio and a low aspect ratio, CEGAN achieves a particularly pronounced improvement in generation accuracy. In contrast, although a considerable accuracy gain is also observed for the extrapolation conditions in the lower-right region, these conditions are shown here primarily for completeness, as they hold limited practical engineering value.

Table~\ref{tab:全部效果} summarizes the quantitative performance of different models in terms of generation accuracy and diversity. The results indicate that for interpolation cases (Cond-A and Cond-B), CEGAN achieves a generation error reduction of at least 36\% compared to CWGAN-GP. The improvement is even more significant for strongly coupled extrapolation cases, with a generation error reduction of at least 46\%. Further analysis incorporating diversity metrics reveals that the improvement in generation accuracy is accompanied by a certain degree of diversity decay; however, this attenuation remains within an acceptable range, indicating that the model significantly enhances generation accuracy while preserving the capability to explore the design space. Meanwhile, as the accuracy of the generator improves, the diversity of the generated data may decrease, since the results become more concentrated around a specific condition vector.

\begin{table}[hbt!]
\caption{\label{tab:全部效果} Generation performance of different models under varying design conditions.}
\centering
\small
\setlength{\tabcolsep}{4pt}
\begin{adjustbox}{max width=\linewidth}
\begin{tabular}{clccccccc}
\hline\hline
\multirow{2}{*}{\makecell{Condition}} & \multirow{2}{*}{\makecell{Model}}
& \multicolumn{6}{c}{MRE (\%) } 
& \multirow{2}{*}{\makecell{MNND}} \\
\cmidrule(lr){3-8} 
& & $S$ & $L$ & $\Lambda$ & $C_M$ & $C_D$ & $C_L$ &  \\
\hline

\multirow{3}{*}{ Cond-A}  
 & CWGAN-GP        & 1.03& 0.19& 0.92& 1.63& 0.52& 1.15& $\bm{4.0813}$\\
 & CEGAN-Unfiltered& 0.90& 0.09& 0.49& 1.41& 0.44& 0.86& 3.4288\\
 & CEGAN           & $\bm{0.44}$& $\bm{0.08}$& $\bm{0.41}$& $\bm{0.48}$& $\bm{0.33}$& $\bm{0.50}$& 3.2568\\
\multirow{3}{*}{ Cond-B}  
 & CWGAN-GP        & 1.22& 0.24& 2.76& 2.47& 0.75& 0.82& $\bm{3.8098}$\\
 & CEGAN-Unfiltered& 0.78& 0.10& 1.27& 2.19& 0.50& 0.78& 3.0420\\
 & CEGAN           & $\bm{0.44}$& $\bm{0.08}$& $\bm{0.49}$& $\bm{0.99}$& $\bm{0.48}$& $\bm{0.43}$& 2.8953\\
\multirow{3}{*}{ Cond-C} 
 & CWGAN-GP        & 1.49& 0.26& 6.32& 4.05& 1.23& 5.32& $\bm{3.8300}$\\
 & CEGAN-Unfiltered& 1.28& 0.20& 2.97& 2.78& 0.63& 3.38& 3.0502\\
 & CEGAN           & $\bm{0.58}$& $\bm{0.14}$& $\bm{1.09}$& $\bm{0.73}$& $\bm{0.31}$& $\bm{2.30}$& 2.9794\\
\multirow{3}{*}{ Cond-D}  
 & CWGAN-GP        & 2.31& 1.17& 54.63& 7.14& 1.51& 4.93& $\bm{4.0697}$\\
 & CEGAN-Unfiltered& 1.68& 0.33& 26.99& 8.42& 1.16& 5.25& 3.8475\\
 & CEGAN           & $\bm{1.79}$& $\bm{0.34}$& $\bm{26.17}$& $\bm{7.00}$& $\bm{1.17}$& $\bm{4.73}$& 3.3953\\
 \hline\hline
 \multicolumn{5}{l}{\small\textit{Note: The best results are marked in \textbf{bold}.}}
\end{tabular}
\end{adjustbox}
\end{table}

\begin{figure}[hbt!]
\centering
\includegraphics[width=\textwidth]{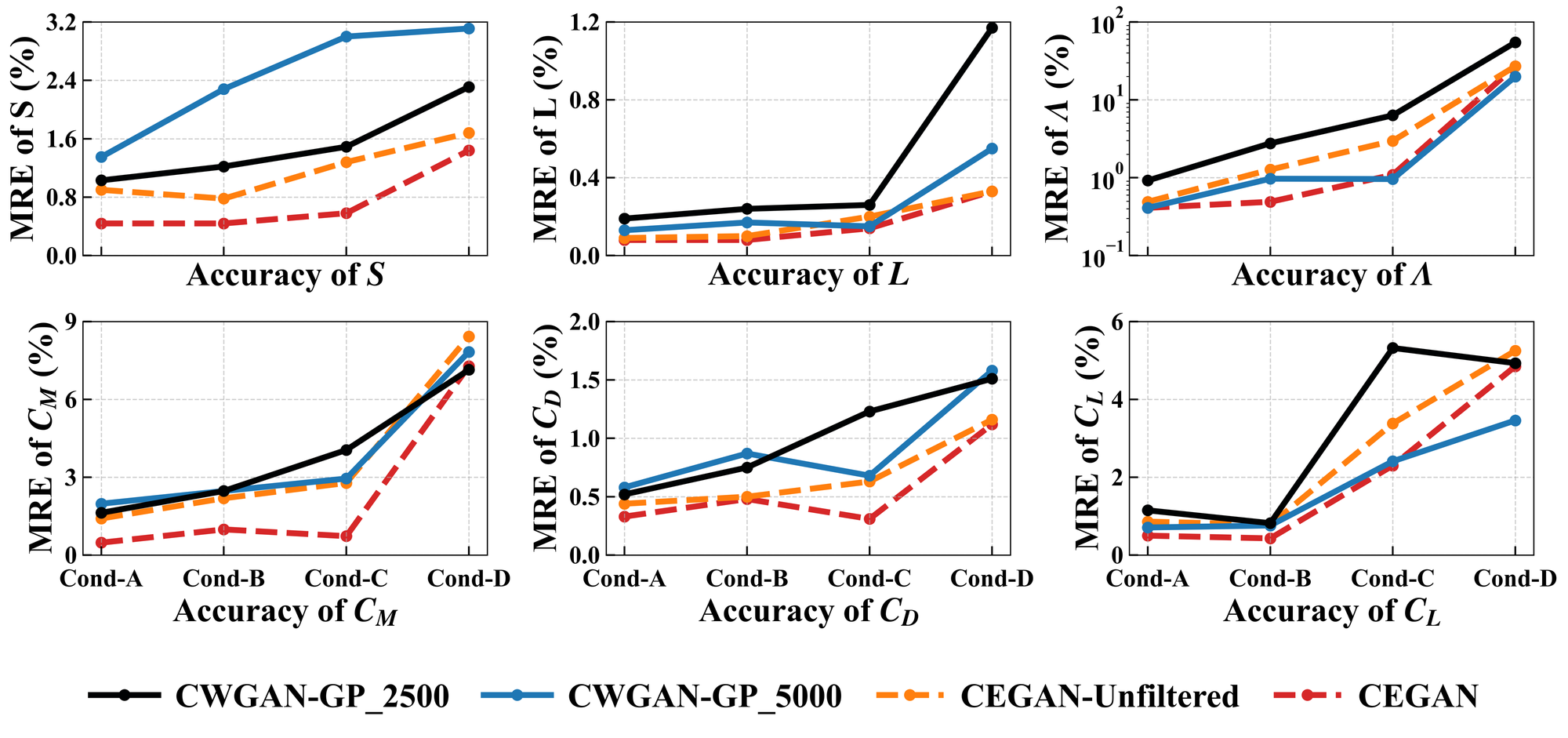}
\caption{Performance comparison of different models.}
\label{fig:不同模型精度}
\end{figure}

As illustrated in Figure~\ref{fig:不同模型精度}, a CWGAN-GP model trained on 5,000 samples is introduced as an additional baseline for comparison. The results indicate that even without the output filtering module, the CEGAN achieves higher generation accuracy than the original CWGAN-GP trained on the same sample size (2,500 samples). Furthermore, the complete CEGAN framework attains accuracy comparable to that of the CWGAN-GP model trained on a larger number of samples (5,000 samples). Consistent improvements in generation accuracy are observed across Cond-A, Cond-B and Cond-C for both geometric parameters ($S$, $L$, $\Lambda$) and aerodynamic coefficients ($C_L$, $C_D$, $C_M$). Moreover, given the inherent training instability of GANs, merely increasing the training dataset size to 5,000 does not guarantee improvement in generation accuracy. In contrast, the proposed conditional enhancement method not only effectively improves generation accuracy for prescribed conditions but also maximizes the utility of the training data, thereby reducing the reliance on large training datasets while ensureing better generation accuracy.

\section{Conclusions}
\label{sec:conclusions}
This study proposes a CEGAN model to address the limited generation accuracy of conditional generative models used for aerodynamic configuration design of aircraft. The main conclusions are as follows:

(1) CEGAN incorporates physical information into the generative model, through a condition-informed loss function, composed of a geometry-guided and an aerodynamic performance-guided enhancement strategies. Furthermore, it employs an output filtering approach that discards physically inconsistent samples during model generation.

(2) CEGAN is tested on the generative design of general aviation aircraft. Compared to the baseline CWGAN-GP model, it reduces the generation error by at least 36\% for interpolation cases (Cond-A and Cond-B) and 46\% for the strongly extrapolated case (Cond-C), while imporoves overall accuracy in the global condition space (Cond-D). 

(3) Beyond accuracy gains, CEGAN preserves reasonable geometric diversity among the generated samples. Moreover, it achieves the same level of accuracy compared to the CWGAN-GP trained with double number of the training samples, indicating efficient and sufficient utilization of the training dataset.

(4) Additional validation shows that the benefit of aerodynamic enhancement depends on surrogate accuracy. Mean validation $R^2$ levels near 0.95 and above give more consistent improvements in the present sensitivity study. This is an empirical observation for the tested configuration and training setup, rather than a universal accuracy threshold.

By establishing a general condition-enhanced generative modeling framework, this work provides a practical and extensible pathway towards deploying generative machine learning in aircraft conceptual design. Future work will focus on developing an adaptive condition-enhancement framework and incorporating high-fidelity training data to further improve generation fidelity with fewer training resources.

\appendix
\renewcommand{\thefigure}{\arabic{figure}}
\renewcommand{\thetable}{\arabic{table}}
\section{Surrogate-Based Output Filtering: Methodology and Preliminary Validation}
\phantomsection

\begin{figure}[htbp]
\centering
\begin{subfigure}{0.48\textwidth}
    \centering
    \includegraphics[width=\linewidth]{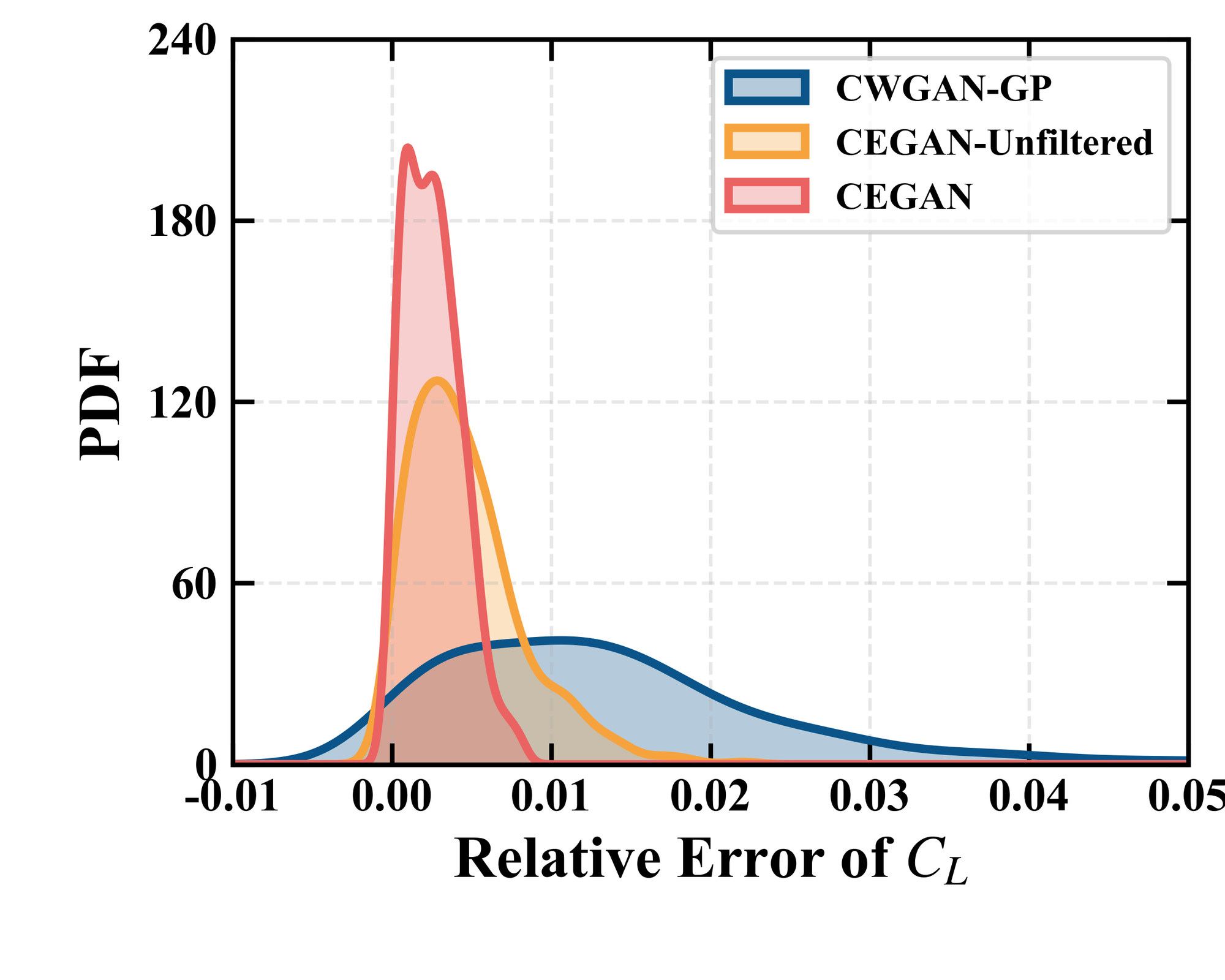}
    \caption{}
    \label{fig:CL}
\end{subfigure}
\hfill
\begin{subfigure}{0.48\textwidth}
    \centering
    \includegraphics[width=\linewidth]{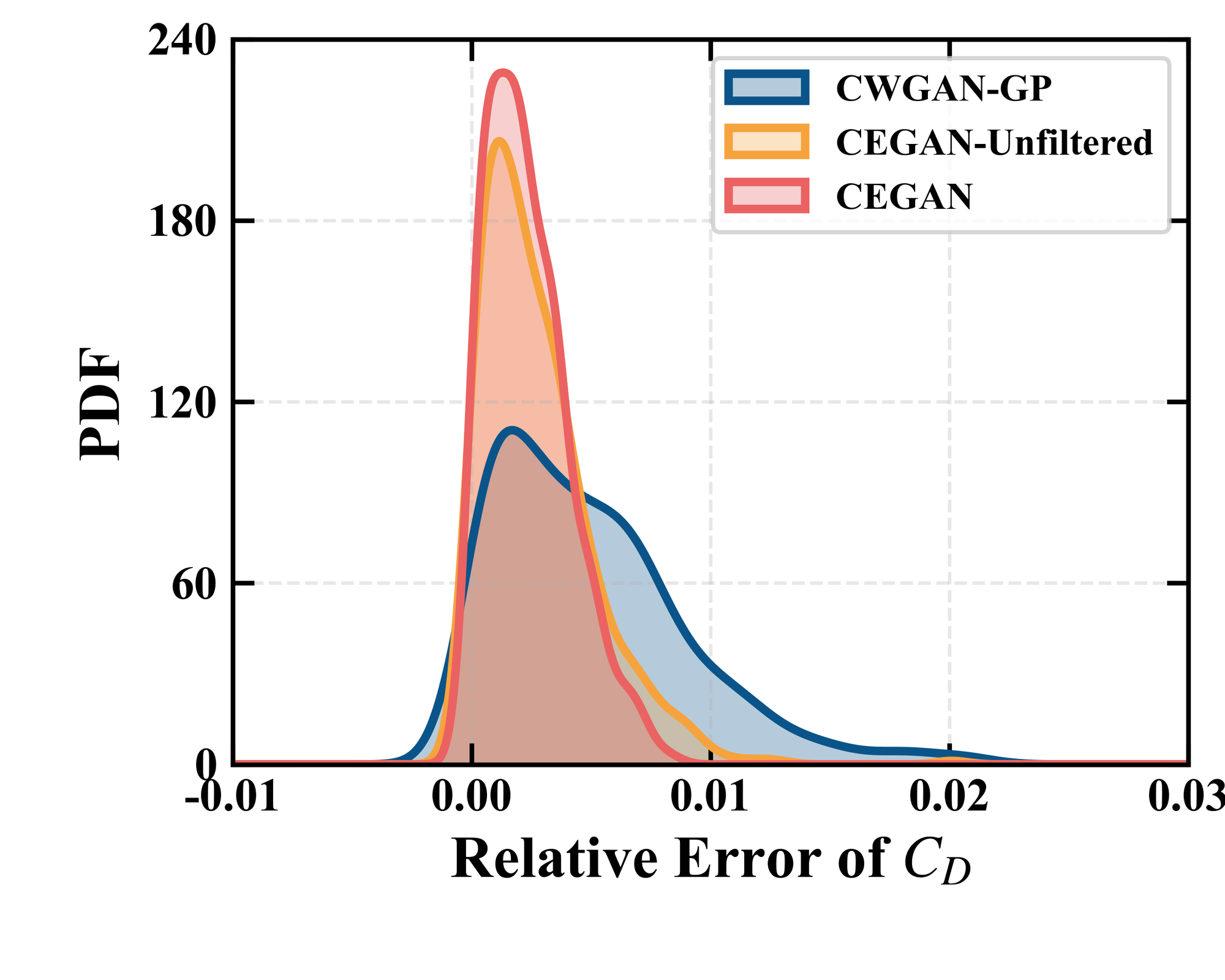}
    \caption{}
    \label{fig:CD}
\end{subfigure}

\vspace{1em} 

\begin{subfigure}{0.48\textwidth}
    \centering
    \includegraphics[width=\linewidth]{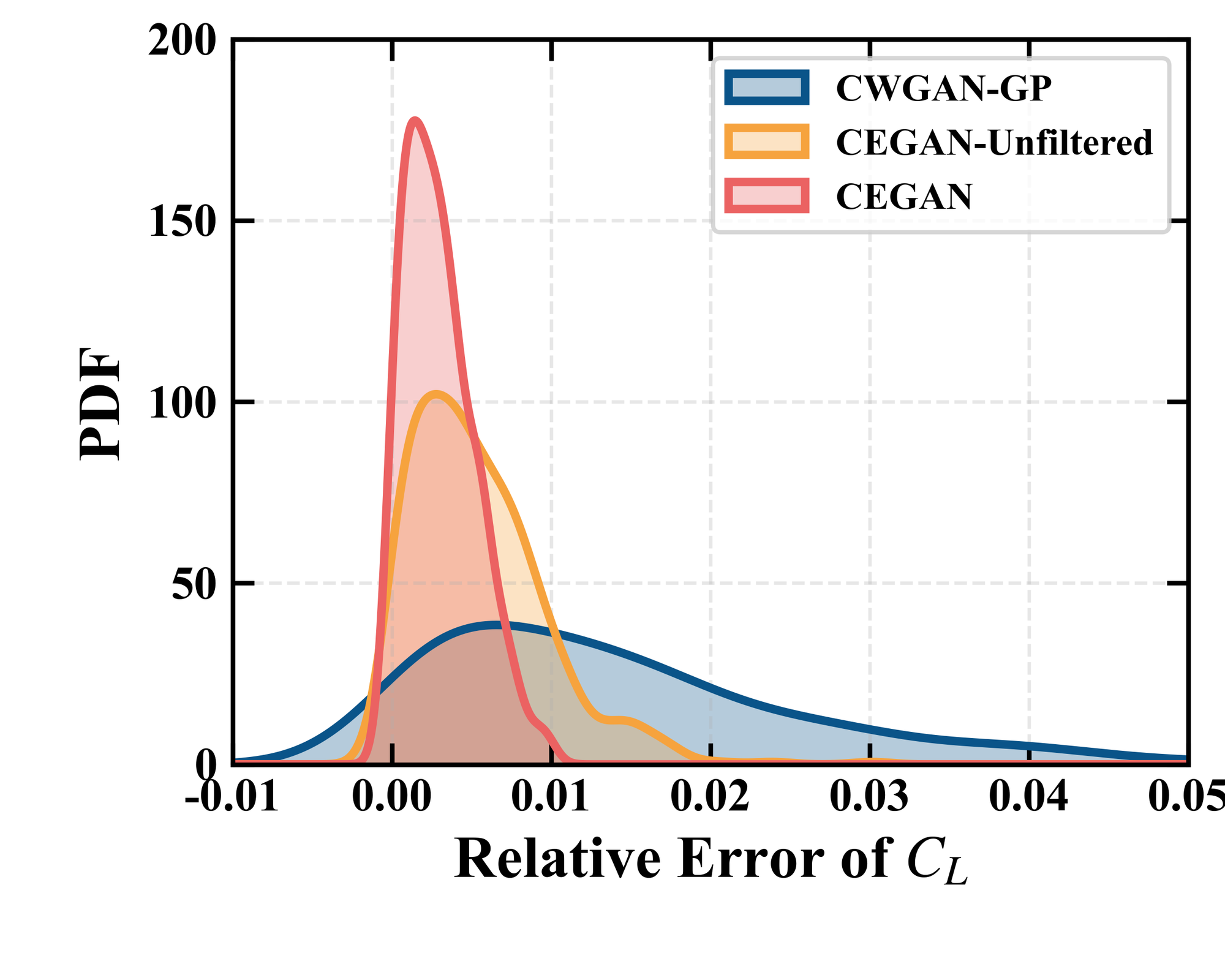}
    \caption{}
    \label{fig:CM}
\end{subfigure}
\hfill
\begin{subfigure}{0.48\textwidth}
    \centering
    \includegraphics[width=\linewidth]{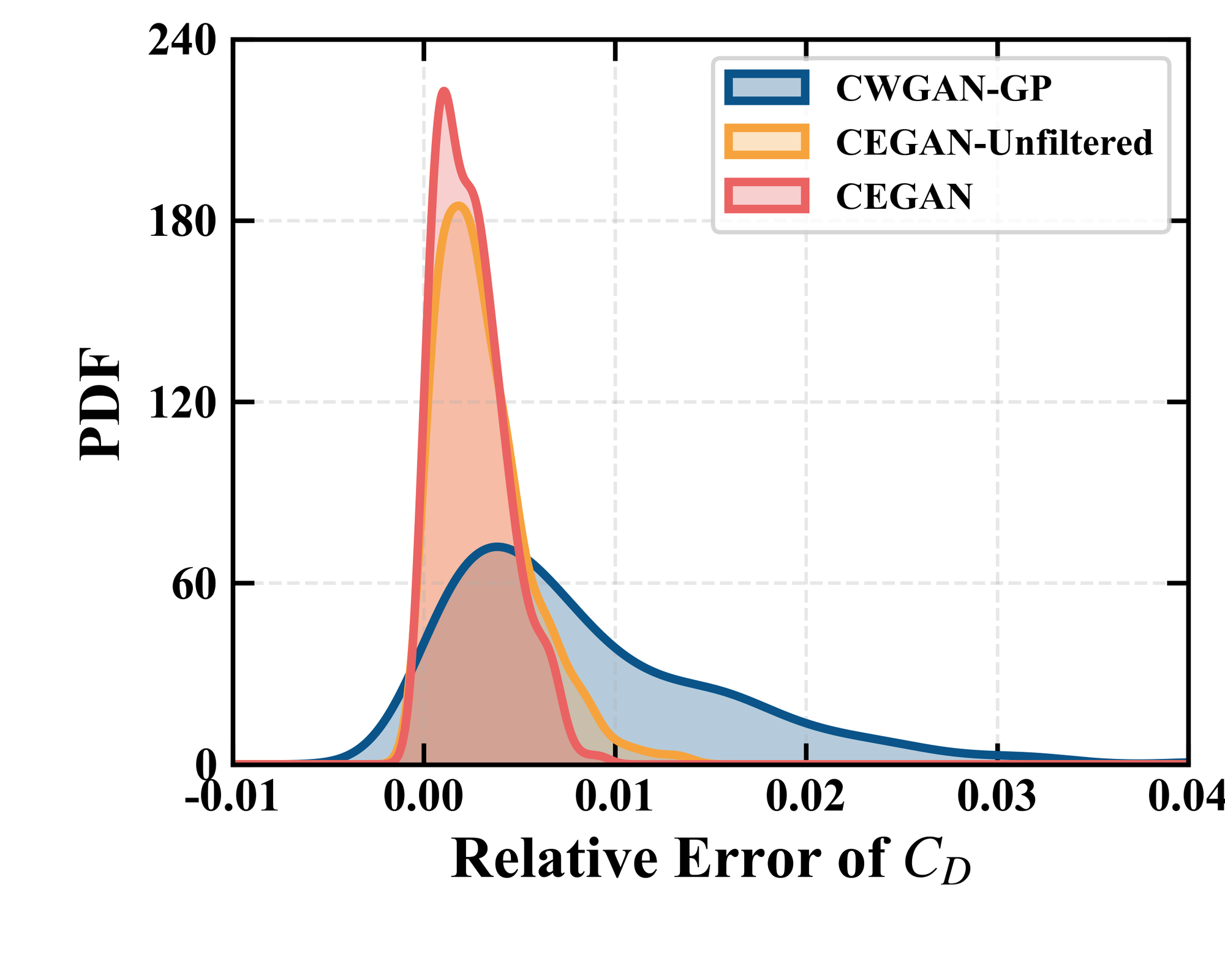}
    \caption{}
    \label{fig:acc_dist}
\end{subfigure}
\caption{
    Validation of the output filtering method. (a)-(b) Conditions: $C_L=0.21$, $C_D=0.0125$; (c)-(d) Conditions: $C_L=0.225$, $C_D=0.012$.
}
\label{fig:four_subfigs}
\end{figure}

We conducted a preliminary validation of the different CEGAN enhancement strategies on a generative model conditioned on two aerodynamic coefficients ($C_L, C_D$), using the same training dataset as described in Sec.~\ref{sec:3}. Two representative test conditions were selected, with one located at the center($C_L=0.21$, $C_D=0.0125$) and the other at the boundary($C_L=0.225$, $C_D=0.012$) of the distribution of the training conditions. The condition-informed loss functions are used for both models. As shown in Figure~\ref{fig:four_subfigs}, the output filtering approach achieves effective accuracy improvements, providing a solid foundation for the more comprehensive investigations presented in the subsequent sections.

\section{Ablation Study on Network Design}

The training results for the different CWGAN-GP architectures are summarized in Table~\ref{tab:Network_design}. Unless otherwise specified (e.g., rows explicitly indicating different activation functions for the generator and discriminator), the activation function listed in each row is used in both the generator ($G$) and the discriminator ($D$). For instance, the entry "Swish" denotes that Swish is applied to all layers of both networks, while an entry such as "$G$: Swish, $D$: LeakyReLU($k = 0.2$)"($k$ denotes the negative slope coefficient of the LeakyReLU function) indicates distinct activation functions for each network. 

For the model comparison, the condition vector specifies the lift coefficient $C_L$ and the drag coefficient $C_D$. For each model, 500 condition vectors were sampled within the range covered by the training samples using LHS. The MRE was then computed over the 500 generated results, with lower MRE values indicating higher generation accuracy and better generation quality. Through balancing generation accuracy and the number of training samples, the generator architecture was selected as 256-128-64, with Tanh as the activation function for the generator and LeakyReLU for the discriminator.

\begin{table}[htbp]
\centering
\caption{Generator architecture and activation function ablation results under different training samples.}
\label{tab:Network_design}
\small
\setlength{\tabcolsep}{4pt}
\begin{adjustbox}{max width=\linewidth}
\begin{tabular}{lccccc}
 \hline\hline
\multirow{2}{*}{\makecell{Generator\\Architecture}} & \multirow{2}{*}{\makecell{Activation\\Function}}
& \multicolumn{2}{c}{MRE (\%) (2,500 samples)} 
& \multicolumn{2}{c}{MRE (\%) (20,000 samples)} \\
\cmidrule(lr){3-4} \cmidrule(lr){5-6}
& & $C_L$ & $C_D$ & $C_L$ & $C_D$ \\
 \hline

\multirow{6}{*}{128-64-32} 
& ReLU    & 4.33 & 5.27 & 0.95 & 0.77 \\
& $\text{LeakyReLU}(k = 0.2)$ & 2.19 & 1.24 & 1.85 & 0.69 \\
& Swish   & 1.82 & 0.95 & 2.27 & 1.07 \\
& Tanh    & 2.18 & 1.19 & 1.65 & 0.58 \\
& \makecell[l]{$G:$Swish\\$D:\text{LeakyReLU}(k = 0.2)$} & 1.87 & 0.68 & 0.39 & 0.49 \\
& \makecell[l]{$G:$Tanh\\$D:\text{LeakyReLU}(k = 0.2)$} & 1.80 & 0.63 & $\bm{0.35}$ & 0.48 \\

 \hline

\multirow{6}{*}{256-128-64} 
& ReLU    & 4.95 & 3.32 & 0.35 & 0.61 \\
& $\text{LeakyReLU}(k = 0.2)$ & 1.57 & 1.02 & 1.15 & 0.47 \\
& Swish   & 1.56 & 1.97 & 3.39 & 1.81 \\
& Tanh    & 1.94 & 0.73 & 1.18 & 0.82 \\
& \makecell[l]{$G:$Swish\\$D:\text{LeakyReLU}(k = 0.2)$} & 3.36 & 4.83 & 1.33 & 0.70 \\
& \makecell[l]{$G:$Tanh\\$D:\text{LeakyReLU}(k = 0.2)$} & $\bm{1.44}$ & $\bm{0.59}$ & 0.59 & $\bm{0.47}$ \\

 \hline

\multirow{6}{*}{512-128-32} 
& ReLU    & 4.88 & 3.54 & 1.13 & 0.68 \\
& $\text{LeakyReLU}(k = 0.2)$ & 2.32 & 1.51 & 2.22 & 0.58 \\
& Swish   & 5.51 & 2.13 & 0.91 & 0.64 \\
& Tanh    & 1.83 & 1.00 & 0.58 & 0.70 \\
& \makecell[l]{$G:$Swish\\$D:\text{LeakyReLU}(k = 0.2)$} & 5.51 & 1.21 & 1.20 & 0.60 \\
& \makecell[l]{$G:$Tanh\\$D:\text{LeakyReLU}(k = 0.2)$} & 1.79 & 0.63 & 0.49 & 0.73 \\

 \hline\hline
 \multicolumn{5}{l}{\small\textit{Note: The best results are marked in \textbf{bold}.}}
\end{tabular}
\end{adjustbox}
\end{table}

\section{Additional Validation Results for Condition B}

\begin{figure}[htbp]
\centering

\begin{subfigure}[b]{0.3\textwidth}
    \centering
    \includegraphics[width=\textwidth]{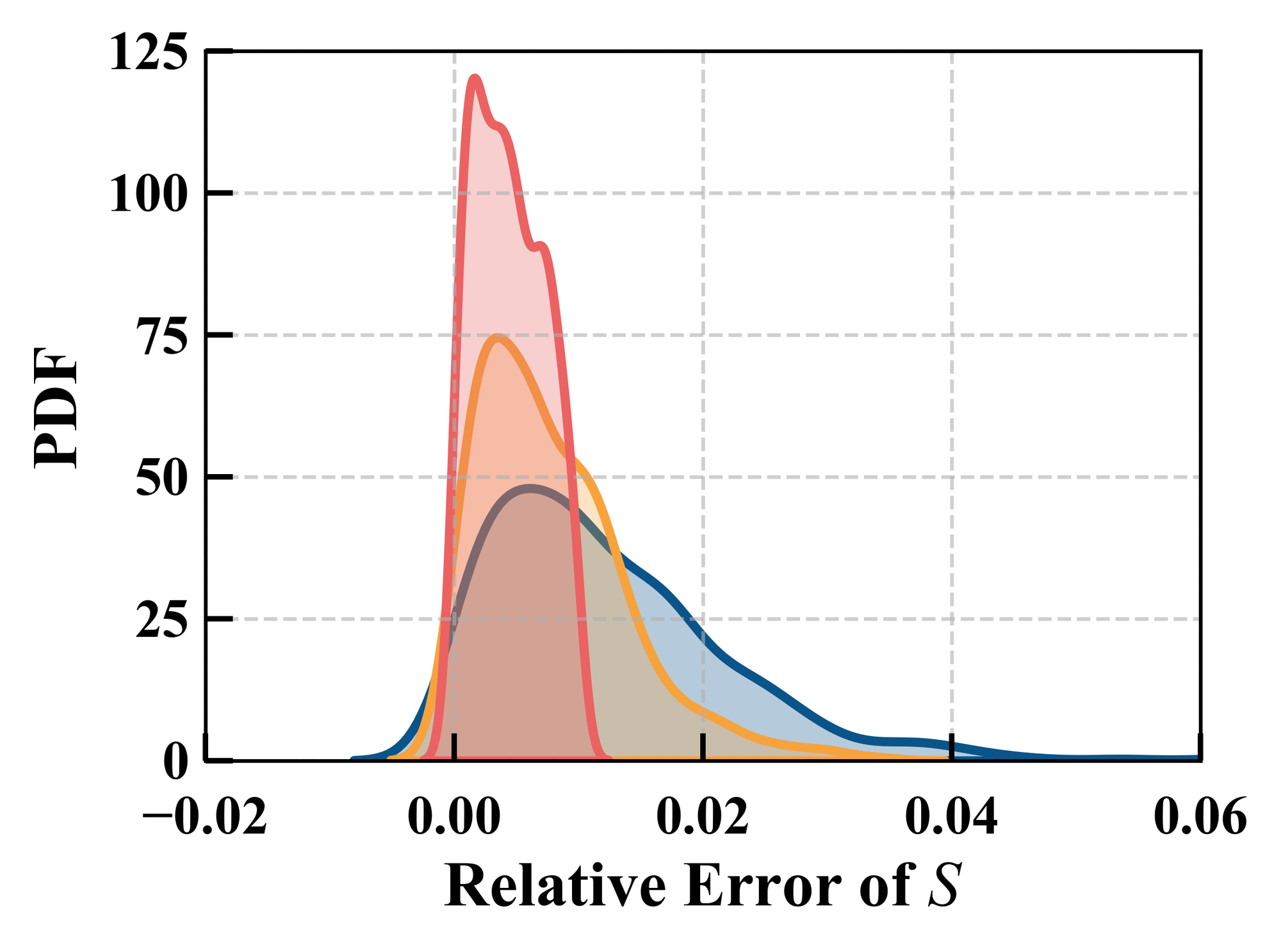}
    \caption{$S$}
    \label{fig:bsura}
\end{subfigure}
\hfill
\begin{subfigure}[b]{0.3\textwidth}
    \centering
    \includegraphics[width=\textwidth]{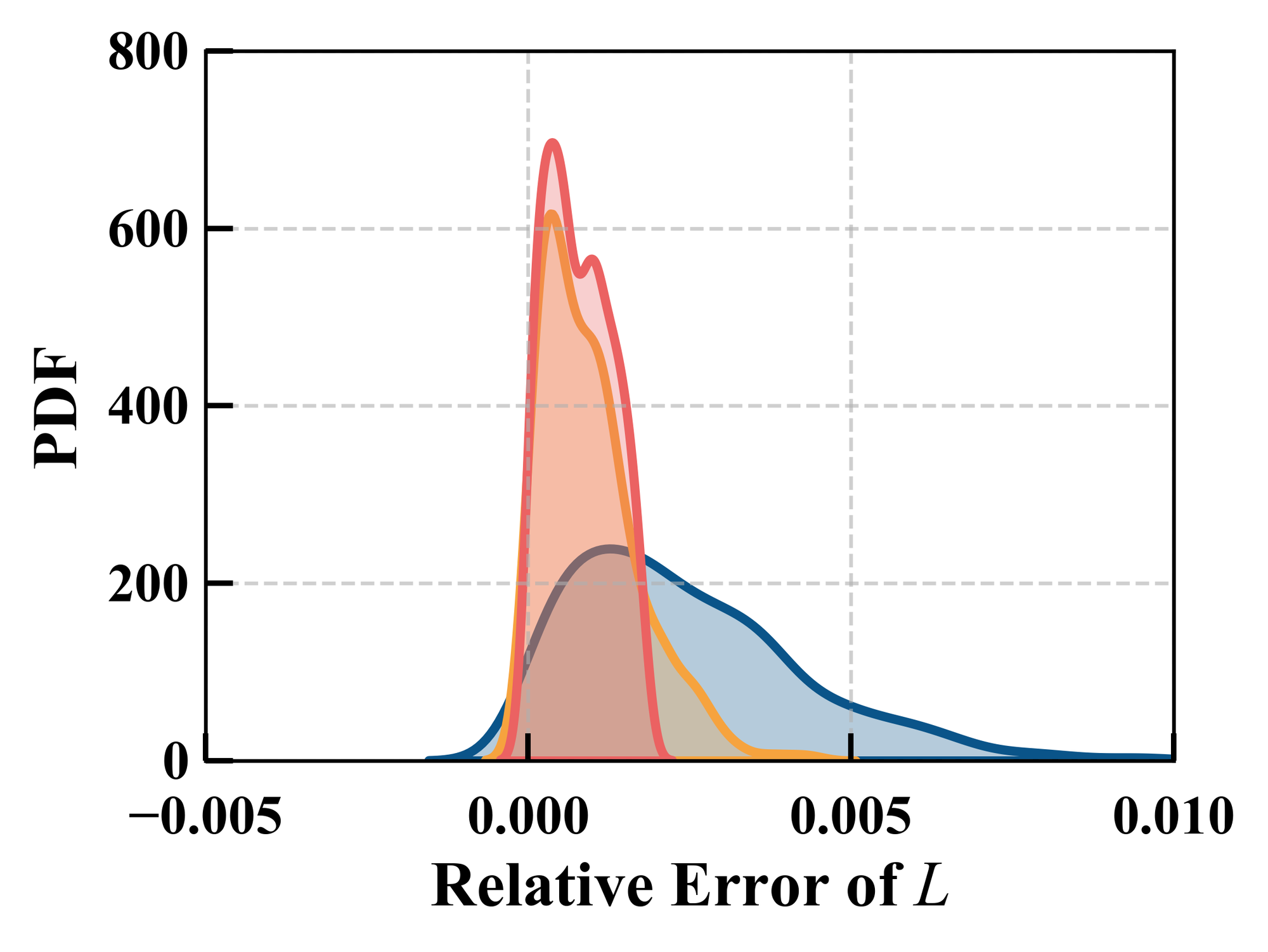}
    \caption{$L$}
    \label{fig:bsurb}
\end{subfigure}
\hfill
\begin{subfigure}[b]{0.3\textwidth}
    \centering
    \includegraphics[width=\textwidth]{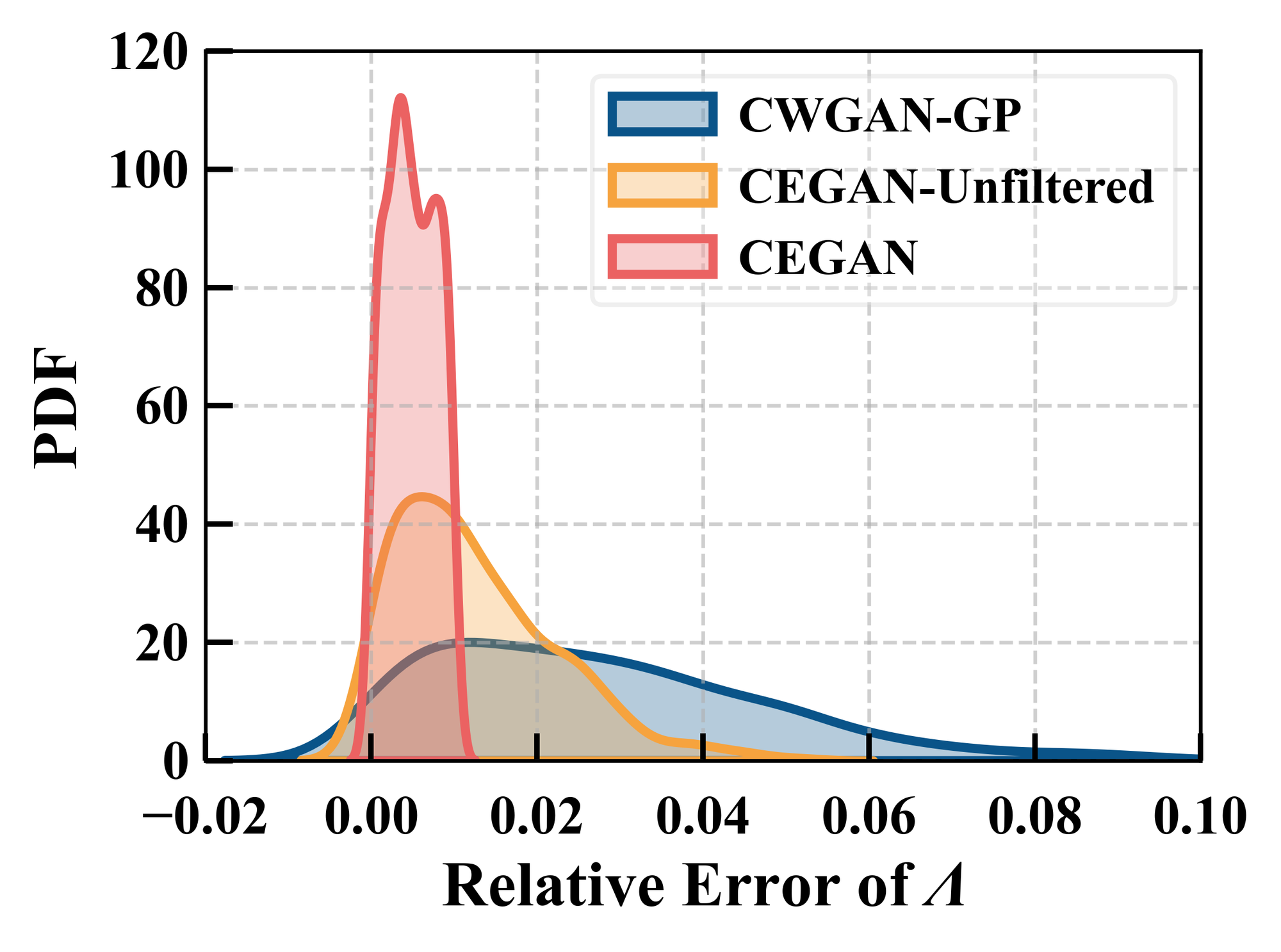}
    \caption{$\Lambda$}
    \label{fig:bsurc}
\end{subfigure}

\vspace{1em} 

\begin{subfigure}[b]{0.3\textwidth}
    \centering
    \includegraphics[width=\textwidth]{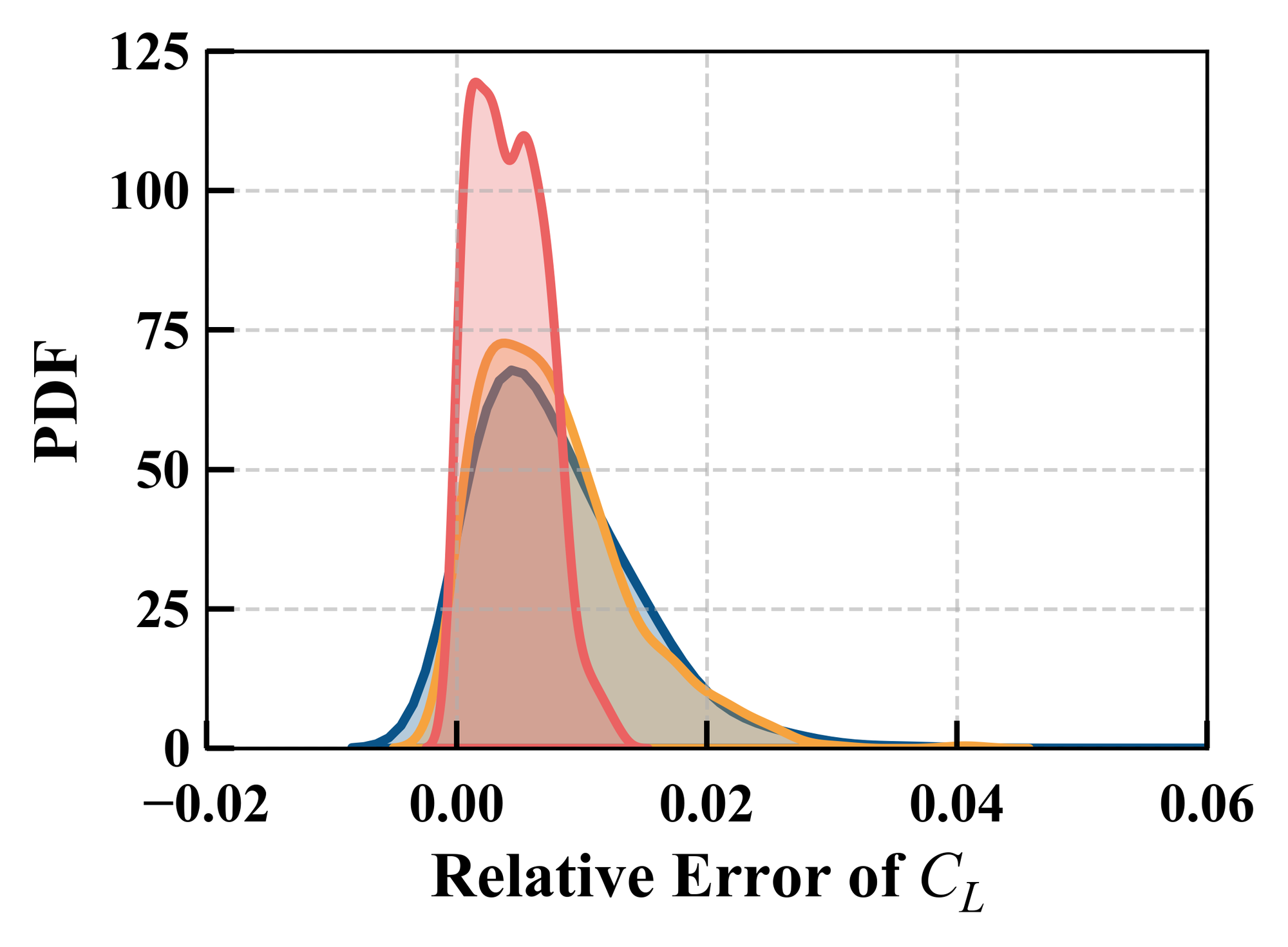}
    \caption{$C_L$}
    \label{fig:bsurd}
\end{subfigure}
\hfill
\begin{subfigure}[b]{0.3\textwidth}
    \centering
    \includegraphics[width=\textwidth]{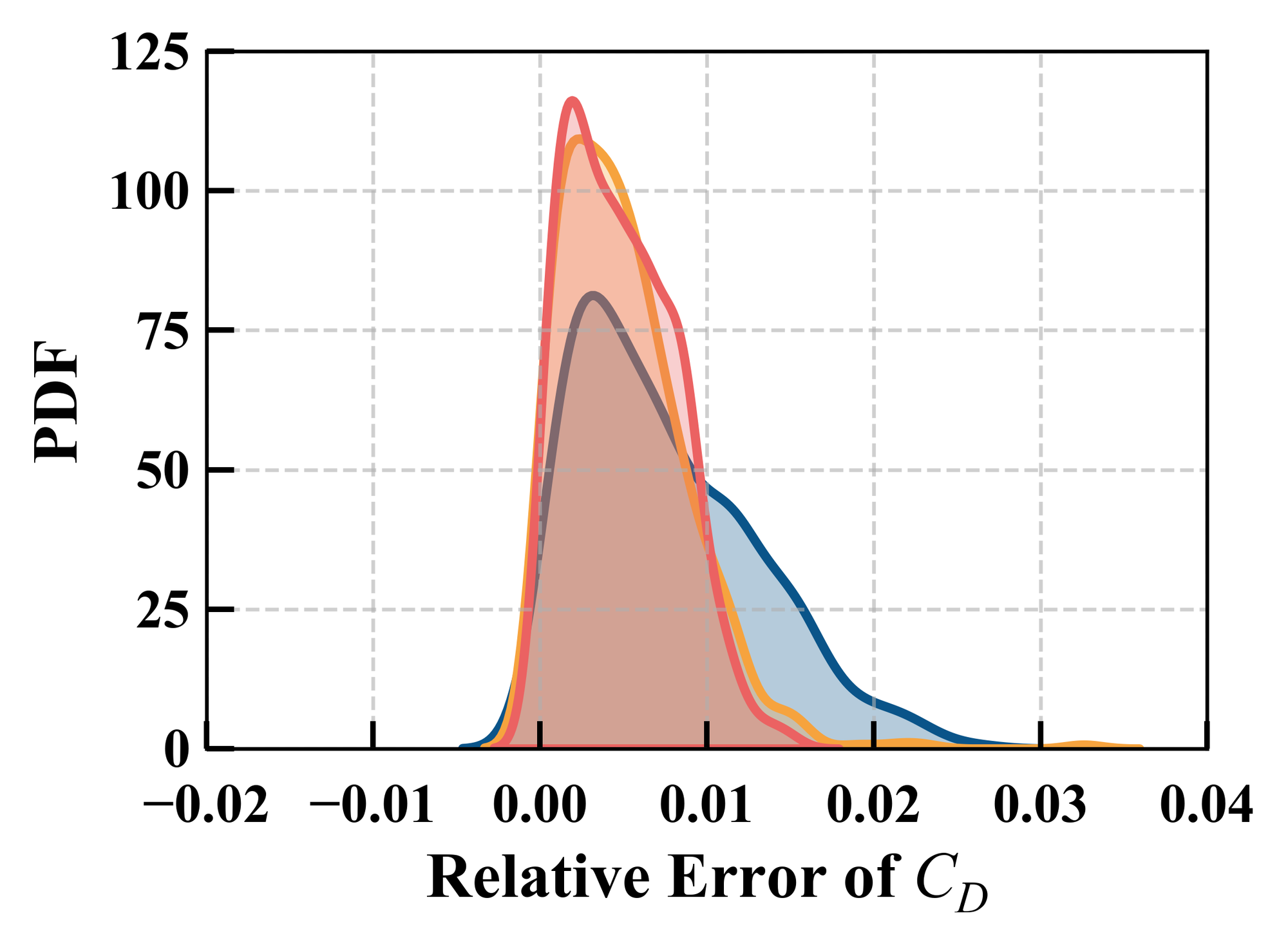}
    \caption{$C_D$}
    \label{fig:bsure}
\end{subfigure}
\hfill
\begin{subfigure}[b]{0.3\textwidth}
    \centering
    \includegraphics[width=\textwidth]{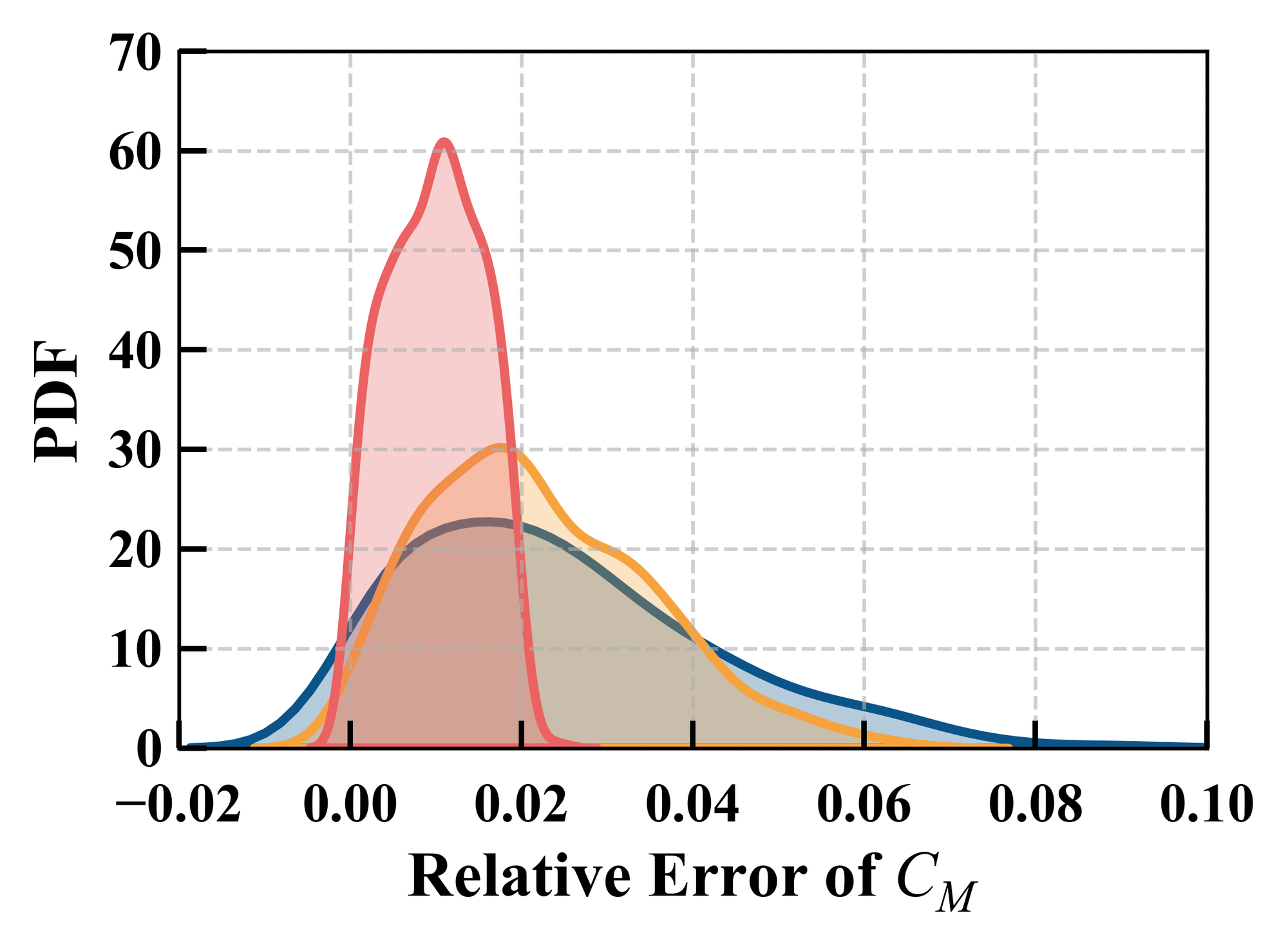}
    \caption{$C_M$}
    \label{fig:bsurf}
\end{subfigure}

\caption{Generation accuracy under interpolated aerodynamic and geometric conditions (Cond-B, on the boundary of the condition space among the training dataset). }
\label{fig:CEGAN框架完整效果-B}
\end{figure}

\begin{figure}[htbp]
\centering

\begin{subfigure}[b]{0.3\textwidth}
    \centering
    \includegraphics[width=\textwidth]{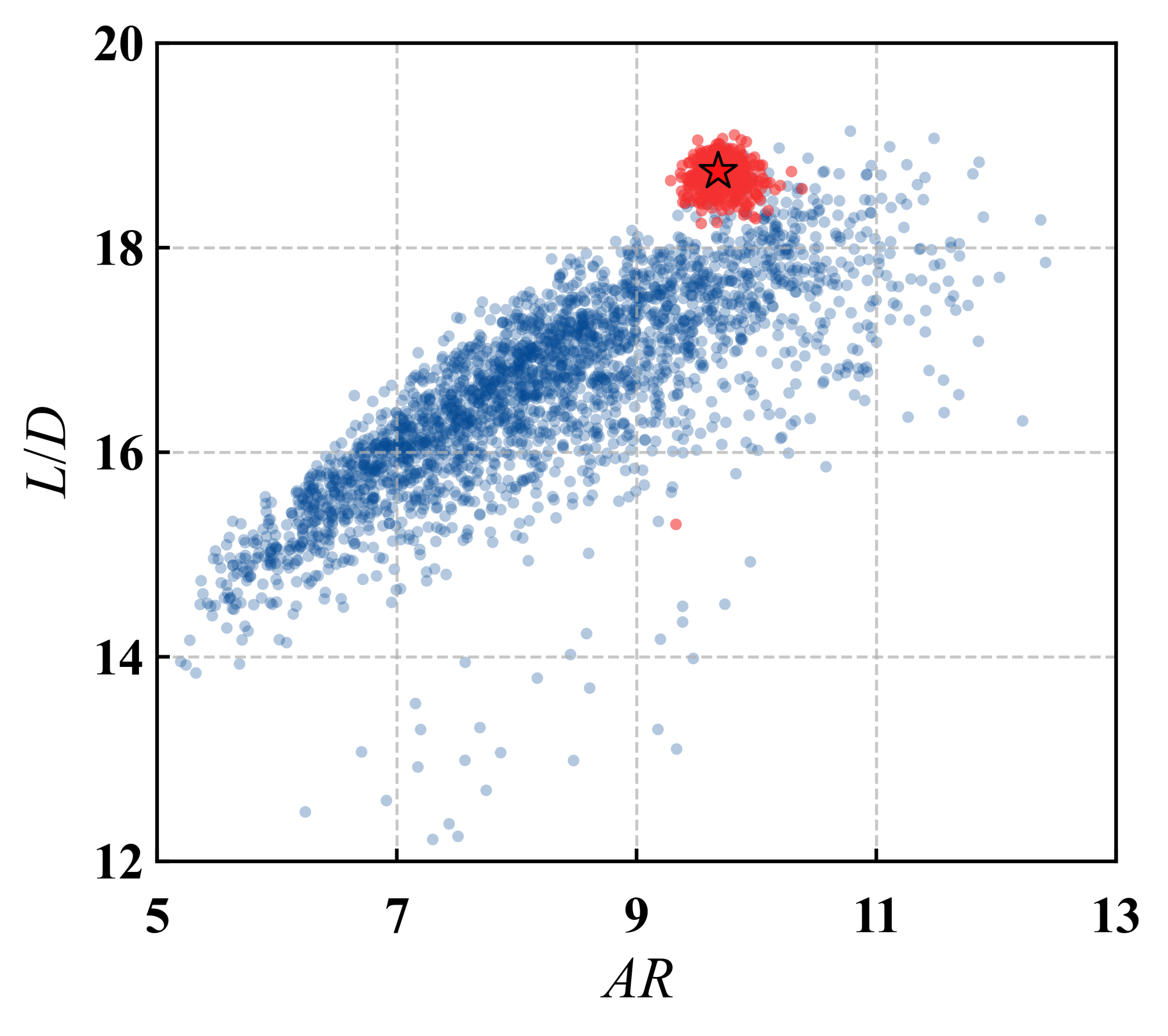}
    \caption{}
    \label{fig:bsandiana}
\end{subfigure}
\hfill
\begin{subfigure}[b]{0.3\textwidth}
    \centering
    \includegraphics[width=\textwidth]{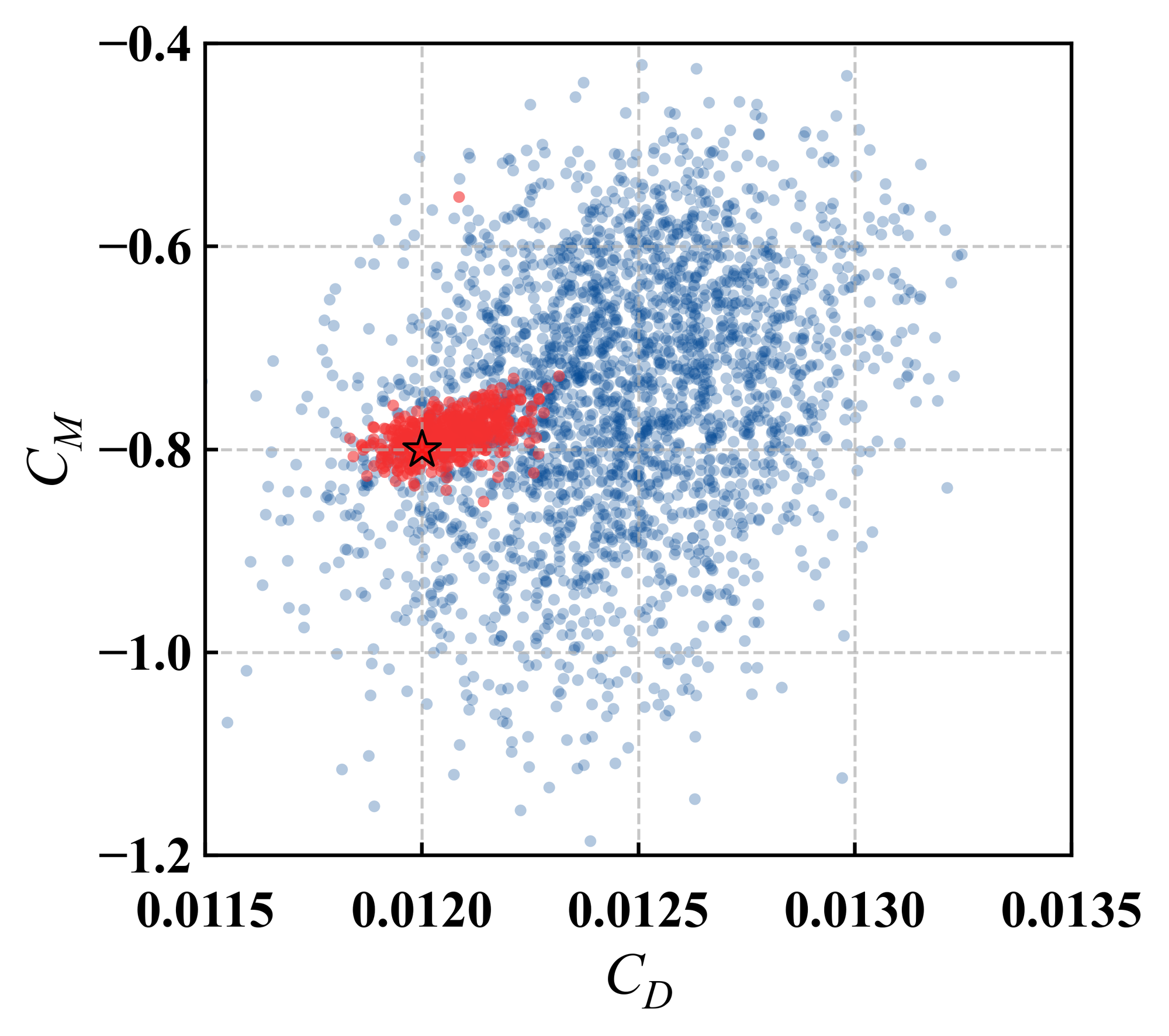}
    \caption{}
    \label{fig:bsandianb}
\end{subfigure}
\hfill
\begin{subfigure}[b]{0.3\textwidth}
    \centering
    \includegraphics[width=\textwidth]{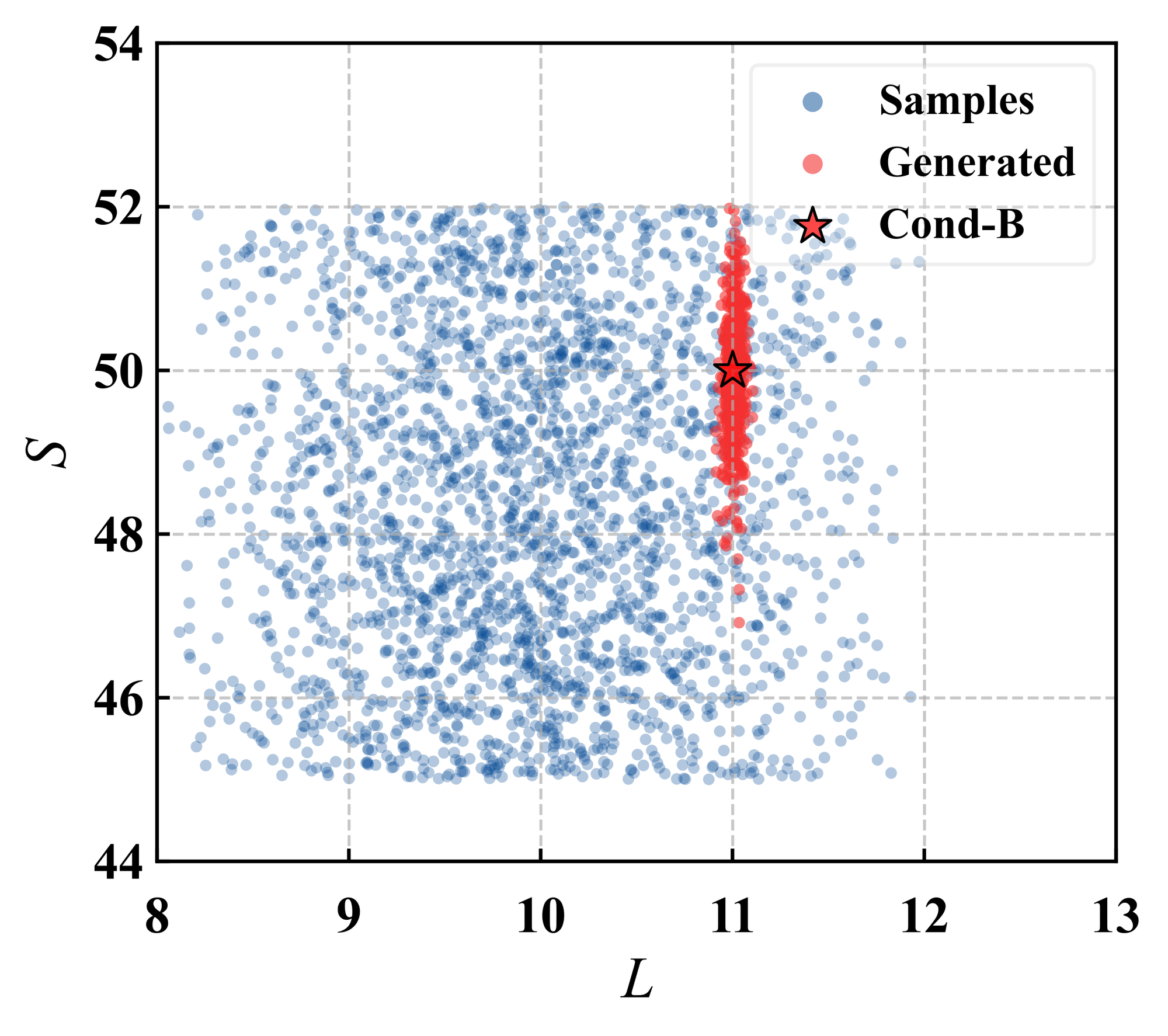}
    \caption{}
    \label{fig:bsandianc}
\end{subfigure}

\vspace{1em} 

\begin{subfigure}[b]{0.3\textwidth}
    \centering
    \includegraphics[width=\textwidth]{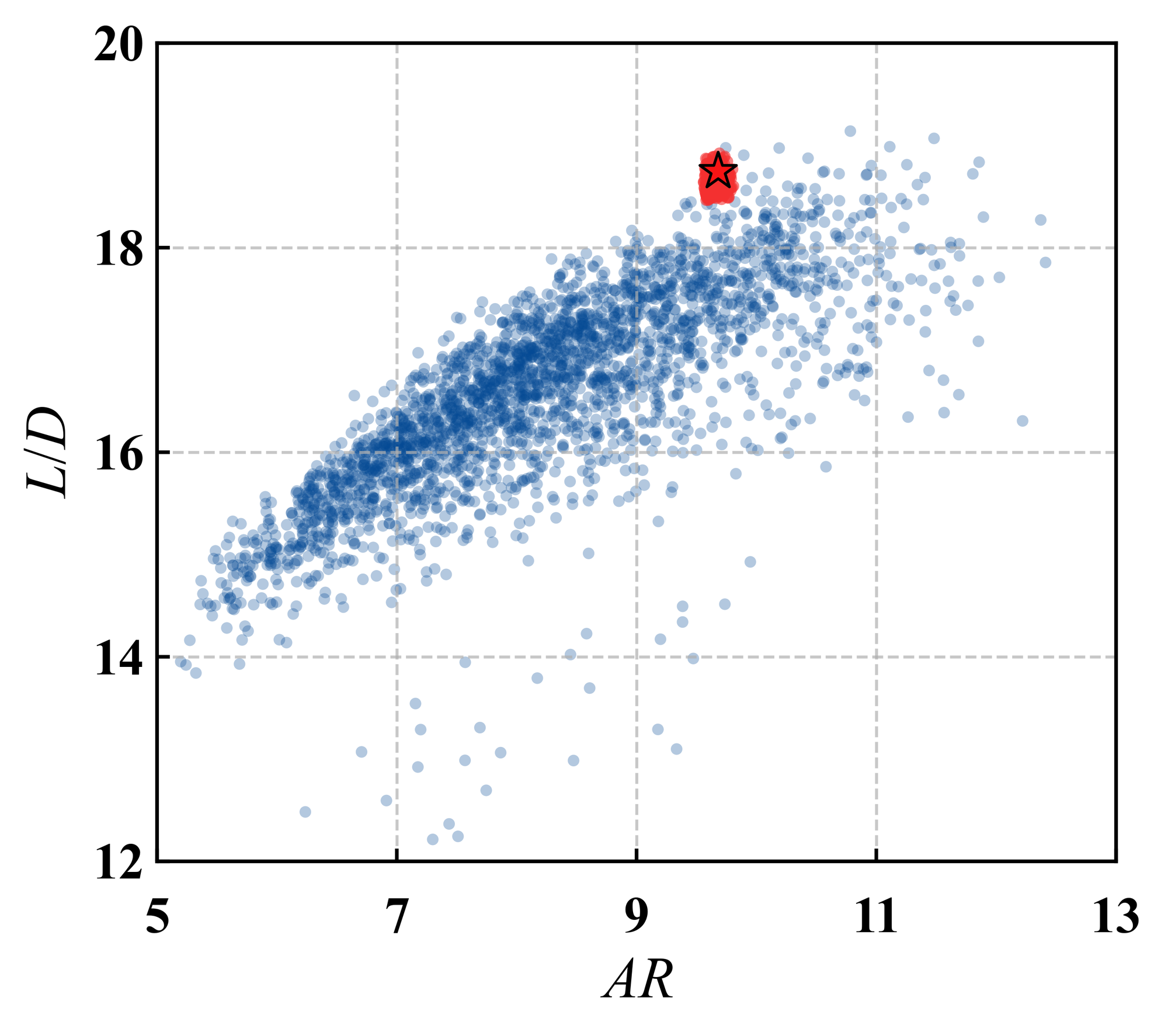}
    \caption{}
    \label{fig:bsandiand}
\end{subfigure}
\hfill
\begin{subfigure}[b]{0.3\textwidth}
    \centering
    \includegraphics[width=\textwidth]{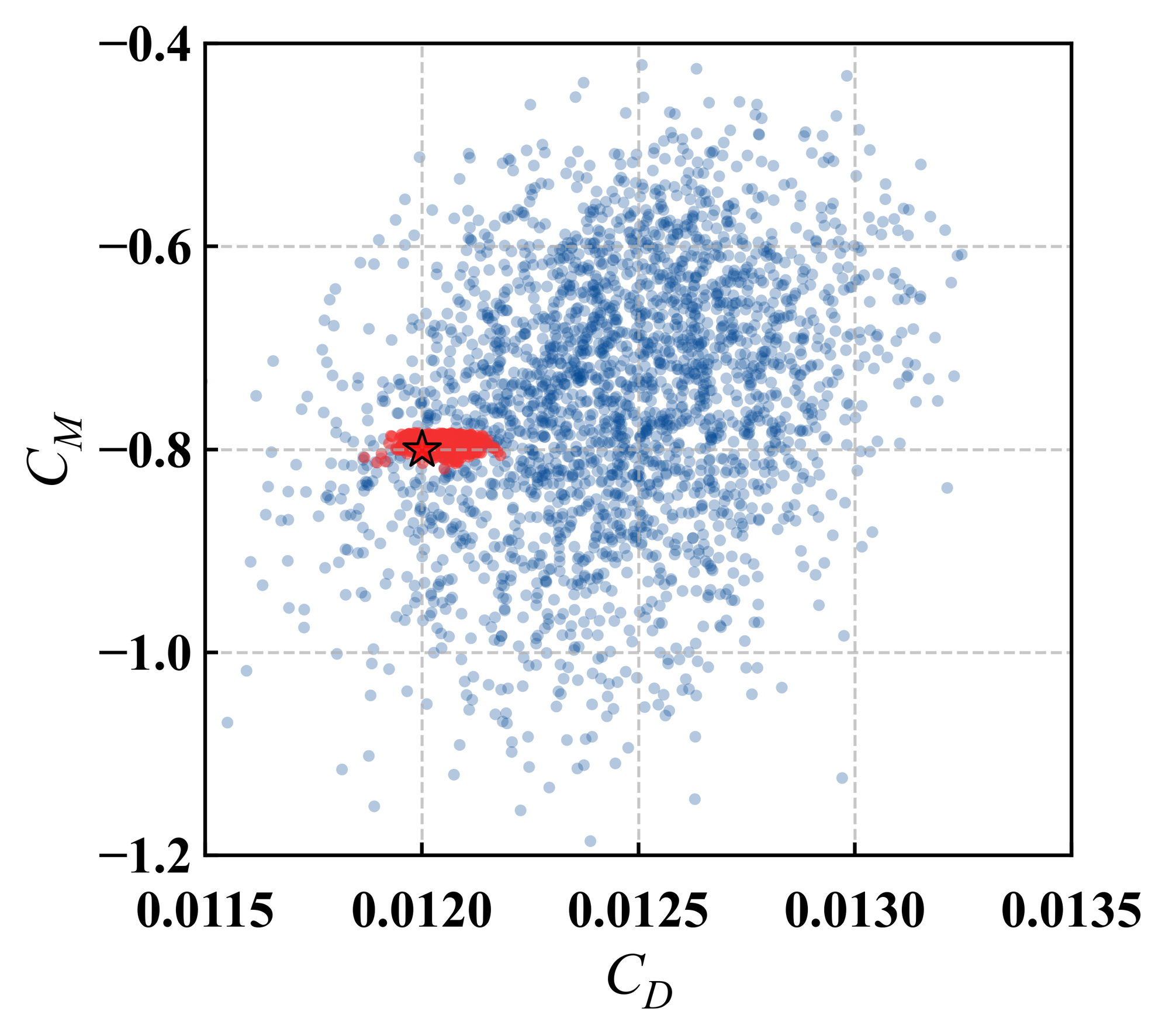}
    \caption{}
    \label{fig:bsandiane}
\end{subfigure}
\hfill
\begin{subfigure}[b]{0.3\textwidth}
    \centering
    \includegraphics[width=\textwidth]{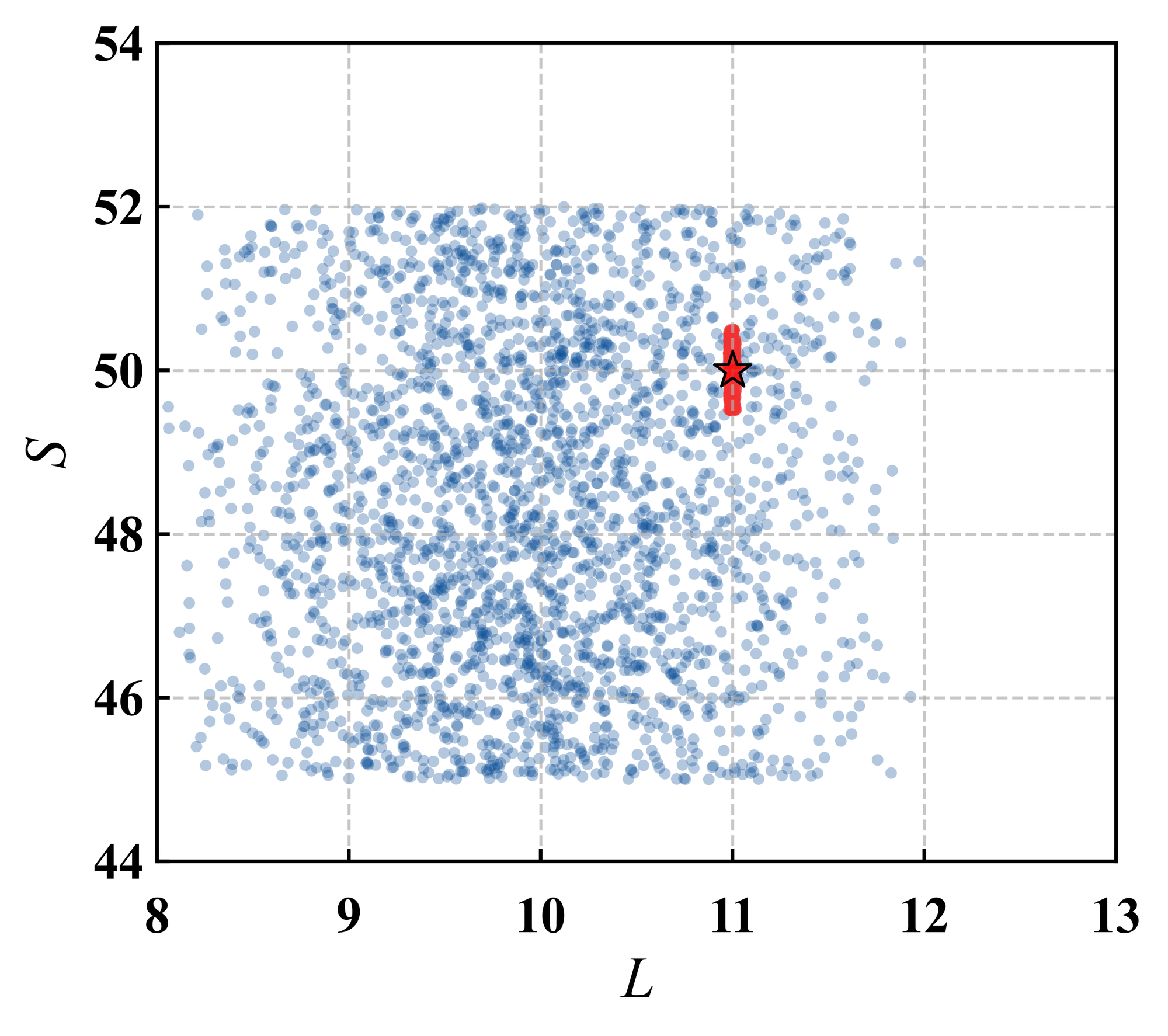}
    \caption{}
    \label{fig:bsandianf}
\end{subfigure}

\caption{Generation results of CEGAN vs. CWGAN-GP 
in different low-dimensional spaces under Cond-B. (a)--(c) CWGAN-GP; (d)--(f) CEGAN.}
\label{fig:极限散点对比-B}
\end{figure}

The validation results for the interpolated condition configuration (Condition B), located in the boundary of the data distribution space, are presented in Figure~\ref{fig:CEGAN框架完整效果-B}. It can be observed that, compare to the CWGAN-GP model, the CEGAN model without the output filtering module has already achieved effective improvements in accuracy across all aspects. The complete CEGAN framework further yields generation results with higher accuracy through the effective filtering of unfeasible outputs of the generator. Figure~\ref{fig:极限散点对比-B} visualizes the generation results for CEGAN and CWGAN-GP under the Condition B from the perspective of different condition spaces, further demonstrating the improved accuracy of CEGAN.

\clearpage
\section{Surrogate Validation and Accuracy Sensitivity}
\label{app:surrogate_validation}
\begingroup

This appendix first evaluates surrogate prediction accuracy on generated aircraft configurations. It then examines how surrogate accuracy affects aerodynamic guidance during generator training. Both tests use the same 28-parameter aircraft representation.

\subsection{Surrogate Accuracy on Generated Configurations}
\label{app:generated_surrogate_accuracy}

The first test evaluates the original surrogate models on the CEGAN configurations used for Cond-A--D in Table~\ref{tab:target}. Cond-A, Cond-B and Cond-C contain 500 generated configurations each. Cond-D contains 2000 configurations associated with its LHS design. Each generated geometry is evaluated by both its aerodynamic surrogate and VSPAERO.

For coefficient $C_i$, where $i\in\{L,D,M\}$, the error of the surrogate model is
\begin{equation}
e_{\mathrm{sur},i}(\bm{x})=
100\frac{\left|\hat C_i(\bm{x})-C_i^{\mathrm{VSP}}(\bm{x})\right|}
{\left|C_i^{\mathrm{VSP}}(\bm{x})\right|+\epsilon},
\qquad \epsilon=10^{-8}.
\label{eq:appD_surrogate_error}
\end{equation}
Here, $\hat C_i(\bm x)=f_i(\bm x)$ is the surrogate prediction and $C_i^{\mathrm{VSP}}(\bm x)$ is the VSPAERO result for the same geometry.

Figure~\ref{fig:appD_surrogate_validation} shows the error PDFs, and Table~\ref{tab:appD_generated_proxy_errors} reports the MREs of the surrogate models. The surrogate models provide useful predictions across the tested conditions. For the more challenging $C_D$ predictions under the LHS conditions of Cond-D, 92.0\% of the generated samples have errors below 10\%.

\begin{figure}[htbp]
\centering
\begin{subfigure}[b]{0.32\linewidth}
\centering
\includegraphics[width=\linewidth]{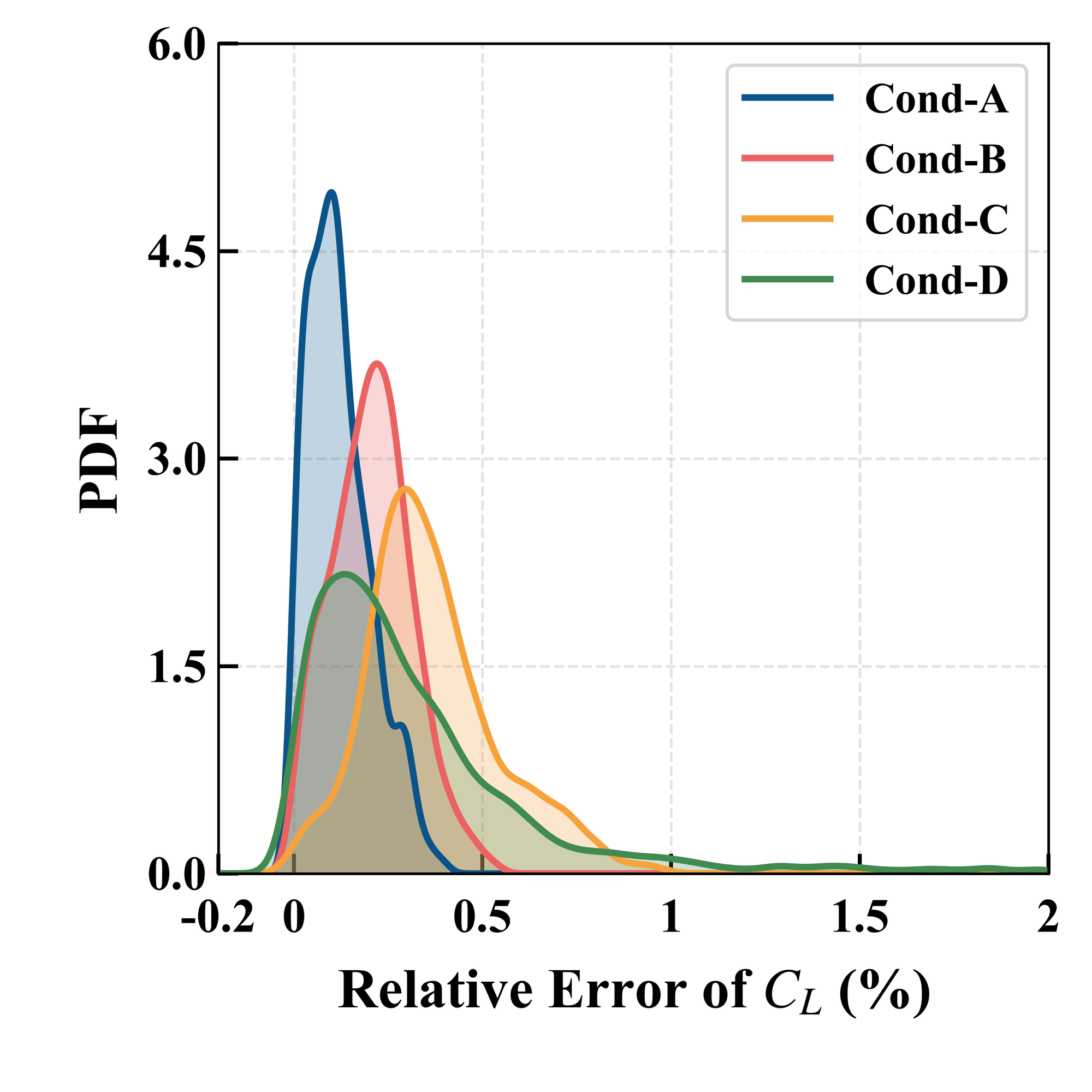}
\caption{$C_L$}
\end{subfigure}\hfill
\begin{subfigure}[b]{0.32\linewidth}
\centering
\includegraphics[width=\linewidth]{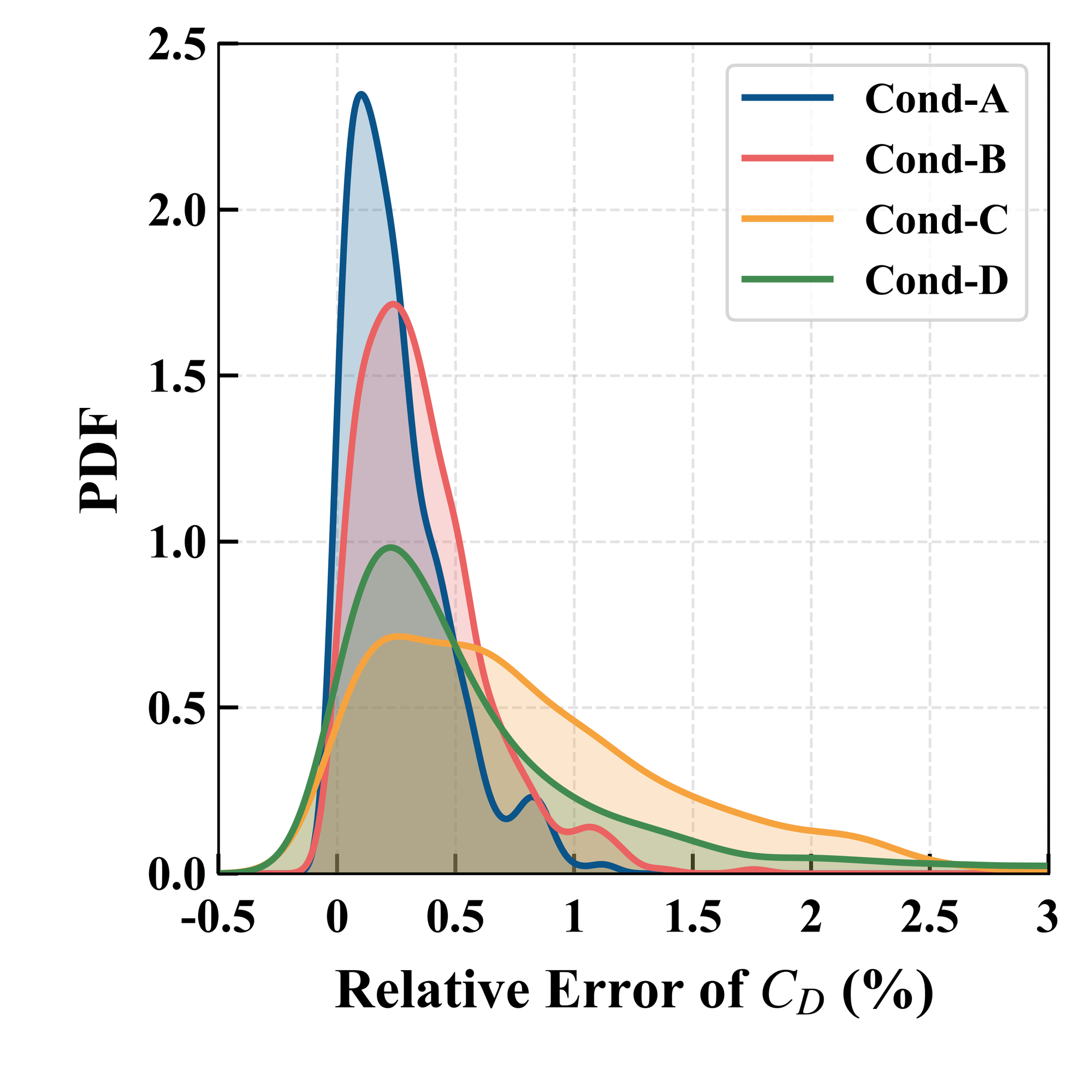}
\caption{$C_D$}
\end{subfigure}\hfill
\begin{subfigure}[b]{0.32\linewidth}
\centering
\includegraphics[width=\linewidth]{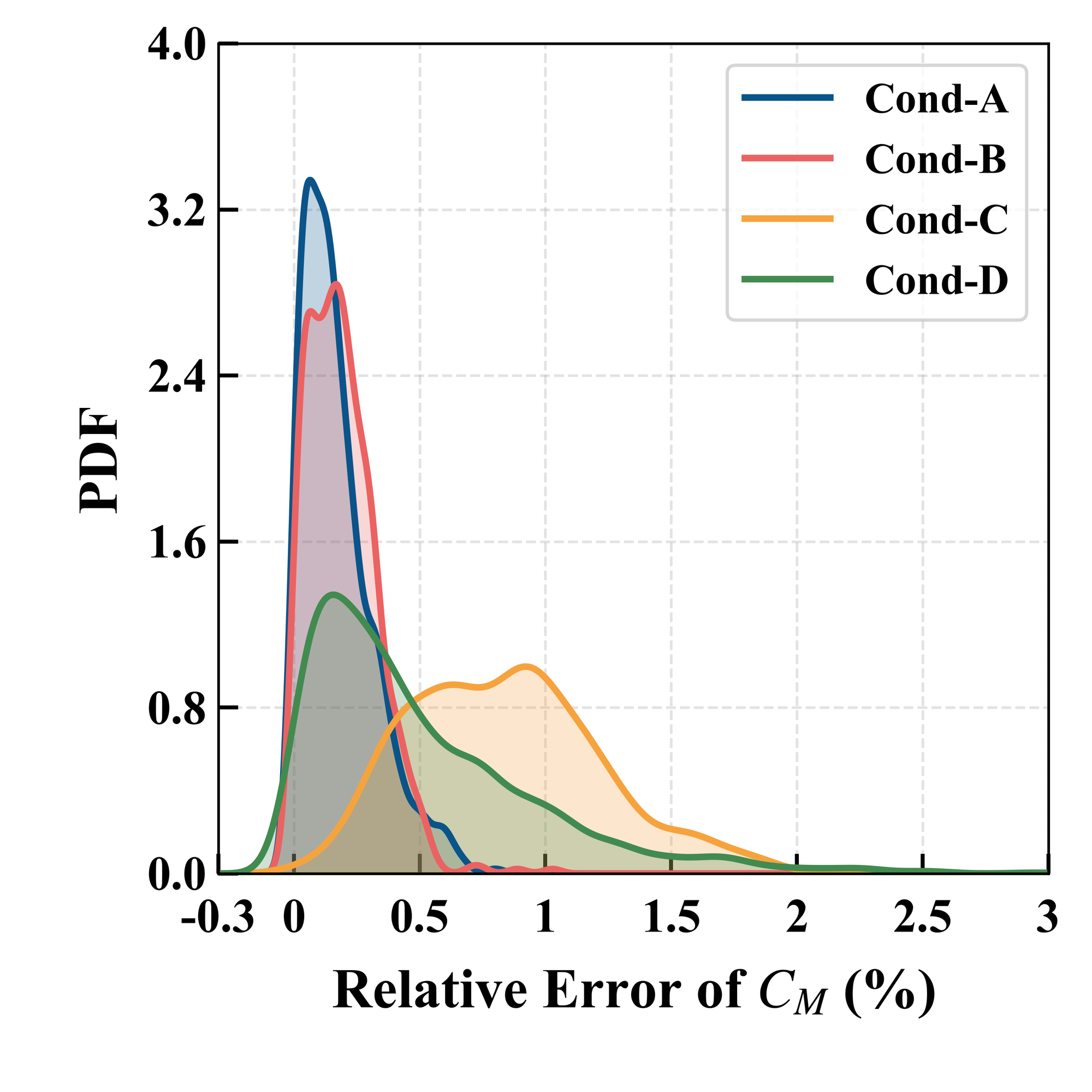}
\caption{$C_M$}
\end{subfigure}
\caption{Errors of the surrogate models relative to VSPAERO for CEGAN configurations under Cond-A--D. The panels show (a) $C_L$, (b) $C_D$ and (c) $C_M$.}
\label{fig:appD_surrogate_validation}
\end{figure}

\begin{table}[htbp]
\centering
\caption{Errors of the surrogate models on generated CEGAN configurations.}
\label{tab:appD_generated_proxy_errors}
\begin{tabular}{lrrrr}
\toprule
Condition & $N$ & \multicolumn{3}{c}{MRE, \%} \\
\cmidrule(lr){3-5}
 & & $C_L$ & $C_D$ & $C_M$ \\
\midrule
Cond-A & 500 & 0.121 & 0.254 & 0.175 \\
Cond-B & 500 & 0.206 & 0.357 & 0.193 \\
Cond-C & 500 & 0.366 & 0.781 & 0.841 \\
Cond-D & 2000 & 0.342 & 7.211 & 0.513 \\
\bottomrule
\end{tabular}
\end{table}

\subsection{Effect of Surrogate Accuracy on CE-aero}
\label{app:surrogate_sensitivity}

The second test uses CE-aero to validate aerodynamic enhancement with surrogate models of different accuracies. Geometry-guided penalties and output filtering are disabled, while all six design conditions remain inputs to the generator. Five separate generators are trained for 20000 epochs using the same architecture, training data, random seed and optimization settings. The aerodynamic loss weights for $C_L$, $C_D$ and $C_M$ are fixed at 150, 1500 and 20, respectively.

Five sets of surrogate models are trained with nominal $R^2$ values of 0.99, 0.95, 0.90, 0.80 and 0.50 for comparison. Table~\ref{tab:appD_proxy_accuracy_tiers} lists the measured $R^2$ values. Here, $\overline{R^2}$ denotes the mean of the three coefficient-wise values.

\begin{table}[htbp]
\centering
\caption{Measured coefficients of determination on the validation set for the surrogate checkpoints used in the CE-aero sensitivity study.}
\label{tab:appD_proxy_accuracy_tiers}
\begin{tabular}{ccccc}
\toprule
Nominal $R^2$ & $R^2_{C_L}$ & $R^2_{C_D}$ & $R^2_{C_M}$ & $\overline{R^2}$ \\
\midrule
0.99 & 0.9983 & 0.9944 & 0.9996 & 0.9974 \\
0.95 & 0.9500 & 0.9499 & 0.9500 & 0.9500 \\
0.90 & 0.9000 & 0.8991 & 0.9000 & 0.8997 \\
0.80 & 0.7999 & 0.7973 & 0.7999 & 0.7991 \\
0.50 & 0.4995 & 0.4960 & 0.4999 & 0.4985 \\
\bottomrule
\end{tabular}
\end{table}

Each model is assessed using the same 500 condition vectors sampled from the LHS design of Cond-D. One configuration is considered for each condition vector, without selecting it by the error of the surrogate model or the error calculated using VSPAERO. The VSPAERO calculations use $Ma=0.6$, $Re=8\times10^6$ and an angle of attack of $2^\circ$, with 20 wake iterations. These flow conditions match the aircraft design case in the main text.

For the generated geometry, the error of the generative model relative to the condition value is
\begin{equation}
e_{\mathrm{gen},i}(\bm{x},\bm{c})=
100\frac{\left|C_i^{\mathrm{VSP}}(\bm{x})-C_{i,\mathrm{cond}}\right|}{|C_{i,\mathrm{cond}}|},
\label{eq:appD_generation_error}
\end{equation}
where $C_{i,\mathrm{cond}}$ is the corresponding aerodynamic condition in $\bm c$. Figure~\ref{fig:appD_surrogate_guidance} shows the relationship between surrogate prediction accuracy and generation accuracy. In this sensitivity test, the error of the surrogate model is $100|\hat C_i(\bm x)-C_i^{\mathrm{VSP}}(\bm x)|/|C_i^{\mathrm{VSP}}(\bm x)|$. All evaluated reference values are nonzero. The horizontal dashed line is the historical mean error of CWGAN-GP over the same 500 condition vectors. For each model, the dark points show the mean errors in ten equal-count groups of samples sorted by surrogate prediction error. The colored lines connect these points to show the overall trend.

\begin{figure}[htbp]
\centering
\begin{subfigure}[b]{0.32\linewidth}
\centering
\includegraphics[width=\linewidth]{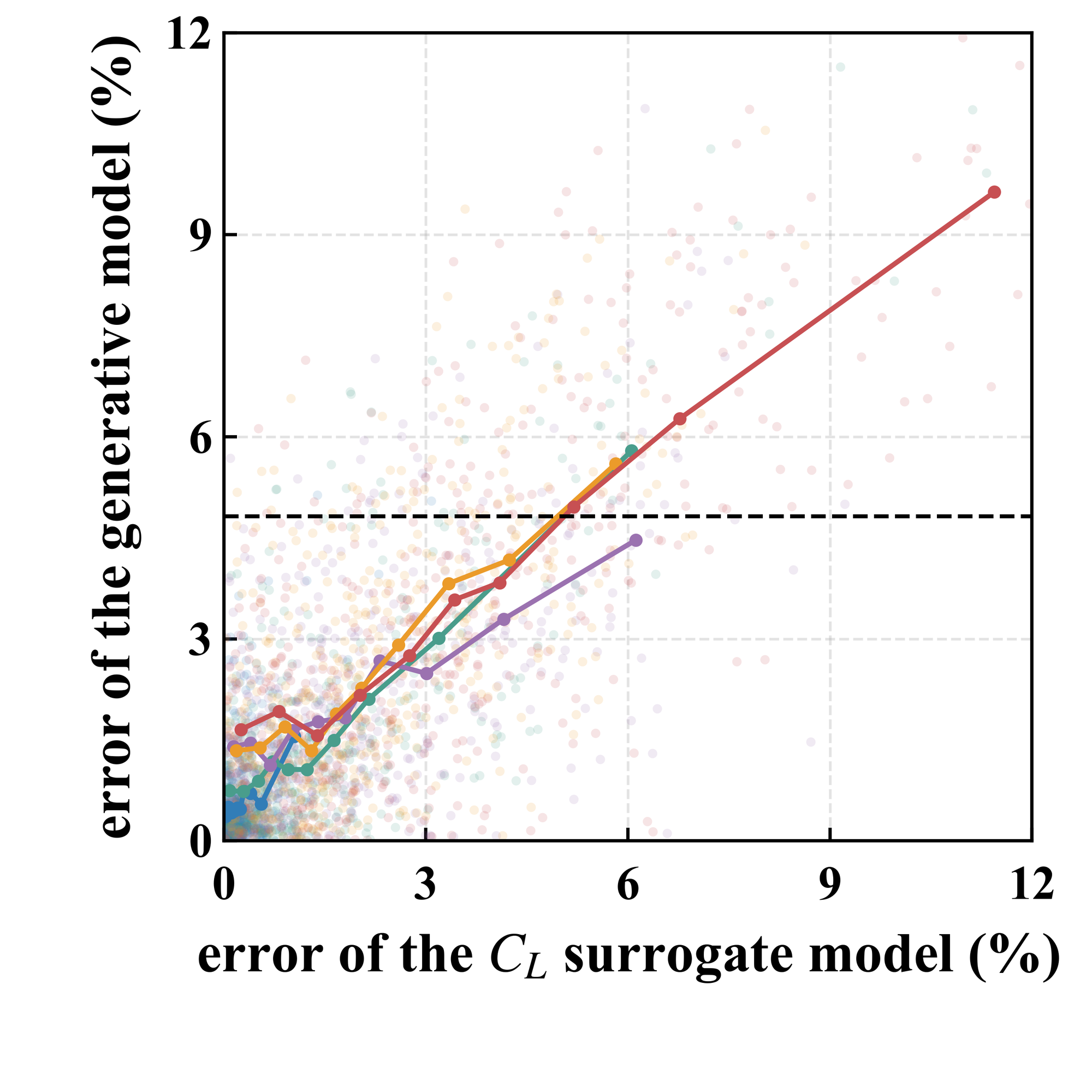}
\caption{$C_L$}
\end{subfigure}\hfill
\begin{subfigure}[b]{0.32\linewidth}
\centering
\includegraphics[width=\linewidth]{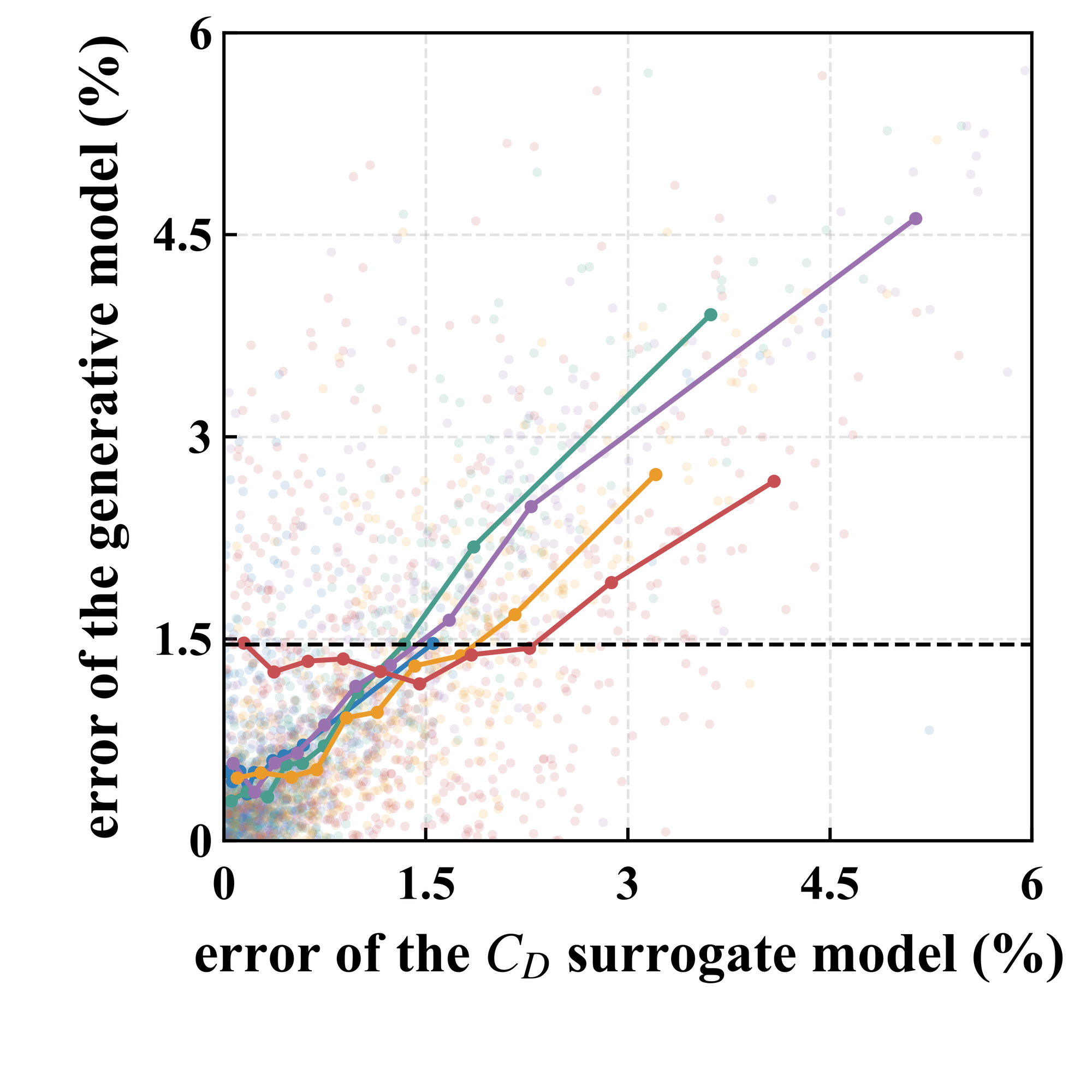}
\caption{$C_D$}
\end{subfigure}\hfill
\begin{subfigure}[b]{0.32\linewidth}
\centering
\includegraphics[width=\linewidth]{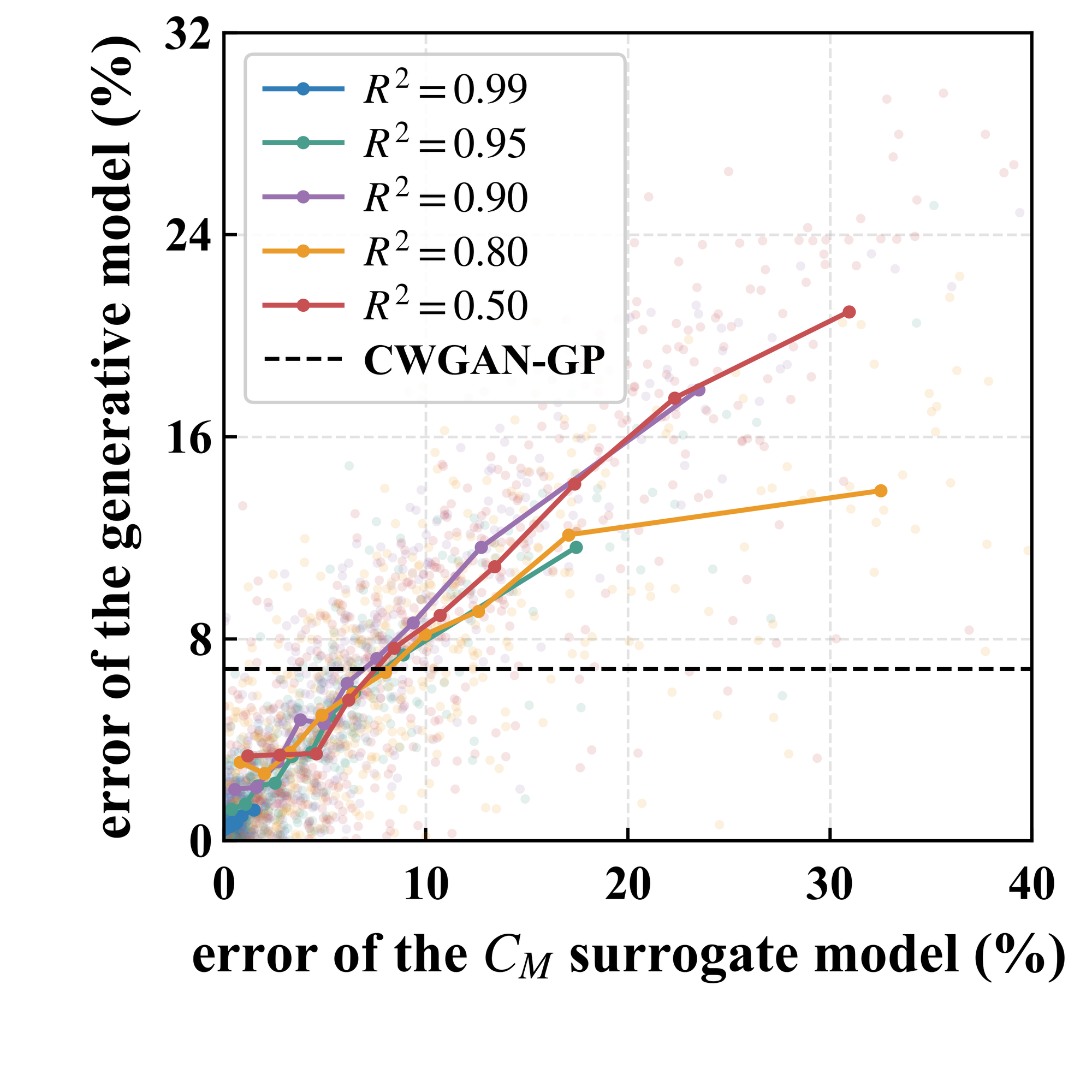}
\caption{$C_M$}
\end{subfigure}
\caption{Relation between the errors of the surrogate models and those of the CE-aero generators for (a) $C_L$, (b) $C_D$ and (c) $C_M$.}
\label{fig:appD_surrogate_guidance}
\end{figure}

These results show that generation accuracy generally improves as surrogate accuracy increases. In this case, surrogate models with $R^2$ values of about 0.95 or higher provide effective guidance for the generator. LHS combines different geometric and aerodynamic conditions. Some combinations may be physically infeasible, leading to poor generation accuracy. This may explain why some samples have accurate surrogate predictions but poor generation accuracy. The overall trend remains evident.

\endgroup

\section*{Acknowledgments}
This work was supported by the National Key Research and Development Program of China (2024YFB3310401), the National Natural Science Foundation of China (U2441211), the foundation of National Key Laboratory of Aircraft Configuration Design (JBGS-202504), the Natural Science Basic Research Program of Shaanxi (2025JC-YBQN-087), and the National Natural Science Foundation of China Excellent Young Scientists Fund (Overseas).

\end{document}